%% file: ShaiBagonPhDThesis.tex
\documentclass[12pt,a4paper]{thesis}

\pdfoutput=1

\input{predef}

\begin{document}

\pagenumbering{roman}

% title page
\includepdf[pages=1,lastpage=1]{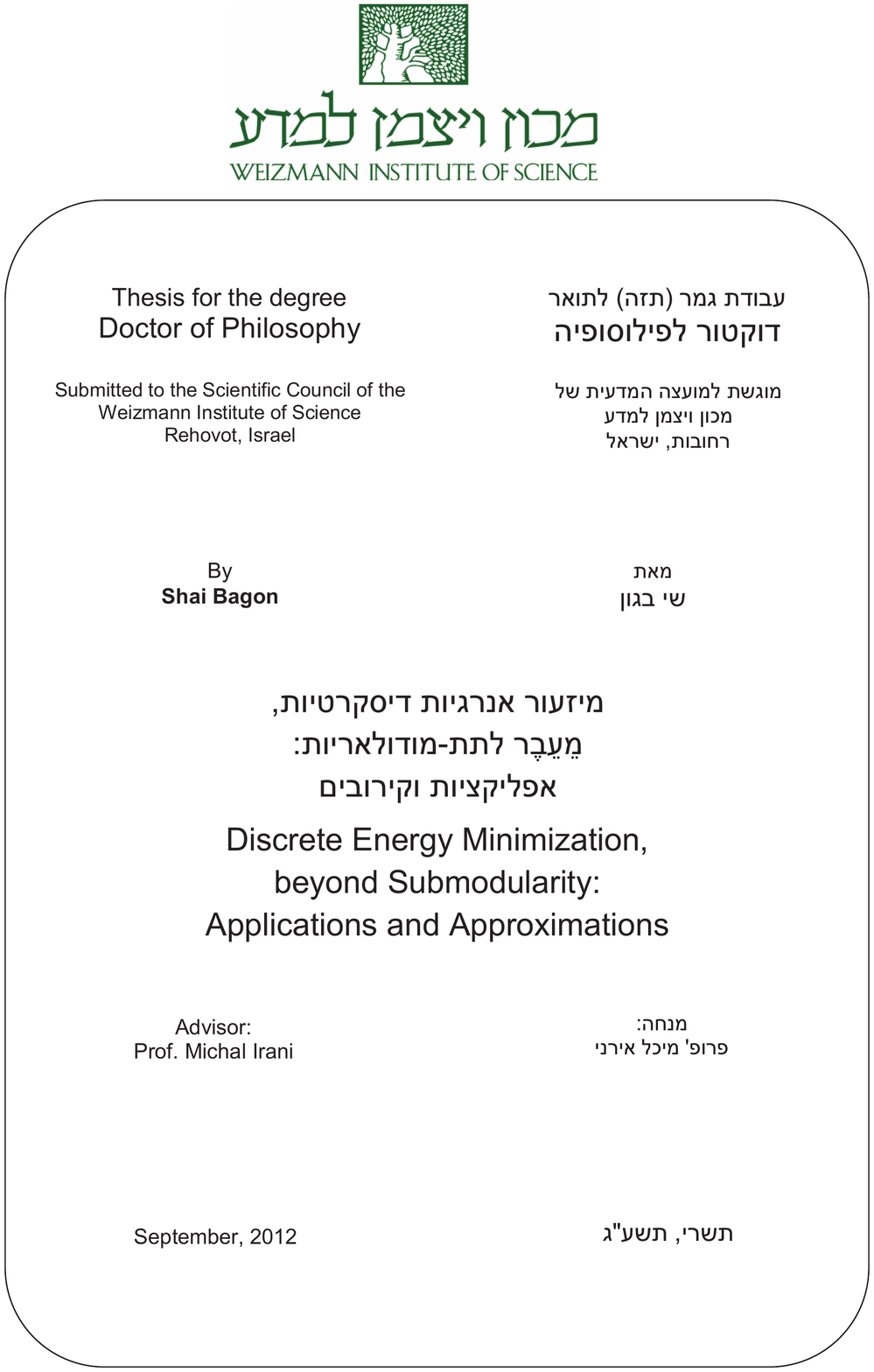}

% dedication here
\includepdf[pages=1]{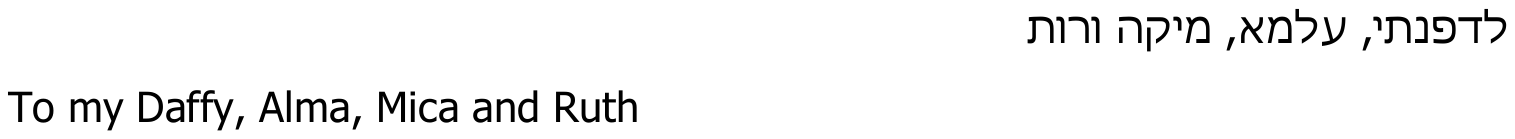}

\clearpage

\tableofcontents

\clearpage

% English abstract
\input{abstract}

% Hebrew abstract + acknowledgements
\includepdf[pages={3,2}]{./abstract_heb.pdf}

\clearpage

% English acknowledgements
% \input{acknowledge}

\setcounter{page}{0}
\pagenumbering{arabic}

%---------------------------------------------------------------------------%
% discrete energy optimization - introduction
% existing methods and approaches
% main research questions
% outline of this work
\input{introduction_main}

%---------------------------------------------------------------------------%
\input{applications_main}

%---------------------------------------------------------------------------%
\input{approximations_main}

%---------------------------------------------------------------------------%
% A short discussion of the ENTIRE study that discusses the work as a whole.
\input{discussion}

%---------------------------------------------------------------------------%
\clearpage

%%%
%{\small
\bibliographystyle{abb-named}
\bibliography{phd_final,disc_opt,dtf,lighting,ms,cc,sketch}
\addcontentsline{toc}{part}{Bibliography}
%}
%---------------------------------------------------------------------------%

\comment{
\appendix
\section*{List of Accepted Publications}
\begin{itemize}
\item {\em Shai Bagon}, {\bf Boundary Driven Interactive Segmentation}.
Published in the proceedings of the 3$^{rd}$ International Conference on Information Science and Applications (ICISA), 2012.

\item {\em Sebastian Nowozin, Carsten Rother, Shai Bagon, Toby Sharp, Bangpeng Yao and Pushmeet Kohli},
{\bf Decision Tree Fields}.
Published in the proceedings of the 13$^{th}$ International Conference on Computer Vision (ICCV), 2011.

\item {\em Shai Bagon, Or Brostovsky, Meirav Galun and Michal Irani},
{\bf Detecting and Sketching the Common}.
Published in the proceedings of the 23$^{rd}$ IEEE Conference on Computer Vision and Pattern Recognition (CVPR), 2010.
\end{itemize}
}

\end{document}

%% file: predef.tex
\usepackage[USenglish]{babel}
\usepackage{graphicx}

\usepackage[usenames]{color}
\definecolor{citecl}{rgb}{0, 0, .8} %color for citation links
\definecolor{linkcl}{rgb}{.1, .6, .1} % color for internal links

\definecolor{Blue}{rgb}{0,0,1}
\definecolor{Red}{rgb}{1,0,0}
\definecolor{Green}{rgb}{0,.8,0}
\definecolor{Magenta}{rgb}{1,0,1}

\definecolor{swap}{rgb}{0,0,1}
\definecolor{expand}{rgb}{0.44,0.44,0.44}
\definecolor{trws}{rgb}{0,1,1}
\definecolor{bp}{rgb}{1,0,0}
\definecolor{icm}{rgb}{0,1,0}
\definecolor{rlp}{rgb}{1,0,1}
\definecolor{coarse}{rgb}{0.753, 0.314, 0.302} % coarse variables in figure
\definecolor{fine}{rgb}{0.310, 0.506, 0.741}   % for fine scale variables in figure
\definecolor{ours}{rgb}{.8, 0, 0} % {.969, .588, .275}   % four our results in tables
\usepackage{makeidx}
\usepackage{picins}
\usepackage{multirow}
\usepackage{amsfonts}
\usepackage{amssymb}
\usepackage{amsmath}
\usepackage{amsthm}
\usepackage{bbm}
\usepackage{natbib}
\usepackage[ruled,vlined]{algorithm2e} % for writing pseudo-code
\usepackage[final]{pdfpages}

\usepackage{stmaryrd} % pretty brackets
\usepackage{floatflt}

\usepackage{slashbox} % for diagonal split table entry

\usepackage{appendix} % for appendix at section level rather than chapter (C. multiscale)

\newcommand{\TT}{\mathcal{T}}

\newcommand{\bx}{\mathbf{x}}
\newcommand{\by}{\mathbf{y}}
\newcommand{\Fx}{F\left(\bx\right)}

\newcommand{\phiij}{\varphi_{ij}}
\newcommand{\phiijx}{\varphi_{ij}\left(x_i,x_j\right)}

\newcommand{\half}{\frac{1}{2}}

\newcommand{\marg}[1]{\mathbb{M}\left({#1}\right)}
\newcommand{\local}[1]{\mathbb{L}\left({#1}\right)}

\newcommand{\dotprod}[2]{\left<{#1},{#2}\right>}

\newcommand{\phix}{\phi\left(\bx\right)}
\newcommand{\comment}[1]{}
\newcommand{\abs}[1]{\mbox{$\left|#1\right|$}}

\newcommand{\norm}[1]{\left\| #1 \right\|}
\newcommand{\uut}{\left[UU^T\right]_{ij}}

\newcommand{\TabCenter}[2]{\begin{minipage}[t]{#1}\centering #2\end{minipage}}
\newcommand{\etal}{\mbox{{\em et~al.}}}
\newcommand{\todo}[1]{\fbox{\bf TODO:} #1.\\ }

\newcommand{\deff}{\mbox{$\stackrel{\mbox{\tiny{def}}}{=}$}}

\newcommand{\GG}{\mbox{$\mathcal{G}$}}
\newcommand{\VV}{\mbox{$\mathcal{V}$}}
\newcommand{\EE}{\mbox{$\mathcal{E}$}}

\newcommand{\calS}{\mbox{$\mathcal{S}$}}
\newcommand{\calpha}{\boldsymbol{\alpha}}
\newcommand{\cbeta}{\boldsymbol{\beta}}
\newcommand{\calV}{\mathcal{V}}
\newcommand{\calX}{\mathcal{X}}
\newcommand{\calY}{\mathcal{Y}}
\newcommand{\calL}{\mathcal{L}}
\newcommand{\calF}{\mathcal{F}}
\newcommand{\w}{\boldsymbol{w}}
\newcommand{\y}{\boldsymbol{y}}
\newcommand{\x}{\boldsymbol{x}}

\newtheorem{definition}{Definition}

\newtheorem{claim}{Claim}[chapter]

%% file: abstract.tex
%---------------------------------------------------------------------------%
\begin{abstract}
In this thesis I explore challenging discrete energy minimization problems that arise mainly in the context of computer vision tasks.
This work motivates the use of such ``hard-to-optimize" non-submodular functionals, and proposes methods and algorithms to cope with the NP-hardness of their optimization.
Consequently, this thesis revolves around two axes: applications and approximations.
The applications axis motivates the use of such ``hard-to-optimize" energies by introducing new tasks.
As the energies become less constrained and structured one gains more expressive power for the objective function achieving more accurate models.
Results show how challenging, hard-to-optimize, energies are more adequate for certain computer vision applications.
To overcome the resulting challenging optimization tasks the second axis of this thesis proposes approximation algorithms to cope with the NP-hardness of the optimization.
Experiments show that these new methods yield good results for representative challenging problems.    

\end{abstract}

%---------------------------------------------------------------------------%

\clearpage

%% file: introduction_main.tex
\part{Introduction}

In this thesis I explore challenging discrete energy minimization problems that arise mainly in the context of computer vision.
From binary energies for figure-ground segmentations through multi-label semantic segmentation, stereo, denoising, to inpainting and image editing (e.g., \cite{Szeliski2008,Pritch2009,Bagon2012icisa}).
These energies usually involve thousands of variables and dozens of discrete states.
Moreover, most of the energies in this domain are pair-wise energies, that is, they only involve interactions between pairs of neighboring variables.

The optimization of these discrete energies is known to be NP-hard in most cases (\cite{Boykov2001}).
Still, despite this theoretical hardness, instances of these energies that have special properties may give rise to polynomial time global optimal algorithms.
Other instances with slightly different properties allow, in practice, for good and efficient approximation schemes.

The next chapter~(Chap.~\ref{cp:related-work}) reviews previous work related to discrete pair-wise energy functions and their optimization.
It outlines the properties and conditions under which global optimization is feasible, and the conditions required for successful practical approximations.
Chapter~\ref{cp:related-work} also surveys several key approximation algorithms.
It provides a brief outline of the properties of the energy that must be met in order for each algorithm to succeed.
The conclusion of this survey is that discrete pair-wise energies may be broadly classify into two categories: \*

\noindent{\bf smoothness-encouraging} energies: energies that favor configurations with neighboring variables taking the same discrete state.\*

\noindent{\bf contrast-enhancing} energies: energies that encourage solutions where neighboring variables take different states.\*

So far the energies mainly used in computer vision tasks are of the first category: smoothness-encouraging (see e.g., \cite{Szeliski2008}).
These smoothness-encouraging energies allow for efficient approximation schemes.
On the other hand, contrast-enhancing energies are far more challenging when it comes to optimization, and are indeed less popular in practice.

\pagebreak

In this thesis I would like to step outside of this ``comfort-zone" of the smoothness-encouraging energies and explore more challenging discrete energies.
This work revolves around two axes:
\begin{enumerate}
\item {\bf Applications:} The first motivates the use of such ``hard-to-optimize" functionals by introducing new  applications.
    As the energies become less constrained and structured one gains more expressive power for the objective function achieving more accurate models.
    Results show how contrast-enhancing, hard-to-optimize, functionals are more adequate for certain computer vision tasks.

\item {\bf Approximations:} To overcome the resulting challenging optimization tasks the second axis of this thesis proposes methods and algorithms to cope with the NP-hardness of this optimization.
    Experiments show that these new methods yield good results for representative challenging problems.
\end{enumerate}
%The second axis proposes {\em approximation} algorithms to cope with the resulting challenging energy minimization problems.
%This work motivates the use of such ``hard-to-optimize" functionals:

%---------------------------------------------------------------------------%
% A short introduction to the ENTIRE study.
\chapter{Discrete Pair-wise Energies -- a Review}
\label{cp:related-work}

Discrete energy minimization is a ubiquitous task in computer vision.
From binary energies for figure-ground segmentations through multi-label semantic segmentation, stereo, denoising, to inpainting and image editing (e.g., \cite{Szeliski2008,Pritch2009,Bagon2012icisa}).
In my thesis I focus on various types of minimization problems of pair-wise energies as they arise in different computer vision applications.
%
% Describe here in brief why the problem is hard, and how my work relates to it
%
These discrete optimization problems are, in general, NP-hard.
Yet, there are cases in which the minimization of a pair-wise energy can be solved {\em exactly} in polynomial time.
In this introductory chapter I survey different properties of discrete pair-wise energies.
I show how these properties of the energies relate to the inherent difficulty of the optimization task.
Some properties entail exact optimization algorithms, while other properties admit efficient approximations.
The most important property is ``smoothness-encouraging": an energy that prefers the labels of neighboring variables to be the same.
For these ``smoothness-encouraging" energies there exist efficient approximate minimization algorithms.
In contrast, energies that encourage neighboring variables to have {\em different} labels
are much more challenging to minimize.
For these ``contrast-enhancing" energies existing algorithms provide poor approximations.
This thesis focuses on the optimization of these challenging ``contrast-enhancing" energies.

%---------------------------------------------------------------------------%
\section{Pair-wise Energy Function}

Before diving into the minimization task, this section presents the discrete pair-wise energy function and the notations that are used in this thesis.
It also provides some insights and motivation for the use of such energies.

A discrete pair-wise energy is a functional of the form
\begin{equation}
\Fx =  \sum_{ij\in\EE} \varphi_{ij}\left(x_i, x_j\right) + \sum_i \varphi_i\left(x_i\right)
\label{eq:pair-wise-gen}
\end{equation}

It is defined over $n$ variables ($x_i$, $i=1,\ldots,n$), each taking one of $l$ discrete labels ($x_i\in\left\{1,\ldots,l\right\}$), where $\EE$ represents a set of neighboring variables.
The term $\varphi_i\left(x_i\right)$ is a unary term reflecting the compatibility of label $x_i$ to variable $i$ (also known as a ``data term").
The pair-wise term, $\phiij\left(x_i,x_j\right)$ reflects the interaction between labels $x_i$ and $x_j$ assigned to variables $i$ and $j$ respectively.

If we were to discard the pair-wise term, minimizing energy~(\ref{eq:pair-wise-gen}) is simply choosing the label that best fit each variable separately.
However, the presence of the pair-wise term introduces dependencies between the different variables and turns the optimization into a much more complicated process.
Despite the local nature of the pair-wise term -- binding the values of only neighboring variables introduces {\em global} effects on the overall optimization.
Propagating the information from the local pair-wise terms to form a global solution is a major challenge for the optimization of Eq.~(\ref{eq:pair-wise-gen}).

%Mathematical properties of the pair-wise terms ($\psi_{ij}$) determine how difficult the optimization is.

The energy function of Eq.~(\ref{eq:pair-wise-gen}) has an underlying structure defined by the choice of interacting neighbors.
It is common to associate with $\Fx$ a graph $\GG=\left(\VV,\EE\right)$, where the set of nodes $\VV=\left\{1,\ldots,n\right\}$ represents the variables, and the edges $\EE$ represents the interacting neighboring pairs.

The following example shows how such discrete functional may arise in a well studied computer vision application.
This example also illustrates the relation between discrete optimization and inference in graphical models.

%Generally speaking, the sum over the pairwise terms measure the extent to which $x$ is not piecewise smooth, while the sum over the unary terms measures the disagreement between $x$ and the observed data.

\begin{example}

\hrulefill

\noindent{\bf Example: Stereo reconstruction via MRF representation}

% This short example illustrates the relation between Markov Random Field (MRF) and discrete functionals via a
% well-known computer vision problem, namely the stereo problem.
Given a rectified stereo pair of images, the goal is to find the disparity of each pixel in the reference image.
The true disparity of each pixel is a random variable denoted by $x_i$ for the pixel at location $i$.
Each variable can take one of $L$ discrete states, which represent the possible disparities at that point.
For each possible disparity value, there is a cost associated with matching the pixel in the reference image to the corresponding pixel in the other image at that disparity value.
Typically, this cost is based on the intensity differences between the two pixels, $y_i$, which is an observed quantity.
We denote this cost by $\Phi(x_i,y_i)$.
It relates how compatible a disparity value $x_i$ is with the observed intensity difference $y_i$.
A second function $\Psi(x_i,x_j)$ expresses the disparity compatibility between neighboring pixels.
This function usually expresses the {\em prior} assumption that the disparity field should be smooth.
Examples of such prior that are commonly used are the Potts model:
\begin{equation}
\Psi(x_i,x_j) = \left\{
\begin{array}{cl}
1 & \mbox{if $x_i = x_j$} \\
0 & \mbox{otherwise}
\end{array}
\right.
\end{equation}
the $\ell_1$ similarity:
\begin{equation}
\Psi(x_i,x_j) \propto e^{-\left|x_i-x_j\right|}
\end{equation}
or its robust (truncated) version:
\begin{equation}
\Psi(x_i,x_j) \propto e^{-\min\left\{\left|x_i-x_j\right|,\tau\right\}}
\end{equation}

With the two functions $\Phi$ and $\Psi$ the joint probability for an assignment of disparities to pixels can be written as:
\begin{equation}
P(\bx,\by) \propto \prod_{ij\in\EE} \Psi(x_i,x_j) \prod_{i} \Phi(x_i,y_i)
\label{eq:intro-exmp-mrf-prob}
\end{equation}
where $\bx$ is an assignment of a disparity value for each pixel $i$.
$x_i$ is the hidden variable (disparity) at location $i$ and $y_i$ is the observed variable (intensity difference) at location $i$.
${ij\in\EE}$ represent a pair of neighboring nodes and is usually taken as a regular 4-connected grid over the image domain.

The resulting graphical model is known as a pairwise Markov Random Field.
Although the compatibility functions only consider adjacent variables, each variable is still able to influence every other variable in the field via these pairwise connections.

We look for a disparity assignment such that the labeling $\bx^\star$ maximizes the joint probability, i.e.,
\begin{equation}
\bx^\star = \underset{\bx}{\operatorname{argmax}} ~~~ P(\bx,\by)
\end{equation}
Assuming uniform prior over all configurations $\bx$, the Maximum A Posteriori (MAP) estimator is equivalent to
\begin{equation}
\bx^\star = \underset{\bx}{\operatorname{argmax}} ~~~ P(\bx\vert\by)
\end{equation}

Maximizing the posterior probability is equivalent to minimizing an energy functional of the form (\ref{eq:pair-wise-gen}).
Taking $-\log$ of (\ref{eq:intro-exmp-mrf-prob}) yields the following function
\begin{equation}
\Fx = \sum_{ij\in\EE} -\log\Psi(x_i,x_j) + \sum_{i} -\log\Phi(x_i,y_i)
\end{equation}
This equation can be expressed as
\begin{equation}
\sum_{ij\in\EE} \varphi_{ij}\left(x_i, x_j\right) + \sum_i \varphi_i\left(x_i\right)
\end{equation}
where $\varphi_{ij}\left(x_i, x_j\right) \triangleq -\log\Psi(x_i,x_j)$ and $\varphi_i\left(x_i\right) \triangleq -\log\Phi(x_i,y_i)$.

Thus MAP (maximum a-posteriori) inference in graphical models such as MRF and conditional random fields (CRF) boils down to the optimization of a discrete pair-wise energy functional of the form Eq.~(\ref{eq:pair-wise-gen}) which is the focus of this thesis.

\hrulefill
\end{example}

Energy functions of the form Eq.~(\ref{eq:pair-wise-gen}) arise in many graphical models (MRFs and CRFs) (see e.g., \cite{Blake2011} and references therein).
However, it is not restricted to that domain and are also encountered in a variety of other domains such as structural learning (e.g., \cite{Nowozin2011structured}), and as the inference part of discriminative models (e.g., structural SVM \cite{Taskar2003,Tsochantaridis2006}).
This thesis explores some challenging instances of these energies and explores new methods for improved minimization approaches for these hard-to-optimize energies.

%---------------------------------------------------------------------------%
\section{Minimization of Discrete Pair-wise Energies}

The energy function of Eq.~(\ref{eq:pair-wise-gen}) presented in the previous section is used in many applications to evaluate how compatible is a certain discrete solution $\bx$ to a given problem.
It is now desired to find the {\em best} solution for the given problem by finding a solution $\bx^\star$ with the lowest energy.
That is, solving the optimization problem
\begin{equation}
\bx^\star = \arg\min_{\bx} \Fx
\label{eq:intro-optimization-prob}
\end{equation}
over all {\em discrete} solutions $\bx$.

In general, the optimization problem~(\ref{eq:intro-optimization-prob}) is NP-hard.
Known hard combinatorial problems such as max-cut, multi-way cut and many others may be formulated in the form of~(\ref{eq:intro-optimization-prob}) (see e.g., \cite{Boykov2001}).
Yet, there are special instances of problem~(\ref{eq:intro-optimization-prob}) which can be optimized exactly in polynomial time.
The main two factors that affect the difficulty of the optimization problem~(\ref{eq:intro-optimization-prob}) are:
\begin{enumerate}
\item The underlying graph structure, $\EE$.
\item Mathematical properties of the pair-wise interactions, $\varphi_{ij}$.
\end{enumerate}

When the underlying graph $\EE$ has no cycles (that is, $\EE$ has a tree structure) optimization of~(\ref{eq:intro-optimization-prob}) is fairly straight-forward:
By propagating information back and forth from the leafs to the root convergence to {\em global} minimum is attained.
This information propagation is often referred to as belief-propagation (BP) (\cite{Pearl1988,Koller2009}).
A cycle-free graph is crucial for this rapid polynomial time convergence of this message-passing scheme:
Every path from root to leaf is unique and therefore consensus along path is attained in a single forward backward  pass.
However, when $\EE$ has cycles, paths between the different variables are no longer unique and may give rise to contradicting messages.
These contradictions introduce an inherent difficulty to the optimization process making it NP-hard in general.

Despite the inherent difficulty of optimizing~(\ref{eq:intro-optimization-prob}) over a cyclic graph $\EE$,
there are instances of problem~(\ref{eq:intro-optimization-prob}) that can still be efficiently minimized when the pair-wise terms, $\phiij$, meet certain conditions.
The next few sections explore these conditions in more detail, providing pointers to existing optimization methods that succeed in exploiting special structures of $\phiij$ to suggest methods and guarantees on the optimization process.
For simplicity, we start in Sec.~\ref{sec:intro-binary} with the special case of discrete binary variables, that is $\bx\in\left\{0,1\right\}^n$.
Then we move on, in Sec.~\ref{sec:intro-multilabel}, to the multilabel case of discrete variables taking one of multiple possible states, $\bx\in\left\{1,\ldots,l\right\}^n$.

%---------------------------------------------------------------------------%

\section{Binary problems}
\label{sec:intro-binary}

Binary optimization problems are of the form (\ref{eq:intro-optimization-prob}) where the solution space is restricted to binary vectors only, i.e., $\bx\in\left\{0,1\right\}^n$.

The most basic property of binary functionals is {\bf submodularity}.
This property is defined as follows:

\begin{definition}[Binary submodular]
A pair-wise function $\Fx$ defined for binary vectors $\bx$ is {\bf submodular} iff $\forall i,j\in\EE$:
$\varphi_{ij}\left(0,0\right)+\varphi_{ij}\left(1,1\right)\le\varphi_{ij}\left(1,0\right)+\varphi_{ij}\left(0,1\right)$
\label{def:binary-submodular}
\end{definition}

\begin{figure}
\centering
\includegraphics[width=.65\linewidth]{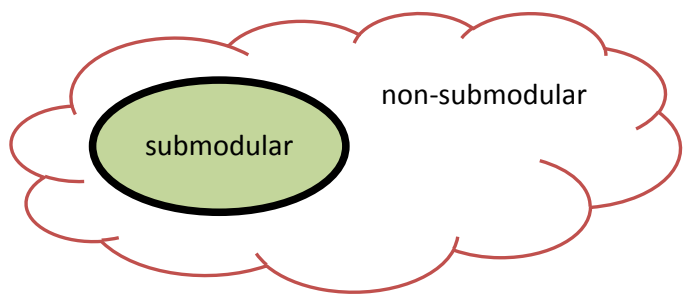}
\caption{
{\bf types of binary $\Fx$:}
{\em submodular vs. non-submodular. Green color indicates energies for which exact minimization can be done in polynomial time.}
}
\label{fig:types-binary}
\end{figure}

%Submodular is a fundamental type of pair-wise functions $\phiij$.
Close inspection of Def.~\ref{def:binary-submodular} reveals that a submodular function assigns lower energy to smooth configurations (i.e. to $x_i=x_j$) than to ``contrastive" configurations (i.e., to $x_i\ne x_j$).
That is the ``smooth" state $\varphi_{ij}\left(0,0\right)+\varphi_{ij}\left(1,1\right)$ has lower energy than the ``contrastive" state $\left(1,0\right)+\varphi_{ij}\left(0,1\right)$.
%This property of submodularity originates from set-functions and is related to convexity of some continuous relaxation of $\Fx$ (\cite{Lovasz1983}).
Fig.~\ref{fig:types-binary} provides an illustration of the space of all pair-wise energies.

\begin{figure}[t!]
\hspace*{-1cm}
\centering
\begin{tabular}{cccc}
\multirow{2}{*}{\includegraphics[width=.25\linewidth]{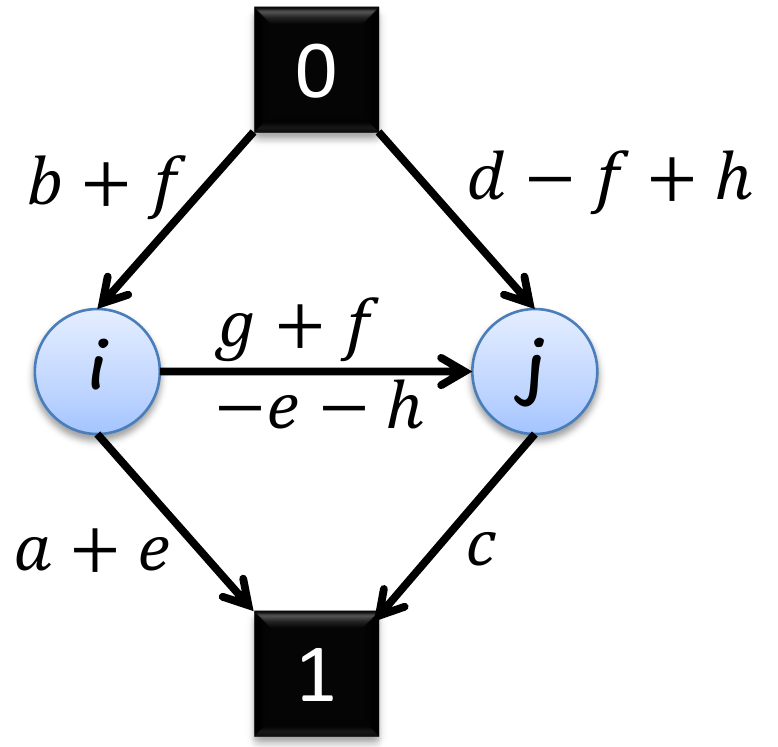}} &
\multicolumn{3}{c}{$\Fx=\varphi_i\left(x_i\right)+\varphi_j\left(x_j\right)+\phiijx$}  \\
& & & \\
&
\begin{tabular}{c|c|}
& $\varphi_i$ \\\hline
$x_i=0$ & $a$ \\\hline
$x_1=1$ & $b$
\end{tabular} &
\begin{tabular}{c|c|}
& $\varphi_j$ \\\hline
$x_j=0$ & $c$ \\\hline
$x_j=1$ & $d$
\end{tabular} &
\begin{tabular}{c|c|c}
$\phiij$ & $x_i=0$ & $x_i=1$\\\hline
$x_j=0$ & $e$ & $f$ \\\hline
$x_j=1$ & $g$ & $h$
\end{tabular} \\
& & & \\
& \multicolumn{3}{c}{\parbox{.73\linewidth}{
Binary energy over two variables $i$ and $j$ parameterized by 8 parameters $a-h$.
}} \\
\end{tabular}
\caption{
{\bf Graph construction:}
{\em
For a binary energy $\Fx$ a weighted graph is constructed in the following way:
Each variable $x_i$ has a corresponding node $i$. In addition, two special nodes (denoted in this figure as \fbox{$0$} and \fbox{$1$}) are added.
Two edges connect each $i^{th}$ node to the special nodes \fbox{$0$} and \fbox{$1$}.
In addition, an edge $(i,j)$ is added for each interacting pair of variables (for which $\phiij$ exists).
The weights of the edges are defined according to the parameters of the energy as illustrated in the figure.
Note that the weight of the $(i,j)$ edge, $g+f-e-h$, is exactly $\phiij(0,1)+\phiij(1,0)-\phiij(0,0)-\phiij(1,1)$, this weight is non-negative iff $\phiij$ is {\em submodular}.
}}
\label{fig:intro-binary-graph}
%\end{figure}

\

\

%\begin{figure}
\hspace*{-1.5cm}
\centering
\begin{tabular}{c|cccc}
\includegraphics[height=3cm]{intro/binary_graph_mapping} & \multicolumn{4}{c}{\includegraphics[height=3cm]{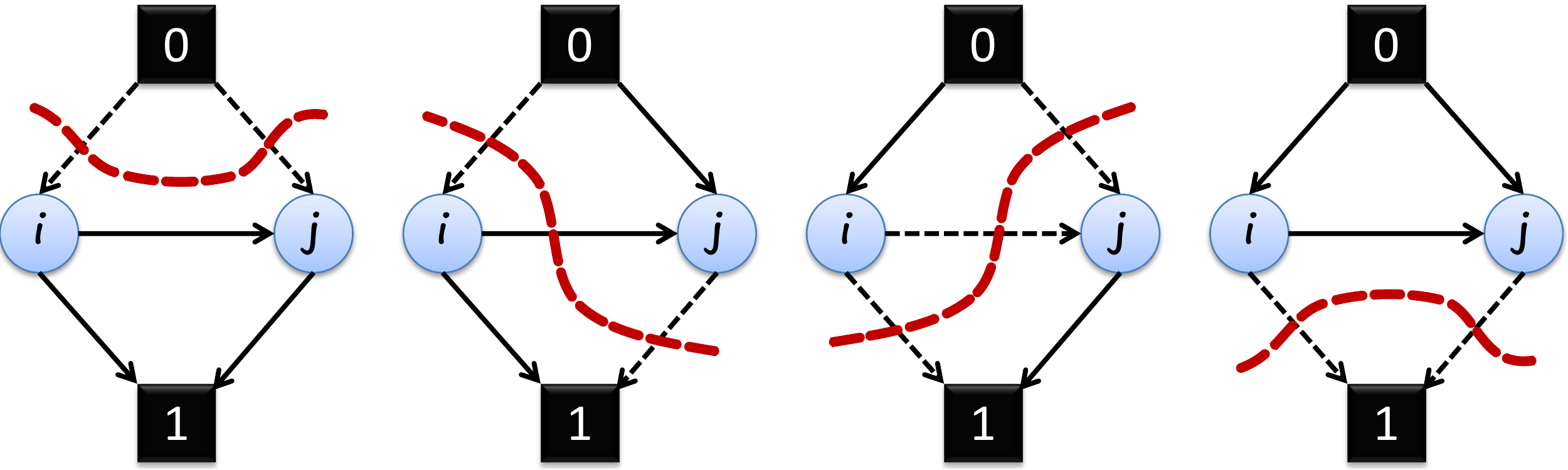}} \\ \hline \hline
cut          &  $\left\{0\right\}$, $\left\{1, i, j\right\}$ &  $\left\{0, j\right\}$, $\left\{1, i\right\}$ & $\left\{0, i\right\}$, $\left\{1, j\right\}$ & $\left\{0, i, j\right\}$, $\left\{1\right\}$ \\ \hline
assignment $\bx$  & $x_i=1$, $x_j=1$ & $x_i=1$, $x_j=0$ & $x_i=0$, $x_j=1$ & $x_i=0$, $x_j=0$ \\ \hline \hline
cut          & $(b+f)$    & $(b+f)$  & $(a+e)+(d-f+h)$  & $(a+e)$ \\
weight       & $+(d-f+h)$ & $+c$     & $+(g+f-e-h)$     & $+c$    \\ \hline
energy $\Fx$ & $b+d+h$    & $b+f+c$  & $a+g+d$          & $a+e+c$ \\
\end{tabular}
\caption{
{\bf Binary energies and graph-cuts:}
{\em
All four possible cuts of the graph constructed in Fig.~\ref{fig:intro-binary-graph}.
A cut (dashed red) separates the two special nodes, \fbox{$0$} from \fbox{$1$}.
Cut edges (edges pointing from the \fbox{$0$} side to the \fbox{$1$} side) are marked with dash lines.
First two rows show the $1:1$ mapping between a cut and an assignment to $\bx$.
Last two rows show the $1:1$ mapping between the weight of the cut (using the weights defined in Fig.~\ref{fig:intro-binary-graph}) and the energy $\Fx$ of the appropriate assignment.}
}
\label{fig:intro-binary-graph-cuts}
\end{figure}

Submodularity is an important property of $\phiij$ since exact optimization of binary submodular functions can be done in polynomial time regardless of the graph structure $\EE$ (\cite{Greig1989}).
One such optimization algorithm identifies a $1:1$ mapping between binary assignments $\bx$ and cuts on a specially constructed graph.
Careful choice of weights for the edges on this special graph gives rise to a $1:1$ mapping between the {\em weight} of a cut and $\Fx$ of the appropriate binary assignment $\bx$.
This construction and choice of weights is illustrated in Fig.~\ref{fig:intro-binary-graph}.
The appropriate correspondences between assignments and graph-cuts, and between cut weight and energy are illustrated in Fig.~\ref{fig:intro-binary-graph-cuts}.
Once this $1:1$ mapping is established, optimizing $\Fx$ is simply finding a minimum cut of the constructed graph.
This can be done in polynomial time {\em provided that all the weights of the edges are non-negative} (\cite{Cormen2001}).
Examining the details of this construction reveals that edge weights are non-negative iff the function $\Fx$ is submodular.
Details of this construction can be found in e.g., \cite{Greig1989,Boykov2001} and in more detail in \cite{KolmogorovZabih}.

However, when $\Fx$ is not submodular
(i.e., there exists at least one pair-wise term $\phiij$ for which the submodularity inequality~(\ref{def:binary-submodular}) does not hold)
the optimization of $\Fx$ becomes NP-hard (\cite{Rother2007}).
In that case exact minimum cannot be guaranteed in polynomial time and an approximation is sought.
One notable approach for approximating non-submodular binary optimization problems is by an extension of the min-cut approach.
This method, called QPBO (Quadratic Pseudo-Boolean Optimization) was proposed by \cite{Hammer1984} and later extended by \cite{Rother2007}.
The main idea behind this approximation scheme is that when the energy function $\Fx$ is non-submodular, the derived graph has edges with negative weights.
Therefore, they propose to construct a redundant graph in which each variable is represented by two nodes (rather than only one as in the original construction).
One node represents the case where the variable is assigned $0$ and the other node represents the case of assigning $1$ to the variable.
This redundant representation eliminates the need for negative edge weights and thus a min-cut of the new graph can be computed in polynomial time.
Looking at the resulting min-cut we can discern two cases for each variable:
The first, in which the cut separates the two complementary nodes representing this variable.
In this case, the cut clearly defines an optimal state for the variable (either $0$ or $1$).
However, there is a second case in which both complementary nodes fall in the same side of the cut.
In this case, we are unable to determine what is the proper assignment for this variable and the variable remains unlabeled.

Therefore, we can conclude that QPBO extends the min-cut approach to handle non-submodular binary energies.
Recovery of the global minimum is no longer guaranteed, but the algorithm may recover a {\em partial} labeling that is guaranteed to be part of some globally optimal solution.
However, in the extreme case it may happen that QPBO is unable to label any variable.

Table~\ref{tab:binary-opt-classification} summarizes the different types of pair-wise binary energy functions and the difficulty they entail on their optimization.

\begin{table}
\centering
\begin{tabular}{r|c|c}
\backslashbox{pair-wise}{structure} & Tree & Cyclic \\\hline
submodular & {\bf Easy:} mincut, BP & {\bf Easy:} mincut \\\hline
non-submodular & {\bf Easy:} QPBO, BP & {\bf Hard} \\
\end{tabular}
\caption{{\bf Hardness of binary optimization:}
{\em the computational ``hardness" of a discrete minimization as a function of the underlying graph structure ($\EE$), and the class of pair-wise interactions ($\phiij$).}}
\label{tab:binary-opt-classification}
\end{table}

%----------------------------------------------------------------------------------------------%
\section{Multilabel problems}
\label{sec:intro-multilabel}

A multilabel discrete problem is the optimization of a discrete function $\Fx$ of the form (\ref{eq:pair-wise-gen}) defined over a finite discrete vector $\bx\in\left\{1,\ldots,L\right\}^n$.
%Functions $\Fx$ over binary vectors are a special case and were presented separately in the previous section.
As was shown in the previous section, properties of the pair-wise terms $\varphi_{ij}$ of $\Fx$ have a crucial effect on the computational complexity of the optimization problem.
For the binary case the only distinctive property was the {\em submodularity} of $\Fx$.
However, when discrete variables over more than two states are considered, there are more subtle types of pair-wise interactions that affect the ability to optimize, or at least efficiently approximate it.
This section describes these various types of $\phiij$ and their effect on the discrete optimization task.

The following definition of multilabel submodularity is given in \cite{Schlesinger2006}.
\begin{definition}[Multilabel submodular]
Assume the labels ($\alpha,\beta,\ldots$) are fully ordered.
Then $\Fx$ is {\bf multilabel submodular} iff $\forall i,j\in\EE$ and $\forall \alpha\le\gamma,\beta\le\delta$
the following inequality holds
\begin{equation}
\varphi_{ij}\left(\alpha,\beta\right) + \varphi_{ij}\left(\gamma,\delta\right) \le
\varphi_{ij}\left(\alpha,\delta\right) + \varphi_{ij}\left(\gamma,\beta\right)
\label{eq:multilabel-submodular}
\end{equation}
\label{def:multilabel-submodular}
\end{definition}

A slightly simpler and equivalent condition for submodularity uses the following condition:
\begin{equation}
\varphi_{ij}\left(\alpha,\beta\right) + \varphi_{ij}\left(\alpha+1,\beta+1\right) \le
\varphi_{ij}\left(\alpha,\beta+1\right) + \varphi_{ij}\left(\alpha+1,\beta\right)
\label{eq:multilabel-submodular-simple}
\end{equation}
$\Fx$ is multilabel submodular iff~(\ref{eq:multilabel-submodular-simple}) holds for all labels $\alpha,\beta$ and for all $\varphi_{ij}$.

The notion of submodularity ~is ~strongly ~related to Monge matrices (\cite{Cechlarova1990}):
A matrix $V$ is a {\em Monge matrix} iff $V_{\alpha,\beta} + V_{\alpha+1,\beta+1} \le V_{\alpha+1,\beta} + V_{\alpha,\beta+1}$, $\forall\alpha,\beta$.
Monge matrices were defined by the French mathematician G. Monge (\citeyear{Monge1781}), and they play a major role in optimal transportation problems and other discrete optimization tasks (see, e.g., \cite{Burkard2007}).
Consider a matrix $V\in\mathbb{R}^{l\times l}$ whose entries are $V_{\alpha\beta} \deff \phiij\left(\alpha,\beta\right)$,
then $\phiij$ is multilabel submodular iff $V$ is a Monge matrix.

A sub class of multilabel submodular functions are functions where $\varphi_{ij}$ are convex on the set of labels.
Convexity on a discrete set is defined in \cite{Ishikawa2003}, as follows:
\begin{definition}[Convexity on a discrete set]
A real valued function $g\left(x\right)$ is {\bf convex} on a set $\mathcal{A}$ iff
\begin{equation}
g\left(t x + (1-t) y \right) \le t g\left(x\right) + (1-t) g\left(y\right)
\end{equation}
for all $x,y\in\mathcal{A}$ and $t\in\left[0,1\right]$ s.t. $t x + (1-t) y \in \mathcal{A}$
\label{def:multilabel-convex-ishikawa}
\end{definition}

When the set of labels is fully ordered and if $\forall i,j$ $\phiijx = g_{ij}\left(x_i - x_j\right)$ and all $g_{ij}$ are convex, then $\Fx$ is convex.
For example $\varphi_{ij}\left(x_i,x_j\right) = \left|x_i - x_j\right|^p$.
Note that a truncated $\ell_p$ norm is no longer convex.

\begin{claim}
Convex is a special case of submodular \cite{Schlesinger2006}.
\end{claim}
\begin{proof}
Let $\varphi\left(\alpha,\beta\right)=g\left(\alpha-\beta\right)$ for some convex $g\left(x\right)$.
Then for all $\alpha,\beta$:
\begin{eqnarray*}
g\left(\half\left(\alpha-\beta-1\right) + \half\left(\alpha-\beta+1\right)\right) & \le & \half g\left(\alpha-\beta-1\right) + \half g\left(\alpha-\beta+1\right) \\
g\left(\alpha - \beta\right) & \le & \half g\left(\alpha-\beta-1\right) + \half g\left(\alpha-\beta+1\right) \\
g\left(\alpha - \beta\right) + g\left(\alpha - \beta\right) & \le &  g\left(\alpha-\beta-1\right) + g\left(\alpha-\beta+1\right) \\
g\left(\alpha - \beta\right) + g\left(\alpha + 1 - \left(\beta + 1\right) \right) & \le &  g\left(\alpha-\left(\beta+1\right)\right) + g\left(\alpha+1 -\beta\right) \\
\varphi\left(\alpha,\beta\right) + \varphi\left(\alpha+1,\beta+1\right) & \le & \varphi\left(\alpha+1,\beta\right) + \varphi\left(\alpha,\beta+1\right) \\
\end{eqnarray*}
From property (\ref{eq:multilabel-submodular-simple}) it follows that a convex $\varphi$ is also submodular.
\end{proof}

To make these definitions more concrete, we can consider a few examples.
Popular pair-wise terms of the form $\phiijx=\left|x_i-x_j\right|$ (also known as $\ell_1$), and the $\ell_2$: $\phiijx=\left(x_i-x_j\right)^2$ are both {\em convex} and therefore {\em multilabel submodular}.
However, the robust (or truncated) version of these $\ell_p$ terms: $\phiijx=\min\left\{\left|x_i-x_j\right|^p,\tau\right\}$, is no longer multilabel submodular.

An important result regarding the minimization of multilabel submodular functions is presented in \cite{Schlesinger2006}. A reduction is made from submodular minimization to st-mincut on a specially constructed graph.
It is shown that when the original energy is multilabel submodular all weights in the resulting graph are non-negative and hence a {\em global} minimum can be found in polynomial time.
This construction generalizes the construction of \cite{Ishikawa2003} that is specific to convex pair-wise functions.

However, submodularity of $\Fx$ is a very restrictive property.
The well known Potts term, and many other pair-wise interactions are not submodular.
Still, there are other important properties for non-submodular functions $\Fx$.
\cite{Boykov2001} derived important approximations that rely on other properties of $\Fx$.
These properties were further relaxed by \cite{KolmogorovZabih}:
\begin{definition}[Relaxed metric]
A function $\Fx$ is a {\bf relaxed metric} iff $\forall i,j\in\EE$ and $\forall \alpha,\beta,\gamma$
\begin{equation}
\phiij\left(\alpha,\alpha\right) + \phiij\left(\beta,\gamma\right) \le \phiij\left(\beta,\alpha\right) + \phiij\left(\alpha,\gamma\right)
\end{equation}
\label{def:large-move-metric}
\end{definition}

The condition of Def.~\ref{def:large-move-metric} resembles the triangle inequality of metric functions in the case $\phiij(\alpha,\alpha) = 0$.
The Potts model and the robust (truncated) $\ell_1$ interaction, mentioned earlier in this section are examples of relaxed-metric pair-wise interaction.
Note that this property of relaxed metric is different than the convexity of Def.~\ref{def:multilabel-convex-ishikawa}.
Another property, less restrictive than relaxed metric is:
\begin{definition}[Relaxed semi-metric]
A function $\Fx$ is a {\bf relaxed semi-metric} iff $\forall i,j\in\EE$ and $\forall \alpha,\beta$
\begin{equation}
\phiij\left(\alpha,\alpha\right) + \phiij\left(\beta,\beta\right) \le \phiij\left(\beta,\alpha\right) + \phiij\left(\alpha,\beta\right)
\end{equation}
\label{def:large-move-semi-metric}
\end{definition}

Examples of relaxed semi-metric functions: $\ell_2$, truncated $\ell_2$.
Clearly, any relaxed metric function is also a relaxed semi-metric.
At this point it may be useful to get some intuition about the meaning of the semi-metric property:
According to Def.~\ref{def:large-move-semi-metric} a function $\Fx$ is semi-metric if the cost of assigning neighboring variables $i$ and $j$ to the {\em same} label (either $\alpha$ or $\beta$) is never greater than the cost of assigning them to different labels.
This property implies that $\Fx$ encourages smoothness of the solution $\bx$.

\begin{figure}
\centering
\includegraphics[width=.9\textwidth]{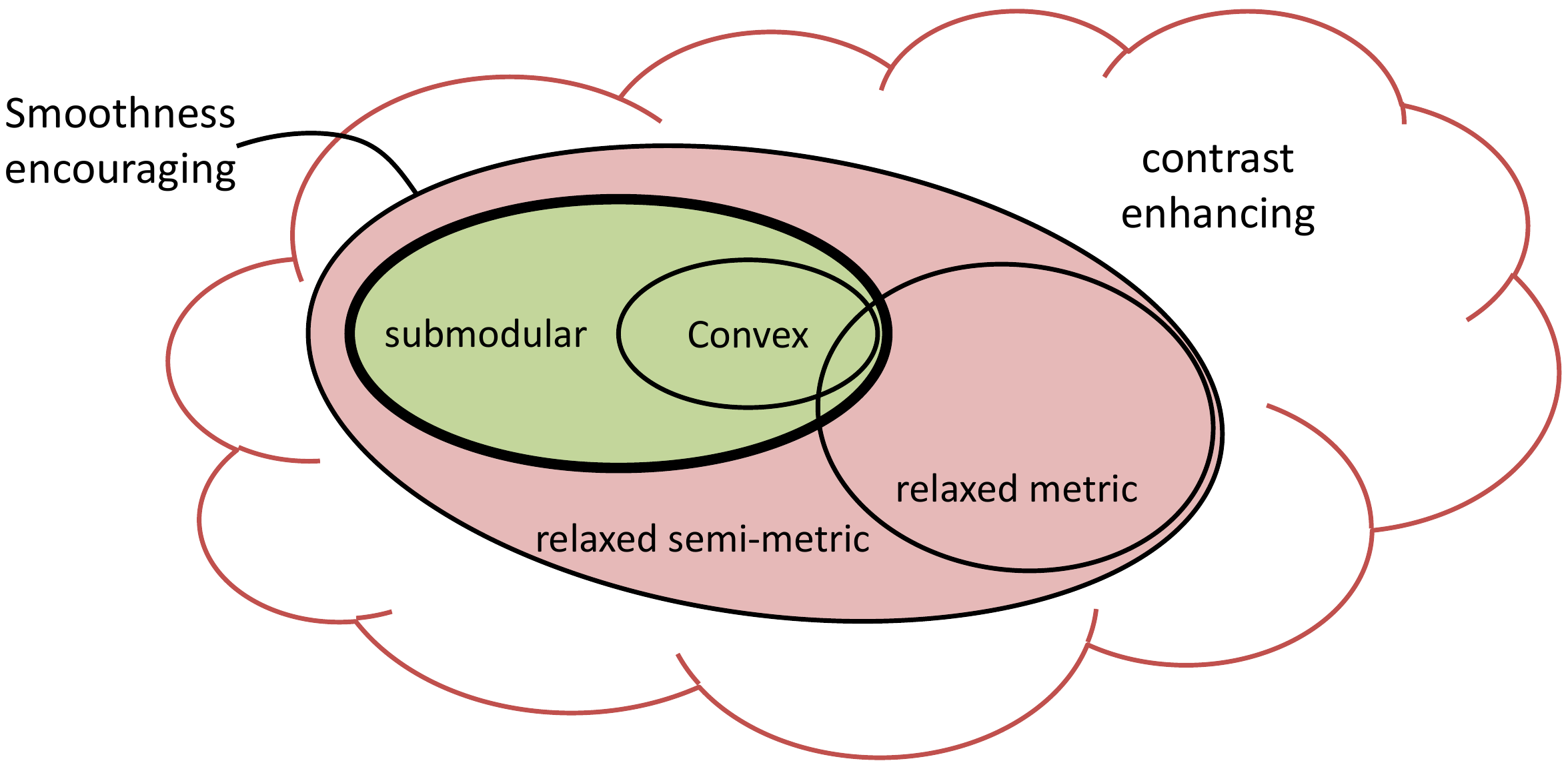}
\caption{
{\bf Different types of multilabel $\Fx$:}
{\em The hierarchy and relations between different types of multilabel energies.
Green indicates the existence of global minimization algorithms.
For energies in red there are good approximation algorithms.}
}
\label{fig:large-move-multilabel-types}
\end{figure}

Figure~\ref{fig:large-move-multilabel-types} shows the relation between the different types of functions $\phiij$.
The most restrictive type is the convex (Def.~\ref{def:multilabel-convex-ishikawa}) which is a subset of submodular (Def.~\ref{def:multilabel-submodular}).
Regarding relaxed metric and submodular: there are submodular functions that are not relaxed metric (e.g., $\ell_2$), and there are relaxed metric that are not submodular (e.g., Potts).
Table~\ref{tab:multilabel-examples-phi} shows examples of popular pair-wise functions and their properties.

\begin{claim}
In general, multilabel submodular (Def.~\ref{def:multilabel-submodular}) is not metric (Def.~\ref{def:large-move-metric}).
\end{claim}
\begin{proof}
Let $\phiij$ be multilabel submodular. Consider three labels $\alpha\le\gamma\le\delta$. Choose $\beta=\gamma$.
We now have $\alpha\le\gamma$ and $\beta\le\delta$.
Submodularity of $\phiij$ (Def.~\ref{def:multilabel-submodular}) yields:
\begin{eqnarray*}
\phiij\left(\alpha,\beta\right)+\phiij\left(\gamma,\delta\right)& \le & \phiij\left(\alpha,\delta\right)+\phiij\left(\gamma,\beta\right) \\
\phiij\left(\alpha,\gamma\right)+\phiij\left(\gamma,\delta\right)& \le & \phiij\left(\alpha,\delta\right)+\phiij\left(\gamma,\gamma\right) \\
\end{eqnarray*}
This inequality is the opposite of the inequality of Def.~(\ref{def:large-move-metric}) (semi-metric).
In general, Equality does not hold and therefore most submodular $\phiij$ are not metric.
However, for $\ell_1$, i.e., $\phiijx=\left|x_i-x_j\right|$ equality holds and thus $\ell_1$ is a special case of $\phiij$ that is both submodular and metric.
\end{proof}

\begin{table}
\centering
\begin{tabular}{cccc}
{\bf convex} & {\bf submodular} & {\bf relaxed metric} & {\bf relaxed semi-metric} \\\hline
$\ell_1$     & $\ell_1$         & truncated $\ell_1$   & $\ell_2$                  \\
$\ell_2$     & $\ell_2$         & $\ell_1$             & truncated $\ell_2$        \\
             &                  & Potts                &                           \\
\end{tabular}
\caption{
{\bf examples of different types of pair-wise functions $\phiij$}}
\label{tab:multilabel-examples-phi}
\end{table}

Unfortunately, the promising results of \cite{Schlesinger2006} regarding the {\em global} minimization of submodular functions does not hold for more general functions.
When $\Fx$ is no longer submodular one can no longer hope to achieve global optimality in polynomial time.
However, for relaxed metric and relaxed semi-metric functions \cite{Boykov2001} showed large move making approximate algorithms that performs quite well in practice (see e.g., \cite{Szeliski2008}).
Large move making algorithms iteratively seek to improve the energy of a current solution by updating large number of variables at once.
Each such large step is carried out by solving a simple binary submodular minimization via st-mincut.
For relaxed metric functions the large move is $\alpha$-expansion.
At each iteration a binary problem is solved: for each variable it can either retain its current label (0) or switch label to $\alpha$ (1).
The relaxed metric property of $\Fx$ ensures that the resulting binary problem is submodular and thus can be solved globally in polynomial time.
The $\alpha$-expansion algorithm iterates over all labels until it converges.
Convergence after finite number of iterations is guaranteed, and in certain cases some theoretical bounds can be proven on the quality of the approximation (see \cite{Boykov2001} for more details).

For relaxed semi-metric functions a slightly different large move is devised.
For each pair of labels $\alpha$ and $\beta$ the large move is called $\alpha\beta$-swap.
At each iteration a binary problem is solved for all variables currently labeled either $\alpha$ or $\beta$:
for each variable it can pick either $\alpha$ (0) or $\beta$ (1).
The relaxed semi-metric property of $\Fx$ ensures that the resulting binary problem is submodular and thus can be solved globally in polynomial time.
The $\alpha\beta$-swap algorithm iterates over all {\em pairs} of labels until it converges.
Convergence after finite number of iterations is guaranteed, however, theoretical bounds on the approximation no longer exists (see \cite{Boykov2001} for more details).
Table~\ref{tab:ml-opt-classification} shows the resulting types of pair-wise energies and the current results on their minimization.
This thesis focuses on the hard optimization of the contrast-enhancing functionals defined on cyclic graphs.
Part~\ref{part:app} shows how introducing energies that contain contrast-enhancing terms gives rise to new applications.
While Part~\ref{part:approx} deals with the methods of approximating these challenging contrast-enhancing energies.

\begin{table}
\centering
\begin{tabular}{r|c|c}
\backslashbox{pair-wise}{structure} & Tree & Cyclic \\\hline
submodular & {\bf Easy:} mincut, BP & {\bf Easy:} mincut \\\hline
semi-metric & {\bf Easy:} BP        & Good Approximations \\\hline
contrast-enhancing  & {\bf Easy:} BP & {\bf Hard} \\
\end{tabular}
\caption{{\bf Hardness of optimization (multilabel):}
{\em the computational ``hardness" of discrete optimization as a function of the underlying graph structure and the class of pair-wise interactions.}}
\label{tab:ml-opt-classification}
\end{table}

%----------------------------------------------------------------------------------------------%
\section{Relation to Linear Programming (LP)}
\label{sec:intro-lp}

This section establishes a connection between the pair-wise energy minimization problem~(\ref{eq:intro-optimization-prob}) and the field of convex optimization, in particular to Linear Programming (LP).
For the following discussion it is useful to introduce some new notations and definitions.
The first useful representation is the {\em overcomplete representation} of the solution vector $\bx$.
This representation is defined as follows \cite{Wainwright2005,wainwright2008graphicalmodels}:

\begin{definition}[Overcomplete representation]
A discrete solution $\bx$ can be represented by an extended binary vector $\phix$, s.t.
\begin{eqnarray}
\phix_{i,\alpha} & = & \delta\left(x_i==\alpha\right) \\
\phix_{ij,\alpha\beta} & = & \delta\left(x_i==\alpha\right) \cdot \delta\left(x_j==\beta\right)
\end{eqnarray}
where $\delta(\cdot)$ is the Kronecker delta function.
\label{def:local-update-overcomplete-representation}
\end{definition}

The overcomplete representation projects a discrete vector $\bx$ of dimension $n$ into a $d$-dimensional binary vector $\phix$.
The index set of vector $\phix$ is defined as $\mathcal{I}=\left\{i,\alpha\right\}\cup\left\{ij,\alpha\beta\right\}$, with $d=\left|\mathcal{I}\right|$.

With the overcomplete representation in mind, it is useful to parameterize the discrete function $\Fx$ of Eq.~(\ref{eq:pair-wise-gen}) using a parameter vector $\theta$:
\begin{subequations}
\begin{equation}
\theta_{i,\alpha}    =  \varphi_i\left(\alpha\right)
\end{equation}
\begin{equation}
\theta_{ij,\alpha\beta}  =  \phiij\left(\alpha,\beta\right)
\end{equation}
\label{eq:local-overcomplete-theta}
\end{subequations}

Combining Def.~\ref{def:local-update-overcomplete-representation} with the parametrization of~(\ref{eq:local-overcomplete-theta}), the functional of Eq.~(\ref{eq:pair-wise-gen}) becomes
\begin{eqnarray}
\Fx & = & \sum_{i,\alpha} \theta_{i,\alpha} \phix_{i,\alpha} + \sum_{ij,\alpha\beta} \theta_{ij,\alpha\beta}\phix_{ij,\alpha\beta} \nonumber \\
    & = & \dotprod{\theta}{\phix}
\label{eq:local-overcomplete-energy}
\end{eqnarray}

The overcomplete representation $\phix$ is defined over a discrete set of points.
However, it is useful to consider a relaxation of this set into a convex continuous domains.

The tightest relaxation of the discrete set $\left\{\phix \vert \bx\in\left\{1,\ldots,l\right\}^n\right\}$ is the marginal polytop $\marg{\EE}$:
\begin{definition}[Marginal polytop]
The marginal polytop of $\Fx$ is the convex combination of the vertices $\phix$ for the $l^n$ discrete solutions $\bx$. This set is formally defined as:
\begin{equation}
\marg{\EE}  =  \left\{\mu\in\mathbb{R}^d \left| \mu=\sum_{\bx}p\left(\bx\right)\phix,\;\mbox{for some distribution $p\left(\cdot\right)$}\right.\right\}
\end{equation}
\label{def:local-marginal-polytop}
\end{definition}
This marginal polytop, $\marg{\EE}$, is defined by finite, yet exponentially large, number of half-spaces.
Therefore, it is convenient to define a relaxed version of the marginal polytop:
\begin{definition}[Local polytop]
The local polytop of $\Fx$ is the convex set:
\begin{equation}
\local{\EE} = \left\{\tau\in\mathbb{R}^d\left|
\begin{array}{rl}
\sum_\alpha \tau_{i,\alpha} = 1 & \forall i \\
\sum_{\alpha\beta} \tau_{ij,\alpha\beta} = 1 & \forall ij\in\EE \\
\sum_{\beta} \tau_{ij,\alpha\beta} = \tau_{i,\alpha} & \forall ij\in\EE,\beta
\end{array}
\right.\right\}
\end{equation}
\label{def:local-local-polytop}
\end{definition}
Unlike the marginal polytop, $\local{\EE}$ is defined using only polynomial number of half-spaces,
and therefore it admits polynomial time optimization schemes.
In fact, $\local{\EE}$ is the first order approximation of $\marg{\EE}$ (\cite{wainwright2008graphicalmodels}).

Note that the geometry of $\marg{\EE}$ and $\local{\EE}$ are affected by the number of variables $n$, the number of states $l$ and by the underlying graph structure $\EE$ defining the interacting pairs of variables.
These polytops are {\em not} affected by the parameters of $\varphi_i$ and $\phiij$.

\begin{figure}
\centering
\begin{tabular}{ccc}
\includegraphics[width=.3\linewidth]{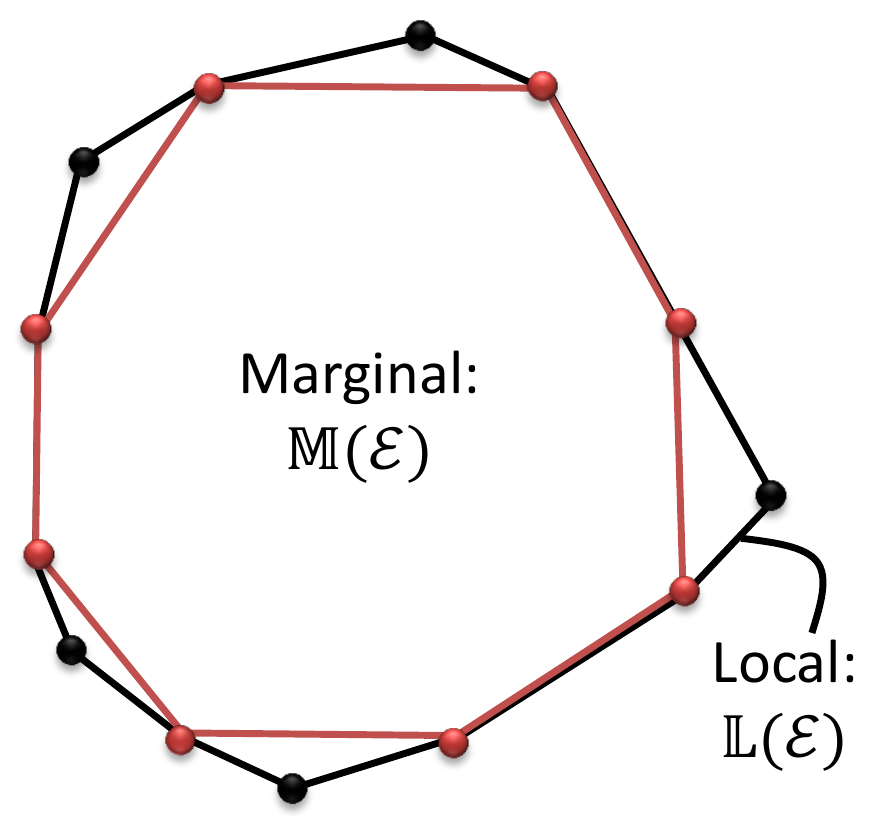} &
\includegraphics[width=.3\linewidth]{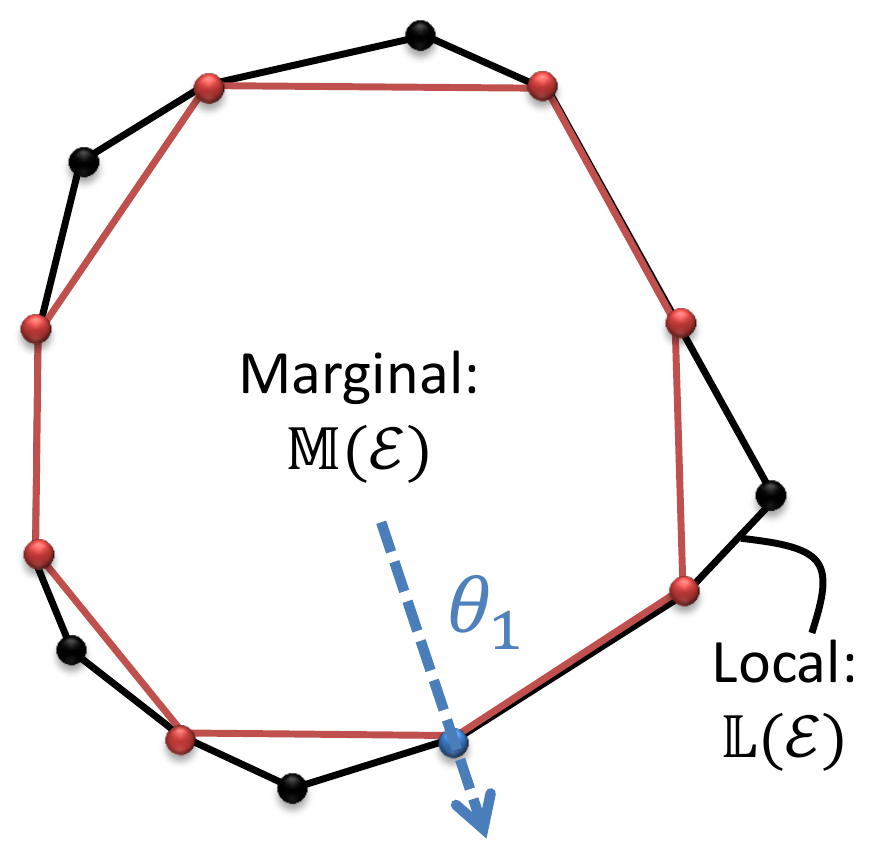} &
\includegraphics[width=.3\linewidth]{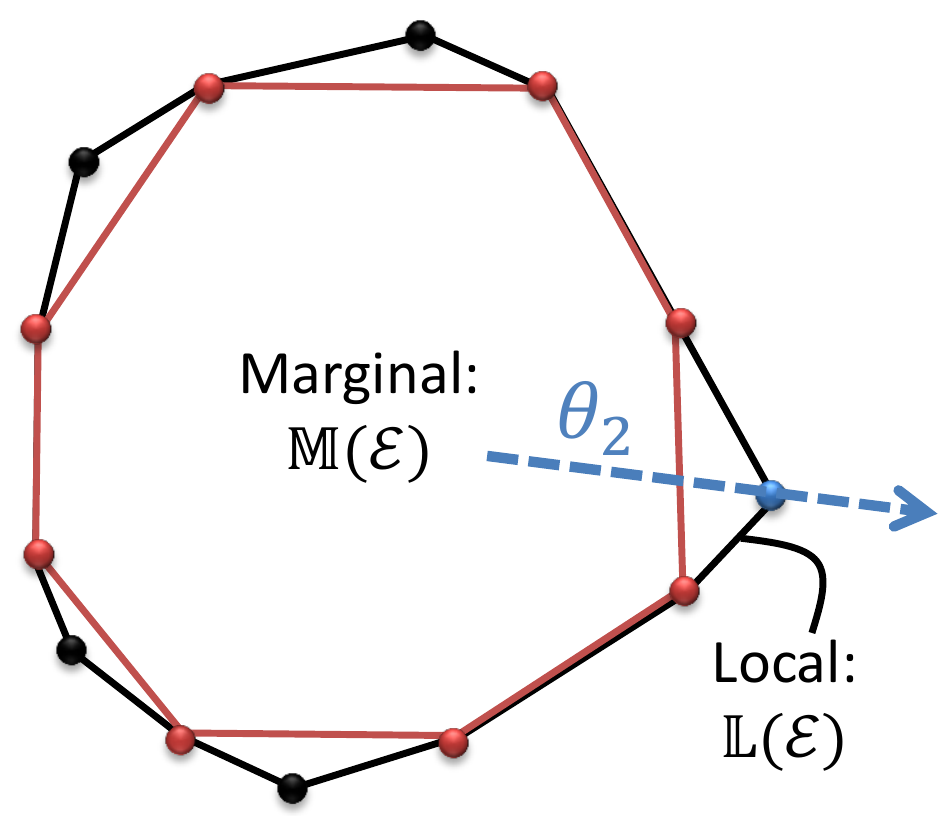} \\
(a) & (b) & (c)
\end{tabular}
\caption{
{\bf The Marginal and Local polytops:}
{\em An illustration of the marginal, $\marg{\EE}$, and local, $\local{\EE}$, polytops.
(a)~$\local{\EE}$ is an outer bound on the exact constraints of $\marg{\EE}$.  The vertex set of $\local{\EE}$ includes all integral vertices of $\marg{\EE}$ (marked in red) . It also includes fractional vertices (marked in black), which are {\em not} vertices of $\marg{\EE}$. (b),~(c)~Solving an LP with cost vector $\theta$ entails translating a hyperplane with normal $\theta$ until it is tangent to the constraint set. In~(b)~the point of tangency of cost vector $\theta_1$ occurs at an integral vertex. In~(c)~the point of tangency of cost vector $\theta_2$ occurs at a fractional vertex, outside $\marg{\EE}$. In this case, there is an integrality gap.
(figure taken from \protect\cite{Wainwright2005}).}
}
\label{fig:intro-marginal-local}
\end{figure}

By standard properties of LP, the optimal value is attained at an extreme point of the constraint set (a vertex of the constraints polytop).
The marginal polytop, $\marg{\EE}$, is a convex set defined by all the possible solutions $\phix$.
Hence, a vertex of $\marg{\EE}$ corresponds to a vector $\phix$ for some {\em discrete} solution $\bx$.
We refer to these vertices as {\em integral} vertices corresponding to integral solutions of the LP.
On the other hand, the local polytop, $\local{\EE}$, may contain more vertices that do not correspond to any discrete solution, $\bx$.
We refer to these vertices as {\em fractional} solutions.
Fig.~\ref{fig:intro-marginal-local} provides an illustration of the marginal and local polytops, the relation between them, and their impact on the optimal solution of LP.
The figure also distinguishes between the integral and fractional vertices of the polytops.
\cite[Example~3]{Wainwright2005} describes in detail a case where fractional solution is optimal.
It is important to note that the parameter vector $\theta$ for which the fractional solution is optimal, in their example, is such that encourages contrast between variables (i.e., $x_i\ne x_j$ for neighboring $i$ and $j$).

The relation between discrete energy minimization (problem~(\ref{eq:intro-optimization-prob})) and convex LP presented in this section lies in the foundation of popular optimization algorithms such as tree-reweighted BP (see Sec.~\ref{sec:intro-trw}).
This relation is also important to illustrate the challenging task of optimizing~(\ref{eq:intro-optimization-prob}):
When $\theta$ represents an energy function that is ``smoothness-encouraging" its optimal value usually corresponds to an integral vertex of $\local{\EE}$ and thus its optimization can be done {\em exactly} via LP over the relaxed constraint set $\local{\EE}$.
However, when the energy has contrast-enhancing terms its optimal solution w.r.t $\local{\EE}$ is {\em fractional} -- the global {\em integral} solution cannot be attained using the relaxed constraints of $\local{\EE}$.
Therefore, the optimization of energies that have contrast-enhancing terms, is a very challenging task.
This thesis focuses on these energies and proposes methods to cope with this inherent difficulty.

\section{Discrete Optimization Algorithms}
In the previous sections we outlined some key properties of the energy function $\Fx$ and their effect on
the minimization process.
We also demonstrated how specific optimization algorithms take advantage of these properties.
In this section we present several prominent optimization algorithms that we refer to later on in this thesis.
These selected representative approaches sketches the main directions at which current discrete optimization research is mainly focused.

%----------------------------------------------------------------------------------------------%
\subsection{Iterated Conditional Modes (ICM)}
\label{sec:local-icm}

ICM is a very simple and basic iterative optimization algorithm proposed by \cite{Besag1986}.
It is an approximate method, acting {\em locally} on the variables, suitable for multilabel functions with arbitrary underlying graph and arbitrary $\phiij$.
At each iteration ICM visits all the variables sequentially, and choose for each variable the best state (with the lower energy) given the current states of all other variables.
This process can be viewed as a greedy coordinate descend algorithm and it bears some analogy to Gauss-Seidel relaxations of the continuous domain (\cite{Varga1962}).

ICM is a local update process and therefore is prone to getting stuck very fast in local minimum.
It is also extremely sensitive to initialization (see e.g., \cite{Szeliski2008}).

When taking a probabilistic point of view, and considering the energy function as a Gibbs energy, that is representing some measure over all possible solutions:
\begin{equation}
Pr\left(\bx\right) \propto \exp \left( - \frac{1}{T} \Fx \right)
\label{eq:intro-gibbs-energy}
\end{equation}
ICM may be viewed as a Gibbs sampler at the temperature limit $T\rightarrow0$.
Therefore, its performance is expected to be inferior to more sophisticated sampling methods such as, e.g., simulated annealing (\cite{Kirkpatrick1983}).

%----------------------------------------------------------------------------------------------%
\subsection{Belief Propagation (BP)}
\label{sec:intro-bp}

Belief-propagation is an optimization algorithm based on local updates.
However, in contrast to the hard assignment ICM performs at each update,
BP maintains ``soft" beliefs for each variable and passes messages between neighboring variables according to their current belief.
A message from variable $i$ to its neighbor $j$, $m_{i\rightarrow j}$, is a vector of length $L$.
That is, the message vector encodes how $i$ ``feels" about assigning state $\alpha$ to $j$.
\begin{equation}
m_{i\rightarrow j}\left(\alpha\right) = \min_{\beta} \left\{\varphi_i\left(\beta\right) + \sum_{ki\in\EE,k\ne j}m_{k\rightarrow i}\left(\beta\right) + \phiij\left(\beta,\alpha\right) \right\}
\label{eq:intro-bp-message}
\end{equation}

The belief of each variable $i$ is also a vector of length $L$ encoding the tendency of $i$ to be assigned to state $\alpha$:
\begin{equation}
b_i\left(\alpha\right) = \varphi_i\left(\alpha\right) + \sum_{ki\in\EE}m_{k\rightarrow i}\left(\alpha\right)
\label{eq:intro-bp-belief}
\end{equation}

BP iteratively passes messages and updates the local belief for each variable.
After its final iteration, each variable is assigned the label with the lowest energy, i.e., $x_i = \arg\min_\alpha b_i(\alpha)$.

Originally, BP was used as an inference algorithm in tree-structured graphical models (\cite{Pearl1988,Koller2009}).
Messages were initialized to zero. Then messages were passed from leafs to root and back to the leafs.
This forward-backward message passing converges to the global optimum when $\EE$ is a tree, regardless of the type of $\phiij$ that can be arbitrary.

When the underlying graph $\EE$ has cycles, BP is no longer guaranteed to converge.
It was proposed to run BP on cyclic graphs, a variant called loopy-BP.
In the loopy case, however, it is not clear how to schedule the messages and how to determine the number of iterations to perform.
Even if the loopy BP converges to some fixed point, it is usually a local optimum with no guarantees on global optimality (\cite{Koller2009,wainwright2008graphicalmodels})

%----------------------------------------------------------------------------------------------%
\subsection{Tree-reweighted Belief Propagation (TRW)}
\label{sec:intro-trw}

A significant development of BP was presented in the works of \cite{Wainwright2005,Kolmogorov2006,Werner2007,wainwright2008graphicalmodels,Komodakis2011}.
These works proposed a new interpretation to the basic message passing operation that BP conducts.
They relaxed the discrete minimization of $\Fx$ to form a continuous linear programming (LP), in the same manner that was presented in Sec.~\ref{sec:intro-lp}.
Then they related message passing to the optimization of the resulting LP.
a It was shown that the relaxation of $\Fx$ forms an LP with very specific structure.
This special structure can, in turn, be exploited to devise a specially tailored optimization scheme that uses message passing as a basic operation.

The tree-reweighted BP approach establishes a relation between discrete optimization and continuous convex optimization of LP.
This relation brings forward interesting results and properties from the continuous optimization domain to the discrete one.
For instance, it allows to use Lagrangian multipliers and formulate a Lagrangian dual to the original problem.
The dual representation provides a lower bound to the sought optimal solution.
If a solution $\bx^\star$ is found with an energy $F\left(\bx^\star\right)$ equals to the lower bound, then a certificate is provided that this $\bx^\star$ is a {\em global} minimum.

It was shown (e.g., \cite{Szeliski2008}) that in practice in many computer vision application TRW was able to recover globally optimal solutions.
These results dealt mainly with relaxed metric energies (see Sec.~\ref{sec:intro-multilabel}).
However, there is no general guarantee on TRW and there are cases involving challenging energies, beyond relaxed metric, for which it was shown that an integrality gap exists and TRW can no longer provide a tight approximation (e.g., \cite{Kolmogorov2006,Bagon2012}).

\subsection{Large Move Methods}

As opposed to local methods such as ICM, BP and TRW,
there is the approach of \cite{Boykov2001} that proposes discrete methods based on combinatorial principles.
The basic observation that lies at the heart of the large-move algorithms is that instead of treating the variables locally one at a time, one may affect the labeling of many variables at once by performing large moves.
These large moves are formulated as a binary step, and the difference between the different ``flavors" of the large-move algorithms is the formulation of these binary steps.
What makes these large move effective and efficient is the fact that binary submodular sub-problems can be solved globally and efficiently.

The two basic large move algorithms, $\alpha$-expand and $\alpha\beta$-swap, were already described in Sec.~\ref{sec:intro-multilabel} in the context of relaxed metric and relaxed semi metric energies.
Recently, another large move making algorithm called fusion-moves was proposed (\cite{Lempitsky2007,Lempitsky2010})
At each iteration of the fusion algorithm a discrete solution is proposed.
The proposed solution is fused into the current solution via a binary optimization: each variable can retain its current label ($0$), or switch to the respective label from the proposed solution ($1$).
However, unlike the swap and expand algorithms, the resulting binary optimization of the fusion step is no longer guaranteed to be submodular and highly depends on the types of proposed solutions.
Therefore, it is often the case that QPBO (\cite{Kolmogorov2007}), which is a non-submodular binary approximation algorithm, is used to perform the binary steps of the fusion algorithm.

\

\

To summarize this brief outline of existing approximation algorithms
one may notice that a lot of effort is put in recent years in developing and improving approximate optimization algorithms.
Research is put into both providing better practical results and into exploring the theoretic aspects of the problem.
Approximation methods are derived and inspired by both the continuous optimization domain (e.g., TRW) and the discrete domain (e.g., graph-cuts).
However, these results mainly focus on the minimization of functions $\Fx$ that have some structure to them: either relaxed metric or relaxed semi-metric (see, for example, the survey of \cite{Szeliski2008}).
For these smoothness-encouraging functions current algorithms succeed in providing good approximations in practice, despite their theoretical NP-hardness.
In contrast, when it comes to arbitrary, {\em contrast-enhancing} functions, little is known in terms of approximation and no method currently exists (to the best of our knowledge) that provides satisfying approximations.
This thesis focuses on the optimization of arbitrary, contrast-enhancing functions.

%---------------------------------------------------------------------------%
\chapter{Outline of this Thesis}

Discrete energy minimization is a ubiquitous task in computer vision and in other scientific domains.
However, in the previous chapter we saw that the optimization of such discrete energies is known to be NP-hard in most cases (\cite{Boykov2001}).
Despite this theoretical hardness, for many ``smoothness encouraging" energies (relaxed semi-metric), approximate optimization algorithms exist and provide remarkable approximations in practice.

However, as tasks become more sophisticated, the models grow more complex:
From tree structured to cyclic graphs and grids,
and from simple smoothing priors to complex arbitrary pair-wise interactions.
As the energies become less constrained and structured one gains more expressive power for the objective function
at the cost of a significantly more challenging optimization task.

In this work I would like to step outside this ``comfort-zone" of the smoothness-encouraging energies and explore more challenging discrete energies.
This step gives rise to two important questions:
\begin{enumerate}
\item Why bother? Why should one consider energy functions beyond semi-metric? What can be gained (in term of expressive power) considering the significant hardness of the entailed optimization task?

\item In case we decide to embark on this challenging task of approximating arbitrary discrete energy, how can we tackle this problem? Can we propose new approaches and directions for the difficult approximation tasks of discrete energies, beyond semi-metric?

\end{enumerate}

These two research questions provide the road map of this thesis.
Consequently, this thesis revolves around two major axes:
applications and approximations.

\section*{Applications}
The first axis of this thesis involves exploring new applications that require arbitrary, contrast-enhancing energies beyond semi-metrics.
These examples demonstrate how the additional expressive power of arbitrary energies is crucial to derive new applications.
We show how utilizing arbitrary energies gives rise to interesting and desirable behaviors for different applications.
%For instance, considering contrast-enhancing objective function for the task of unsupervised clustering may introduce a solution not only to the clustering problem, but also may help in determining the underlying number of clusters.
%This clustering objective function is commonly known as ``Correlation Clustering" (\cite{Bansal2004}).
We present these new applications in Part~\ref{part:app} of this thesis.

Chapter~\ref{cp:negaff-sketch} shows an image sketching application that provides a binary sketch from a small collection of images of similar objects.
In this application the binary sketch is described via the interactions between neighboring pixels in the corresponding images.
Neighboring sketch bits corresponding to similar image pixels are encourage to have similar value (i.e., submodular, smoothness-enhancing term).
In contrast, neighboring sketch bits corresponding to {\em dis}similar image pixels are encourage to have {\em different} value (i.e., non-submodular, contrast-enhancing term).
The binary sketch is then the output of the resulting non-submodular energy minimization.

The sketching application may be thought of as a special case of binary image segmentation.
Considering contrast-enhancing objective function for the task of unsupervised segmentation or clustering may introduce a solution not only to the clustering problem, but also may help in determining the underlying number of clusters.
This clustering objective function is commonly known as ``Correlation Clustering" (\cite{Bansal2004}).
Chapter~\ref{cp:negaff-cc} explores the Correlation Clustering functional and its underlying ``model-selection" capability.

Image segmentation and clustering are not the only examples for contrast-enhancing energies.
Chapter~\ref{cp:lighting} describes a 3D surface reconstruction from multiple images under different known lighting.
The reconstruction takes into account the changes in appearance of the surface due to the change in lighting directions.
These changes amounts to an implicit partial differential equation (PDE) that describes the unknown surface.
In this work we propose to pose the solution of the resulting PDE as a discrete optimization task.
Incorporating integrability prior on the unknown surface, the resulting discrete energy has contrast-enhancing terms.

Modeling and prior knowledge are not the only sources for contrast-enhancing terms in energies.
In many cases, the exact parameters of an energy are learned from training data.
Chapter~\ref{cp:dtf} presents Decision Tree Fields (DTF): an example of such a learning scheme.
DTF learns, in a principled manner, a pair-wise energy from labeled training examples.
Since the learning process is not constraint to smoothness-encouraging energies,
it is often the case that the resulting energy has contrast-enhancing terms.
In its training phase DTF seeks parameters that (approximately) maximize the likelihood of the data.
Therefore, the resulting contrast-enhancing terms are better suited to describe the underlying ``behavior" of the data.
Restricting the model to smoothness-encouraging terms only would prohibit DTF from accurately predicting results at test time.

\section*{Approximate Optimization}
The enhancement in descriptive power gained by considering arbitrary energies comes with a price tag:
we no longer have good approximation algorithms at hand.
Therefore, the second axis of this thesis explores possible directions for approximating the resulting challenging arbitrary energies.
In part~\ref{part:approx} of this work I propose practical methods and approaches to approximate the resulting NP-hard optimization problems.
In particular, in Chapter~\ref{cp:CC}, I propose a discrete optimization approach to the aforementioned correlation clustering optimization.
This approach scales gracefully with the number of variables, better than existing approaches (e.g., \cite{Vitaladevuni2010}).

Chapter~\ref{cp:multiscale} concludes this part with a more general perspective on discrete optimization.
This new perspective is inspired by multiscale approaches and suggests to cope with the NP-hardness of discrete optimization using the {\em multiscale landscape} of the energy function.
Defining and observing this multiscale landscape of the energy, I propose methods to explore and exploit it to derive a coarse-to-fine optimization framework.
This new perspective gives rise to a unified multiscale framework for discrete optimization.
Our proposed multiscale approach is applicable to a diversity of discrete energies, both smoothness-encouraging as well as arbitrary, contrast-enhancing functions.

%% file: applications_main.tex
\part{Applications}  	
\label{part:app}

This part concentrate on the first axis of this thesis.
This direction explores new applications which require arbitrary energies.
We start with an unsupervised clustering objective function, Correlation Clustering (CC).
Chapters~\ref{cp:negaff-sketch} and~\ref{cp:negaff-cc} show several applications all revolving around the correlation clustering energy.
This energy is hard to optimize:
It has both smoothness-encouraging as well as contrastive pair-wise terms;
it has no data term to guide the optimization process,
and the number of discrete labels is not known a-priori.
We analyze an interesting property of the correlation clustering functional: its ability to recover the underlying number of clusters.
This interesting property is due mainly to the usage of terms that enhance contrast, rather than smoothness, in the solution $\bx$.

Another application that requires arbitrary energy is 3D reconstruction of surfaces.
We show, in chapter~\ref{cp:lighting}, how under certain conditions 3D reconstruction may be posed as a solution to a partial differential equation (PDE).
Solving this PDE to recover the 3D surface can be done through discretization of the solution space.
The discrete version of the PDE yields pair-wise terms that are beyond semi-metric.

Finally, the parameters of the energy function defining the terms $\varphi_i$ and $\phiij$ may not be fixed a-priori.
It may happen that one would like to learn the energy function from training data for various applications.
Chapter~\ref{cp:dtf} shows an example of such an energy learning framework.
The resulting learned energy is no longer guaranteed to be ``well-behaved".
In fact, experiments show that when the learning procedure is not constrained it is often the case that the resulting energy is arbitrary and does not yield any known structure.

%---------------------------------------------------------------------------%
\chapter[Sketching the Common]{Sketching the Common\protect\footnotemark{}}\protect\footnotetext{This is joint work with Or Brostovsky, Meirav Galun and Michal Irani. It was published in the 23$^{rd}$ International Conference on Computer Vision and Pattern Recognition (CVPR), \protect\citeyear{Bagon2010}.}
\label{cp:negaff-sketch}

Given very few images containing a common object of interest under
severe variations in appearance, we detect the common object and
provide a compact visual representation of that object, depicted
by a binary sketch. Our algorithm is composed of two stages:
(i)~Detect a mutually common (yet non-trivial) ensemble of
`self-similarity descriptors' shared by all the input images. \
(ii)~Having found such a mutually common ensemble, `invert' it to
generate a compact sketch which best represents this ensemble.
This provides a simple and compact visual representation of the
common object, while eliminating the background clutter of the
query images. It can be obtained from {\em very few} query images.
Such clean sketches may be useful for detection, retrieval,
recognition, co-segmentation, and for artistic graphical purposes.

The `inversion' process that generates the sketch is formulated as
a discrete optimization problem of a binary, non-submodular energy function.

% sketch the common

\input{sketching}

\chapter[Negative Affinities]{Negative Affinities\protect\footnotemark{}}\protect\footnotetext{This is joint work with Meirav Galun}
\label{cp:negaff-cc}

Clustering is a fundamental task in unsupervised learning.
The focus of this chapter is the Correlation Clustering (CC) functional which combines positive and negative affinities between pairs of data points.
In this chapter we provide a theoretical analysis of the CC functional.
Our analysis suggests a probabilistic generative interpretation for the functional, and
justifies its intrinsic ``model-selection" capability.
In addition we suggest two new applications that utilize the ``model-selection" capability of CC:
unsupervised face identification and interactive multi-object segmentation by rough boundary delineation.

The resulting CC energy is arbitrary and is very difficult to approximate.
We defer the discussion on our approximate minimization algorithms for the CC energy to chapter~\ref{cp:CC} in part~\ref{part:approx}, which deals with approximation schemes for arbitrary energies.

% "application" part of correlation clustering - includes theoretic interpretation, faces and interactive segmentation
\input{cc_app}

%---------------------------------------------------------------------------%
\chapter[3D Shape Reconstruction by Combining Motion and Lighting Cues]{3D Shape Reconstruction by Combining Motion and Lighting Cues\protect\footnotemark{}}\protect\footnotetext{This is joint work with Meirav Galun and Ronen Basri}
\label{cp:lighting}

In this chapter we consider the problem of reconstructing the 3D shape of a moving object while accounting for the change of intensities due to a change in orientation with respect to the light sources. We assume that both the lighting and motion parameters are given. Two methods are presented. First, for lambertian objects illuminated by a point source we derive a PDE that is quasilinear and implicit in the surface shape. We propose to solve this PDE by continuation (characteristic curves), extending an existing method to allow for large motion. Secondly, we formulate the reconstruction problem as an MRF and solve it using discrete optimization techniques. The latter method works with fairly general reflectance functions and can be applied to sequences of two or more images. It can also incorporate prior information and boundary conditions. We further discuss a method for extracting boundary conditions. We demonstrate the performance of our algorithms by showing reconstructions of smooth shaped objects and comparing these reconstructions to reconstructions with laser scans.

\input{lighting}
\chapter[Learning Discrete Energies]{Learning Discrete Energies\protect\footnotemark{}}\protect\footnotetext{This is joint work with Sebastian Nowozin, Carsten Rother, Toby Sharp, Bangpeng Yao and Pushmeet Kohli. It was published in the 13$^{th}$ International Conference on Computer Vision (ICCV), \citeyear{Nowozin2011}.}
\label{cp:dtf}

This chapter introduces a new formulation for discrete image labeling tasks,
the Decision Tree Field (DTF), that combines and generalizes random forests
and conditional random fields (CRF) which have been widely used in computer
vision.
%CRF based
%methods typically work by integrating likelihood potentials defined
%on individual pixels with potentials defined on pairs of neighboring
%pixels.
In a typical CRF model the unary potentials are derived from
sophisticated random forest or boosting based classifiers, however, the
pairwise potentials are assumed to (1) have a simple parametric form
with a pre-specified and fixed dependence on the image data, and (2)
to be defined on the basis of a small and fixed neighborhood. In contrast, in
DTF, local interactions between multiple variables are determined by
means of decision trees evaluated on the image data, allowing the
interactions to be adapted to the image content.
This results in powerful graphical models which are able to represent
complex label structure.
Our key technical contribution is to show that the DTF model
can be trained efficiently and jointly using a convex approximate
likelihood function, enabling us to learn over a million free model
parameters.
We show experimentally that for applications which have a rich and complex
label structure, our model outperforms state-of-the-art approaches.

\input{dtf}

%% file: sketching.tex
%\begin{abstract}
%Given very few images containing a common object of interest under
%severe variations in appearance, we detect the common object and
%provide a compact visual representation of that object, depicted
%by a binary sketch. Our algorithm is composed of two stages:
%(i)~Detect a mutually common (yet non-trivial) ensemble of
%`self-similarity descriptors' shared by all the input images. \
%(ii)~Having found such a mutually common ensemble, `invert' it to
%generate a compact sketch which best represents this ensemble.
%This provides a simple and compact visual representation of the
%common object, while eliminating the background clutter of the
%query images. It can be obtained from {\em very few} query images.
%Such clean sketches may be useful for detection, retrieval,
%recognition, co-segmentation, and for artistic graphical purposes.
%\end{abstract}

\section{Introduction}
\label{sec:sketch-intro}

%%Consider the following scenario: A user provides {\em a
%%few} (e.g., 3-5) example images containing an object of interest
%%with varying appearances, and wants to retrieve new images
%%containing this object (e.g., from a database, or from the web).
%%The problem addressed in this paper is motivated by this scenario
%%(although does not address the retrieval problem). }}
%

\begin{figure}
(a)
\fbox{\includegraphics[width=.7\linewidth]{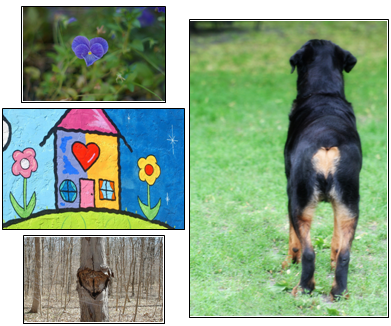}}
\ (b)
\includegraphics[width=.12\linewidth]{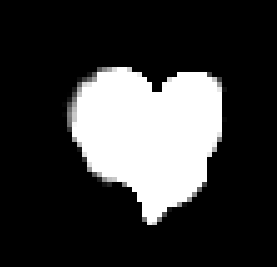}
\caption{{\bf Detecting and sketching the common:}
{\em (a) The 4 input images provided to the algorithm. 
(b) The least trivial common part (the heart) is detected and
sketched by the algorithm. }}
\label{fig:hearts}
\end{figure}

Given {\em very few images} (e.g., 3-5) containing a common object
of interest, possibly under severe appearance changes, we detect
the common object and provide a simple and compact visual
representation of that object, depicted by a binary sketch (see
Fig.~\ref{fig:hearts}).
The input images may contain additional distracting objects and
clutter, the object of interest is at unknown image locations, and
its appearance may significantly vary across the images (different
colors, different textures, and small non-rigid deformations). We
do assume, however, that the different instances of the object
share a {\em very rough} common geometric shape, of roughly the
same scale ($\pm 20\%$) and orientation ($\pm 15^\circ$). Our
output sketch captures this rough common shape.

The need to extract the common of {\em very few} images
%
%%is not a contrived one, and
%
occurs in various application areas, including: \ (i)~object
detection in large digital libraries. For example, a user may
provide very few (e.g., 3) example images containing an object of
interest with varying appearances, and wants to retrieve new
images containing this object from a database, or from the web. \
(ii)~ Co-segmentation of a few images.
%
%%(but unlike existing cosegmentation methods~\cite{REFS+NewRef},
%%we do not assume high visual similarity across the different instances of the object).
%
\ (iii)~Artistic graphical uses.
%
%

% HEARTS

%%%
%%%Our method is based on densely computed {\em Local Self-Similarity
%%%Descriptors}~\cite{Shechtman2007}. This descriptor
%%%%
%%%%(Fig.~\ref{fig:self_sim_desc})
%%%%
%%%implicitly captures local shape information, is invariant to
%%%severe changes in photometric properties (color, texture, etc.),
%%%and is {\em in}sensitive to small affine and non-rigid
%%%deformations. It was  empirically shown by~\cite{Horster2008} that
%%%the local self-similarity descriptor has a stronger descriptive
%%%power than SIFT.
%%%
%%%

Our method is based on densely computed {\em Local Self-Similarity
Descriptors}~\cite{Shechtman2007}. Our algorithm is composed of
two main steps: (i)~Identify the common object by detecting a
similar (yet ``non-trivial'') {\em ensemble of self-similarity
descriptors}, that is shared by all the input images.
Corresponding descriptors of the common object across the
different images should be similar in their descriptor values, as
well as in their relative positions within the ensemble.
(ii)~Having found such a mutually common ensemble of descriptors,
our method ``inverts'' it to generate a compact binary sketch
which best represents this ensemble. \

It was shown in~\cite{Shechtman2007} that given a {\em single
query image} of an object of interest (with very little background
clutter), it is possible to detect other instances of that object
in other images by densely computing and matching their local
self-similarity descriptors. The query image can be a real or
synthetic image, or even a {\em hand-drawn sketch} of the object.

In this paper we extend the method of~\cite{Shechtman2007} to
handle {\em multiple query images}. Moreover, in our case those
images are not centered around the object of interest (its
position is unknown), and may  contain also other objects and
significant background clutter. Our goal is to detect the {\em
``least trivial'' common part} in those query images, and generate
as clean as possible (region-based) sketch of it, while
eliminating the background clutter of the query images. Such clean
sketches can be obtained from {\em very few} query images, and may
be useful for detection, retrieval, recognition, and for artistic
graphical purposes. Some of these applications are illustrated in
our experiments.

Moreover, while~\cite{Shechtman2007} {\em received as an input} a
clean hand-drawn sketch of the object of interest (and used it for
detecting other instances of that object),  we {\em produce} a
sketch as one of our {outputs}, thereby also solving the
``inverse'' problem, namely: Given several images of an object, we
can generate its sketch using the self-similarity descriptor.

%
%%Our approach is based on detecting {\em common regions} (as
%%opposed to common edges), using densely computed {\em Local
%%Self-Similarity Descriptors}~\cite{Shechtman2007}. This is a local
%%{\em region-based feature}. It captures local shape, is invariant
%%to severe changes in photometric properties (color, texture,
%%etc.), and is {\em in}sensitive to small affine and non-rigid
%%deformations. It was further shown by~\cite{Horster2008} that the
%%local self-similarity descriptor has a strong descriptive power
%%(outperforming SIFT).
%

A closely related research area to the problem we address is that
of 'learning appearance models' of an object category, an area
which has recently received growing attention
(e.g.,~\cite{Chum2007,Chum2009,Ferrari2009,Jojic2005,karlinsky2008,Lee2009,Nguyen2009,Wu2009,Zhu2008},
to name just a few). The goal of these methods is to discover
common object shapes within collections of images. Some methods
assume a single object category
(e.g.,~\cite{Chum2007,Ferrari2009,karlinsky2008,Jojic2005,Nguyen2009,Wu2009,Zhu2008}),
while others assume multiple object categories
(e.g.,~\cite{Chum2009,Lee2009}). These methods, which rely on
weakly supervised learning ({\bf WSL}) techniques, typically
require tens of images in order to learn, detect and represent an
object category. What is unique to the problem we pose and to our
method is the ability to depict the common object from {\em very
few images}, despite the large variability in its appearance. This
is a scenario no WSL method (nor any other method, to our best
knowledge) is able to address. Such a small number of images
(e.g., $3$) does not provide enough 'statistical samples' for WSL
methods. While our method cannot compete with the performance of
WSL methods when many (e.g., tens) of example images are provided,
%%%%
%%%it {\em provides very good performance in scenarios that
%%%no existing method can address} --
%%%%
it outperforms existing methods when only few images with large
variability are available.
We attribute the strength of our method to the use of {\em densely
computed region-based information} (captured by the local
self-similarity descriptors), as opposed to commonly used {\em
sparse and spurious edge-based information} (e.g., gradient-based
features, SIFT descriptors, etc.) Moreover, the sketching step in
our algorithm provides an {\em additional global constraint}.

Another closely related research area to the problem addressed
here is `co-segmentation'
(e.g.,~\cite{Rother2004,Bagon2008,Mukherjee2009}). The aim of
co-segmentation is to segment out an object  common to a few
images ($2$ or more), by seeking segments in the different images
that share common properties (colors, textures, etc.) These common
properties are not shared by the remaining backgrounds in the
different images. While co-segmentation methods extract the common
object from {\em very few images}, they usually assume a much
higher degree of similarity in appearance between the different
instances of the object than that assumed here (e.g., they usually
assume similar color distributions, similar textures, etc.)

The rest of the paper is organized as follows:
Sec.~\ref{sec:formulation} formulates the problem and gives an
overview of our approach. Sec.~\ref{sec:detection} describes the
component of our algorithm which detects the `least trivial'
common part in a collection of images, whereas
Sec.~\ref{sec:sketching} describes the sketching component of our
algorithm. Experimental results are presented in
Sec.~\ref{sec:sketch-results}.

\section{Problem Formulation}
\label{sec:formulation}

\begin{figure}
\centering
\includegraphics[width=.8\linewidth]{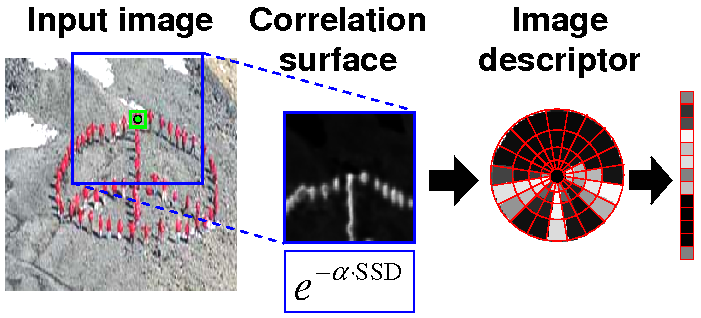}
\caption{{\bf The Local Self Similarity
Descriptor:} (Figure taken from~\cite{Shechtman2007}.) \
{\it  The self-similarity descriptor for any given point
(e.g., the green point in the left image), is computed by
measuring the similarity of a $5 \times 5$ patch around the point
with the surrounding $60 \times 60$ image region. This results in
a `correlation' surface (middle image). The correlation surface is
quantized into a compact log-polar representation of 45 bins (15
angles, 3 radial intervals) to achieve invariance against small
local affine and non-rigid deformations. The maximum value in
each bin constitutes the value at the corresponding descriptor
entry  (right most image).
}}
\label{fig:self_sim_desc}
\end{figure}

Let $I_1,...,I_K$ be $K$ input images containing a common object
under widely different appearances. The object may appear in
different colors, different textures, and under small non-rigid
deformations. The backgrounds are arbitrary and contain
distracting clutter. The images may be of different sizes, and the
image locations of the common object are unknown. We do assume,
however, that the different instances of the object share a {\em
very rough} common geometric shape, of roughly the same scale and
orientation. Our output sketch captures this rough common shape.

%%%The region-based descriptor we employ (and will be used in our
%%%problem formulation) is the {\em local self-similarity descriptor}
%%%of~\cite{Shechtman2007}. This descriptor (illustrated in
%%%Fig.~\ref{fig:self_sim_desc}) captures local shape information in
%%%the image vicinity where it is computed, while being invariant to
%%%its photometric properties (color, texture, etc.) Because of its
%%%log-polar representation, this descriptor is also {\em
%%%in}sensitive to small affine and non-rigid deformations in the
%%%image (up to $\pm 20\%$ in scale, and $\pm 15^\circ$). It was
%%%further shown empirically by~\cite{Horster2008} that the local
%%%Self-Similarity Descriptor has a strong descriptive power
%%%(outperforming SIFT).
%%
Our approach is thus based on detecting {\em 'common regions'} (as
opposed to 'common edges'), using densely computed {\em Local
Self-Similarity Descriptors}~\cite{Shechtman2007}. This descriptor
(illustrated in Fig.~\ref{fig:self_sim_desc}) captures local shape
information in the image vicinity where it is computed, while
being invariant to its photometric properties (color, texture,
etc.) Its log-polar representation makes this descriptor {\em
in}sensitive to small affine and non-rigid deformations (up to
$\pm 20\%$ in scale, and $\pm 15^\circ$). It was further shown
by~\cite{Horster2008} that the local self-similarity descriptor
has a strong descriptive power (outperforming SIFT).
The use of local self-similarity descriptors allows our method to
handle much stronger variations in appearance (and in much fewer
images) than those handled by previous methods.
We densely compute the self-similarity descriptors in images
$I_1,...,I_K$ (at every $5$-th pixel). `Common' image parts across
the images will have similar arrangements of self similarity
descriptors.

\begin{figure}
(a)\includegraphics[width=.6\linewidth]{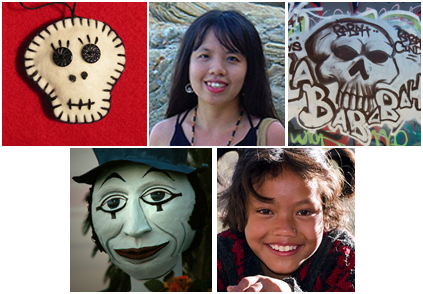}
(b)\includegraphics[width=.2\linewidth]{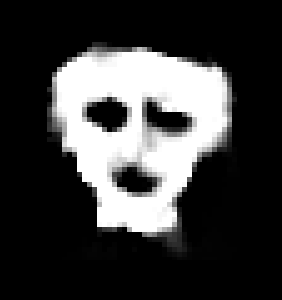}
\caption{{\bf Sketching:} \ {\small \it (a) Five
input images. \  (b) Their joint sketch.}}
\label{fig:faces}
\end{figure}

Let $c_1,...,c_K$ denote the unknown locations of the common
object in the $K$ images. Let $I_k^{c_k}$ denote a $w \times h$
subimage of $I_k$ centered at $c_k$,  containing the common object
($k=1,...,K$) (need not be tight). For short, we will denote it by
$\tilde{I_k}$.
%%%%
%%%$w \times h$ is unknown, needs {\em not} be tight around the
%%%common object, but is assumed to be the same for all images.
%%%%
The sketch we seek is a binary image $S$ of size $w \times h$
which best captures the rough characteristic shape of the common
object shared by $\tilde{I_1},...,\tilde{I_K}$.
More formally, we seek a binary image $S$ whose {local
self-similarity descriptors} match as best as possible the local
self-similarity descriptors
%extracted densely from
of $\tilde{I_1},...,\tilde{I_K}$. The descriptors should match in
their {\em descriptor values}, as well as in their {\em relative
positions} with respect to the centers $\{c_k\}$:
\begin{eqnarray}
\label{eq:Score}
Score(S|\tilde{I_1},...,\tilde{I_K}) & = & \sum_{k=1}^{K}match(S,\tilde{I_k}) \\
                    & = & \sum_{k=1}^{K}\sum_{i=1}^{w \cdot h} sim\left(
                    d_i^S,d_i^k
                    \right) \nonumber
\end{eqnarray}
where $d_i^S$ is the $i$-th {self-similarity descriptor} computed
at image location $l_i$ in the sketch image $S$, $d_i^k$ is the
self-similarity descriptor computed {\em at the same relative
position} $l_i$ (up to small shifts) in the $w \times h$ subimage
$\tilde{I_k}$, and $sim(d_1,d_2)=-\parallel d_1-d_2
\parallel_p$ measures how similar two descriptor vectors are (we
experimented with $L_p$ norms for $p=1,2$).
Thus, the {\em binary sketch} we seek is:
\begin{equation}
\label{eq:argmax}
\hat{S}=argmax \{  Score(S|\tilde{I_1},...,\tilde{I_K}) \} \  \
s.t. \ \ S(l)\in\{-1,1\}
%\forall (x,y) \in S
\end{equation}
%
%(where $l$ denote pixel locations).
where $S(l)$ is the value of $S$ at pixel $l$. This process is
described in detail in Sec.~\ref{sec:sketching}, and results in a
sketch of the type shown in Fig~\ref{fig:faces}.

\begin{figure}
\begin{tabular}{ccc}
\includegraphics[width=.28\linewidth]{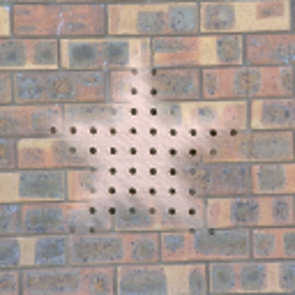}&
\includegraphics[width=.28\linewidth]{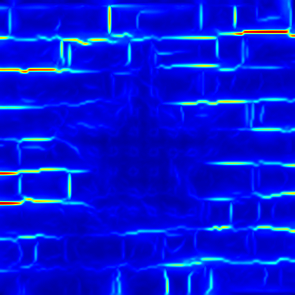}&
\includegraphics[width=.28\linewidth]{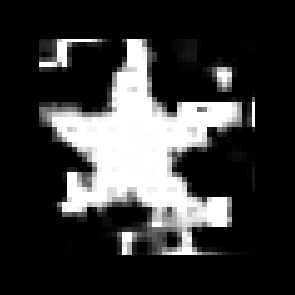}\\
(a) & (b) & (c)
\end{tabular}
\caption{{\bf Regions  vs. Edges:}{\small \it \ \
(a) a single input image. \ (b) The edge map generated by the
method of~\cite{Maire2008}. \ (c) The binary sketch generated by
our method when applied to the \underline{single input image}
(using all the self-similarity descriptors densely computed in
that image). \ This illustrates the concept that region-based
information is much richer than sparse edge-based information, and
therefore appears to be more powerful for detection and for
sketching. }}
\label{fig:star}
\end{figure}

While edge-based detection and/or
sketching~\cite{Lee2009,Zhu2008,Ferrari2009} requires many input
images, our region-based detection and sketching can be recovered
from very few images.
Edges tend to be very spurious, and are very prone to clutter
(even sophisticated edge detectors like~\cite{Maire2008} -- see
Fig.~\ref{fig:star}.b). Edge-based approaches thus require a
considerable number of images, to allow for the consistent
edge/gradient features of the object to stand out from the
inconsistent background clutter. In contrast, region-based
information is much less sparse (area vs. line-contour), less
affected by clutter or by misalignments, and is not as sensitive
to the existence of strong clear boundaries. Much larger image
offsets are required to push two corresponding regions out of
alignment than to misalign two thin edges. Thus, region-based cues
require fewer images to detect and represent the common object.
Indeed, our method can provide good sketches from as few as $3$
images. In fact, in some cases our method produces a meaningful
sketch even from a {\em single} image, where edge-based sketching
is impossible to interpret -- see example in Fig.~\ref{fig:star}.

In the general case, however, the locations $c_1,...,c_K$ of the
object within the input images $I_1,...,I_K$, are unknown. We seek
a binary image $S$
%%%
%%%(of pre-defined frame size $w \times h$),
%%%
which sketches the {\em `least trivial'} object (or image
part) that is {\em `most common'} to all those images. The {\em
`most common'} constraint is obvious:
in each image $I_k$ there should be a location $c_k$ for which \
$match \left(S,I_k^{c_k} \right)$  \ is high (where
$\tilde{I_k}=I_k^{c_k}$ is the subimage centered at $c_k$). \
However, there are many image regions that are {\em trivially}
shared by many natural images. For example, {\em uniform regions}
(of uniform color or uniform texture) occur abundantly in natural
images. Such regions share similar self-similarity descriptors,
even if the underlying textures or colors are different (due to
the invariance properties of the self-similarity descriptor).
Similarly, strong vertical or horizontal edges (e.g., at
boundaries between two different uniformly colored/textured
regions) occur abundantly in  images.
We do not wish to identify such trivial (insignificant) common
regions in the images as the `common object'.

Luckily, since such regions have good image matches in lots of
locations, the {\em statistical significance}
of their good matches tends to be low (when measured by how many
standard deviations its peak match values are away from its mean
match value in the collection of images).
In contrast, a {\em non-trivial}  common part (with non-trivial
structure) should have at least one good match in each input image
(could also have a few matches in an image), but these matches
would be `statistically significant' (i.e., this part would not be
found `at random' in the collection of images).

Thus, in the general case, we seek a {binary sketch} $S$ and
locations $c_1,...,c_K$ in images $I_1,...,I_K$, such that:

\noindent {\bf (i)  $S$  is `most common'}, in the sense that it
maximizes $Score(S|I_1^{c_1},..,I_K^{c_K}) =
\sum_{k=1}^{K}match(S,I_k^{c_k})$ of Eq.~(\ref{eq:Score}).

\noindent  {\bf (ii)  $S$ is `least trivial'}, in the sense that
its matches at $c_1,...,c_K$  are {\em statistically significant},
i.e., it maximizes $\sum_{k=1}^{K} StatSignificance \left(
match(S,I_k^{c_k}) \right)$,\  where the significance of a match
of $S$ is measured by how many standard deviations it is away from
the mean match value of $S$.

Our optimization algorithm may iterate between these two
constraints: \  (i)~Detect the locations $\{c_k\}_{k=1}^K$ of the
least trivial common image part in $\{I_k\}_{k=1}^K$
(Sec.~\ref{sec:detection}). \ (ii)~Sketch the common object given
those image locations (Sec.~\ref{sec:sketching}). \ \ The overall
process results in a sketch image, which provides a simple compact
visual representation of the common object of interest in a set of
query images, while eliminating any distracting background clutter
found in those images.

\section{Detecting the Common}
\label{sec:detection}

We wish to detect image locations $c_1,...,c_K$ in $I_1,...,I_K$,
such that corresponding subimages centered at those locations,
$I_k^{c_k}$, share as many self-similarity descriptors with each
other as possible, yet their matches to each other are non-trivial
(significant). The final sketch $S$ will then be obtained from
those subimages (Sec.~\ref{sec:sketching}).

Let us first assume that the dimension  $w \times h$ of the
subimages is given. We will later relax this assumption.
Let $\tilde{I}$ be a $w \times h$ image segment (this could be the
final sketch $S$, or a subimage extracted from one of the $K$
input images in the iterative process). We wish to check if
$\tilde{I}$ has a good match in each of the input images
$I_1,...,I_K$, and also check the statistical significance of its
matches.
We `correlate' $\tilde{I}$ against all the input images (by
measuring the similarity of its underlying self-similarity
descriptors\footnote{We use the same algorithm employed
by~\cite{Shechtman2007} to match ensembles of self-similarity
descriptors, which is a modified version of the efficient
``ensemble matching'' algorithm of~\cite{Boiman2007}. This
algorithm employs a simple probabilistic ``star graph'' model to
capture the relative geometric relations of a large number of
local descriptors, up to small non-rigid deformations.}).
In each image $I_k$ we find the highest match value of
$\tilde{I}$:  $maxMatch(\tilde{I},I_k)$. The higher the value, the
stronger the match. However, not every high match value is
statistically significant. The {\em statistical significance}
of $maxMatch(\tilde{I},I_k)$ is measured by how many standard
deviations it is away from the mean match value of $\tilde{I}$ in
the entire collection of images, i.e.,:
\begin{equation*}
\left(maxMatch(\tilde{I},I_k) - avgMatch(\tilde{I}) \right) /
stdMatch(\tilde{I})
\end{equation*}
where $avgMatch(\tilde{I})$ is the mean of all match values of
$\tilde{I}$ in the collection $I_1,...,I_K$, and
$stdMatch(\tilde{I})$ is their standard deviation. We thus define
the `Significance' of a subimage  $\tilde{I}$ as:
\begin{equation*}
Significance(\tilde{I}|I_1,...,I_K)=\frac{1}{K} \sum_{k=1}^K
StatSignificance \left( maxMatch(\tilde{I},I_k) \right)
\end{equation*}

Initially, we have no candidate sketch $S$. However, we can
measure how `significantly common' is each $w \times h$ subimage
of $I_1,...,I_K$, when matched against all locations in all the
other $K-1$ images.
We can assign a significance score to each {\em pixel} $p \in I_k$
($k=1,..,K$), according to the `Significance' of its surrounding
$w \times h$ subimage: $Significance(I_k^p|I_1,...,I_K)$.

We set $c_k$ to be the pixel location with the {\em highest}
significance score in image $I_k$, i.e., $c_k = argmax_{p \in I_k}
\{ Significance(I_k^p|I_1,...,I_K) \} $.

The resulting $K$ points (one per image), $c_1,...,c_K$, provide
the centers for $K$ candidates of `non-trivial' common image
parts.
We generate a sketch $S$ from these image parts (using the
algorithm of Sec.~\ref{sec:sketching}).

\begin{figure}
\includegraphics[width=\linewidth]{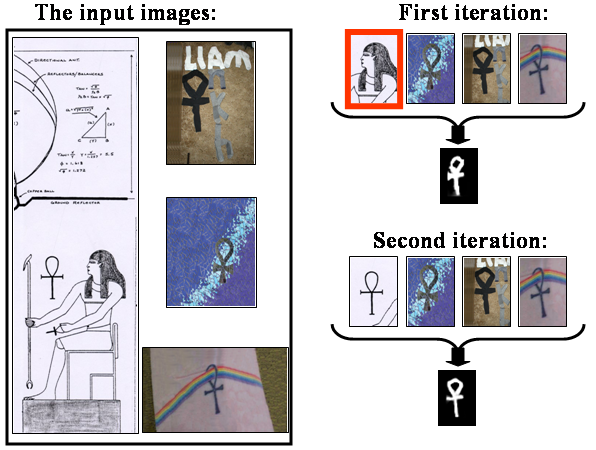}
\caption{{\bf Iterations of Detection \&
Sketching:} {\small \it {\bf Left:} The $4$ input images. \ {\bf
Right:} The first iteration of the detection algorithm results in
$4$ detected image regions, of which $3$ are correct and one is an
outlier (marked by red). The resulting sketch produced from these
regions is reasonably good (due to the robustness of the sketching
to outliers -- see Secs.~\ref{sec:sketching}
and~\ref{sec:sketch-experiments}), and is used for refining the detection
in the input images. This results in $4$ correct detections in the
second iteration, and an improved sketch. }}
\label{fig:iterations}
\end{figure}

% LARGE FACES
\begin{figure}
\includegraphics[width=\linewidth]{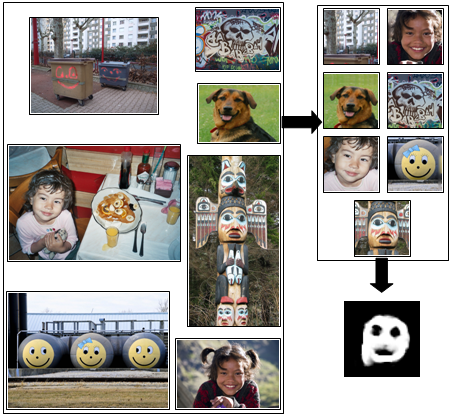}
\caption{{\bf Detecting and sketching the
common:} {\it \small  (Left) The input images. \  (Upper-Right)
The detected image regions of the common object, including one
outlier. \  (Lower-Right) The resulting sketch.}}
\label{fig:large-faces}
\end{figure}

We repeat the above process, this time for $\tilde{I}=S$, to
detect its best matches in $I_1,...,I_K$. This should lead to
improved detection and localization of the common object
($c_1,...,c_K$), and accordingly to an improved sketch $S$. This
algorithm can be iterated several times.
%%%%
%%%In practice, we found $2$-$3$ times to be enough.
%%%%
In practice, in all our experiments a good sketch $S$ was
recovered already in the first iteration. An additional iteration
was sometimes useful for improving the detection.
Fig.~\ref{fig:iterations} shows two iterations of this process,
applied to $4$ input images. More results of the detection can be
seen in Fig.~\ref{fig:large-faces}.

\noindent {\bf Handling unknown w $\times$ h:}
%%%
%%%\ \ {[ADD HERE 1-2 SHORT PARAGRAPHS] } \\
%%%
\ \ In principle, when $w \times h$ is unknown, we can run the
above algorithm {\em ``exhaustively''} for a variety of
$w=w_{min},..,w_{max}$ and $h=h_{min},..,h_{max}$, and choose
``the best'' $w \times h$ (with maximal significance score). In
practice, this is implemented more efficiently using ``integral
images'', by integrating the contributions of individual
self-similarity descriptors into varying window sizes $w \times
h$.

\noindent {\bf Computational Complexity:}
\ \ The detection algorithm is implemented coarse-to-fine. The
first step of the algorithm described above is quadratic in the
size of the input images. However, since the number of images is
typically small (e.g., $3-5$), and since the quadratic step occurs
only in the coarsest/smallest resolutions of the images, this
results in a computationally efficient algorithm.

%%%%
%%%(e.g., an order of $1-2$ minutes for the entire detection and
%%%sketching for $5$ images).

\section{Sketching the Common}
\label{sec:sketching}
Let $\tilde{I_1},\ldots,\tilde{I_K}$ be the $w \times h$ subimages
centered around the common object (detected and extracted from the
input images using the algorithm of Sec.~\ref{sec:detection}). The
goal of the sketching process is to produce a {binary image} $S$,
which best captures the rough characteristic shape of the object
shared by $\tilde{I_1},...,\tilde{I_K}$, as posed by
Eq.~(\ref{eq:argmax}). Namely, find $S$ whose ensemble of
self-similarity descriptors is as similar as possible to the
ensembles of  descriptors extracted from
$\tilde{I_1},\ldots,\tilde{I_K}$. If we were to neglect the binary
constraint $S(l)\in\{-1,1\}$ in Eq.~(\ref{eq:argmax}), and the
requirement for consistency between descriptors of an image, then
the {\em optimal solution} for the collection of self-similarity
descriptors of $S$, \ $\{d_i\}_{i=1}^{w \cdot h}$, \ could be
explicitly computed as:
\begin{eqnarray}
\label{eq:median}
d_i  = & \textmd{median}_k \{ d_i^k \}   & ~~\mbox{if }L_1\mbox{-norm}\\
d_i  = & \textmd{mean}_k   \{ d_i^k \}   & ~~\mbox{if
}L_2\mbox{-norm} \nonumber
\end{eqnarray}
We use the $L_1$-norm to generate these `combined' descriptors
$\{d_i\}_{i=1}^{w \cdot h}$, because of the inherent robustness of
the median operator to outliers in the descriptors (also confirmed
by our empirical evaluations in Sec~\ref{sec:sketch-results}).
Having recovered such a collection of descriptors for $S$, we
proceed and solve the ``inverse'' problem -- i.e., to generate the
image $S$ from which these descriptors emanated. However, the
collection of descriptors $\{d_i\}_{i=1}^{w \cdot h}$ generated
via a `median' or `average' operations is no longer guaranteed to
be a valid collection of self-similarity descriptors of any real
image (binary or not). We thus proceed to recover the simplest
possible image $S$ whose self-similarity descriptors best
approximate the `combined' descriptors $\{d_i\}_{i=1}^{w \cdot h}$
obtained by Eq.~(\ref{eq:median}).

%%%We next describe how we determine  $S$ that best approximates the
%%%combined descriptors.
%%%%

Self-similarity descriptors cover large image regions, with high
overlaps. As such, the similarity and dissimilarity between two
image locations (pixels) of $S$ are \textit{implicitly} captured
by multiple self-similarity descriptors and in different
descriptor entries. The self-similarity descriptor as defined
in~\cite{Shechtman2007} has values in the range $[0,1]$, where $1$
indicates high resemblance of the central patch to the patches in
the corresponding log-polar bin, while $0$ indicates high {\em
dis}similarity of the central patch to the corresponding log-polar
bin. For our purposes, we stretch the descriptor values to the
range $[-1,1]$, where $1$ signifies ``attraction'' and $-1$
signifies ``repulsion'' between two image locations.

\begin{figure}
\centering
%\protect\parpic(.6\linewidth,4cm)(1cm,1cm)[r][t]{}
\includegraphics[width=.5\linewidth]{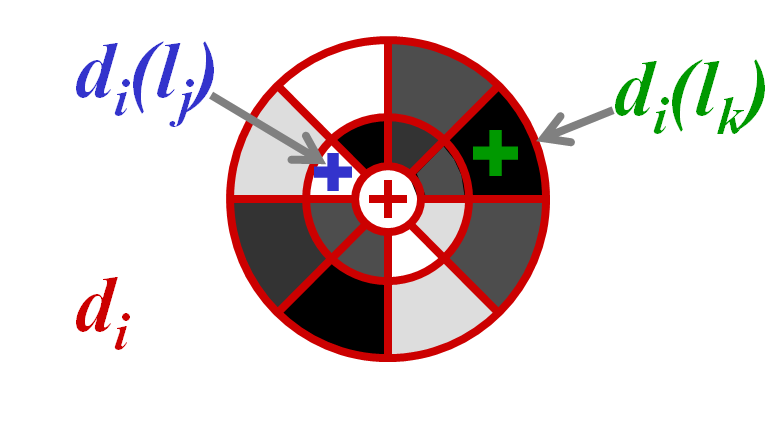}
\caption{
{\bf Computing attraction/repulsion matrix $W$:}
{\em  The log-polar self-similarity
descriptor $d_i$ is located at $l_i$ (red cross). White bins
signify image areas of high similarity to the central patch, dark
bins signify image areas of dissimilarity to the central patch.
The point $l_j$ (blue cross), which is the center of descriptor
$d_j$ (not drawn), falls in a white bin of descriptor $d_i$ (i.e.,
$0 < d_i(l_j) \leq 1$). The entry $w_{ij}$ in the matrix $W$ is
determined accordingly: $w_{ij} =  \alpha_{ij} \left( d_i(l_j) +
d_j(l_i) \right)/2$, where $\alpha_{ij}$ (the certainty assigned
to this entry), is inversely proportional to the distance
$\parallel l_i-l_j \parallel$ (the distance between the red and
blue crosses). Similarly, the point $l_k$ (green cross), which is
the center of another descriptor $d_k$ (also not drawn), falls in
a dark bin of descriptor $d_i$, i.e., $-1 \leq d_i(l_k) < 0$, and
$\alpha_{ik} < \alpha_{ij}$ (because the green cross falls farther
away from the center of $d_i$, hence lower certainty).
%Similarly, $w_{ik} = \left( d_i(l_k) + $d_j(l_k) \right) \alpha_{ik}$.
%
}}
\label{fig:self-sim-correlation}
\end{figure}

Let $W$ be a  $wh \times wh$ matrix capturing the
attraction/repulsion between every two image locations, as induced
by the collection of the `combined' self-similarity descriptors
$\{d_i\}_{i=1}^{w \cdot h}$ of Eq.~(\ref{eq:median}). Entry
$w_{ij}$ in the matrix is the degree of attraction/repulsion
between image locations $l_i$ and $l_j$, determined by the
self-similarity descriptors $d_i$ and $d_j$ centered at those
points. $d_i(l_j)$ is the value of the bin containing location
$l_j$ in descriptor $d_i$ (see
Fig.~\ref{fig:self-sim-correlation}). Similarly, $d_j(l_i)$ is the
value of the bin containing location $l_i$ in descriptor $d_j$.
The entry $w_{ij}$ gets the following value:
\begin{equation}
w_{ij} =  \alpha_{ij} \left( d_i(l_j) +    d_j(l_i) \right) / 2
\end{equation}
where $\alpha_{ij} = \alpha_{ji}$ is inversely proportional to the
distance $\parallel l_i - l_j
\parallel$ between the two image locations (we give higher weight to bins that are closer to
the center of the descriptor, since they contain more
accurate/reliable information).

% HORSES
\begin{figure}
(a) \fbox{\includegraphics[width=.7\linewidth]{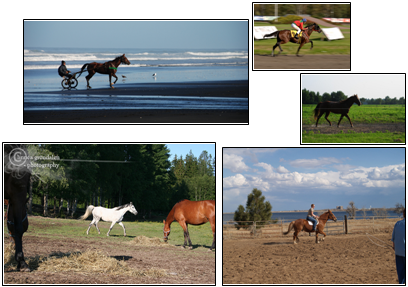}}
(b) \includegraphics[width=.14\linewidth]{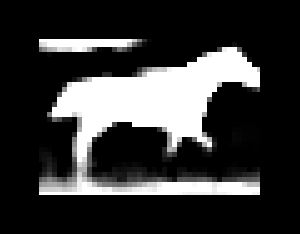}
\caption{{\bf Detecting and sketching the
common:} {\it \small (a) Five input images. (b) The resulting
sketch.}}
\label{fig:horses}
\end{figure}

Note that a `pure' attraction/repulsion matrix $W$ of a true
binary image $S$ contains only $3$ types of values $w_{ij}$: $-1,
0, 1$. If $l_i$ and $l_j$ belong to the same region in $S$ (i.e.,
both in foreground or both in background), then $w_{ij}=1$; if
$l_i$ and $l_j$ belong to different regions in $S$, then
$w_{ij}=-1$, and if the points are distant (out of descriptor
range), then $w_{ij}=0$.  In the general case, however, the
entries span the range $[-1,1]$, where $1$ stands for ``strong''
attraction, $-1$ for ``strong'' repulsion and $0$ means ``don't
care''. The closer the value of $w_{ij}$ to $0$, the lower its
attraction/repulsion confidence; the closer it is to $\pm 1$, the
higher the attraction/repulsion confidence.

Note that $W$ is different from the classical affinity matrix used
in spectral clustering or in min-cut, which use non-negative
affinities, and their value $0$ is {\em ambiguous} -- it signifies
both {\em high-dissimilarity} as well as {\em low-confidence}. The
distinction between `attraction', `repulsion', and
`low-confidence' is critical in our case, thus we cannot resort
to the max-flow algorithm or to spectral clustering in order to
solve our problem. An affinity matrix with positive and negative
values was used by~\cite{Yu2001} in the context of the normalized-cut
functional. However, their functional is not appropriate for our
problem (and indeed did not yield good results for $S$ when
applied to our $W$). We therefore define a different functional
and optimization algorithm in order to solve for the binary sketch
$S$.

The  binary image $S$ which {\em best} approximates the
attraction/repulsion relations captured by $W$,
will minimize the following functional:
\begin{equation}
\label{eq:functional_opt}
\min_S \sum_{i,j} w_{ij} (S(l_i) - S(l_j))^2 ~~~~~\mbox{subject
to}~~ S(l) \in \{-1,1\}
\end{equation}
where $S(l)$ is the value of $S$ at pixel $l$.
Note that for a binary image, the term $(S(l_i) - S(l_j))^2$ can
obtain only one of two values:  $0$ (if both pixels belong to
foreground, or both belong to background), {\em or} \  $4$ (if one
belongs to the foreground, and one to the background). Thus, when
$w_{ij}$ is positive (attraction), $S(l_i)$ and $S(l_j)$ should
have the same value (both $1$ or both $-1$), in order to minimize
that term $w_{ij} (S(l_i) - S(l_j))^2$. The larger $w_{ij}$
(stronger confidence), the stronger the incentive for $S(l_i)$ and
$S(l_j)$ to be the same. Similarly, a negative $w_{ij}$
(repulsion) pushes {\em apart} the values $S(l_i)$ and $S(l_j)$.
Thus, $S(l_i)$ and $S(l_j)$ should have opposite signs in order to
minimize that term $w_{ij} (S(l_i) - S(l_j))^2$.  When
$w_{ij}\approx 0$ (low confidence), the value of the functional
will not be affected by the values $S(l_i)$ and $S(l_j)$ (i.e.,
``don't care'').  It can be  shown that in the `ideal' case, i.e.,
when $W$ is generated from a binary image $S$,  the global minimum
of Eq.~(\ref{eq:functional_opt}) is obtained at $S$.

\noindent {\bf Solving the constrained optimization problem:} \
The min-cut problem where only non-negative values of $w_{ij}$ are
allowed can be solved by the max-flow algorithm in polynomial
time. However,  the weights $w_{ij}$ in the functional of
Eq.~(\ref{eq:functional_opt}) can obtain both positive and
negative values, turning our `cut' problem as posed above into an
NP-hard problem. We therefore {\em approximate}
Eq.~(\ref{eq:functional_opt}) by reposing it as a quadratic
programming problem, while relaxing the binary constraints.

Let $D$ be a diagonal matrix with $D_{ii} = \sum_j w_{ij}$, and
let $L=D-W$ be the graph Laplacian of $W$. Then $\frac{1}{2}
\sum_{i,j} w_{ij} (S(l_i) - S(l_j))^2 = S^T L S$. Thus, our
objective function is a quadratic expression in terms of $S$. The
set of binary constrains are relaxed to the following set of
linear constraints $-1 \leq S(l) \leq 1$, resulting in the
following quadratic programming problem:
\begin{equation}
\hat{S} = \arg \min_S S^T L S  \ \ \ \mbox{s.t.} \ \ -1 \leq S(l)
\leq 1
\end{equation}
Since $L$ is not necessarily positive semi-definite, we do not
have a guarantee regarding the approximation quality  (i.e., how
far is the achieved numerical solution from the optimal solution).
Still, our empirical tests demonstrate good performance of this
approximation. We use Matlab's optimization toolbox (quadprog) to
solve this optimization problem and obtain a sketch $\hat{S}$. In
principle, this does not yield a binary image. However, in
practice, the resulting sketches look very close to binary images,
and capture well the rough geometric shape of the common objects.

The above sketching algorithm is quite robust to outliers (see
Sec.~\ref{sec:sketch-results}), and obtains good sketches from very few
images. Moreover, if when constructing the attraction/repulsion
matrix $W$ we replace the `combined' descriptors of
Eq.~(\ref{eq:median}) with the self-similarity descriptors of a
{\em single image}, our algorithm will produce `binary' sketches
of a single image (although these may not always be visually
meaningful). An example of a sketch obtained from a single image
(using all its self-similarity descriptors) can be found in
Fig.~\ref{fig:star}.

\section{Experimental Results}
\label{sec:sketch-results} \label{sec:sketch-experiments}
%
% TRIKONA
\begin{figure}
\vspace*{1.5mm}
(a) \fbox{\includegraphics[width=.69\linewidth]{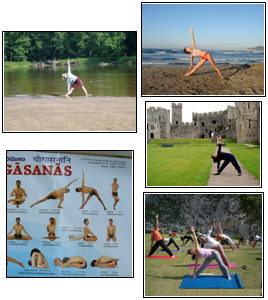}}
(b) \includegraphics[width=.15\linewidth]{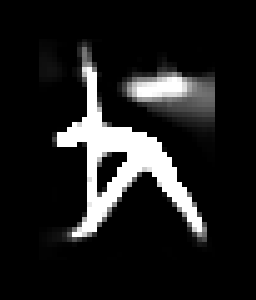}
\caption{{\bf Detecting and sketching the
common:} {\small \it (a) The input images. (b) The resulting
sketch.
% least trivial common part (the Trikonasana pose)  is detected and sketched by our algorithm.
}}
\label{fig:trikonasana}
\end{figure}

%%% ETHZ figure
\begin{figure}
\begin{tabular}{|c|c||c|c|}
\hline
\rule{0cm}{6.5mm}Input images &\parbox[b]{7ex}{Output sketch}& Input images &\parbox[b]{7ex}{Output sketch}\\
\hline
\rule{0cm}{20mm}\includegraphics[width=.25\linewidth]{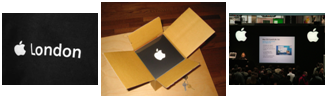}&
\raisebox{2mm}{\includegraphics[width=.05\linewidth]{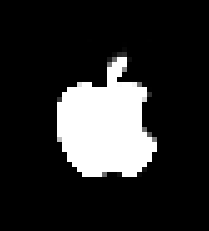}}&
\includegraphics[width=.52\linewidth]{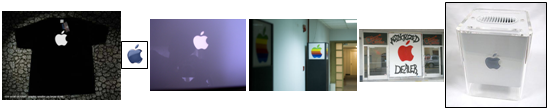}&
\raisebox{2mm}{\includegraphics[width=.05\linewidth]{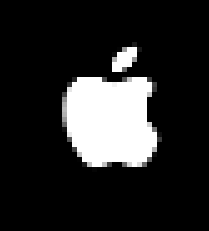}}\\
\hline
\rule{0cm}{18mm}\includegraphics[width=.25\linewidth]{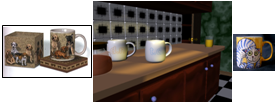}&
\raisebox{2mm}{\includegraphics[width=.05\linewidth]{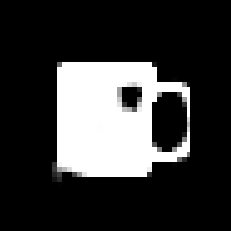}}&
\includegraphics[width=.52\linewidth]{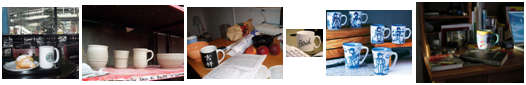}&
\raisebox{2mm}{\includegraphics[width=.05\linewidth]{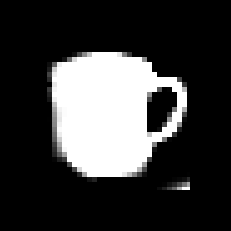}}\\
\hline
\rule{0cm}{21mm}\includegraphics[width=.25\linewidth]{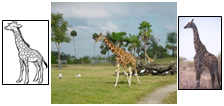}&
\raisebox{2mm}{\includegraphics[width=.06\linewidth]{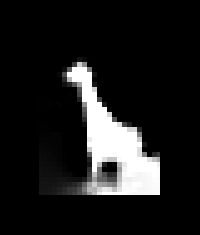}}&
\includegraphics[width=.52\linewidth]{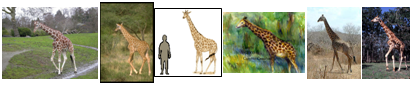}&
\raisebox{2mm}{\includegraphics[width=.06\linewidth]{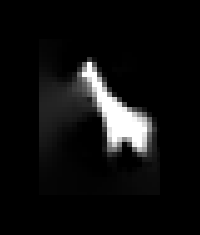}}\\
\hline
\rule{0cm}{21mm}\includegraphics[width=.25\linewidth]{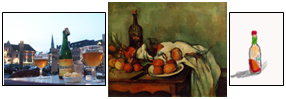}&
\raisebox{2mm}{\includegraphics[width=.03\linewidth]{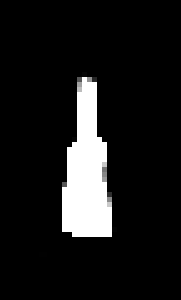}}&
\includegraphics[width=.52\linewidth]{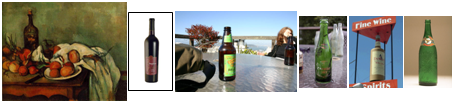}&
\raisebox{2mm}{\includegraphics[width=.03\linewidth]{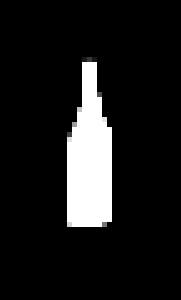}}\\
\hline
\rule{0cm}{20mm}\includegraphics[width=.25\linewidth]{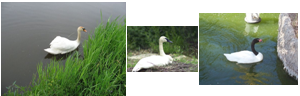}&
\raisebox{2mm}{\includegraphics[width=.05\linewidth]{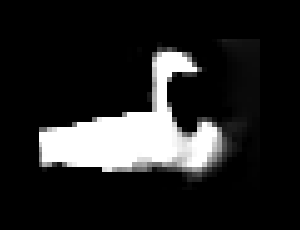}}&
\includegraphics[width=.52\linewidth]{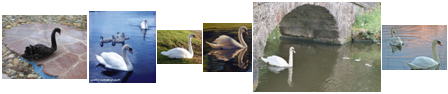}&
\raisebox{2mm}{\includegraphics[width=.05\linewidth]{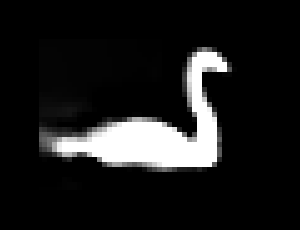}}\\
\hline
\end{tabular}
\caption{{\bf Sample results on ETHZ
shapes \cite{Ferrari2006} dataset:} \ {\small \it Detection and
sketching using only 3 images (left), and using 6 images
(right).}}
\label{fig:ethz-sketches}
\end{figure}

\begin{figure}
\begin{tabular}{cc}
\includegraphics[width=.45\linewidth]{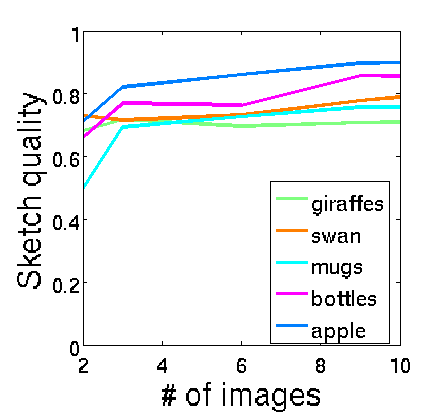}  &
\parbox[b]{.45\linewidth}{\caption{{\bf
Evaluating sketch quality:} {\em \small Mean values of
$Quality(S)$ as a function of the number of input
%%%`query images'
images ($K=2,...,10$) randomly sampled from each set of ETHZ shape
dataset~\cite{Ferrari2006}.\vspace*{5mm}}}
\label{fig:graph-sketch-vs-gt}
}
\end{tabular}
\begin{tabular}{cl}
\includegraphics[width=.45\linewidth]{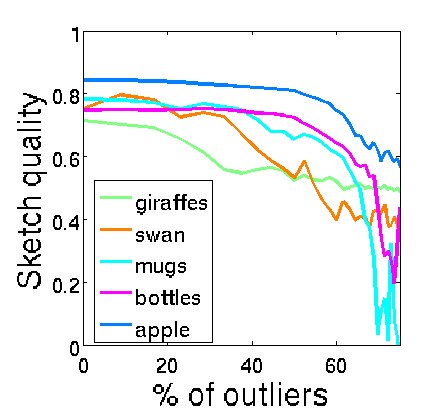}&
\parbox[b]{.45\linewidth}{
\caption{{\bf Sketching in
presence of outliers:} {\em \small We ``corrupt" a set of 10
``inlier" with $n$ randomly chosen natural images. Graph shows
mean values of $Quality(S)$ as a function of the percent of
outlier images in the input set, i.e., $n/(10+n)$.}}
\label{fig:graph-sketch-with-outliers}
}
\end{tabular}
\begin{tabular}{cl}
\includegraphics[width=.45\linewidth]{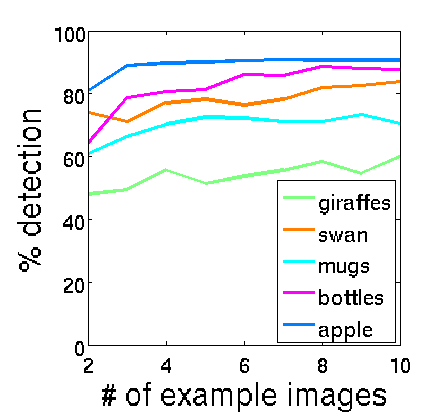}&
\parbox[b]{.45\linewidth}{
\caption{{\bf Detection in new
images:} {\em \small We empirically evaluated how well the sketch
generated form very few images ($K=2,..10$) performs in detecting
the common shape in \underline{new} images.\vspace*{5mm}}} 
\label{fig:graph-detect-in-novel}}
\end{tabular}
\end{figure}

%

%%%%We tested our detection and sketching algorithm both qualitatively
%%%%and quantitatively.

Figs.~\ref{fig:hearts},\ref{fig:faces},\ref{fig:large-faces},\ref{fig:horses},\ref{fig:trikonasana},\ref{fig:ethz-sketches}
show qualitative results on various image sets. In all of these
examples the number of input images was very small ($3-7$), with
large variability in appearance and background clutter. Our
algorithm was able to detect and produce a compact representation
(a sketch) of the common content.

We further conducted empirical evaluations of the algorithm using
ETHZ shape dataset~\cite{Ferrari2006}. This dataset consists of
five object categories with large variability in appearance:
Applelogos, Bottles, Giraffes, Mugs and Swans (example images can
be seen in Fig.~\ref{fig:ethz-sketches}). There are around $50$
images in each set, with ground-truth information regarding the
location of the object in each image, along with a single
hand-drawn ground truth shape for each category.  In order to
assess the quality of our algorithm (which is currently not scale
invariant, although it can handle up to $\pm 20\%$ scale
variation, and $\pm 15^\circ$ rotations), we scaled the images in
each dataset to have {\em roughly} the same object size (but we
have not rotated the images, nor changed their aspect ratios).

\noindent {\bf Sketch quality score:}
%%%%
%%%%To asses the quality of our produced sketch $S$, it is first
%%%%aligned to the ground-truth sketch $S_{GT}$ (to account for small
%%%%affine transformations between the input images and the single
%%%%hand-drawn ground-truth sketch).
%%%%
Because our sketch $S$ is continuous in the range $\left[-1,\
1\right]$, we stretch the values of the ground-truth sketch
$S_{GT}$ also to this range, and multiply the two sketches
pixel-wise.  Our sketch quality score is: \
\mbox{$Quality(S)={<S,S_{GT}>}/{(\#~of~pixels)}.$}
In places where both sketches agree in their sign (either white
regions or black) the pixel-wise product is positive, while in
places where the sketches disagree, the product is negative. This
produces a sketch quality score with values ranging between $-1$
(lowest quality) to $+1$ (highest quality). Note that even if our
sketch displays a perfect shape, its quality will be smaller than
$1$, because it is not a perfect binary image. From our
experience, sketch quality $\geq 0.8$ are usually
excellent-looking sketches.

We first assessed the quality of our algorithm to identify and
sketch the common object correctly, as a function of the number of
input images $K$ ($K=2,3,..,10$). We randomly sampled $K$ images
out of an object category set, applied our detection and sketching
algorithm to that subset, and compared the resulting sketch $S$ to
the ground-truth $S_{GT}$. We repeated this experiment $15$ times
for each $K$, and computed mean sketch quality scores.
Fig.~\ref{fig:graph-sketch-vs-gt} displays plots of the mean
quality score for the $5$ categories. It can be seen that from
relatively few images ($K=3$) we already achieve sketches of good
quality, even for challenging sets such as the giraffes (although,
with the increased number of example images, its legs tend to
disappear from the sketch because of their non-rigid
deformations). Examples for sketching results for some of these
experiments can be seen in Fig.~\ref{fig:ethz-sketches}.

We next evaluated the robustness of the sketching component of our
algorithm to outliers. Such robustness is important, since the
detection algorithm often produces outlier detections (see
Fig.~\ref{fig:iterations}). We used $10$ ``inlier'' images which
alone generate a good sketch with high sketch quality score. We
then added to them $n=1,...,30$ outlier images (cropped at random
from natural images). For every such $10+n$ image set we generated
a sketch, and compared it to the ground-truth. Each experiment was
repeated $15$ times. Fig.~\ref{fig:graph-sketch-with-outliers}
displays plots of sketch quality vs. percent of outliers
$n/(10+n)$. Our sketching method is relatively robust to outliers,
and performs quite well even in presence of $50\%$ outliers (as
expected due to the median operation in Eq.~(\ref{eq:median})).

In addition to sketch quality evaluation we tested the performance
of our algorithm in the scenario described in the Introduction:
given a very small number of example images, how useful is the
output of our automatical detection \& sketching algorithm for
successfully detecting that object in {\em new images}. For
$K=2,3,...,10$, we randomly sampled $K$ images out of an object
category set, applied our detection \& sketching algorithm to that
subset, and used the resulting sketch to detect the object in the
{\em remaining} $50-K$ images of that category set. We consider an
object in image $I_n$ as ``detected'' if the location of
$maxMatch(S,I_n)$ (the detected center $c_n$ of the object) falls
{\em no farther away} than 1/4 of the width or height of the
bounding-box from the ground-truth center.
We repeated each experiment $40$ times and plotted the average
detection rates in Fig.~\ref{fig:graph-detect-in-novel}. For the
Apples, Bottles, and Swans we get high detection rates (for as few
as $K=3$ example images; a scenario no WSL method can handle to
the best of our knowledge). However, our detection rates are not
as good in the Giraffe set, since the giraffes undergo strong
non-rigid deformations (they sometimes tilt their necks down, and
their legs change positions). Our current algorithm cannot handle
such strong non-rigid deformations.

%% file: cc_app.tex
\section{Introduction}

One of the fundamental tasks in unsupervised learning is clustering: grouping data points into coherent clusters.
In clustering of data points, two aspects of pair-wise affinities can be measured: (i)~{\em Attraction} (positive affinities), i.e.,  how likely are points $i$ and $j$  to be in the same cluster, and
(ii)~{\em Repulsion} (negative affinities), i.e., how likely are points $i$ and $j$ to be in different clusters.

% existing algorithms
Indeed, new approaches for clustering, recently presented by \cite{Yu2001} and \cite{Bansal2004}, suggest to combine attraction and repulsion information.
Normalized cuts was extended by \cite{Yu2001} to allow for negative affinities.
However, the resulting functional provides sub-optimal clustering results in the sense that it may lead to fragmentation of large homogeneous clusters.

The Correlation Clustering functional ({\bf CC}), proposed by \cite{Bansal2004}, tries to maximize the intra-cluster agreement (attraction)
and the inter-cluster disagreement (repulsion).
Contrary to many clustering objectives, the CC functional has an inherent
``model-selection" property allowing to {\em automatically} recover the underlying number of clusters (\cite{Demaine2003}).

Sec.~\ref{sec:cc-theory} focuses on a theoretical probabilistic interpretation of the CC functional.
The subsequent sections (Sec.~\ref{sec:cc-app-uimos} and~\ref{sec:cc-app-faces}) present two new applications.
Both these applications build upon integrating attraction and repulsion information
between large number of points, and require the robust recovery of the underlying number of clusters $k$.

This chapter focuses on the CC functional, its properties and derived applications.
We defer to chapter~\ref{cp:CC} our novel approach to CC optimization.
Experimental results presented in this chapter were produced using our algorithms, which are described in more detail in chapter~\ref{cp:CC}.
%
%-------------------------------------------------------------------------------------------------------------------%

%-------------------------------------------------------------------------------------------------------------------%
% CC - NOTATIONS
\section*{Correlation Clustering (CC) Functional}

Let $W\in\mathbb{R}^{n\times n}$ be an affinity matrix
combining attraction and repulsion: for $W_{ij}>0$ we say that $i$ and $j$ attract each other with certainty $\abs{W_{ij}}$, and for $W_{ij}<0$ we say that $i$ and $j$ repel each other with certainty $\abs{W_{ij}}$. Thus the sign of $W_{ij}$ tells us if the points attract or repel each other and the magnitude of $W_{ij}$ indicates our certainty.

Any $k$-way partition of $n$ points can be written as $U\in\left\{0,1\right\}^{n\times k}$ s.t. $U_{ic}=1$ iff point $i$ belongs to cluster $c$. $\sum_c U_{ic}=1\;\forall i$ ensure that every $i$ belongs to {\em exactly} one cluster.

% the CC explicitly
The CC functional
maximizes the intra-cluster agreement (\cite{Bansal2004}).
Given a matrix $W$\footnote{Note that $W$ may be sparse.
The ``missing" entries are simply assigned ``zero certainty" and therefore they do not affect the optimization.},
an optimal partition $U$ minimizes:
\begin{eqnarray}
\mathcal{E}_{CC}\left(U\right) &=&  - \sum_{ij} W_{ij}\sum_c U_{ic}U_{jc} \label{eq:CorrClust}\\
 & s.t. & U_{ic}\in \left\{0,1\right\} ,\; \sum_c U_{ic}=1 \nonumber
\end{eqnarray}
Note that $\sum_c U_{ic}U_{jc}$ equals 1 iff $i$ and $j$ belong to the same cluster.
For brevity, we will denote $\sum_c U_{ic}U_{jc}$ by $\uut$ from here on.

%
%-------------------------------------------------------------------------------------------------------------------%

%-------------------------------------------------------------------------------------------------------------------%
%
\section{Probabilistic Interpretation}
\label{sec:cc-theory}

This section provides a  probabilistic interpretation for the CC functional.
This interpretation allows us to provide a theoretic justification for the ``model selection" property of the CC functional.
Moreover, our analysis exposes the underlying implicit prior that this functional assumes.

We consider the following probabilistic generative model for matrix $W$.
Let $U$ be the true unobserved partition of $n$ points into clusters.
Assume that for some pairs of points $i,j$ we observe their pairwise similarity values $s_{ij}$.
These values are random realizations from either a distribution $f^+$ or $f^-$,
depending on whether points $i,j$ are in the same cluster or not. Namely,
\begin{eqnarray*}
p\left(s_{ij}=s\left|\uut=1\right.\right) &=& f^+\left(s\right) \\
p\left(s_{ij}=s\left|\uut=0\right.\right) &=& f^-\left(s\right)
\end{eqnarray*}

Assuming independency of the pairs, the likelihood of observing similarities $\left\{s_{ij}\right\}$ given a partition $U$ is then
\[
\mathcal{L}\left(\left\{s_{ij}\right\}\left|U\right.\right) = \prod_{ij} f^+\left(s_{ij}\right)^{\uut}\cdot f^-\left(s_{ij}\right)^{\left(1-\uut\right)}
\]
To infer a partition $U$ using this generative model we look at the posterior distribution:
\[
Pr\left(U\left|\left\{s_{ij}\right\}\right.\right)  \propto  \mathcal{L}\left(\left\{s_{ij}\right\}\left|U\right.\right) \cdot Pr\left(U\right)
\]
where $Pr\left(U\right)$ is a prior. Assuming {\em a uniform prior} over all partitions, i.e., $Pr\left(U\right)=const$, yields:
\[
Pr\left(U\left|\left\{s_{ij}\right\}\right.\right)  \propto \prod_{ij} f^+\left(s_{ij}\right)^{\uut}\cdot f^-\left(s_{ij}\right)^{\left(1-\uut\right)}
\]
Then, the negative logarithm of the posterior is given by
\begin{eqnarray*}
-\log Pr\left(U\left|\left\{s_{ij}\right\}\right.\right) &=& \hat{C} + \sum_{ij}\log f^+\left(s_{ij}\right)\uut \\
& & + \sum_{ij}\log f^-\left(s_{ij}\right)\left(1-\uut\right)
\end{eqnarray*}
where $\hat{C}$ is a constant not depending on $U$.

Interpreting the affinities as log odds ratios $W_{ij} = \log\left(\frac{f^+\left(s_{ij}\right)}{f^-\left(s_{ij}\right)}\right)$,
the posterior becomes
\begin{eqnarray}
-\log Pr\left(U\left|\left\{s_{ij}\right\}\right.\right) & = &
C - \sum_{ij} W_{ij}\uut
\label{eq:CorrClustPij}
\end{eqnarray}
That is, Eq.~(\ref{eq:CorrClustPij}) estimates the log-posterior of a partition $U$.
Therefore, a partition $U$ that minimizes Eq.~(\ref{eq:CorrClustPij}) is the {\bf MAP} (maximum a-posteriori) partition.
Since Eq.~(\ref{eq:CorrClust}) and Eq.~(\ref{eq:CorrClustPij}) differ only by a constant they share the same minimizer: the MAP partition.

%---------------------------------------------%
\subsection{Recovering $k$ (a.k.a. ``model selection")}

We showed that the generative model underlying the CC functional has a {\em single} model for all partitions, regardless of $k$.
Therefore, optimizing the CC functional one need not select between different generative models to decide on the optimal $k$.
Comparing partitions with different $k$ is therefore straight forward and does not require an additional ``model complexity" term (such as BIC, MDL, etc.)

\begin{figure}
\centering
\includegraphics[width=.42\linewidth]{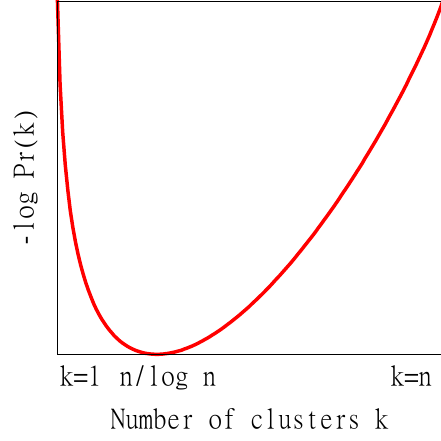}
\caption{ {\bf Prior on the number of clusters $k$:}
{\em Graph shows $-\log Pr\left(k\right)$, for uniformly distributed $U$.
The induced prior on $k$ takes a non-trivial shape:
it assigns very low probability to the trivial solutions of $k=1$ and $k=n$,
while at the same time gives preference to partitions with non-trivial $k$.
The mode of this prior is when $U$ has roughly $\frac{n}{\log n}$ clusters.
}}
\label{fig:stirling}
\end{figure}

As described in the previous section the CC functional assumes a uniform prior over all partitions.
This uniform prior on $U$ induces a prior on the number of clusters $k$,
i.e., what is the a-priori probability of $U$ having $k$ clusters: $Pr\left(k\right)=Pr\left(U\mbox{ has $k$ clusters}\right)$.
We use Stirling numbers of the second kind (\cite{Rennie1969}) to compute this induced prior on $k$.
Fig~\ref{fig:stirling} shows the non-trivial shape of this induced prior on the number of clusters $k$.

%
%-------------------------------------------------------------------------------------------------------------------%

%-------------------------------------------------------------------------------------------------------------------%
%
\section[Interactive multi-object segmentation]{Interactive multi-object segmentation {\color{red}(Patent Pending)\protect\footnotemark{}}}\footnotetext{This work was published in the 3$^{rd}$ International Conference on Information Science and Applications (ICISA), \citeyear{Bagon2012icisa}.}
\label{sec:cc-app-uimos}

%----------------------------------------------------------------------%
\begin{figure}
\centering
\begin{tabular}{p{.45\linewidth}p{.45\linewidth}}
\includegraphics[width=\linewidth]{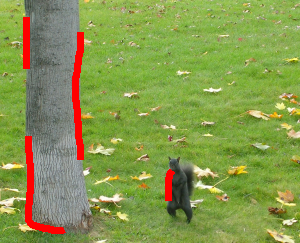}&
\includegraphics[width=\linewidth]{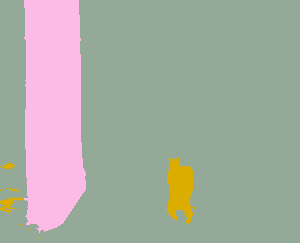}\\
\TabCenter{\linewidth}{(a) Input image and boundary scribbles (red)}&\TabCenter{\linewidth}{(b) Resulting segmentation}
\end{tabular}
\caption{\label{fig:interactive_result006}
{\bf Interactive multi-object segmentation:}
{\em (a)~The user provides only crude and partial indications to the locations of boundaries between the relevant objects in an image (red). (b)~The output of our algorithm correctly segments the image into multiple segments. Image was taken from \protect\cite{alpert2007}.}
}\vspace*{-5mm}
\end{figure}
%----------------------------------------------------------------------%
%----------------------------------------------------------------------%
\begin{figure}
\centering
\includegraphics[width=\linewidth]{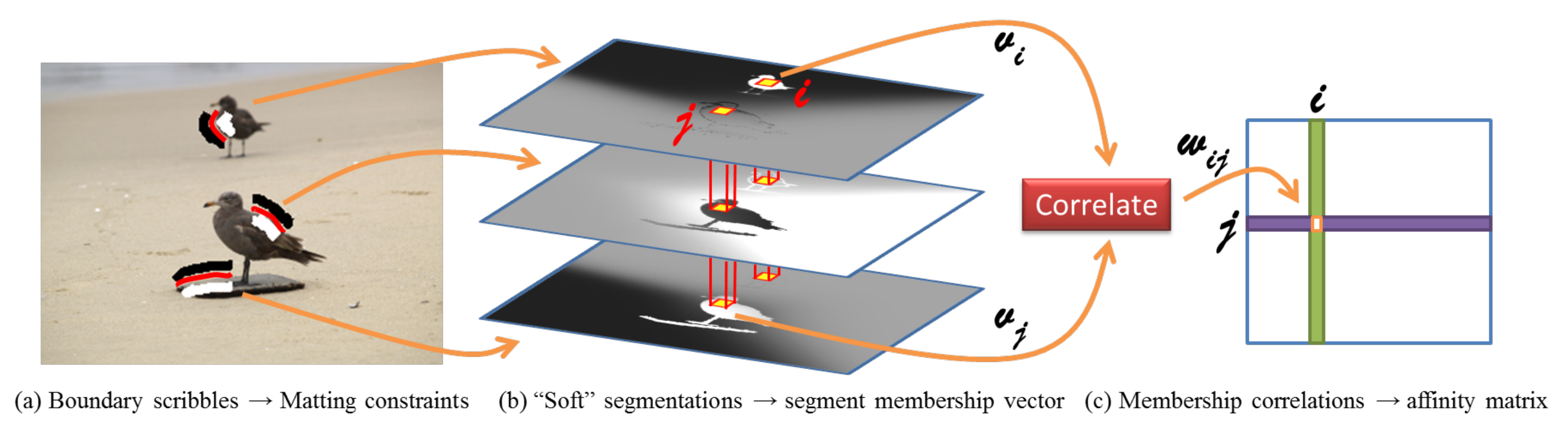}
\caption{\label{fig:neg_aff_ala_stein}
{\bf From boundary scribbles to affinity matrix:}
{\em (a)~A boundary scribble is drawn by the user (red), inducing ``figure/ground" regions on its opposite sides (black and white regions). (b)~For each scribble we use the method of \protect\cite{Levin2008} to generate a soft segmentation of the image into two segments: pixel values in the soft segmentation are in the range $\left[-1,1\right]$. Pixels far away from the scribble are assigned 0 as it is uncertain to what segment they should belong to. Each pixel $i$ is described using a segmentation membership vector $v_i$ with an entry corresponding to its assignment at each soft segmentation (red columns). (c)~A non-zero entry $w_{ij}$ in the sparse affinity matrix is the {\em correlation} between normalized vectors $v_i$ and $v_j$: $w_{ij}=v_i^Tv_j/\norm{v_i}\cdot\norm{v_j}$. We also add strong repulsion across each scribble.}
}
\vspace*{-3mm}
\end{figure}
%----------------------------------------------------------------------%

Negative affinities in image segmentation may come very naturally from boundary information: pixels on the same side of a boundary are likely to be in the same segment (attraction), while pixels on opposite sides of a boundary are likely to be in different segments (repulsion). We use this observation to design a novel approach to interactive multi-object image segmentation. Instead of using $k$ different ``strokes" for the different objects (e.g., \cite{Santner2011}), the user applies a {\em single} ``brush" to indicate parts of the boundaries between the different objects. Using these {\em sparse and incomplete} boundary hints we can correctly complete the boundaries and extract the desired number of segments. Although the user does not provide at any stage the number of objects $k$, our method is able to automatically detect the number of segments using only the {\em incomplete} boundary cues.
Fig.~\ref{fig:interactive_result006} provides an example of our novel interactive multi-object segmentation approach.

\noindent{\bf Computing affinities:} Fig.~\ref{fig:neg_aff_ala_stein} illustrates how we use sporadic user-provided boundary cues to compute a {\em sparse} affinity matrix with both positive and negative entries.
Note that this is a modification of the affinity computation presented by \cite{Stein2008}: (i)~We use the interactive boundary cues to drive the computation, rather than some boundaries computed by unsupervised technique. (ii)~We only compute a small fraction of all entries of the matrix, as opposed to the full matrix of Stein~\etal (iii)~Most importantly, we end up with both positive and negative affinities in contrast to Stein~\etal\ who use only positive affinities.

The sparse affinity matrix $W$ is very large ($\sim100k\times100k$). Existing methods for optimizing the correlation clustering functional are unable to handle this size of a matrix.
% thesis addition
Chapter~\ref{cp:CC} describes in detail our novel approach to CC optimization that enables us to optimize such large scale problems.
We applied our Swap-and-Explore algorithm (Alg.~\ref{alg:ab-swap}, described in~\ref{sec:alg}) to this problem and it provides good looking results with only several minutes of processing per image.

\begin{figure}
\newlength{\imh}
\setlength{\imh}{2cm}
\centering
\begin{tabular}{c|c|c|c|c}
\includegraphics[height=\imh]{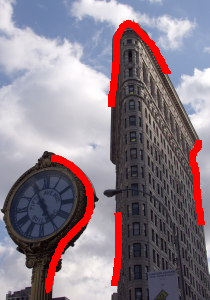}&
\includegraphics[height=\imh]{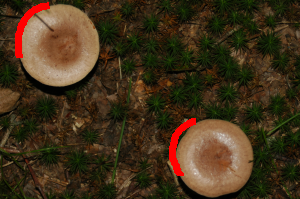}&
\includegraphics[height=\imh]{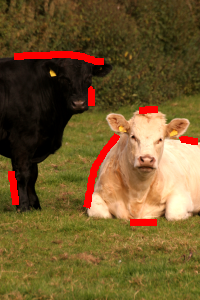}&
\includegraphics[height=\imh]{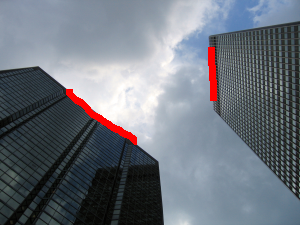}&
\includegraphics[height=\imh]{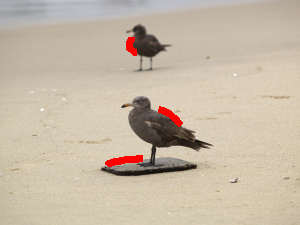}\\
\includegraphics[height=\imh]{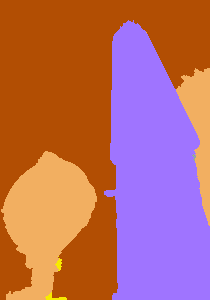}&
\includegraphics[height=\imh]{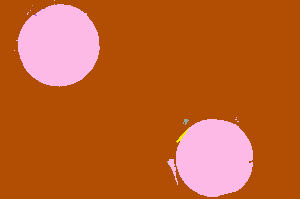}&
\includegraphics[height=\imh]{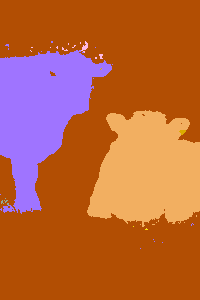}&
\includegraphics[height=\imh]{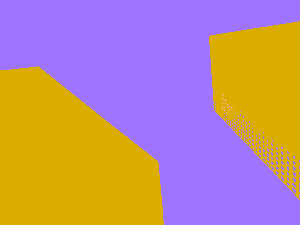}&
\includegraphics[height=\imh]{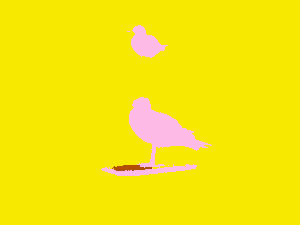}\\\hline
\end{tabular}

\begin{tabular}{c|c|c|c}
\hline
\includegraphics[height=\imh]{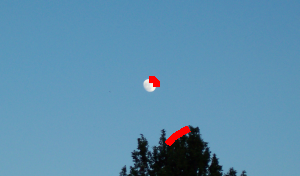}&
\includegraphics[height=\imh]{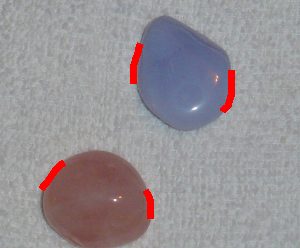}&
\includegraphics[height=\imh]{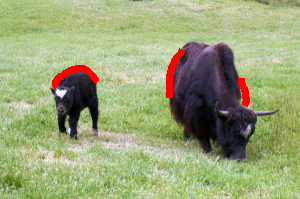}&
\includegraphics[height=\imh]{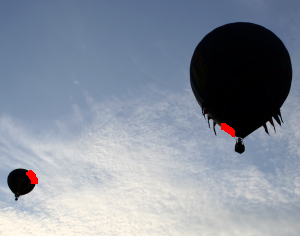}\\
\includegraphics[height=\imh]{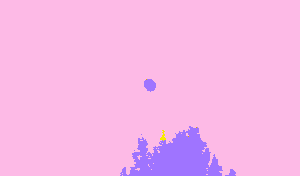}&
\includegraphics[height=\imh]{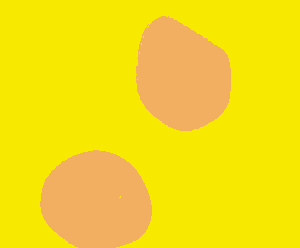}&
\includegraphics[height=\imh]{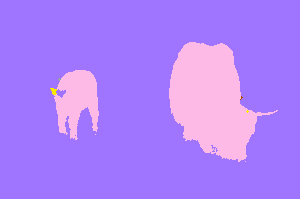}&
\includegraphics[height=\imh]{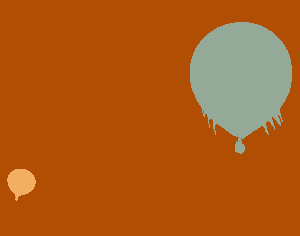}
\end{tabular}
\caption{ {\bf Interactive segmentation results. }{\em  Input image and user boundary cues (top), our result (bottom). Images were taken from \protect\cite{alpert2007}.} }
\label{fig:uimos}
\end{figure}

Fig.~\ref{fig:uimos} shows input images and user marked boundary cues used for computing the affinity matrix. Our results are shown at the bottom row.

The new interface allows the user to segment the image into several coherent segments without changing brushes and without explicitly enumerate the number of desired segments to the algorithm.

%-------------------------------------------------------------------%
\section{Clustering and face identification}
\label{sec:cc-app-faces}
%
%

% GOAL of experiment: correctly identifying the number of persons in the test set
% connection to probabilistic formulation
This application shows that detecting the underlying number of clusters $k$ can be an important task on its own.
Given a collection of face images we expect the different clusters to correspond to different persons. Identifying the different people requires not only high purity of the resulting clusters but more importantly to {\em  correctly discover the appropriate number of clusters}.
This experiment is an extension of existing work on the problem of ``same/not-same" learning. Following recent metric learning approach (e.g., \cite{Guillaumin2009,Guillaumin2010}) we learn a {\em single}
classifier that assigns a probability to each pair of faces: ``how likely is this pair to be of the same person".
Then, using this classifier, we are able to determine {\em the number of persons} and cluster the faces of {\em unseen people}. That is, given a new set of face images of several {\em unseen} people, our clustering approach is able to automatically cluster and identify how many different people are in the new set of face images of {\em never seen before} people.

% describing dataset
For this experiment we use PUT face dataset \cite{Kasinskiput2008}. The dataset consists of 9971 images of 100 people (roughly 100 images per person). Images were taken in partially controlled illumination conditions over a uniform background. The main sources of face appearance variations are changes in head pose, and facial expression.

% describe method
We use the same method as \cite{Guillaumin2009} to describe each face. SIFT descriptors are computed at fixed points on the face at multiple scales. We use the annotations provided in the dataset to generate these keypoints.
 % for computing the descriptors.
Given a training set of labeled faces $\left\{x_i,y_i\right\}_{i=1}^N$ we use a state-of-the-art method by \cite{Guillaumin2010} to learn a Mahalanobis distance $L$ and threshold $b$ such that:
\[
\hspace*{-3mm}
Pr\left(y_i=y_j\vert x_i,x_j;L,b\right)=\sigma\left(b-\left(x_i-x_j\right)^TL^TL\left(x_i-x_j\right)\right)
\]
where $\sigma(z)=(1-e^{-z})^{-1}$ is the sigmoid function.

% training and test
For each experiment we chose $k$ people for test (roughly $100\cdot k$ images), and used the images of the other $100-k$ people for training.
The learned distance is then used to compute $p_{ij}$, the probability that faces $i$ and $j$ belong to the same person, for all pairs of face images of the $k$ people in the test set.
The affinities are set to $W_{ij}=\log\frac{p_{ij}}{1-p_{ij}}$.
We apply our clustering algorithm to search for an optimal partition, and report the identified number of people $k^\prime$ and the purity of the resulting clusters.
We experimented with $k=15, 20, \ldots, 35$. For each $k$ we repeated the experiments for several different choices of $k$ different persons.

% describe comparing methods: connected components and spectral gap
In these settings all our algorithms, described in Sec.~\ref{sec:alg}, performed roughly the same in terms of recovering $k$ and the purity of the resulting clustering. However, in terms of running time adaptive-label ICM completed the task significantly faster than other methods.
We compare Swap-and-Explore (Alg.~\ref{alg:ab-swap} of Sec.~\ref{sec:alg}) to two different approaches: (i)~{\em Connected components:} Looking at the matrix of probabilities $p_{ij}$, thresholding it induces $k^\prime$ connected components. Each such component should correspond to a different person. At each experiment we tried 10 threshold values and reported the best result. (ii)~{\em Spectral gap:} Treating the probabilities matrix as a {\em positive} affinity matrix we use NCuts \cite{shi2000} to cluster the faces. For this method the number of clusters $k^\prime$ is determined according to the spectral gap: Let $\lambda_i$ be the $i^{th}$ largest eigenvalue of the normalized Laplacian matrix, the number of clusters is then $k^\prime=\arg\max_i\frac{\lambda_i}{\lambda_{i+1}}$.

\begin{figure}
\centering
\includegraphics[width=.48\linewidth]{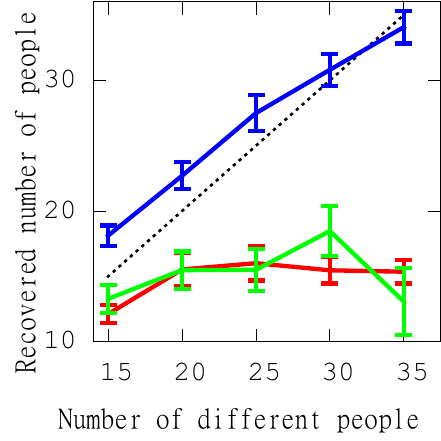}
\includegraphics[width=.48\linewidth]{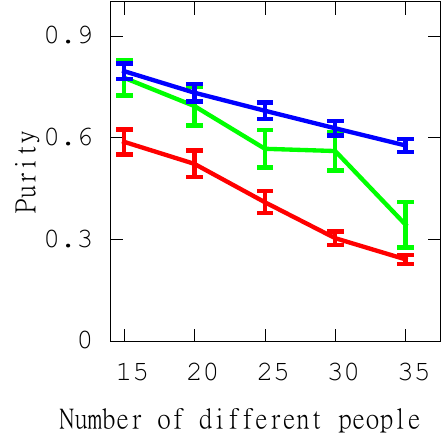}
\caption{
{\bf Face identification:} {\em Graphs showing {\color{Blue} our result (Swap)}, {\color{Green} spectral} and {\color{Red} connected components}. Left: recovered number of people ($k^\prime$) vs. number of people in the test set. Dashed line shows the true number of people. Right: purity of resulting clusters.}}\protect\vspace*{-4mm}
\label{fig:PUT_faces}
\end{figure}

% describe results in fig
Fig.~\ref{fig:PUT_faces} shows cluster purity and the number of different persons $k^\prime$ identified as a function of the actual number of people $k$ for the different methods. Our method succeeds to identify roughly the correct number of people (dashed black line) for all sizes of test sets, and maintain relatively high purity values.

%
%-------------------------------------------------------------------------------------------------------------------%

%-------------------------------------------------------------------------------------------------------------------%
%
\section{Conclusion}
\label{sec:cc-concl}

This chapter provides a generative probabilistic interpretation for the Correlation Clustering functional,
justifying its intrinsic ``model selection" capability.
Using a generative probabilistic formulation allows
for a better understanding of the functional,
underlying assumptions it makes, including the prior it imposes on the solution.

% detecting k
Optimizing large scale CC and robustly recovering the underlying number of clusters allows us to propose new applications: interactive multi-label image segmentation and unsupervised face identification.

% new applications

%
%-------------------------------------------------------------------------------------------------------------------%

%% file: lighting.tex
\section{Introduction}

Moving objects change their appearance as they change their orientation with respect to a light source. Such changes make it difficult to identify corresponding points across images, and as a result complicate the task of motion recovery and subsequently also of 3D reconstruction. Existing approaches to motion analysis largely assume either brightness constancy, or rely on extracting distinctive features. These approaches successfully handle both polygonal shapes and objects with noticeable surface markings, but they are less suitable for handling objects with smooth shapes and gradual albedo variations.

In this paper we propose a framework for reconstructing the shape of such objects that explicitly takes lighting effects into account. Our work is based on the observation that the changes in both the location and intensity of a moving object across images are coupled by its motion. Therefore, these two sources of information can be combined to constrain the matching of points across images, allowing us to achieve a veridical reconstruction. A number of previous studies (see Section~\ref{sec:prev-work}) used this observation to propose algorithms for shape recovery either in multi-view settings or in restricted settings (single directional light source and small motion,~\cite{BasriFrolova}) when only two views are available. In this study we propose a method that can work with two or more images and that can deal with general lighting conditions and fairly large motion.

We consider the case of an object that moves rigidly with respect to the camera and the light source. We assume that the object is lambertian, and that both the lighting conditions and the motion parameters are known. We present two algorithms. First we address the problem of shape reconstruction from two views when lighting is composed of a single directional source, and derive a partial differential equation (PDE) that can be solved to recover the shape of the object using continuation (characteristic curve). We further show how we can derive boundary conditions for this method. This formulation extends the work of~\cite{BasriFrolova} to objects undergoing a large motion.

Our second algorithm uses the PDE formulation to construct a cost function on a graph, representing the domain to be reconstructed. The cost function takes the form of a Markov Random Field (MRF). We then use off-the-shelf algorithms to solve for the sought shape. The MRF formulation offers several advantages over previous work. (1) It can work with fairly general reflectance functions. (2) It can be applied to pairs of images as well as to sequences of three or more images. (3) It is more robust to errors than continuation solutions, which can accumulate errors in integration. Finally, (4) prior information can be incorporated; in particular, the method can use boundary conditions, but as we show in our experiments boundary conditions are not essential. Experiments with real smooth objects demonstrate the utility of our formulation.

The paper is divided as follows. Section~\ref{sec:prev-work} provides a brief summary of related work. Section~\ref{sec:continuation} defines the reconstruction problem and derives a continuation solution in the case of large motion. Section~\ref{sec:MRF} casts the problem as an MRF optimization and discusses its solution by discrete techniques. Section~\ref{sec:bc} explains how we compute boundary conditions, and Section~\ref{sec:light-experiments} shows the results of our experiments.

\section{Previous work}  \label{sec:prev-work}

The majority of 3D reconstruction techniques use either motion or shading techniques, but rarely combine both cues. Shape from shading~\cite{ShapeFromShadingBOOK} and photometric stereo~\cite{Woodham_1980} utilize shading cues to reconstruct objects from single or multiple images of static objects. Motion is often handled with the assumption of brightness constancy~\cite{Horn:optical-flow,Lucas1981} or by matching sparse feature points (e.g.,~\cite{Crandall}). Some authors proposed to reconstruct static objects by combining shading with stereo cues~\cite{Cryer_Tsai_Shah,FuaLeclerc_StereoShading_IJCV95}. Note that in a stereo setup lambertian objects retain their intensities across images. Another set of studies seek to generalize the brightness constancy assumption to account for local lighting variations~\cite{BlackFleet_ChangesInAppearance_CVIU00,Negahdaripour_OpticalFlow_PAMI1998,HausseckerFleet_OpticalFlow_PAMI2001}.

Studies that use motion and shading cues simultaneously~\cite{CarceroniKutulakos_SurfelSampling_ICCV01,Jin&etal2008,Maki_Watanabe_Wiles,Mukawa_ShapeReflectionIllumination_ICCV90,Simakov_Frolova_Basri,Weber_Blake_Cipolla,Zhang&etal,Joshi_Kriegman_2007,ChenChenHung,Moses_Shimshoni_2006}. typically require a multi-frame setting (typically at least 4 images). These studies also estimate the light source direction along with the shape of the object. An exception is~\cite{BasriFrolova}, which requires only two images, but assumes small motion.
%Two recent studies that combine motion and shading cues used the method of continuation to recover shape. Of these methods  \cite{Moses_Shimshoni_2006} requires at least three images, while Both methods handle lambertian objects illuminated by a single directional source.
Our work improves over these methods by proposing reconstruction methods that can handle large motion and can work both with image pairs and with larger sequences of images. In addition, our second algorithm allows for general lighting settings, which include point and extended light sources, through the use of spherical harmonic representations~(\cite{BasriJacobsLinSubsp,RamamoorthiHanrahanDetermIllum}).

\section{Problem definition and solution by continuation}  \label{sec:continuation}

We consider a pair of images taken by a stationary camera. The images depict a lambertian object moving rigidly in space and illuminated by constant lighting. Denote these images by $I(x,y)$ and $J(x,y)$. Let ${\bf P}=(x,y,z(x,y))$ be a point on the object's surface described in the coordinate frame of $I$, and let ${\bf p}$ and ${\bf q}$ denote its projections onto $I$ and $J$. Assuming a weak perspective projection ${\bf p}=(x,y)$, and ${\bf q}=s R {\bf P} + {\bf t}$ in $J$, where the scale $s>0$, rotation $R$ (represented by a $2 \times 3$ matrix with orthonormal rows) and translation ${\bf t} \in \Re^2$ describe the (known) motion of the object. In general, since the motion is assumed to be known the location of ${\bf q}$ depends on the unknown depth value of ${\bf P}$, $z$. We therefore often emphasize this dependence by writing ${\bf q}(z)$.

Denote the normal to the surface at ${\bf p}$ in $I$ by $\hat {\bf n} \in S^2$,
\[ \hat {\bf n} = \frac{1}{\sqrt{z_x + z_y + 1}} (-z_x, -z_y, 1), \]
where $z_x=\partial z/\partial x$, $z_y=\partial z/\partial y$, and denote the albedo at $p$ by $\rho \in \Re$. Thus, the normal at ${\bf q}(z)$ is given by $R\hat {\bf n}$. We assume that the intensity of ${\bf P}$ in the two images can be expressed as $I({\bf p}) = \rho r(\hat {\bf n})$ and $J({\bf q}) = \rho r(R\hat {\bf n})$, where $r:S^2 \rightarrow \Re$, commonly referred to as the {\em reflectance function}, is a (known) function of the surface normal. This expression can be used to model a variety of materials. In particular, it can model lambertian objects illuminated by a directional (``point'') source by setting $r(\hat {\bf n})=\max({\bf l}^T \hat {\bf n},0)$. Here the direction of the point source is given by ${\bf l}/\|{\bf l}\|$ and its intensity by $\|{\bf l}\|$. Likewise, under more general lighting conditions (that may include multiple point and extended sources) $r(\hat {\bf n})$ can be expressed as an inner product between a vector of coefficients ${\bf b} \in \Re^d$ and the spherical harmonic functions evaluated at $\hat {\bf n}$, denoted $Y(\hat {\bf n}) \in \Re^d$. $d$ is set to either 4 or 9 if respectively the first or second order harmonic approximation is used, see~\cite{BasriJacobsLinSubsp} for details of this model.

Under these assumptions we can eliminate the albedo $\rho$ by taking the ratio between the intensities of corresponding points, namely
\begin{equation}  \label{eq:main}
\frac{J({\bf q}(z))}{I({\bf p})} = \frac{r(R\hat {\bf n})}{r(\hat {\bf n})}.
\end{equation}
This PDE captures the relation between the shape of the moving object, its motion and the environment lighting. The shape of the object is captured both implicitly by the correspondence between ${\bf p}$ and ${\bf q}(z)$ (in other words, this equation is implicit in $z$), and explicitly by the surface normal (i.e., the partial derivatives of $z$).

For a lambertian object illuminated by a single directional light source ${\bf l}$ Eq.~\eqref{eq:main} can be written as follows. Let ${\bf n}=(-z_x,-z_y,1)^T$ (so that $\hat{\bf n}={\bf n}/\|{\bf n}\|$) then
\begin{equation}  \label{eq:point-source}
\frac{J({\bf q}(z))}{I({\bf p})} = \frac{{\bf l}^T R {\bf n}}{{\bf l}^T {\bf n}}.
\end{equation}
Note that in this equation the nonlinear term $\|{\bf n}\|$ cancels out. Rearranging this equation we obtain
\begin{equation}  \label{eq:main-point-source}
{\bf a}^T {\bf n} = 0,
\end{equation}
where ${\bf a}(x,y,z) = J({\bf q}(z)) {\bf l} - I({\bf p}) R^T {\bf l}$.
The equation is quasi-linear (i.e., linear in the derivatives $z_x$ and $z_y$), although implicit in $z$

\cite{BasriFrolova} made this equation explicit in $z$ by using a Taylor approximation for $J({\bf q}(z))$ and used the obtained expression to recover the shape of Lambertian objects illuminated by a single directional source. As their method approximates~\eqref{eq:main} to first order, it can handle only objects undergoing very small motion. Specifically, assuming $I$ and $J$ are rectified such that their epipolar lines are horizontal, they showed that~\eqref{eq:main} can be written as an explicit, quasilinear PDE
\begin{equation}  \label{eq:1st-order}
a z_x + b z_y = c,
\end{equation}
where
\begin{eqnarray*} \vspace{-2mm}
a(x,y,z) &=& l_1 (I_\theta - z J_x) - l_3 I\\
b(x,y,z) &=& l_2 (I_\theta - z J_x)\\
c(x,y,z) &=& l_3 (I_\theta - z J_x) + l_1 I.
\end{eqnarray*}
In this equation $J_x=\partial J/\partial x$,  $I_\theta=(J-I)/\theta$ and $\theta$ denotes the angle of rotation about the vertical axis. They further described a solution to~\eqref{eq:1st-order} using the method of continuation (characteristic curves) and showed a method to extract Dirichlet boundary conditions along the bounding contour of an object.

Our first contribution is to note that Basri and Frolova's algorithm can be extended also to handle large motion. This is because~\eqref{eq:main-point-source} is quasi-linear even without the Taylor approximation, and so it too can be solved by the method of continuation. Note that, due to~\eqref{eq:main-point-source}, ${\bf a}$ lies on the tangent to the surface $z(x,y)$ at ${\bf p}$. Therefore, any curve $\gamma(t) \subset \Re^3$ whose tangent at every $t \in [0,1]$ is given by ${\bf a}(\gamma(t))$ is characteristic to~\eqref{eq:main-point-source}, and if $\gamma(0)$ happens to lie on $z(x,y)$ the entire curve will lie on $z(x,y)$. To recover $z(x,y)$ the continuation method traces a family of such curves $\{\gamma(t)\}$ by integrating the vectors ${\bf a}(\gamma(t))$ starting from a 1D set of 3D points given as Dirichlet boundary conditions. Unfortunately, as~\eqref{eq:main-point-source} is implicit in $z$, extracting boundary conditions can be complicated; we suggest a method to do this in Section~\ref{sec:bc}.

\begin{figure}
\begin{center}
\begin{tabular}{|c|c|c|}
\hline
\includegraphics[width=0.25\linewidth]{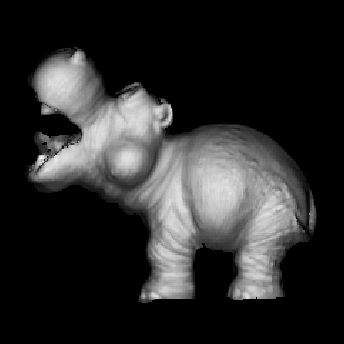} &
\includegraphics[width=0.25\linewidth]{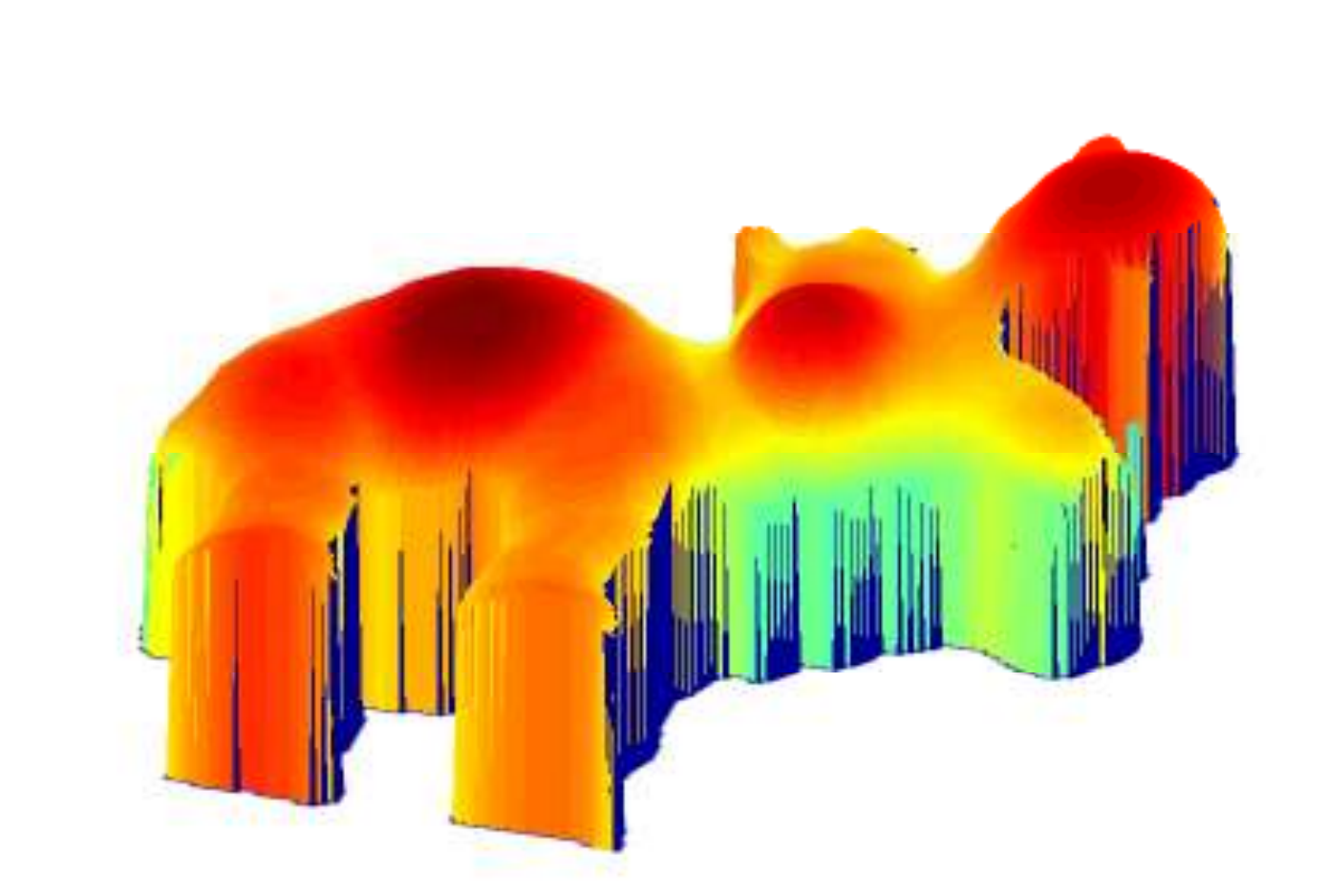} &
\includegraphics[width=0.25\linewidth]{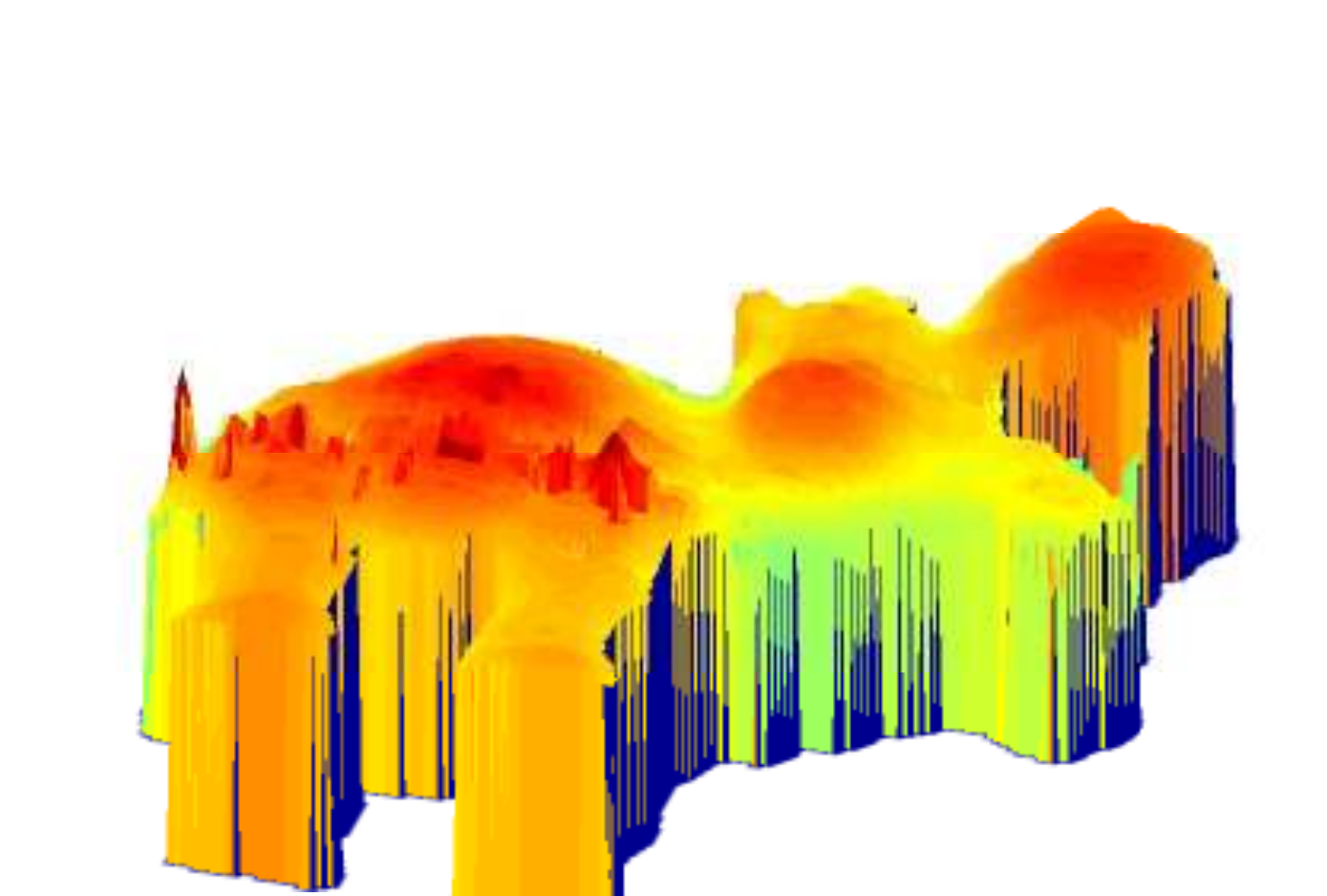} \\

\includegraphics[width=0.25\linewidth]{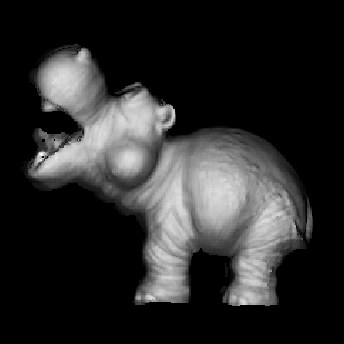} &
\includegraphics[width=0.25\linewidth]{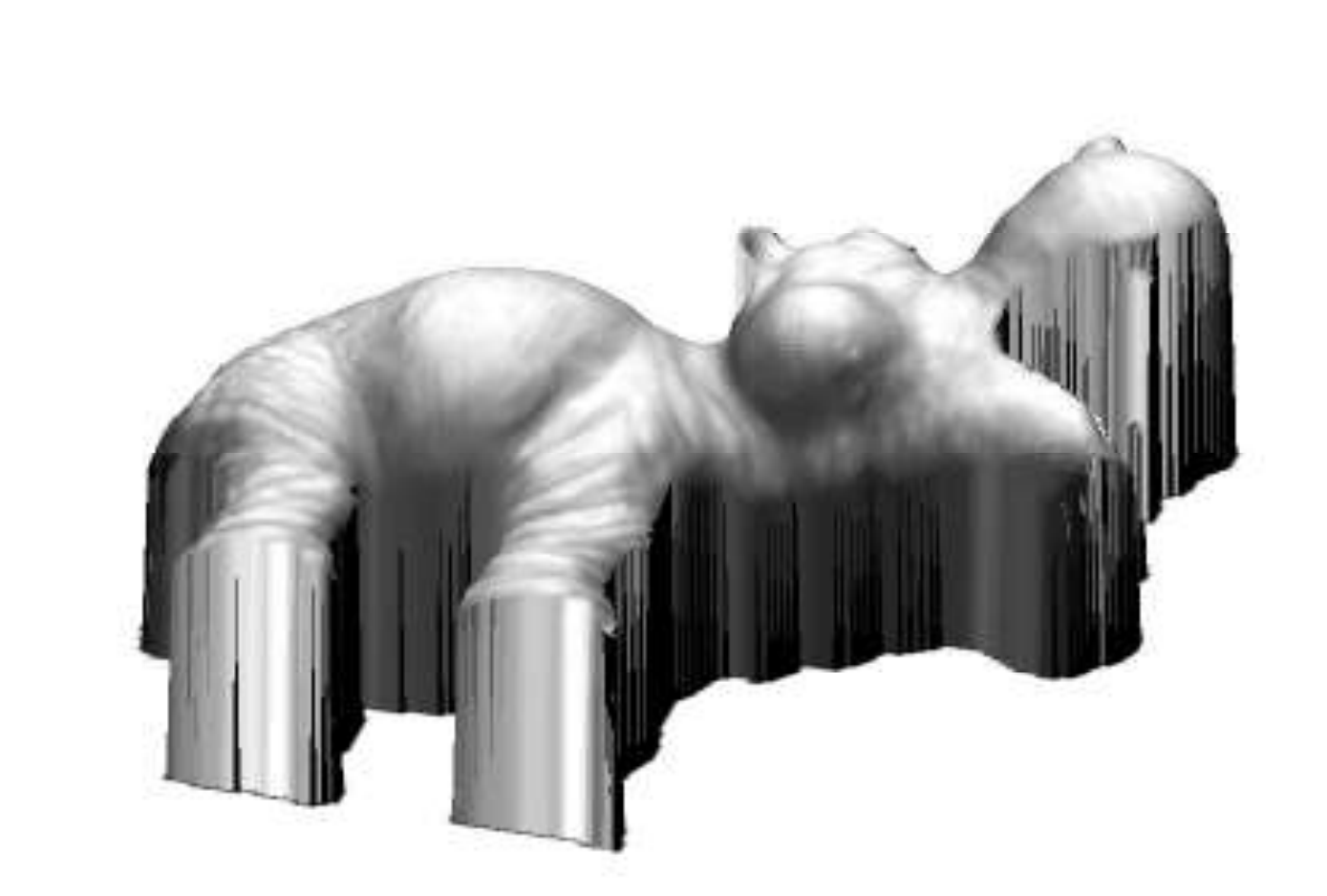}&
\includegraphics[width=0.25\linewidth]{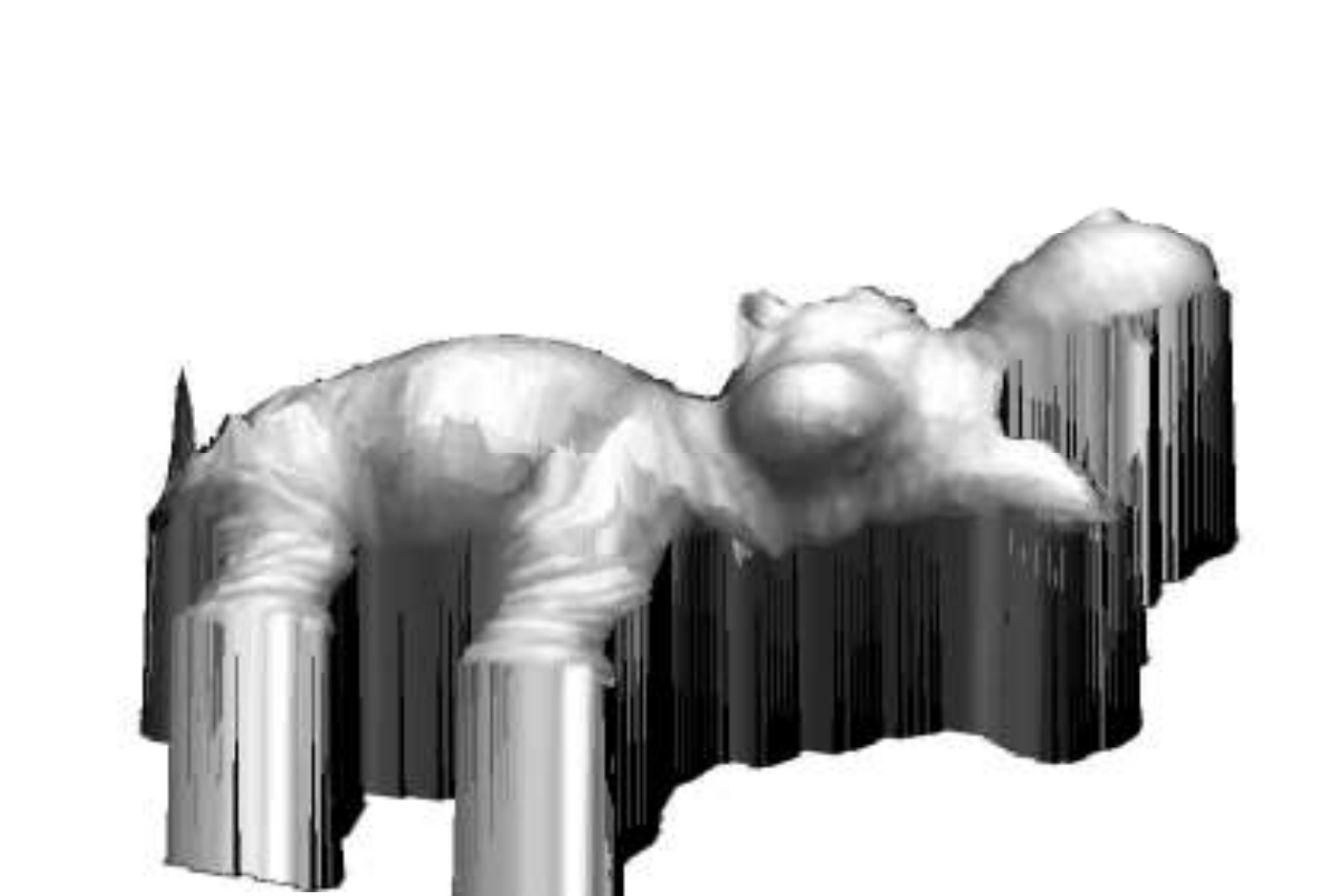} \\

\hline
\end{tabular}
\end{center}
\caption{Reconstructions from rendered images of a toy hippo, using continuation. The left column shows the original images (rotation of $4^\circ$), the middle column shows the 3D model and the right column shows the reconstructed surface. The top row shows the surface (colormap represents depth values) and the bottom row shows the surface painted with intensity values.}
\label{fig:cont}
\end{figure}

Finally, we note that the characteristic curves traced with this procedure are all plane curves that lie in parallel planes. This can be readily seen by noticing that ${\bf a}$ is a linear combination of ${\bf l}$ and $R{\bf l}$, and so ${\bf a}^T ({\bf l} \times R{\bf l}) = 0$. in general, unless ${\bf l}$ points in the direction of the axis of rotation (denoted ${\bf u}$) these parallel planes are perpendicular to ${\bf u}$ and are {\em independent} of ${\bf l}$. Note however that generally the planes that contain the characteristics do not coincide with the epipolar planes except when the axis of rotation is parallel to the image plane. In the case that ${\bf l}$ coincides with ${\bf u}$,~\eqref{eq:point-source} becomes degenerate since its right hand side (substituting ${\bf u}$ for ${\bf l}$) is 1. Consequently, $J({\bf q}(z))=I{\bf p}$, and methods that assume constant brightness can be applied. Figure~\ref{fig:cont} shows a reconstruction obtained with our suggested method for continuation. A toy model was rendered using point source light and rotated by angle of about $4^\circ$. The continuation method was applied with exact boundary condition values.

\section{Discrete optimization}  \label{sec:MRF}

The continuation method suffers from several shortcomings. First, the method accumulates errors in the integration. Moreover,~\eqref{eq:main} is quasi-linear when the lighting is composed of a single directional light source, but is non-linear when more realistic lighting models are used. Continuation can in principle be applied also for non-linear equations, but it is then significantly less robust. In addition, the method relies on boundary conditions that are difficult to compute, and at the same time does not allow for the inclusion of other prior information. Finally, it does not provide a way to integrate information from more than two images when more images of the object are available.

To overcome all of these shortcomings we cast the problem as an MRF optimization and solve it using discrete optimization techniques. Our objective is to find a valid surface $z(x,y)$ that satisfies~\eqref{eq:main}. We therefore define a cost function composed of unary and binary terms. Our MRF is defined over a grid overlaid on the first image $I(x,y)$. Each grid point $(x,y)$ is associated with a state vector determining the parameters of the surface element ({\em surfel}) observed at that pixel. These parameters include both the depth value and the surface normal, i.e., $(z,z_x,z_y)$, of the surfel. Note that such state vectors cannot represent points along the silhouette contour as the derivatives of $z$ at these points diverge. Our cost function combines unary and binary terms over the grid points. The unary term is a function of the residual of~\eqref{eq:main}. Specifically, let
\begin{equation}  \label{eq:T}
T = I({\bf p}) r(R \hat {\bf n}) - J({\bf q}) r(\hat {\bf n}).
\end{equation}
We define the unary term as
\begin{equation}  \label{eq:light-unary}
E^u_{\bf p} = \exp(\alpha T^2)-1,
\end{equation}
where $\alpha$ is a constant ($\alpha=8$ in our experiments). The binary term is set to prefer integrable surfaces. For ${\bf p}=(x,y)$ we use
\begin{equation}  \label{eq:binary}
E^b_{\bf p} = (z(x,y) + z_x(x,y) - z(x+1,y))^2 + (z(x,y) + z_y(x,y) - z(x,y+1))^2.
\end{equation}
Our final energy function is given by
\begin{equation}  \label{eq:energy}
E = \sum_{{\bf p} \in \Omega_I} E^u_{\bf p} + \lambda E^b_{\bf p},
\end{equation}
where $\Omega_I$ denotes the silhouette of the object in $I$ and $\lambda$ is constant. (We used values around 0.001.)

In general, when two images are used for reconstruction we need to supply boundary conditions in the form of $z(x,y)$ along a 1D curve $\gamma_0$. In Section~\ref{sec:bc} we describe a technique to estimate the depths near the bounding contours of the object. This procedure provides both the depth values and the normals along a 1D contour in $\Omega_I$. Given these boundary conditions we simply modify the unary cost along this contour to vanish in states $(z,z_x,z_y)$ that coincide with the depth values and normals predicted by the boundary conditions. As the extraction of boundary conditions can be noisy we then let the optimization modify these depth values. When the optimization is applied to three or more images the additional images further constrain the solution, and so boundary conditions are not necessary.

The binary term is clearly non-submodular. We therefore optimize~\eqref{eq:energy} by a sequence of alpha expansion iterations implemented with the QBPO algorithm~\cite{Rother2007}. Each iteration solves a binary decision problem in which each variable is allowed to either change its current state to a fixed state $(z^\alpha,z^\alpha_x,z^\alpha_y)$ or retain its current state. For submodular energies such an iteration is implemented using the st-mincut algorithm~\cite{KolmogorovZabih,Boykov2001}. The QPBO algorithm extends the st-mincut algorithm to handle negative edge capacities arising in the case of non-submodular pairwise terms. The algorithm constructs a graph with two complimentary vertices for each variable, explicitly representing its two possible binary states. By using a redundant variable representation, the resulting graph can be constructed to have positive capacities in all of its edges, and therefore the st-mincut algorithm can be applied to it. The following rule is used to label the variables at the end of the st-mincut step. Each variable for which its two complimentary vertices fall in different sides of the cut is labeled according to the assignment induced by the cut. Variables for which both complimentary vertices fall in the same side of the cut remain unlabeled. As minimizing non-submodular energies generally is NP-hard, QBPO is not guaranteed to label all the variables after each st-mincut step, and some of the variables may remain unlabeled. These unlabeled variables retain their original label from the previous iteration. This procedure is repeated for every possible choice of a label $(z^\alpha,z^\alpha_x,z^\alpha_y)$. The entire batch of iterations is then repeated until convergence. The optimization is initialized by setting the initial state $(z,z_x,z_y)$ for every point to the state that minimizes the unary term~\eqref{eq:light-unary}.

The discrete optimization solver is followed by a continuous quadratic optimization, in order to relax the quantized labeling (discrete values) of the surface. We solve the following quadratic optimization functional, maintaining the integrability constraint while remaining close to the quantized solution:
\begin{equation}
%Ec{z} =  \sum_{{\bf p} \in \Omega_I} (z(x,y) + z_x(x,y) - z(x+1,y))^2 + (z(x,y) + z_y(x,y) - z(x,y+1))^2 + \mu (z(x,y) - \tilde{z}(x,y))^2,
Ec{z} =  \sum_{{\bf p} \in \Omega_I} (z(x,y) - \tilde{z}(x,y))^2 + \mu  E^b_{\bf p}
\end{equation}
where $\tilde{z}(x,y)$ is the solution obtained by the discrete optimization, and%$\mu = 0.2$
$\mu = 5$ in our experiments.

Finally, note that this procedure can readily be applied also when the lighting changes between images, by using different reflectance functions for $I$ and $J$ in~\eqref{eq:T}.

\section{Boundary conditions}  \label{sec:bc}

Using the differential relation~\eqref{eq:main} to reconstruct a shape from two images requires boundary conditions in the form of depth values -- one depth value is required along each characteristic curve. This is essential for the continuation method and can be useful for the discrete optimization framework to further constrain the solution. Obtaining depth values can be complicated, as they require finding the correspondence between pixels on a smooth objects. This was bypassed in~\cite{BasriFrolova} since under a small motion and with the Taylor expansion~\eqref{eq:1st-order} becomes explicit in $z$, allowing one to compute the depths at the silhouette contours of the object from the intensities along the contour. Below we describe a method that computes depth values near the silhouette contours when $z$ is implicit.

Denote the object silhouette in $I$ and $J$ respectively by $\Omega_I$ and $\Omega_J$. We further denote by $\partial_v \Omega_I$ the portion of the silhouette contour in $I$ that remains visible in $J$ and likewise by $\partial_v \Omega_J$ the portion of the silhouette contour in $J$ that remains visible in $I$. Let ${\bf P}$ be a rim point in the coordinate frame of $I$ such that its projection onto $I$, denoted ${\bf p}$, lies on $\partial_v \Omega_I$. Since ${\bf p}$ lies on the bounding contour of $\Omega_I$ the surface normal at ${\bf P}$ should be parallel to the normal to the occluding contour at ${\bf p}$, and so it can be derived from the image. Let $\hat {\bf n}({\bf p})=(\cos\beta,\sin\beta,0)^T$. Plugging this normal in~\eqref{eq:main} gives us a scalar for the right hand side, and using the known value $I({\bf p})$  we compute the anticipated intensity at the corresponding point $J({\bf q})$. We can then use this intensity to search along the epipolar line on $J$ to determine ${\bf q}$. This for itself can be complicated, since there may be multiple points in along the epipolar line in $J$ with the anticipated intensity, but in addition we face two problems: (1) the intensity values along the bounding contour are unreliable, and (2) neither the continuation nor our discrete optimization scheme can use normals that are parallel to the image plane.
% Thus for each ${{\bf p_i} \in \partial_v \Omega_I}$, we have a corresponding point ${\bf q_i^0} \in J$

To overcome these problems we move from ${\bf p}$ one pixel in the direction of the normal into a point ${\bf p}'$ inside $\Omega_I$. We assume that the surface normal at ${\bf p}'$ can be expressed by $\hat {\bf n}({\bf p}') = (\cos\beta\cos\psi,\sin\beta\cos\psi,\sin\psi)^T$ for some unknown angle $\psi$. Our aim then is to find a point ${\bf q}'$ in $J$ along the epipolar line of ${\bf p}'$, denoted  $e({\bf p}'_i)$, that satisfies~\eqref{eq:main} with this normal. This equation has two unknowns, the corresponding point ${\bf q}'$ in Image $J$  and $\psi$. We therefore regularize the problem by requiring the curve near $\partial_v \Omega_I$ to be mapped to a continuous curve in $J$.

We define the following optimization function.
\begin{eqnarray*}
\min_{\{{\bf q}'_i,\psi_i \}} \sum_{{\bf p}_i \in \partial_v \Omega_I} T({\bf p}'_i,{\bf q}'_i,\psi_i)^2 + \mu_1 \|{\bf q}'_i -{\bf q}_i^0\|^2 + \mu_2 \mathrm{dist}(e({\bf p}'_i),{\bf q}'_i)^2 +  \\
\mu_3 \|{\bf q}'_i-{\bf q}_{i-1}'\|^2 + \mu_4 (\cos\psi_i - \cos\psi_{i-1})^2,
\end{eqnarray*}
where $T(.)$ denotes the residual defined in~\eqref{eq:T} for a pair of points ${\bf p}'_i$ and ${\bf q}'_i$ and normal $\hat {\bf n}({\bf p}')$ parameterized by $\psi$, and $\mathrm{dist}(e({\bf p}'_i),{\bf q}'_i)$ denotes the Euclidean distance between the point ${\bf q}'_i$ and the epipolar line $e({\bf p}'_i)$. For ${\bf q}_i^0$ we use the point along $e({\bf p}_i)$ closes to the boundary (at the same side as ${\bf p}$) that has the intensity anticipated from~\eqref{eq:main}. This heuristic is useful with reasonable rotations. To solve this minimization we substitute for $J({\bf q}'_i)$ in $T({\bf q}'_i,\psi_i)$~\eqref{eq:T} its Taylor expansion around ${\bf q}_i^0$. The resulting system of equations is non-linear and we solve it by alternate minimization.

\begin{figure}[h]
\begin{center}
\includegraphics[width=0.45\linewidth]{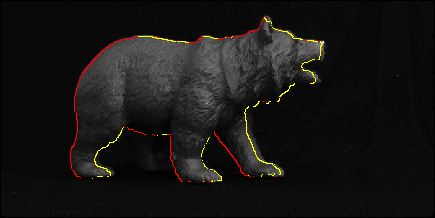}~
\includegraphics[width=0.45\linewidth]{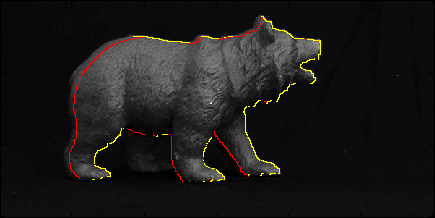}
\end{center}
\caption{The figure shows boundary conditions computed for two images of a bear toy. The red curve segments depict boundary points ${\bf p}'$ in the left image and their selected corresponding points in the right image. The yellow curve segments depict correspondences computed in the opposite direction.}
\label{fig:bc}
\end{figure}

We repeat this procedure for $\partial_v \Omega_J$ to obtain boundary conditions from both sides of the object's silhouette. Figure~\ref{fig:bc} shows an example for boundary conditions obtained with our method.

\section{Experiments}  \label{sec:light-experiments}

We tested our algorithm on two sets of real images and compared them to reconstructions with laser scans. Each image is taken with dark background to allow segmentation of the foreground object. In each image we estimated the motion parameters by manually marking points on the image and the 3D mesh object. As the objects are smooth and contain almost no clear surface markings our motion estimates are not perfect. Subsequently, using the mesh we estimated the environment lighting conditions by calculating the 4 coefficients of the first order harmonic representation of the lighting, representing the ambient lighting and a directional source. The obtained motion and lighting parameters were used as input to our software. The figures below show reconstructions obtained with our optimization algorithm (Section~\ref{sec:MRF}). We measure the quality of the matches obtained using the RMSE error in pixels. The figures show input images of a bear and elephant toys, reconstructions from pairs of images with and without boundary conditions (Section~\ref{sec:bc}), and reconstructions from multiple images. Our experiments cover a range of rotations between $4^\circ$ to about $21^\circ$. It can be seen that for the tested our reconstructions achieve low RMSE values. Interestingly, there is no noticeable advantage to using boundary conditions.

\section{Conclusion}

We described methods for reconstructing the shape of a moving object that accounts for intensities changes due to a change in orientation with respect to the light sources. In particular, we presented a continuation method and a method based on discrete optimization. Our experiments demonstrates the utility of our methods. The setting requires knowledge of the motion and lighting parameters. We plan in the future to seek ways to relax those limiting assumptions.

\begin{figure}
\begin{center}
 \includegraphics[width=0.60\linewidth]{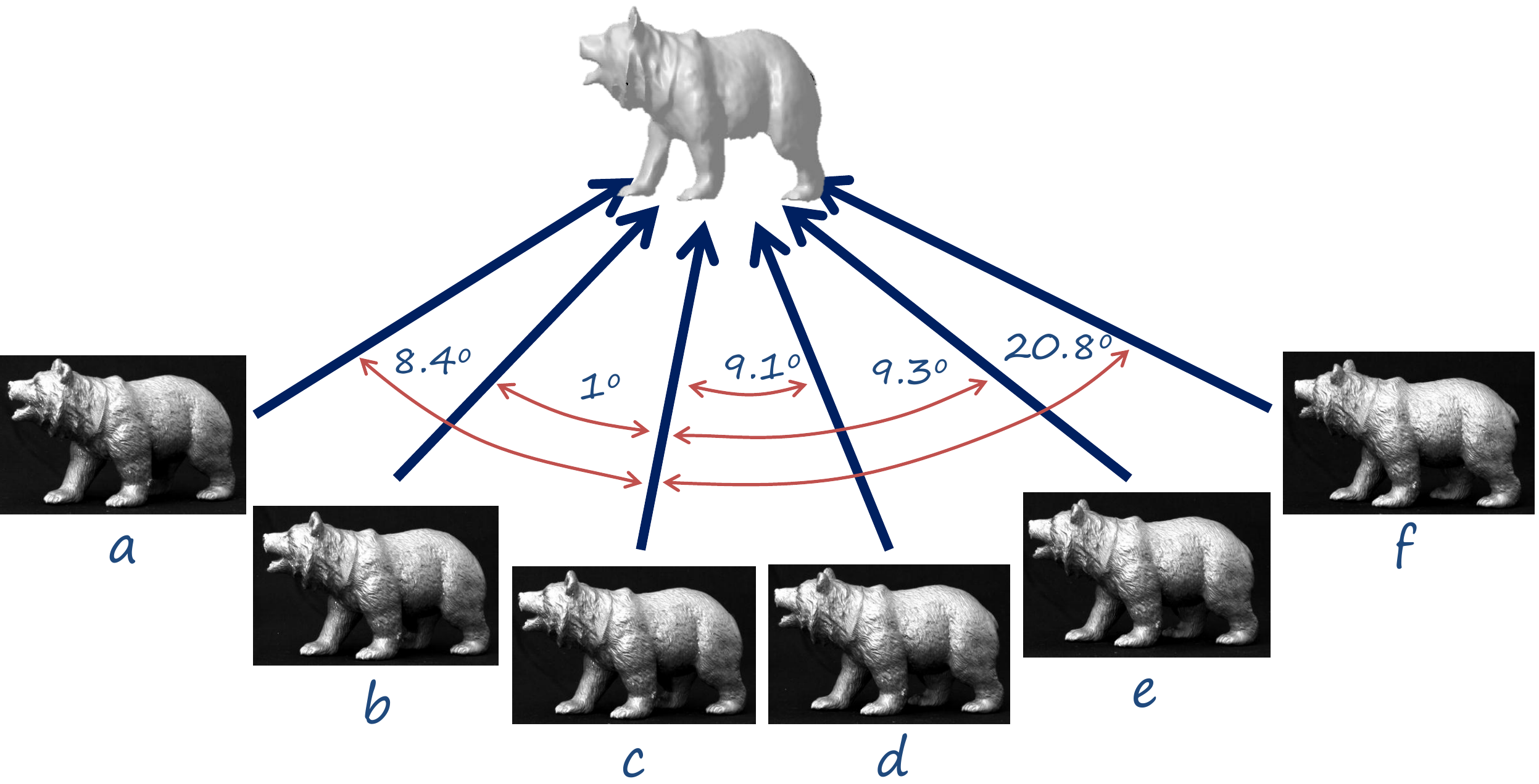}
\end{center}
\caption{Input images of a bear toy. A laser scan is shown at the center, and relative viewing angles are provided.}
\label{fig:bear}
\end{figure}

\begin{figure}
\begin{tabular}{|c|c|}
\hline
 \includegraphics[width=0.15\linewidth]{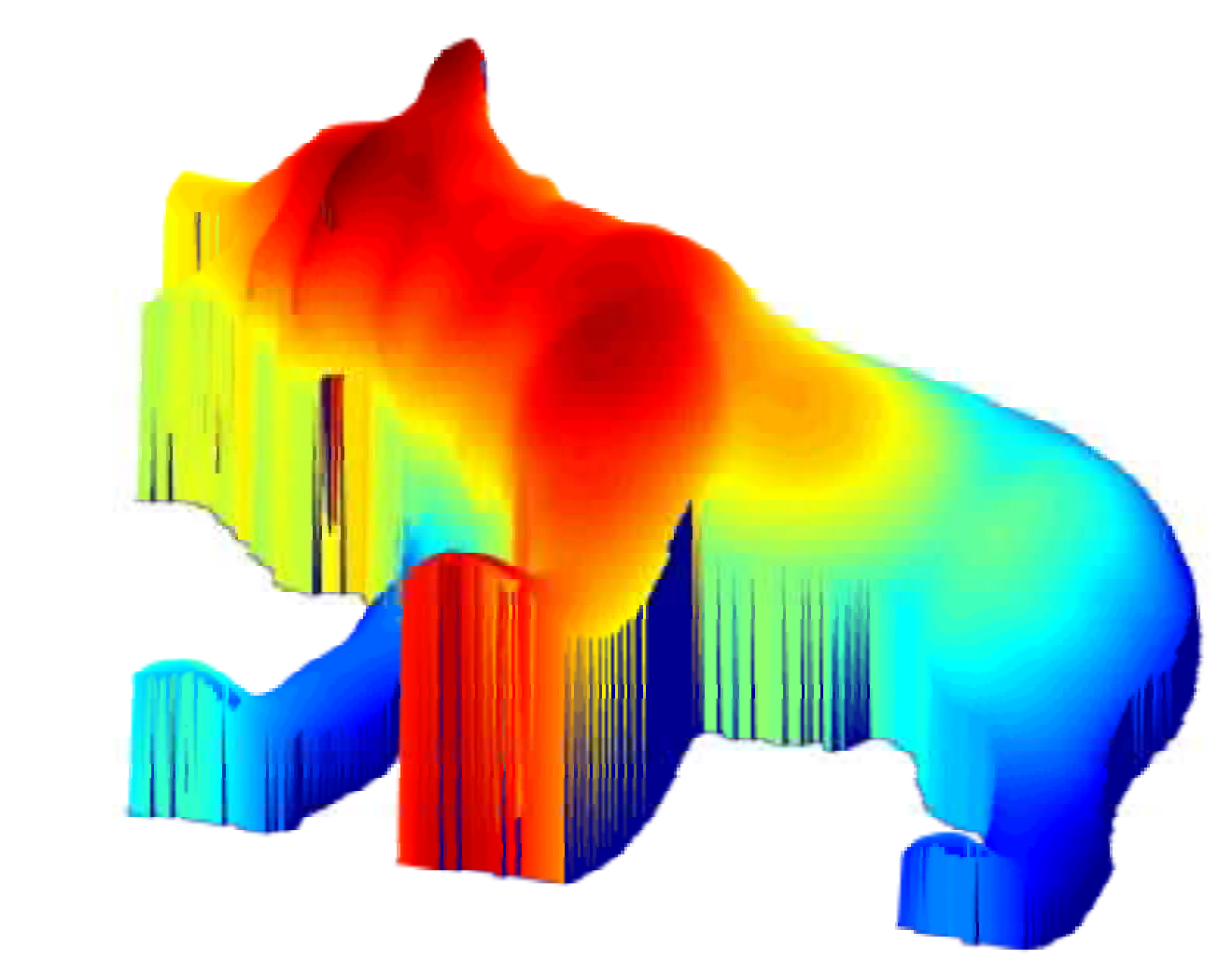}
 \includegraphics[width=0.15\linewidth]{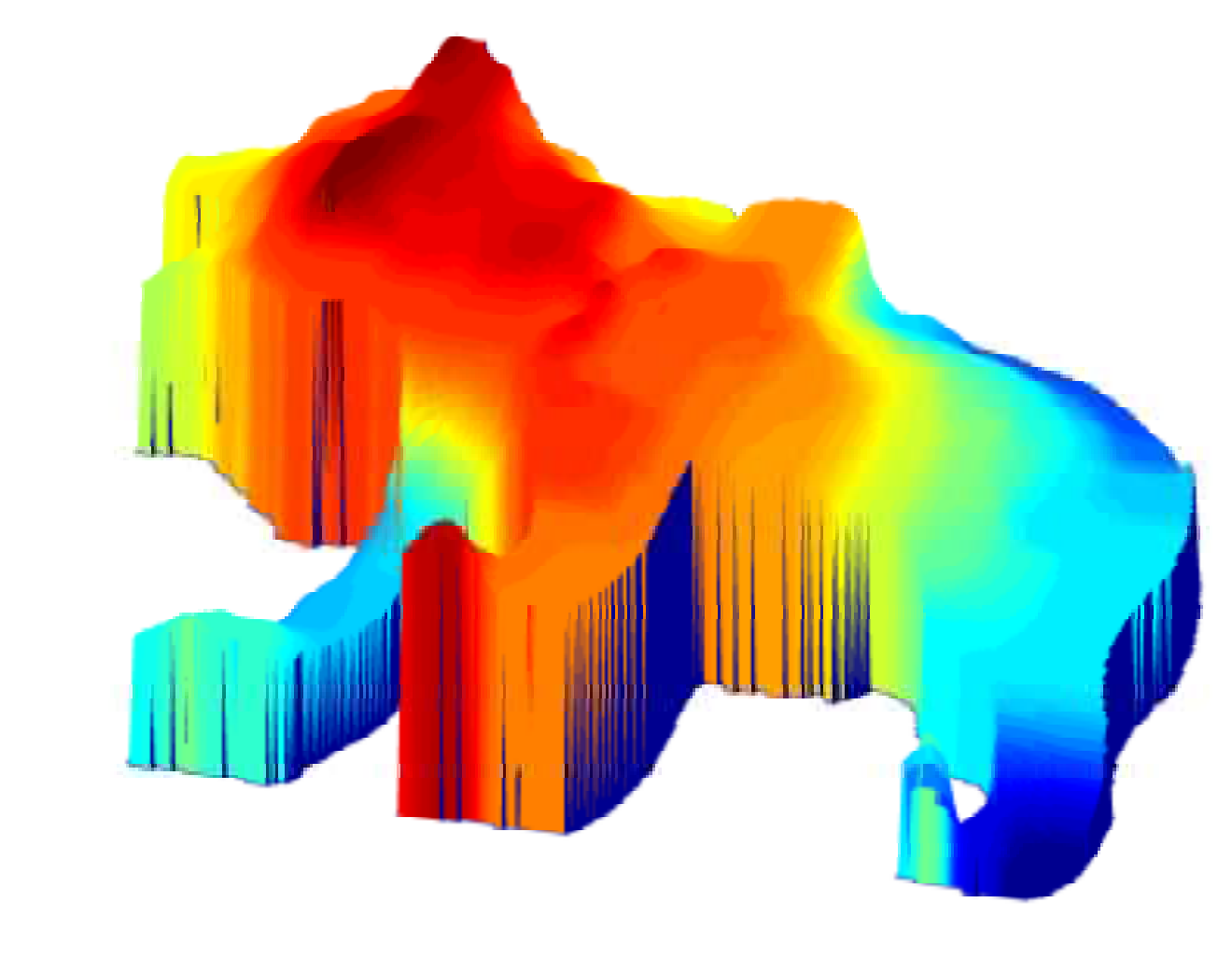}
 \includegraphics[width=0.15\linewidth]{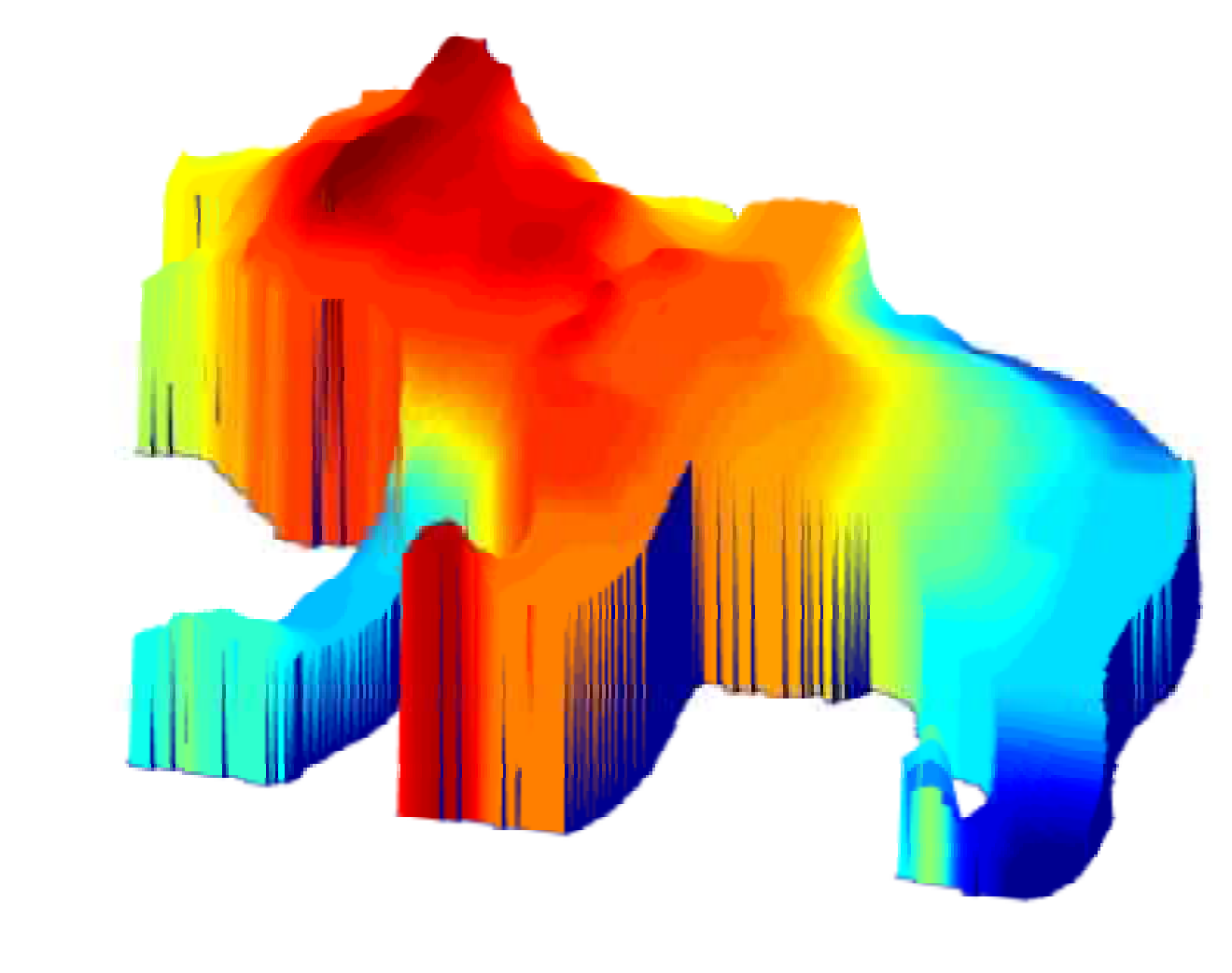} &

 \includegraphics[width=0.15\linewidth]{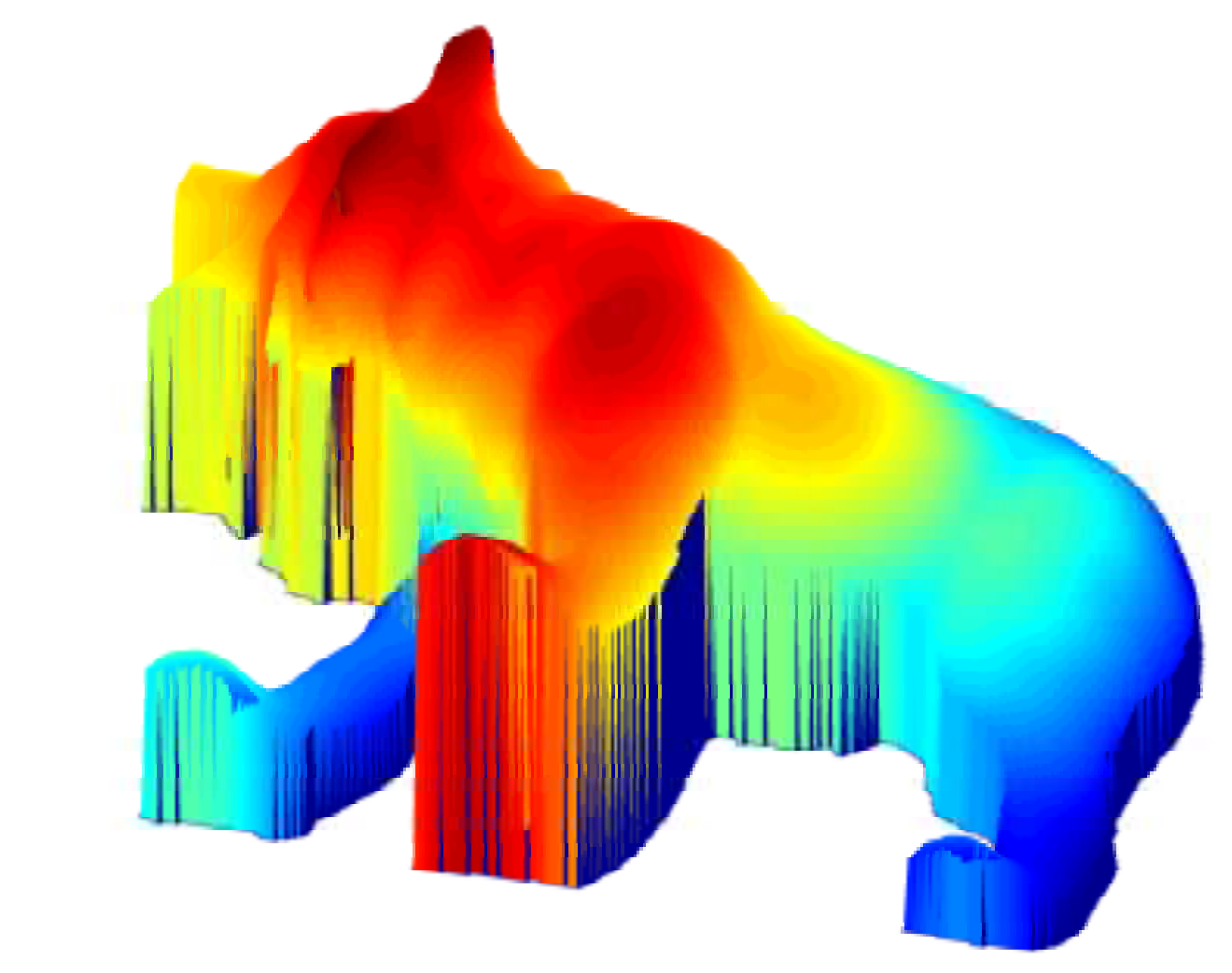}
 \includegraphics[width=0.15\linewidth]{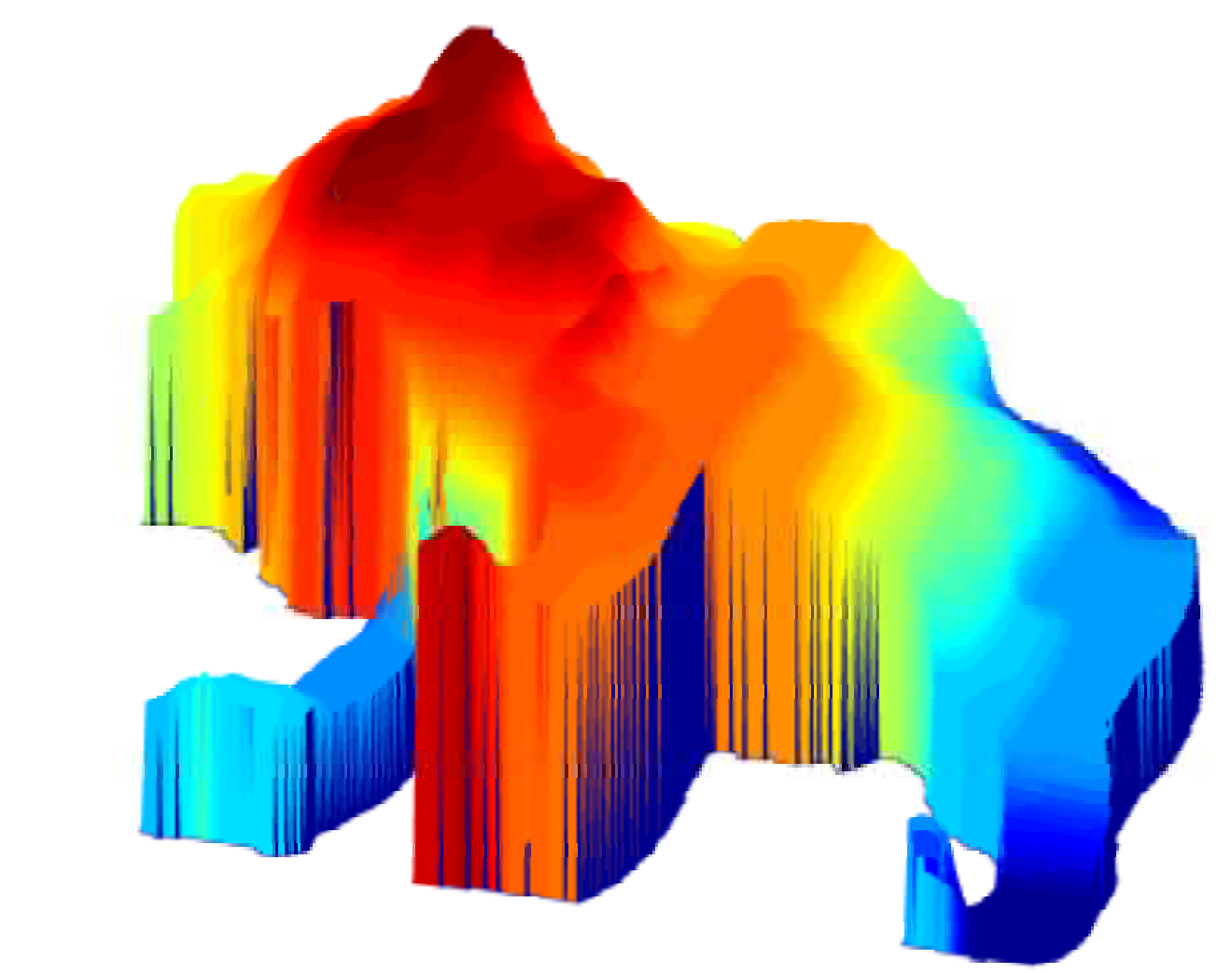}
 \includegraphics[width=0.15\linewidth]{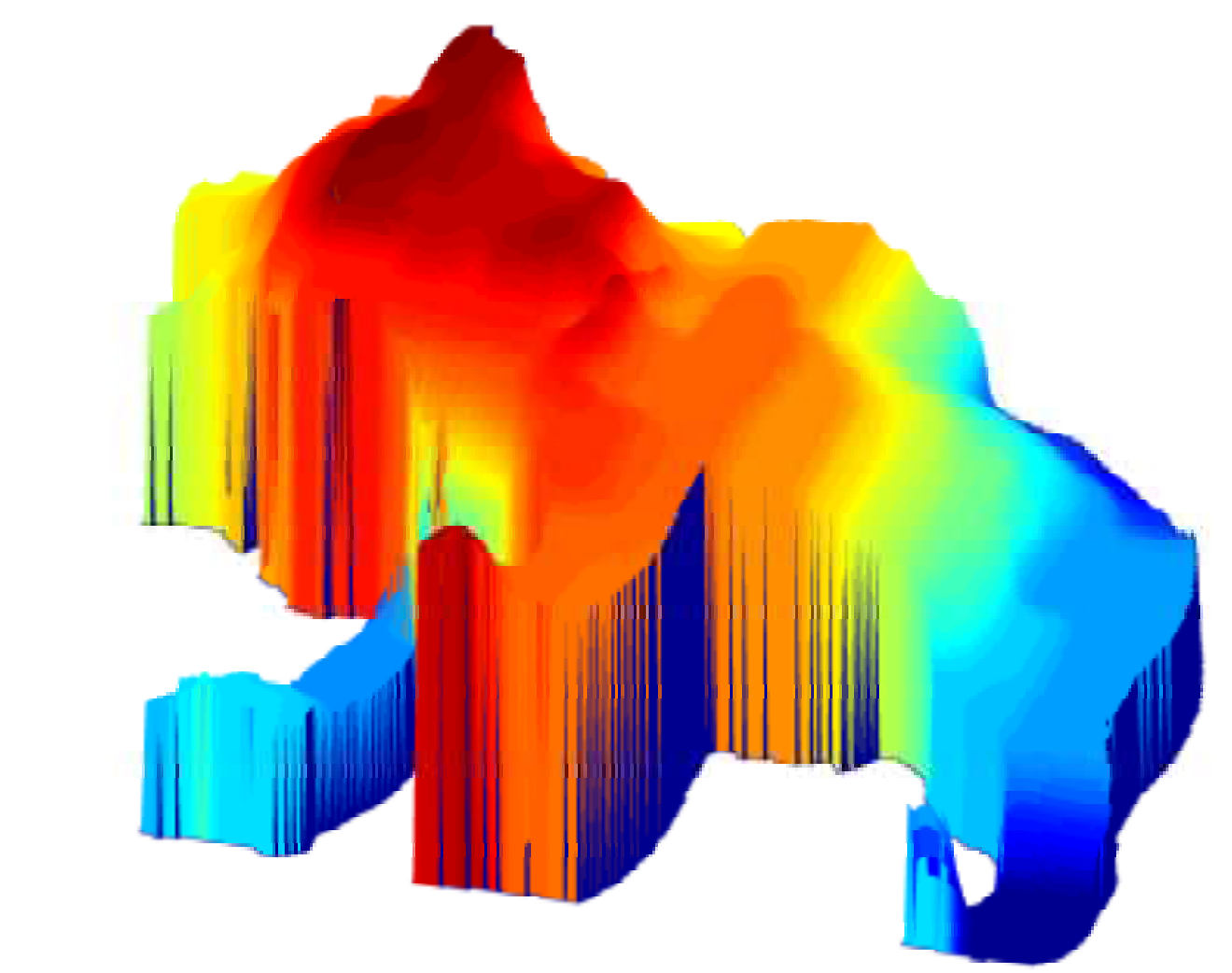} \\

 \includegraphics[width=0.15\linewidth]{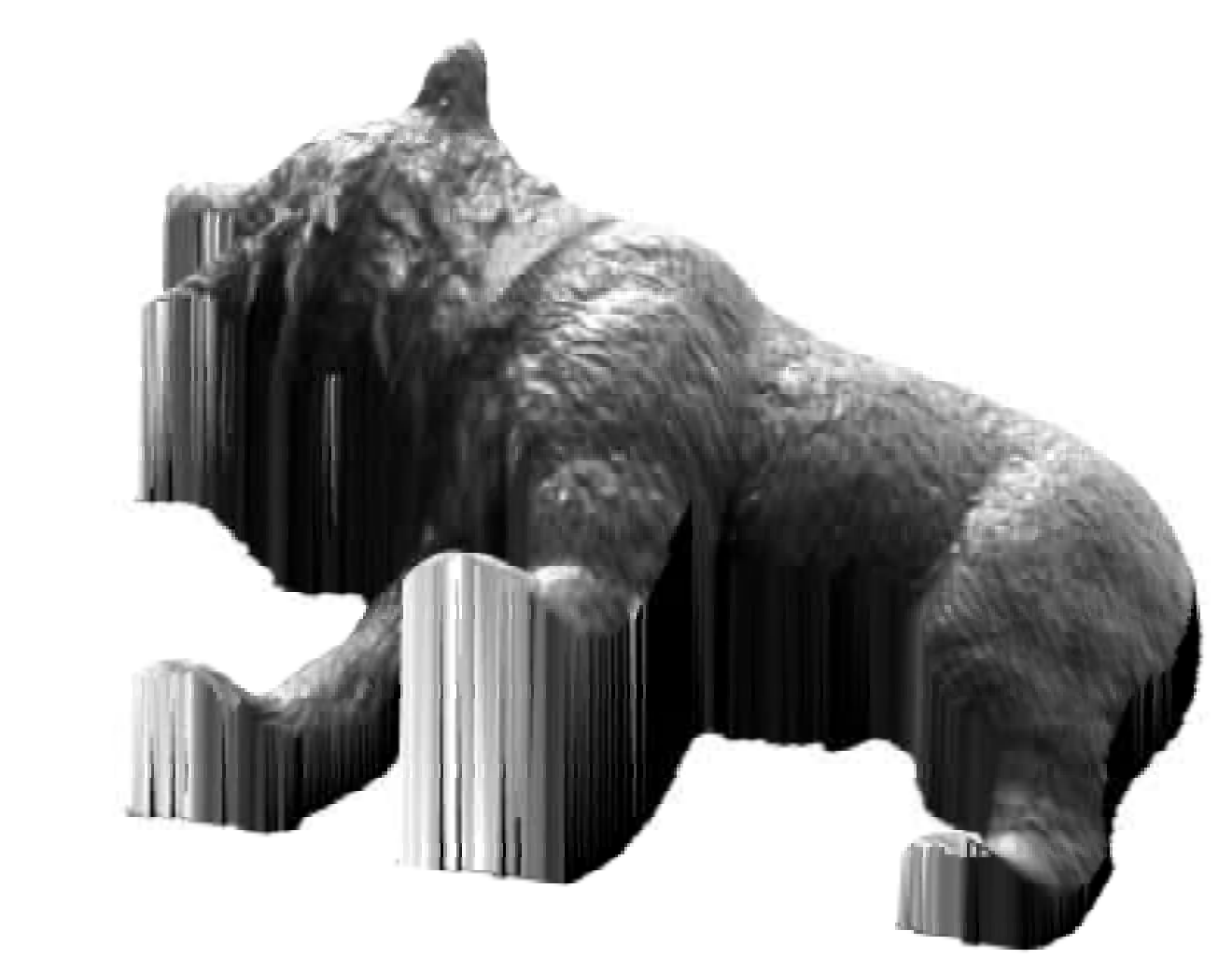}
 \includegraphics[width=0.15\linewidth]{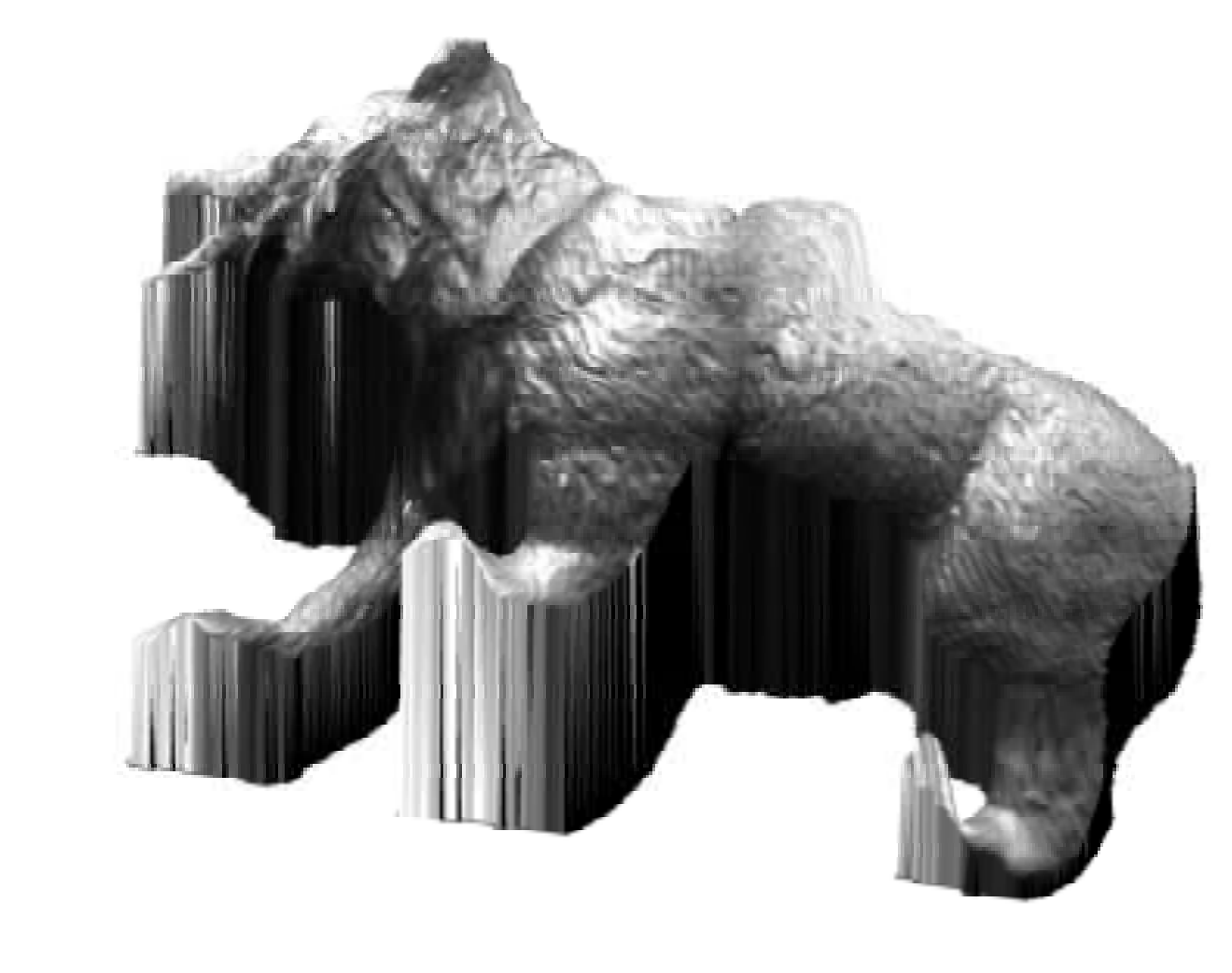}
 \includegraphics[width=0.15\linewidth]{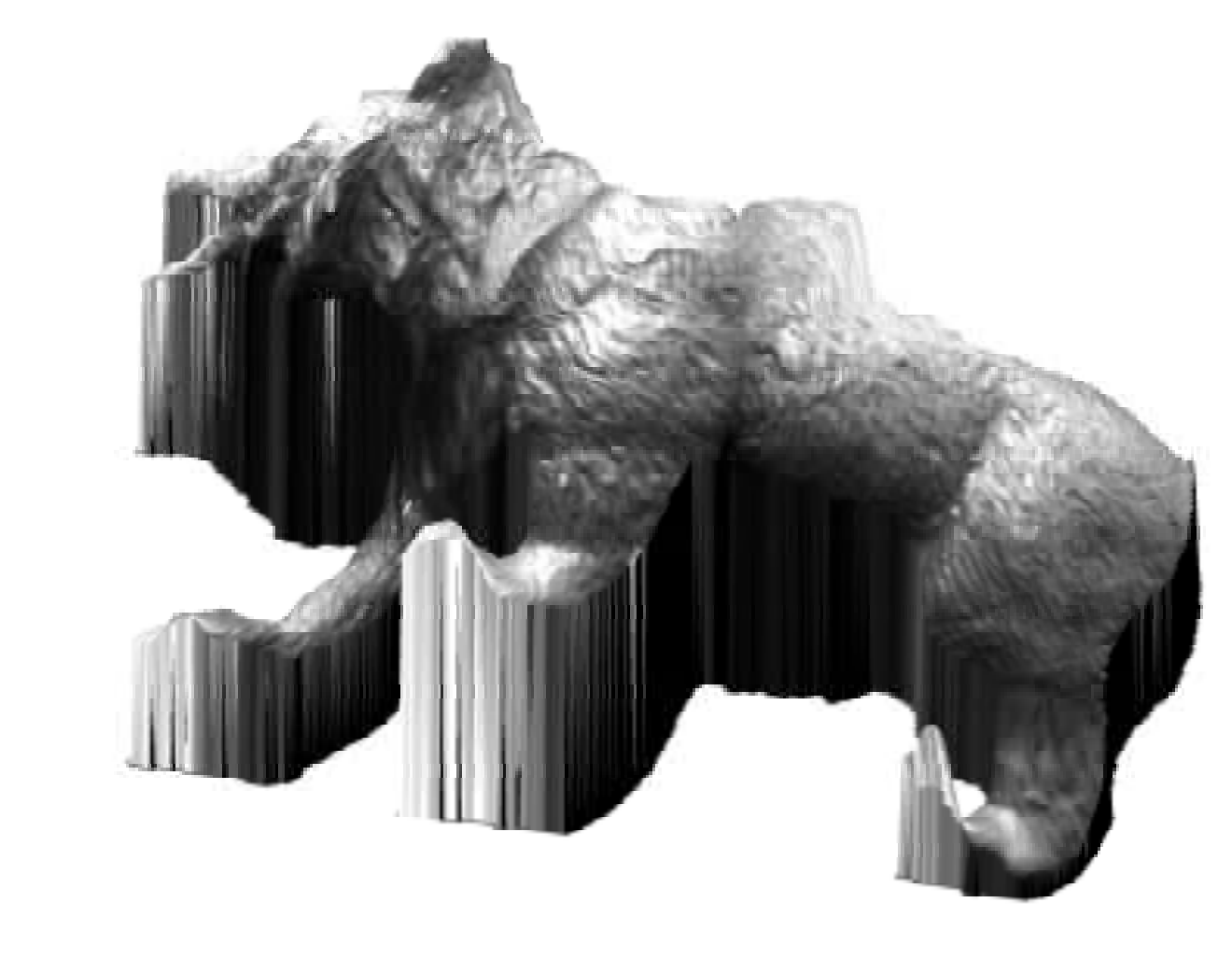} &

 \includegraphics[width=0.15\linewidth]{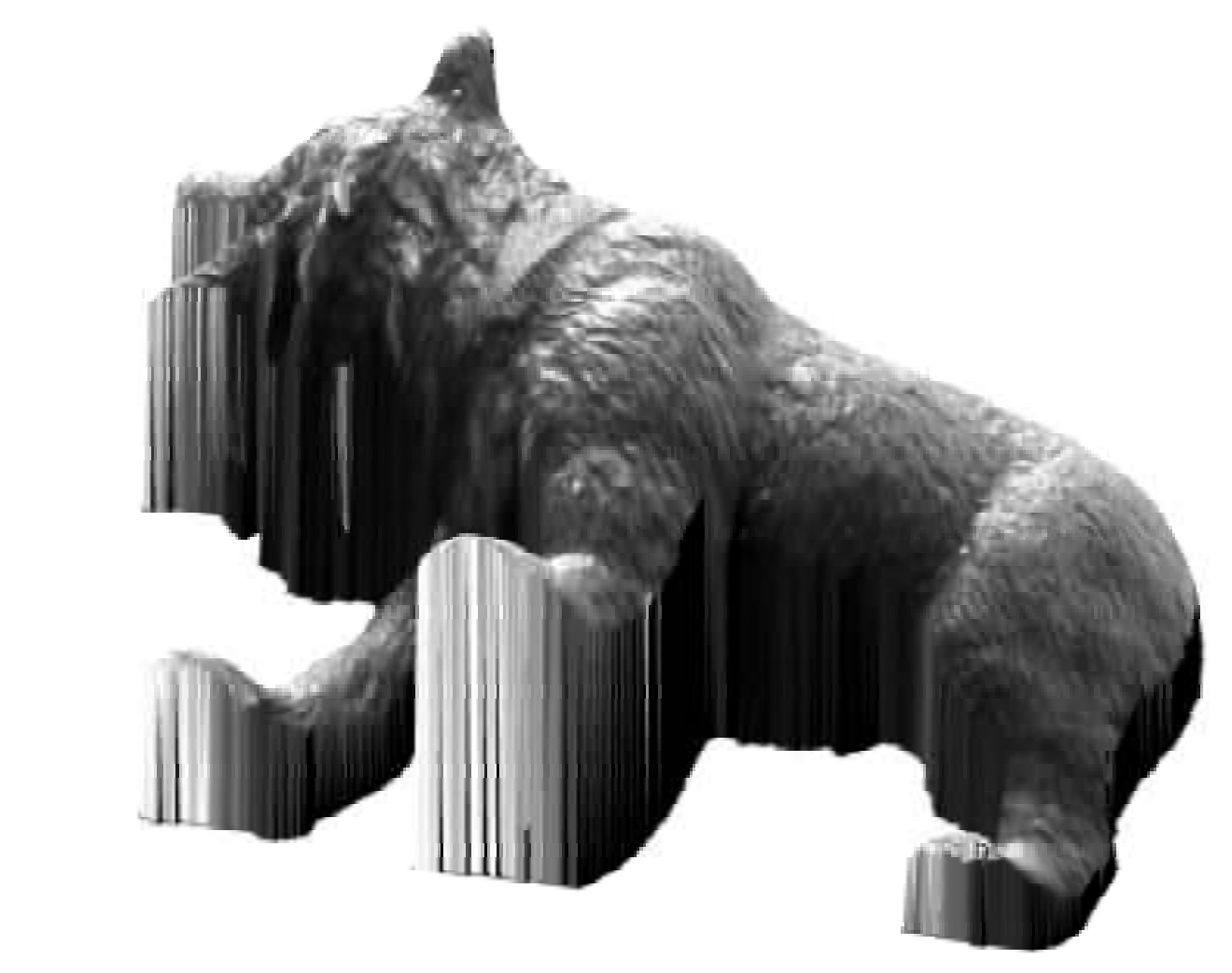}
 \includegraphics[width=0.15\linewidth]{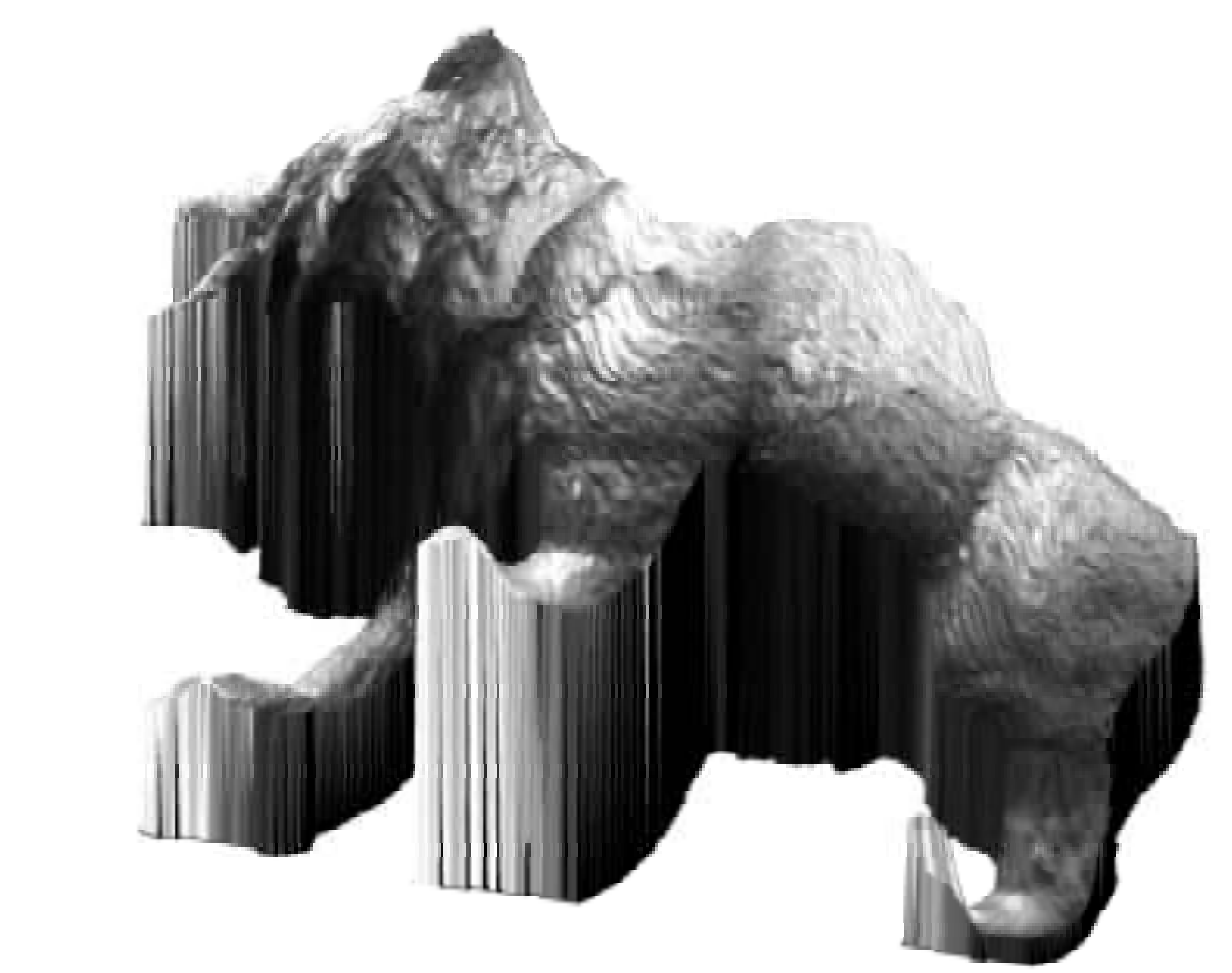}
 \includegraphics[width=0.15\linewidth]{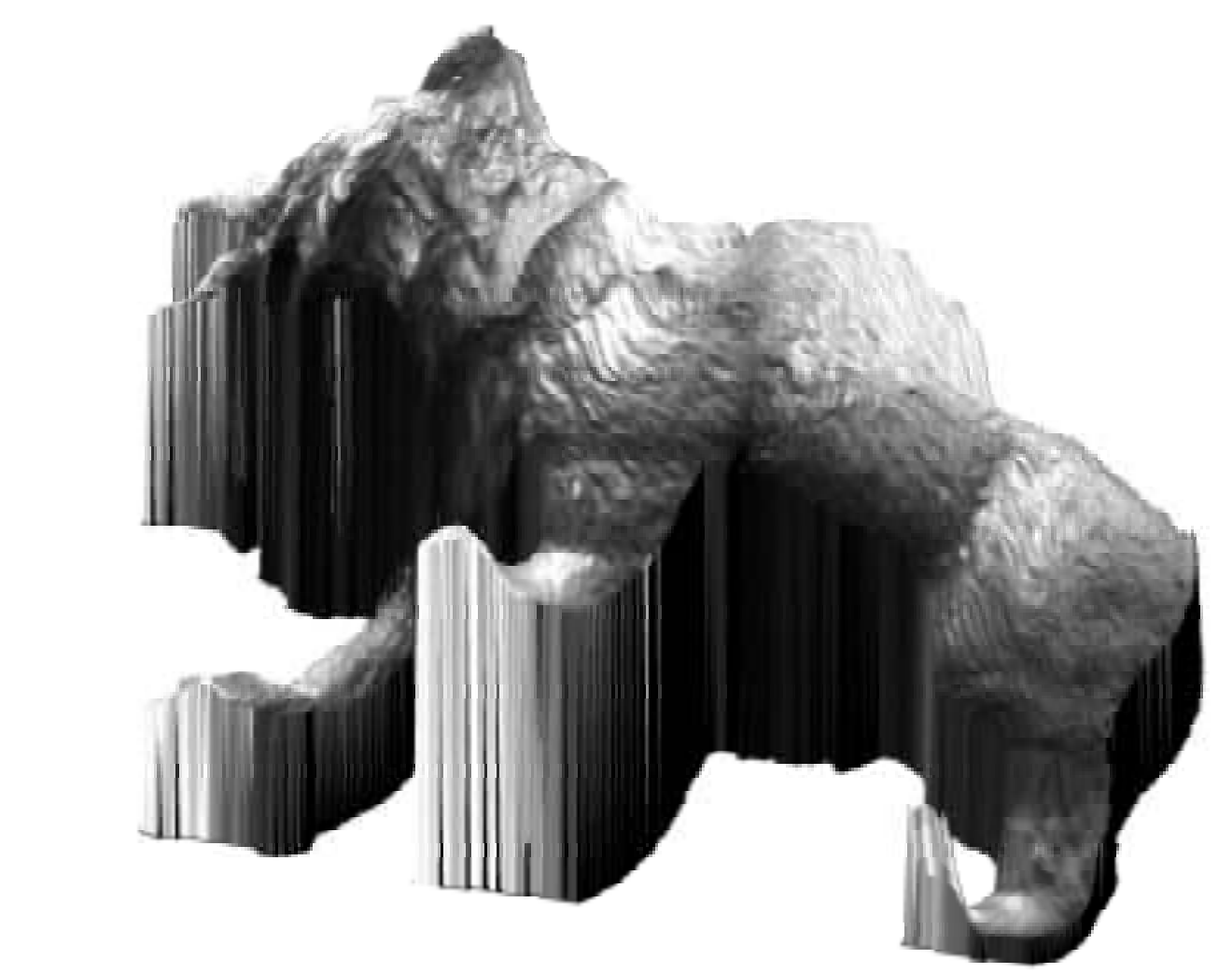} \\

\hline

 \includegraphics[width=0.15\linewidth]{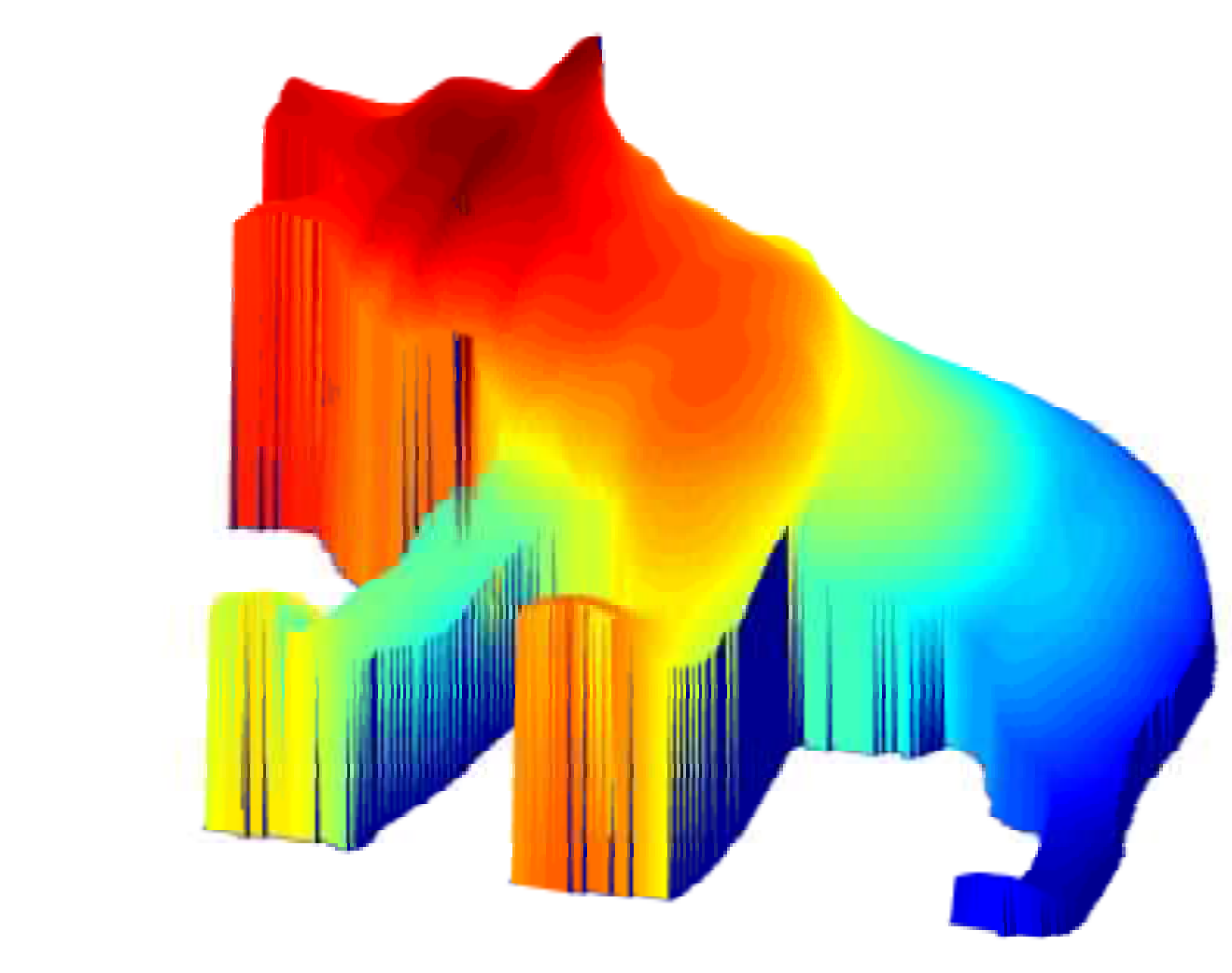}
 \includegraphics[width=0.15\linewidth]{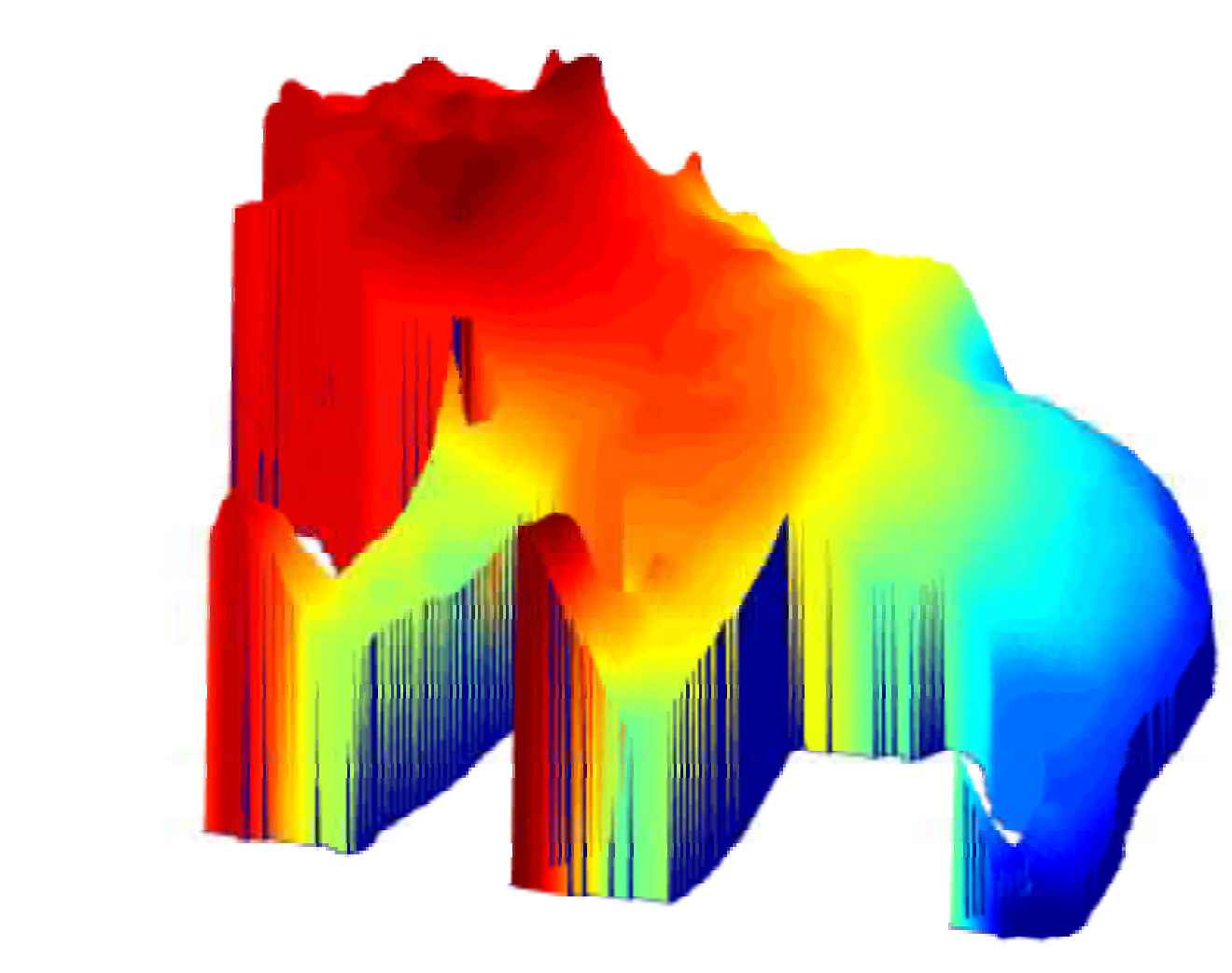}
 \includegraphics[width=0.15\linewidth]{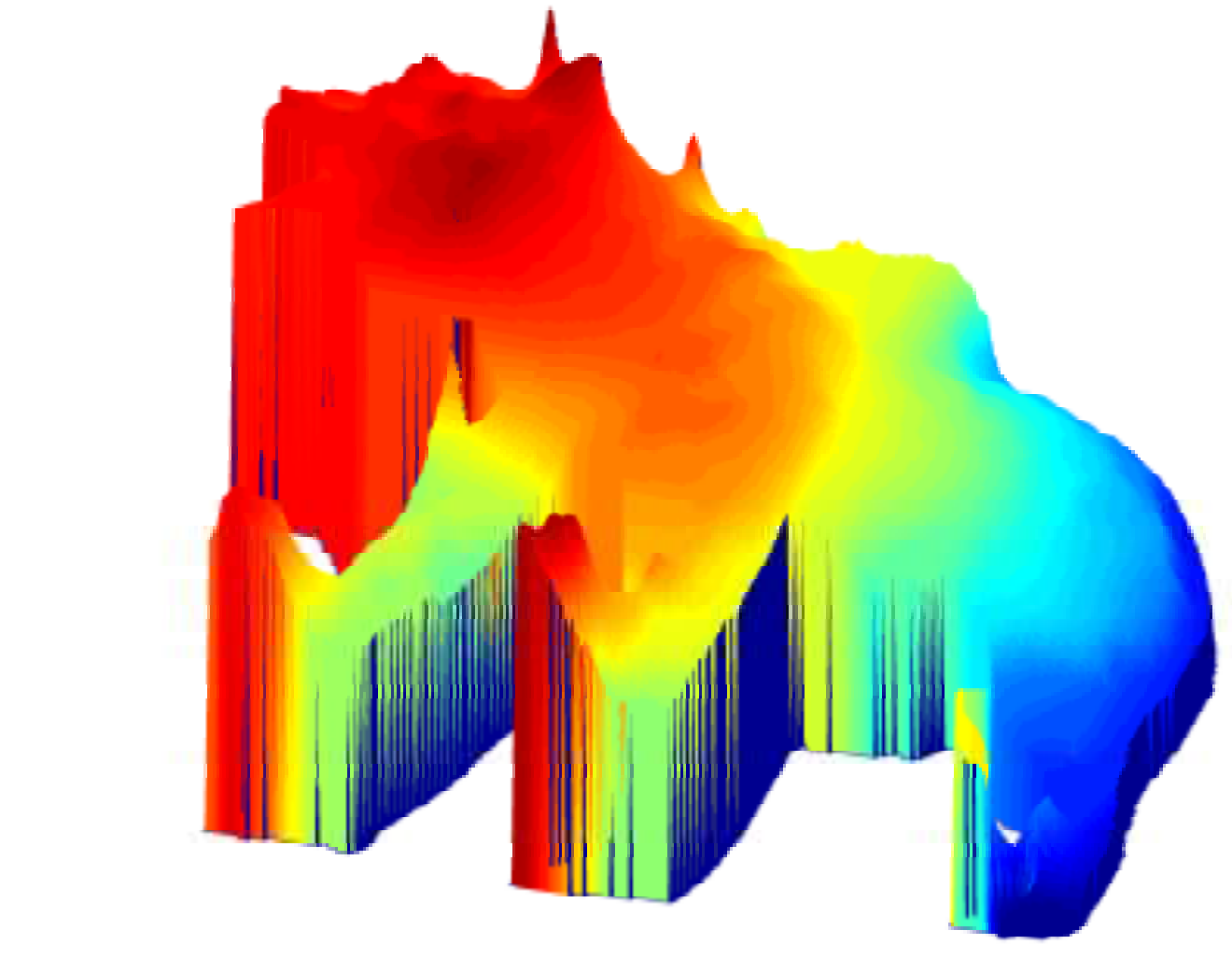} &

 \includegraphics[width=0.15\linewidth]{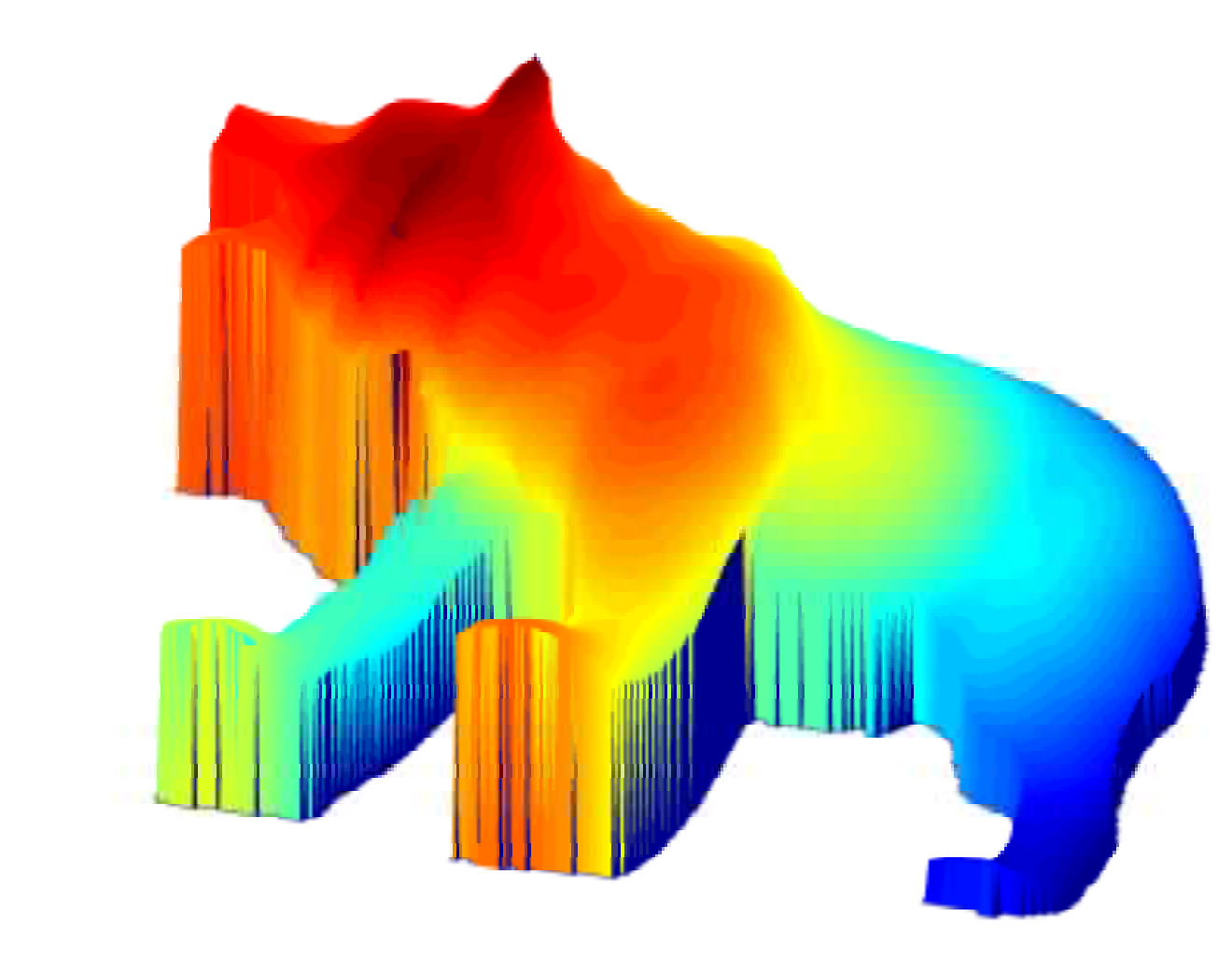}
 \includegraphics[width=0.15\linewidth]{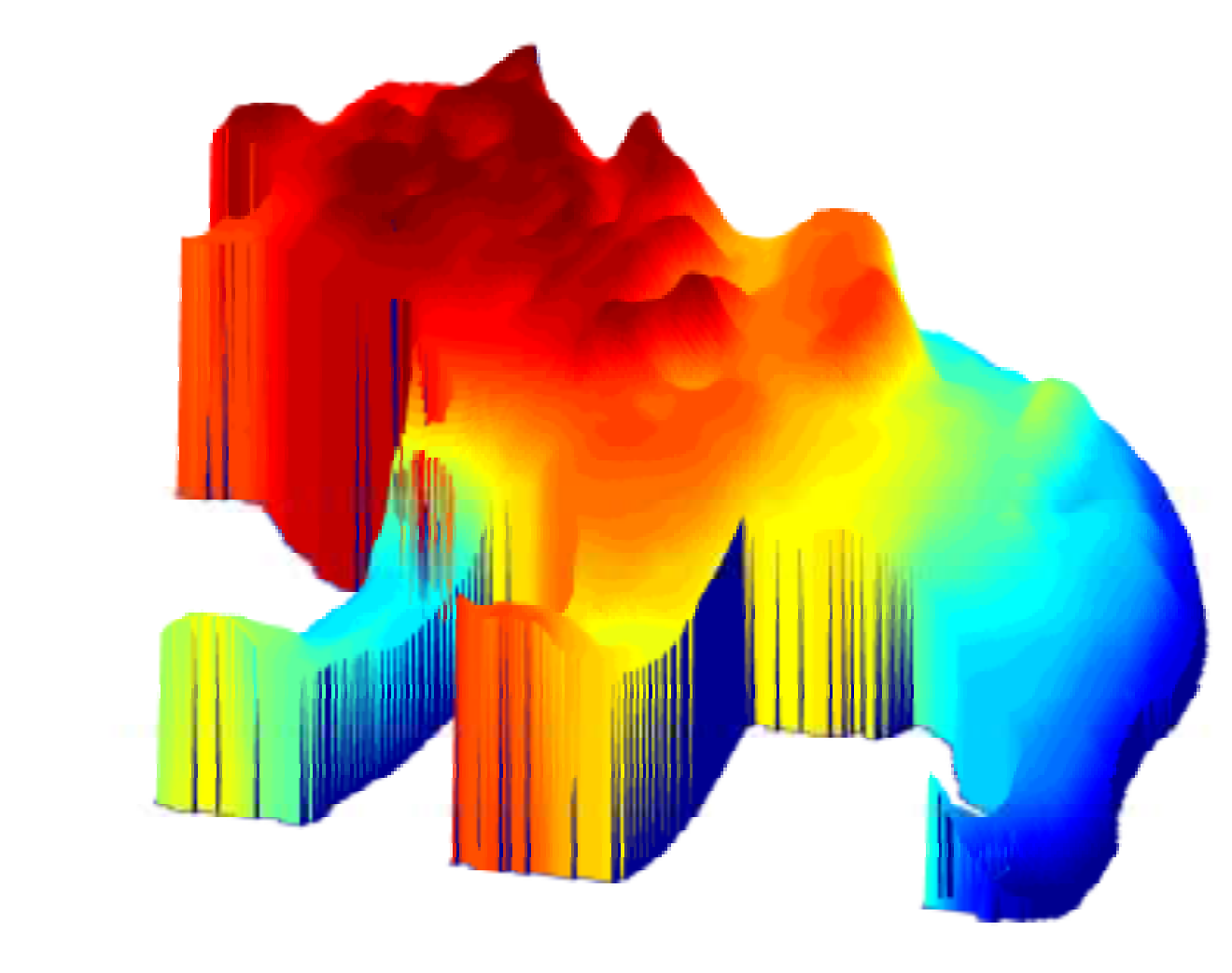}
 \includegraphics[width=0.15\linewidth]{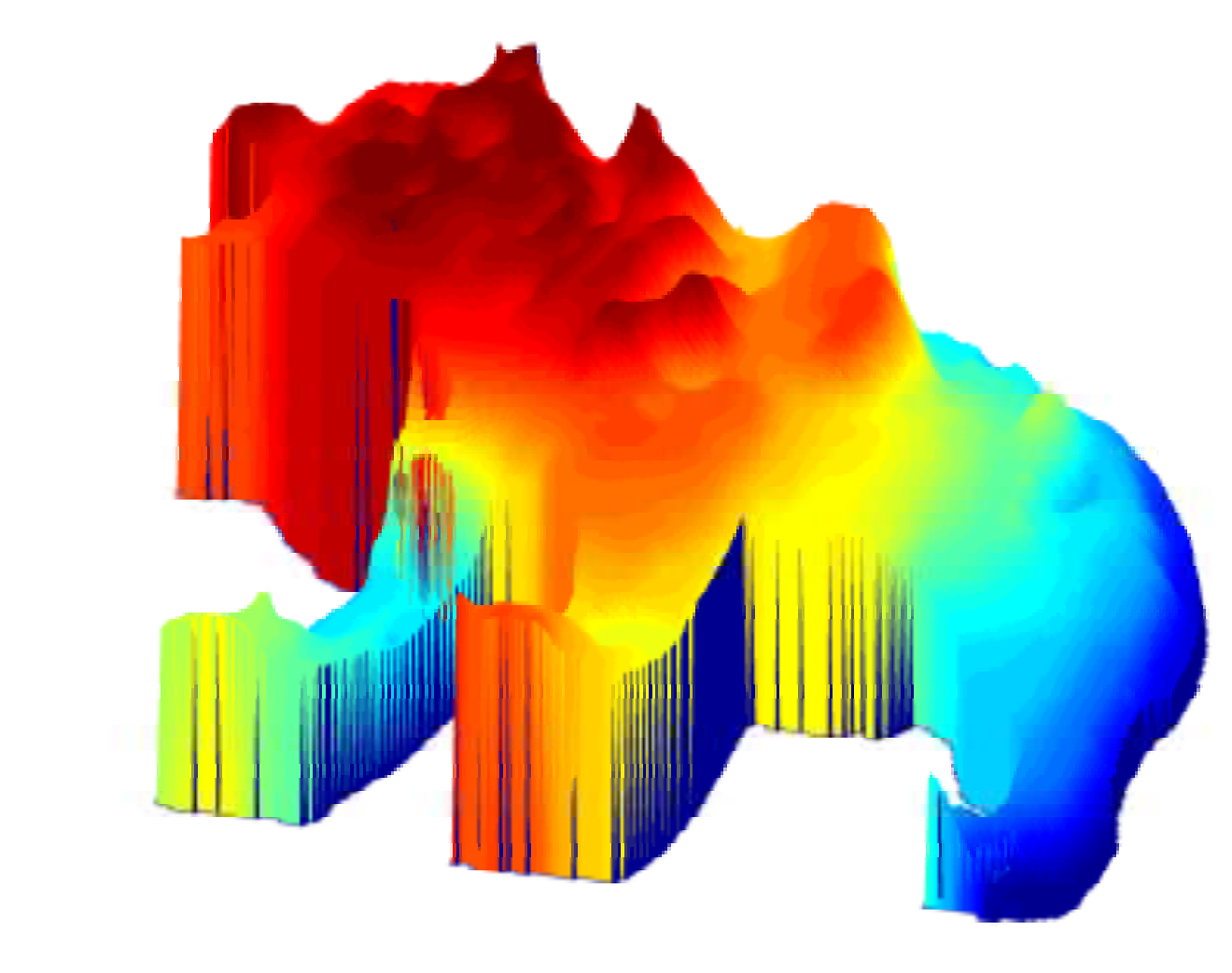} \\

 \includegraphics[width=0.15\linewidth]{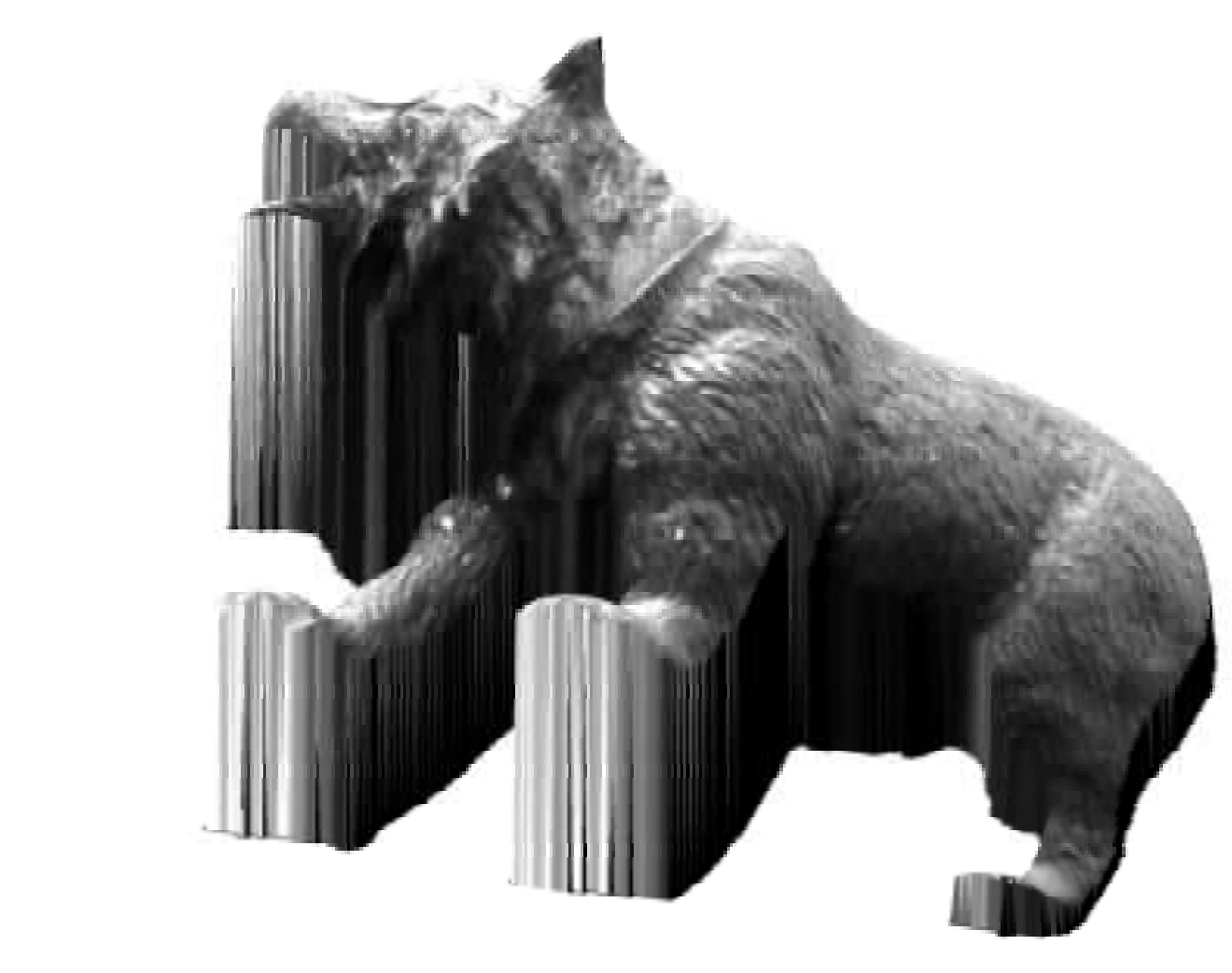}
 \includegraphics[width=0.15\linewidth]{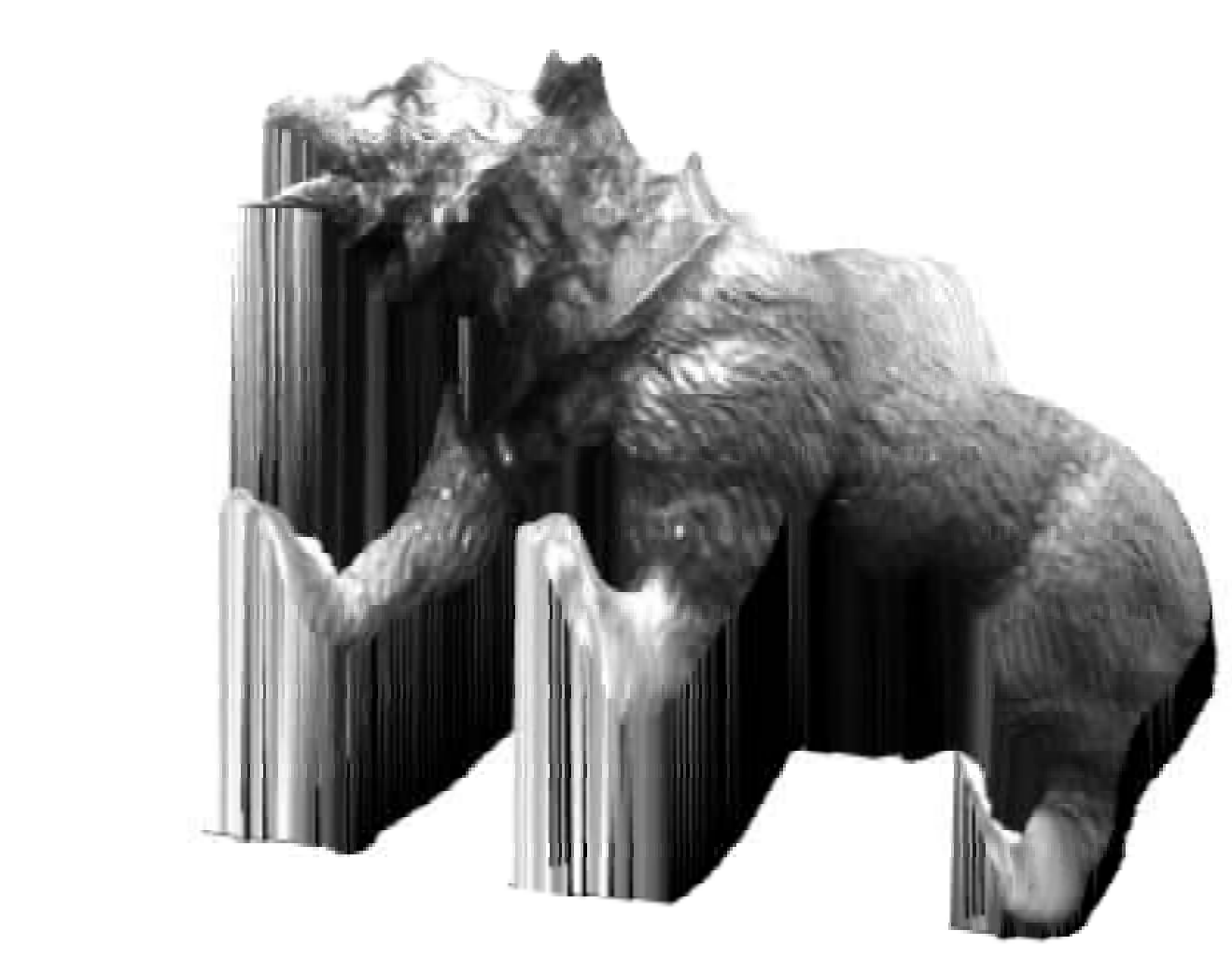}
 \includegraphics[width=0.15\linewidth]{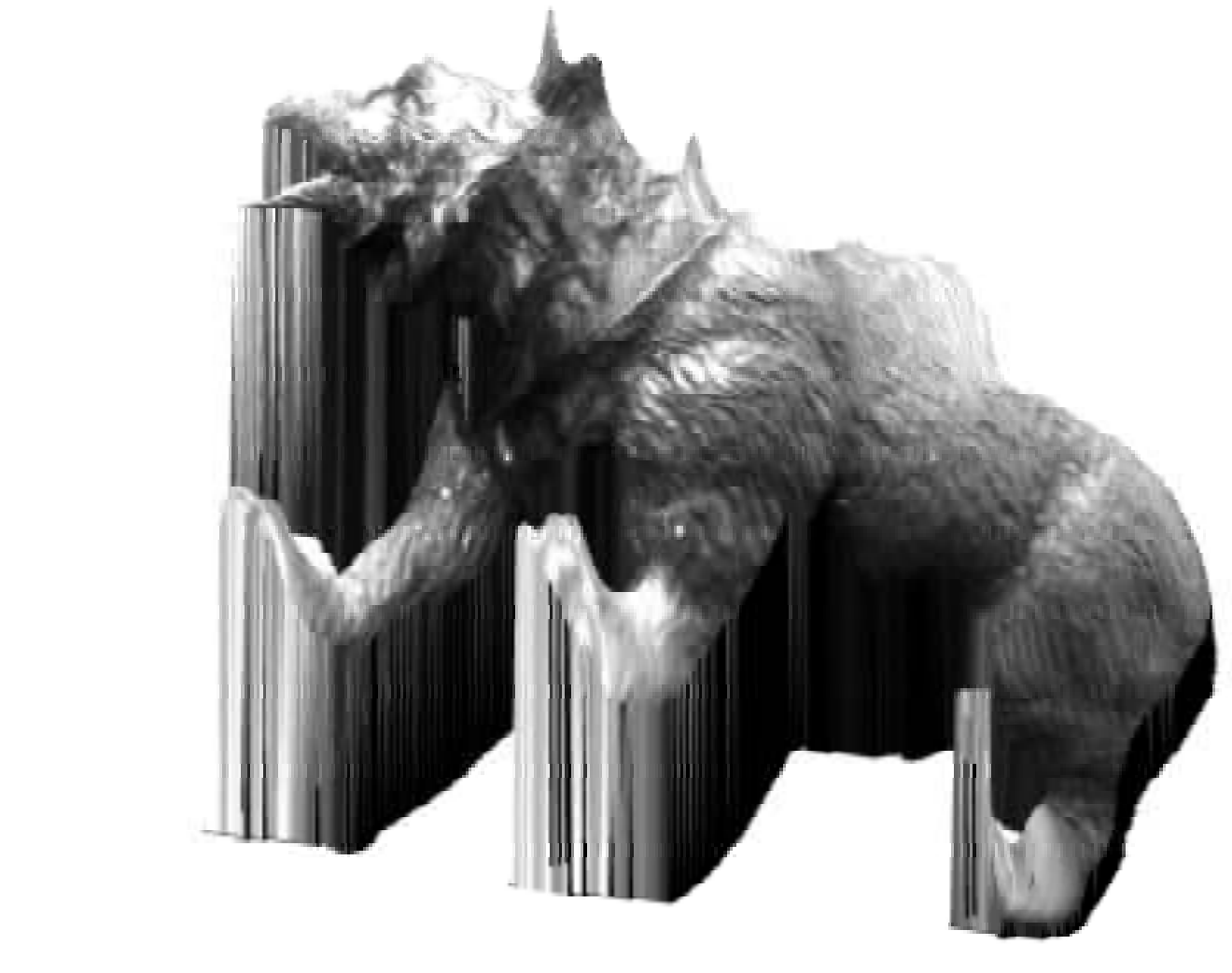} &

 \includegraphics[width=0.15\linewidth]{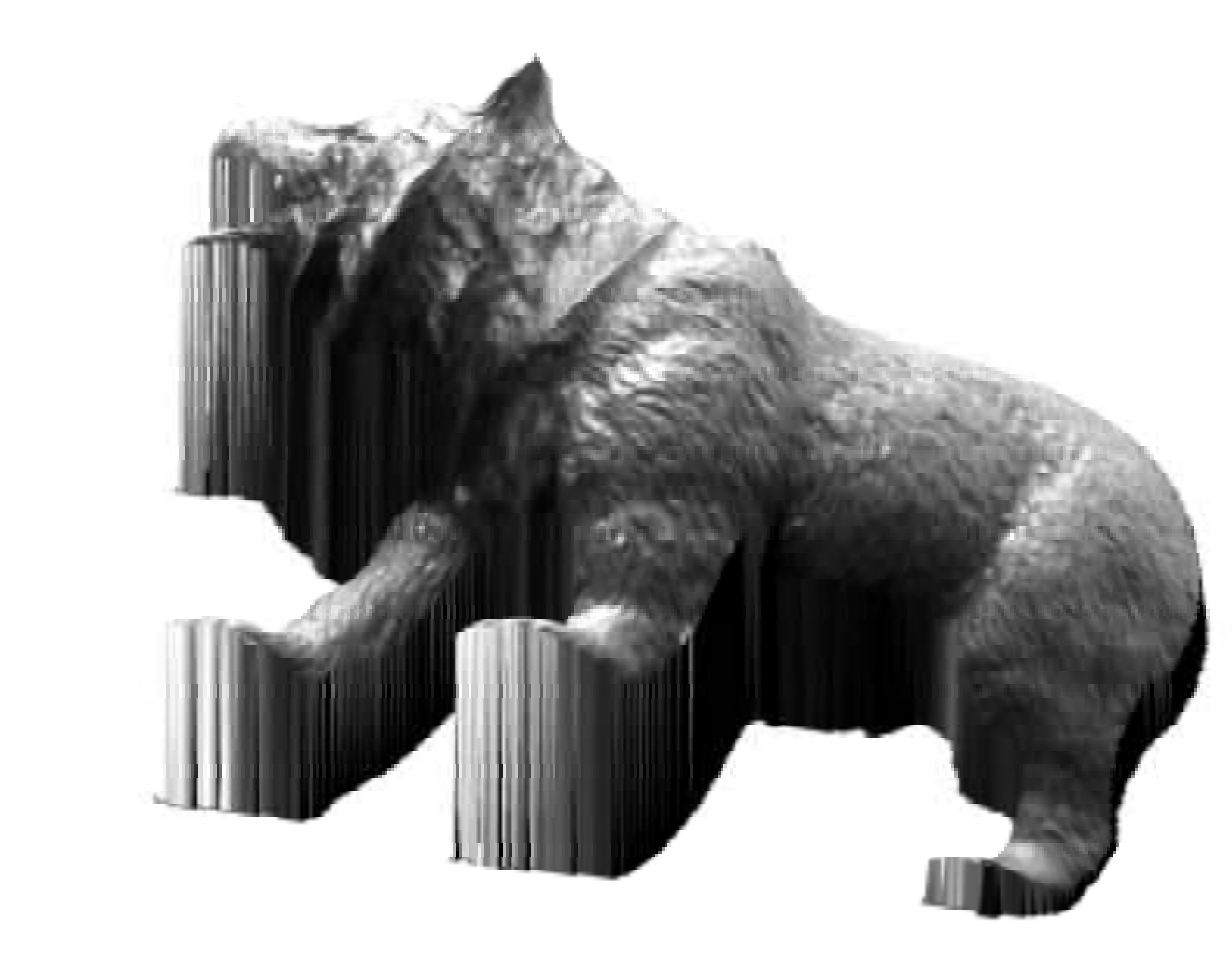}
 \includegraphics[width=0.15\linewidth]{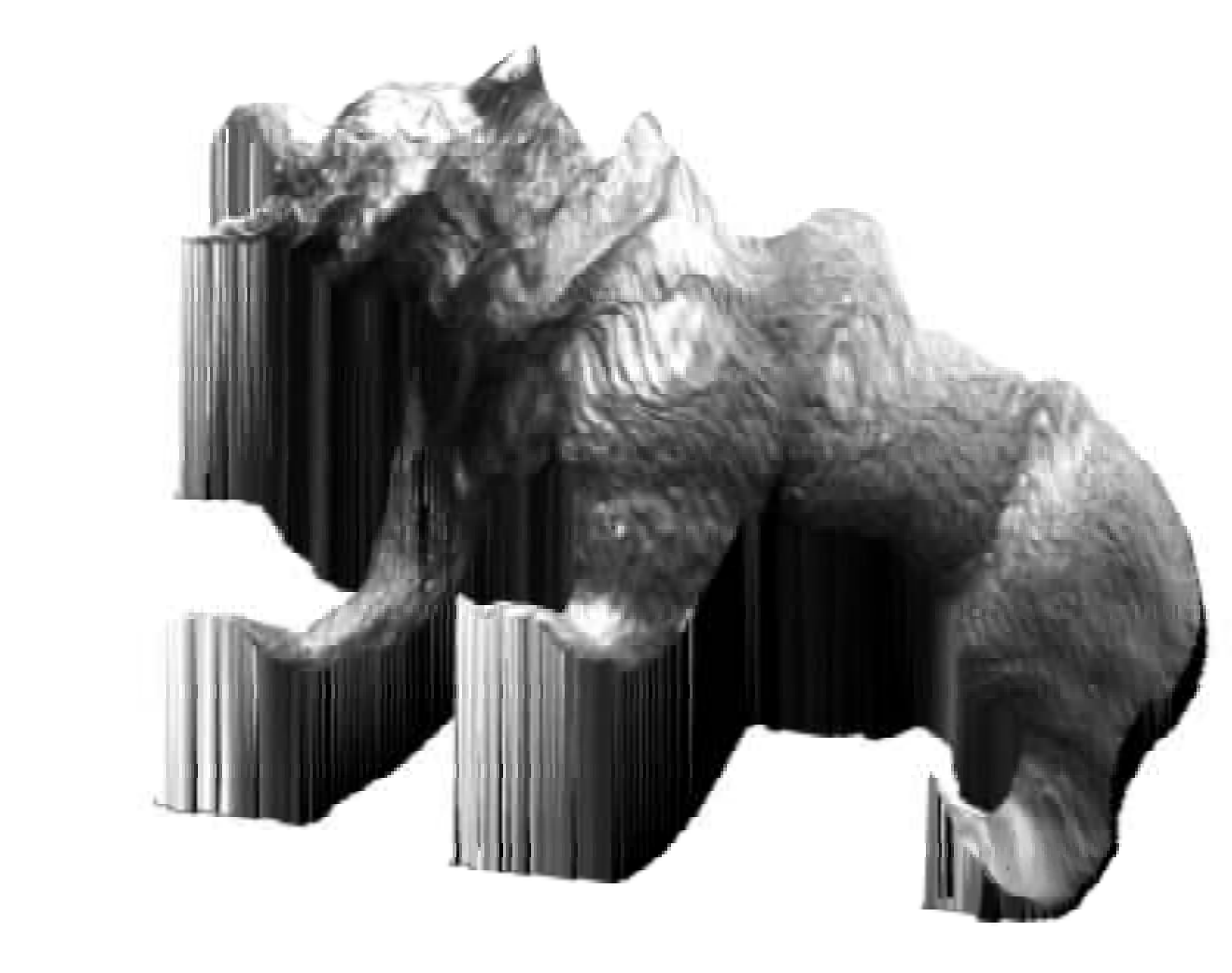}
 \includegraphics[width=0.15\linewidth]{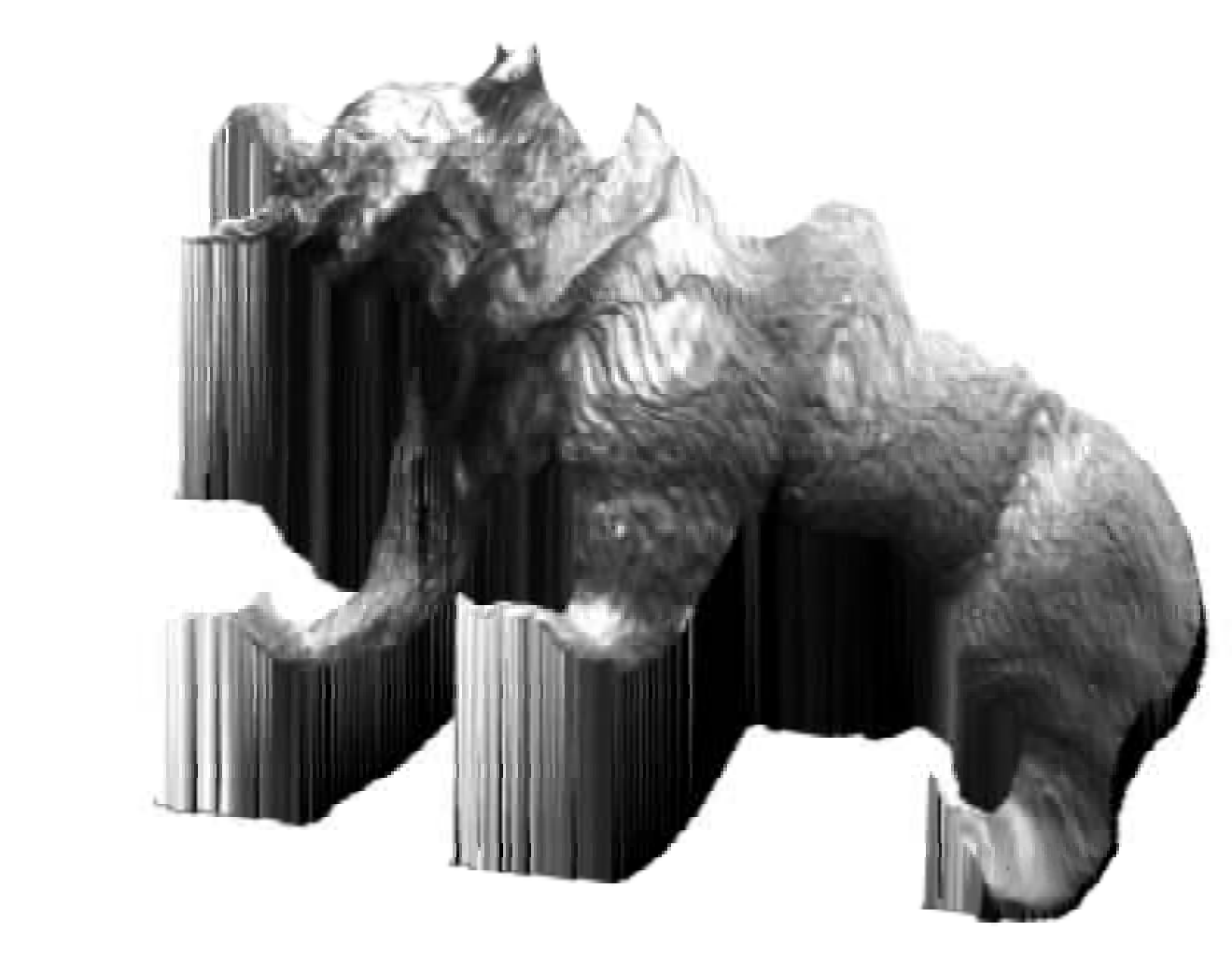} \\

 \hline
\end{tabular}
\caption{Reconstructions from image pairs. Each box shows from left to right the laser scan, reconstruction with and without boundary conditions. The top row in each box shows the surface (colormap represents depth values) and the bottom row shows the surface painted with intensity values. Reconstructions are shown for images e and g ($9.3^\circ$) both achieving RMSE of 1.30 pixels (top left box), images d and f ($10.1^\circ$) both achieving RMSE of 1.52 pixels (top right), a and b ($12.1^\circ$) achieving 2.24 and 2.20 pixels, and c and e ($16.3^\circ$) achieving 3.89 and 3.87.}
\end{figure}

\begin{figure}
\begin{center}
 \includegraphics[width=0.60\linewidth]{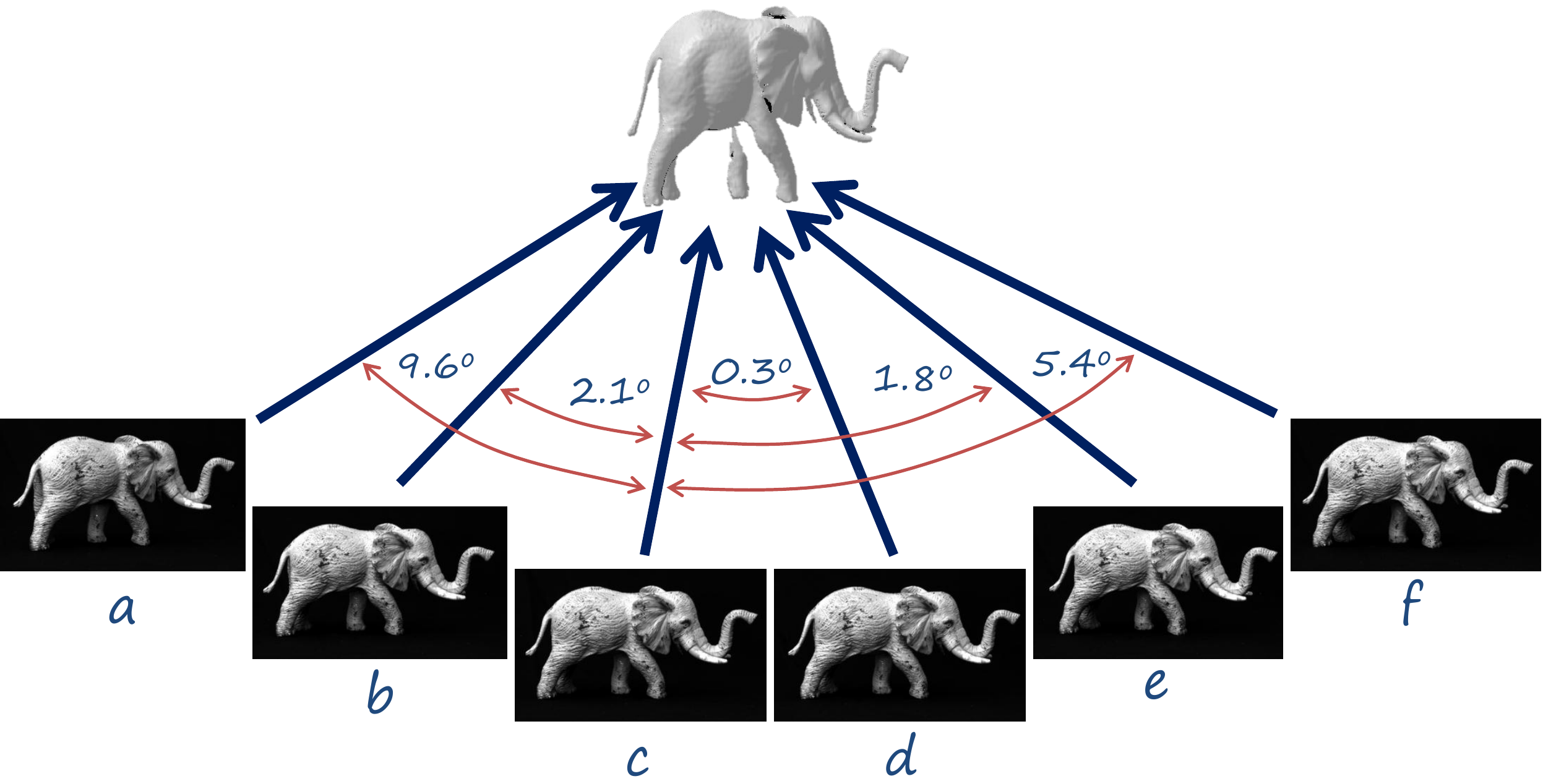}
\end{center}
\caption{Input images of a elephant toy. A laser scan is shown at the center, and relative viewing angles are provided.}
\end{figure}

\begin{figure}
\begin{tabular}{|c|c|}
\hline
\includegraphics[width=0.15\linewidth]{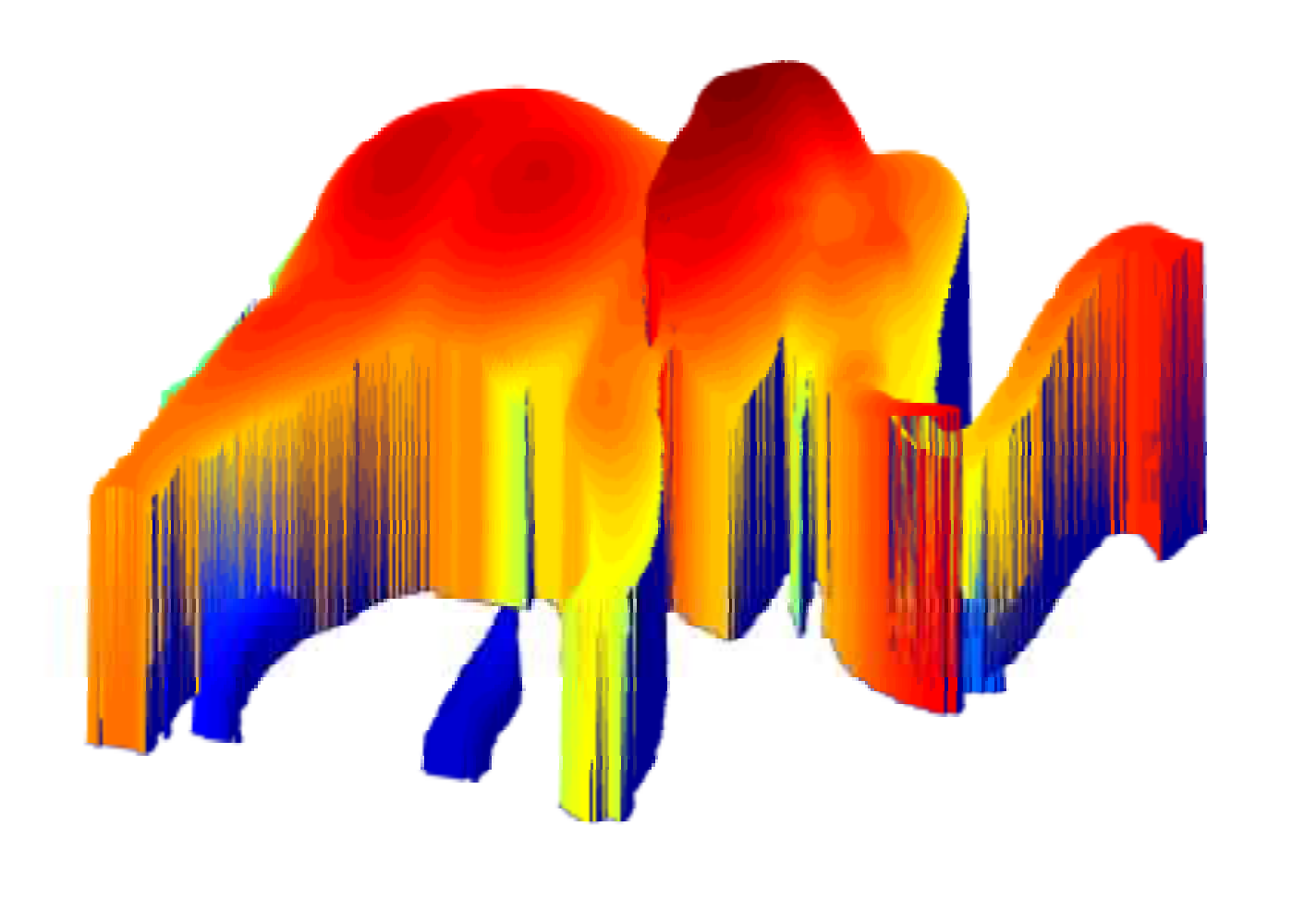}
\includegraphics[width=0.15\linewidth]{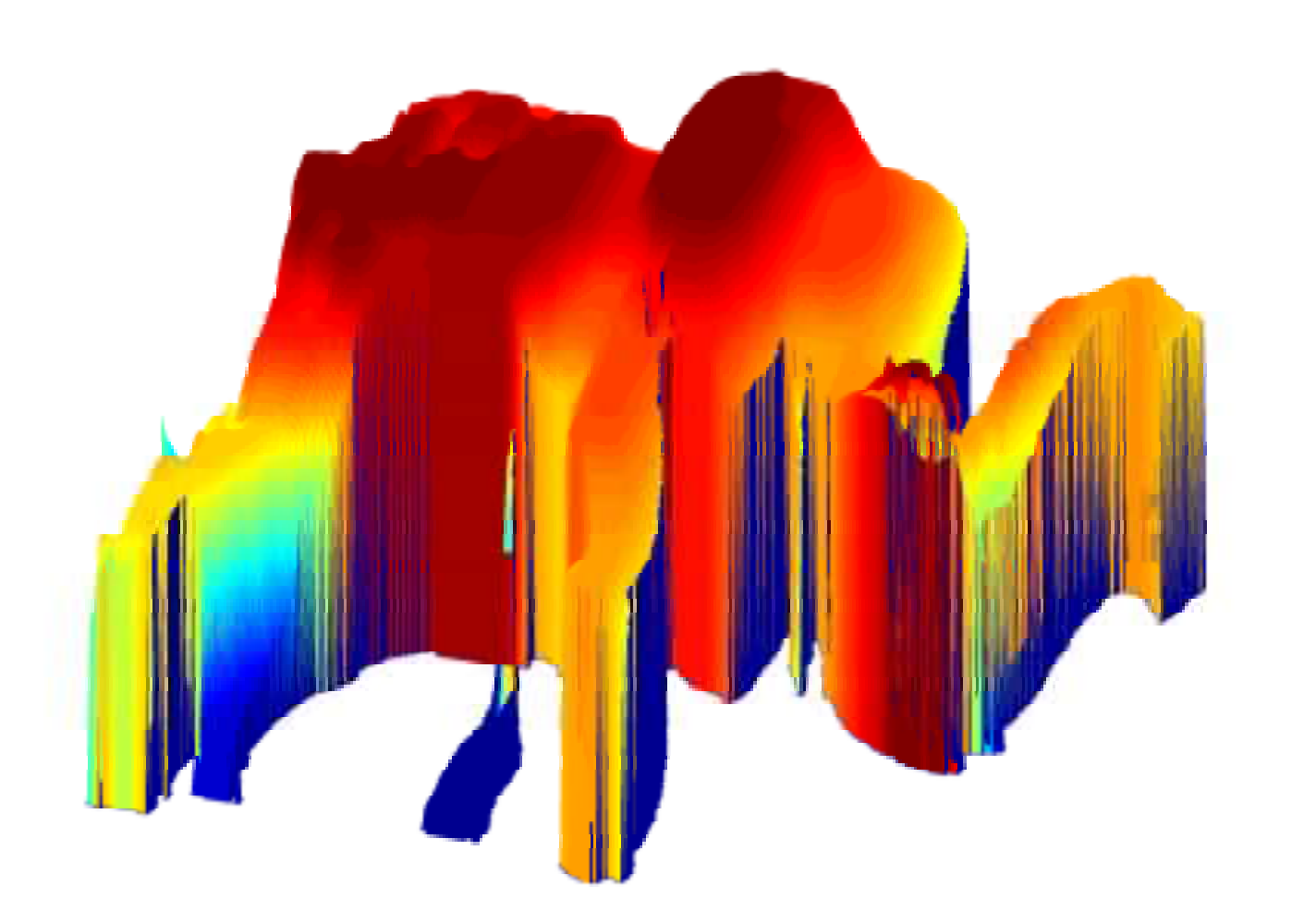}
\includegraphics[width=0.15\linewidth]{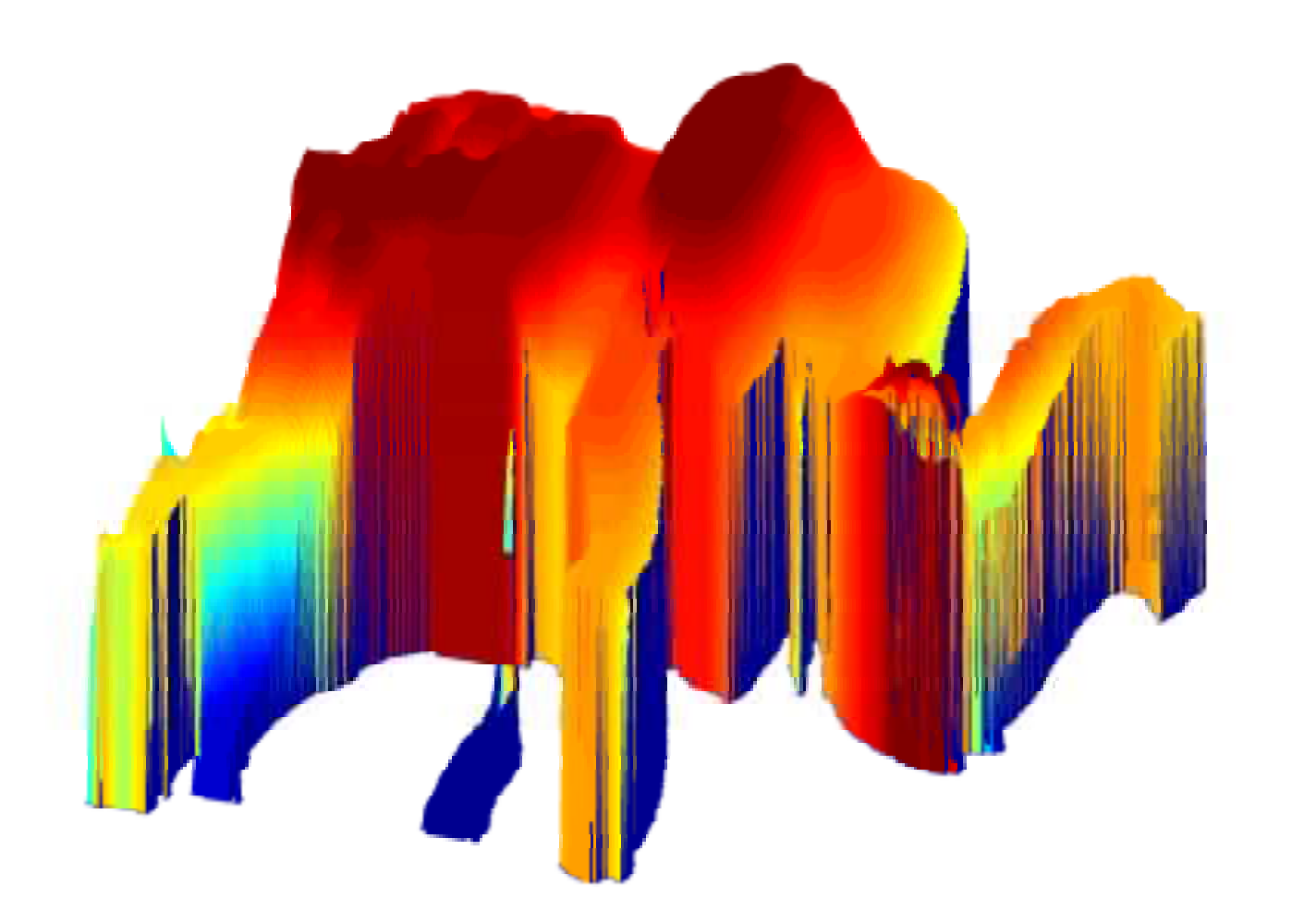} &

\includegraphics[width=0.15\linewidth]{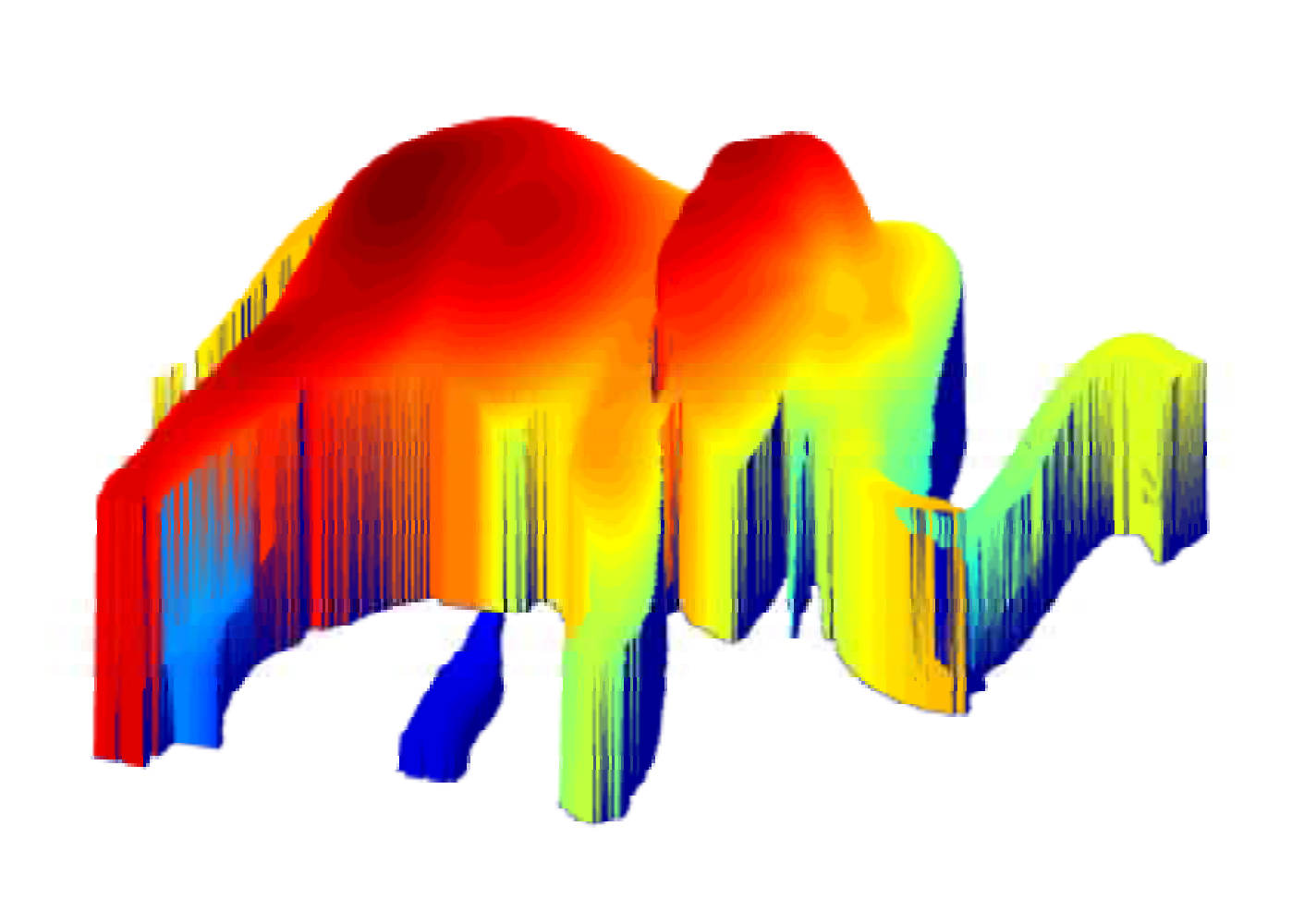}
\includegraphics[width=0.15\linewidth]{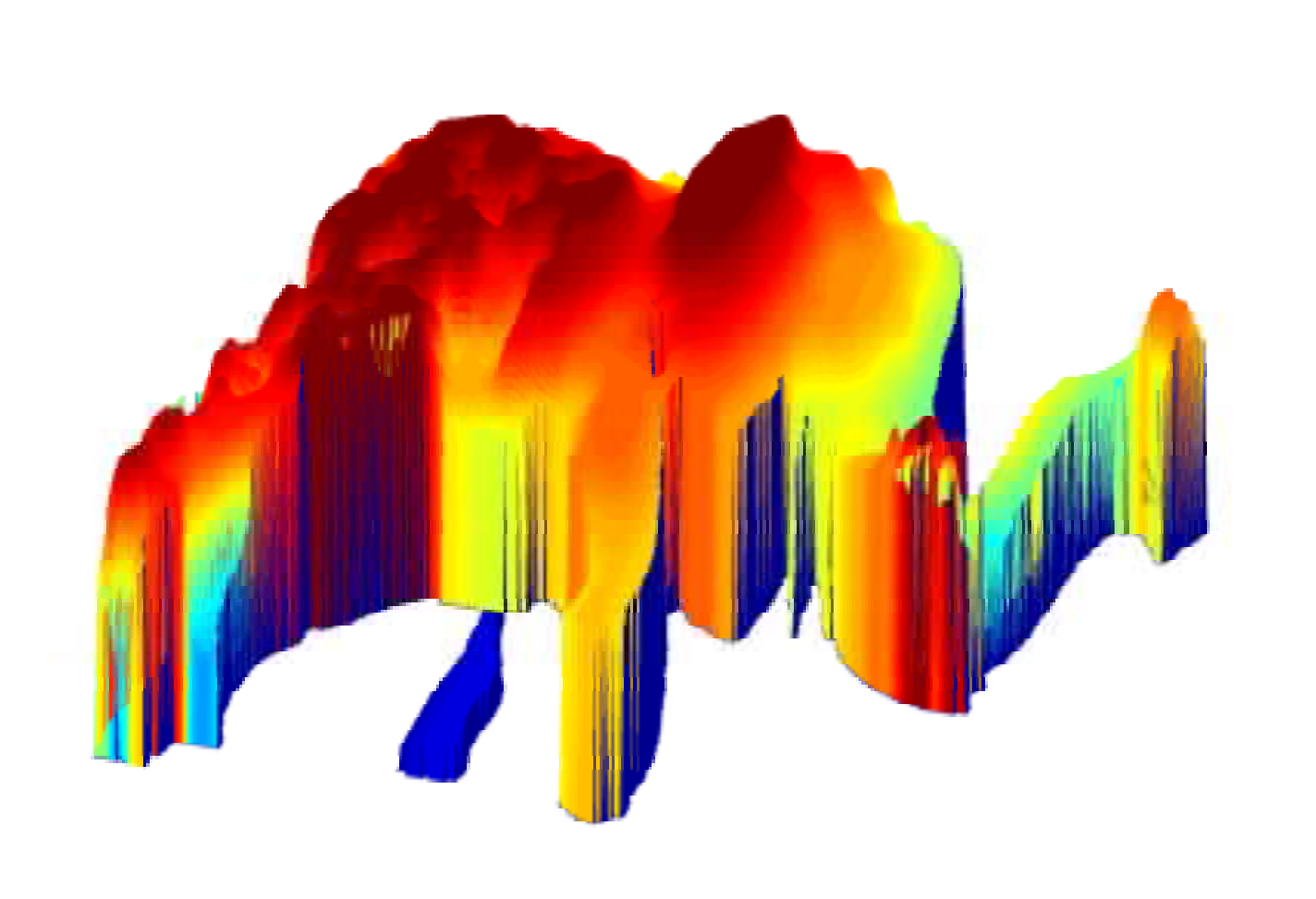}
\includegraphics[width=0.15\linewidth]{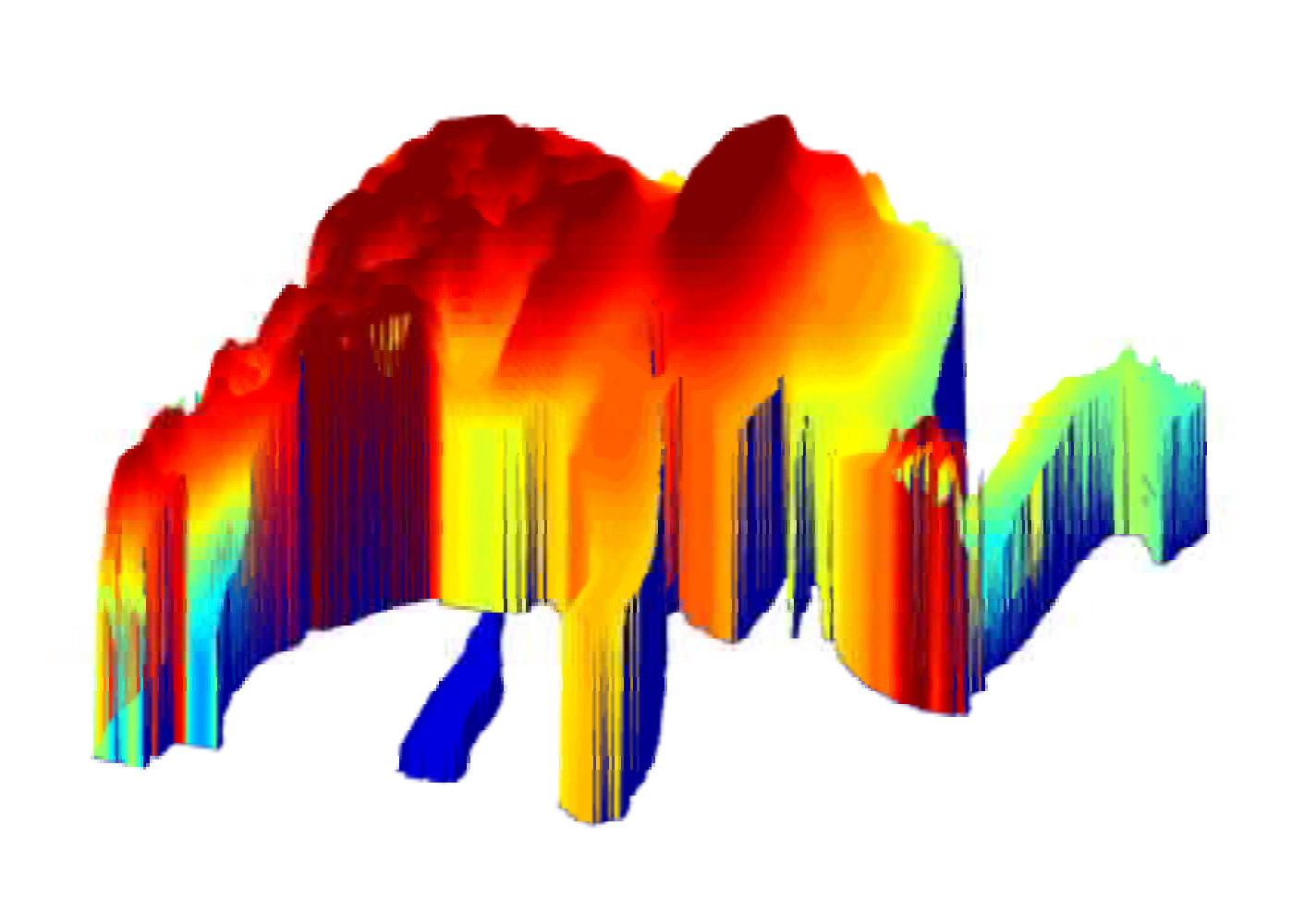} \\

\includegraphics[width=0.15\linewidth]{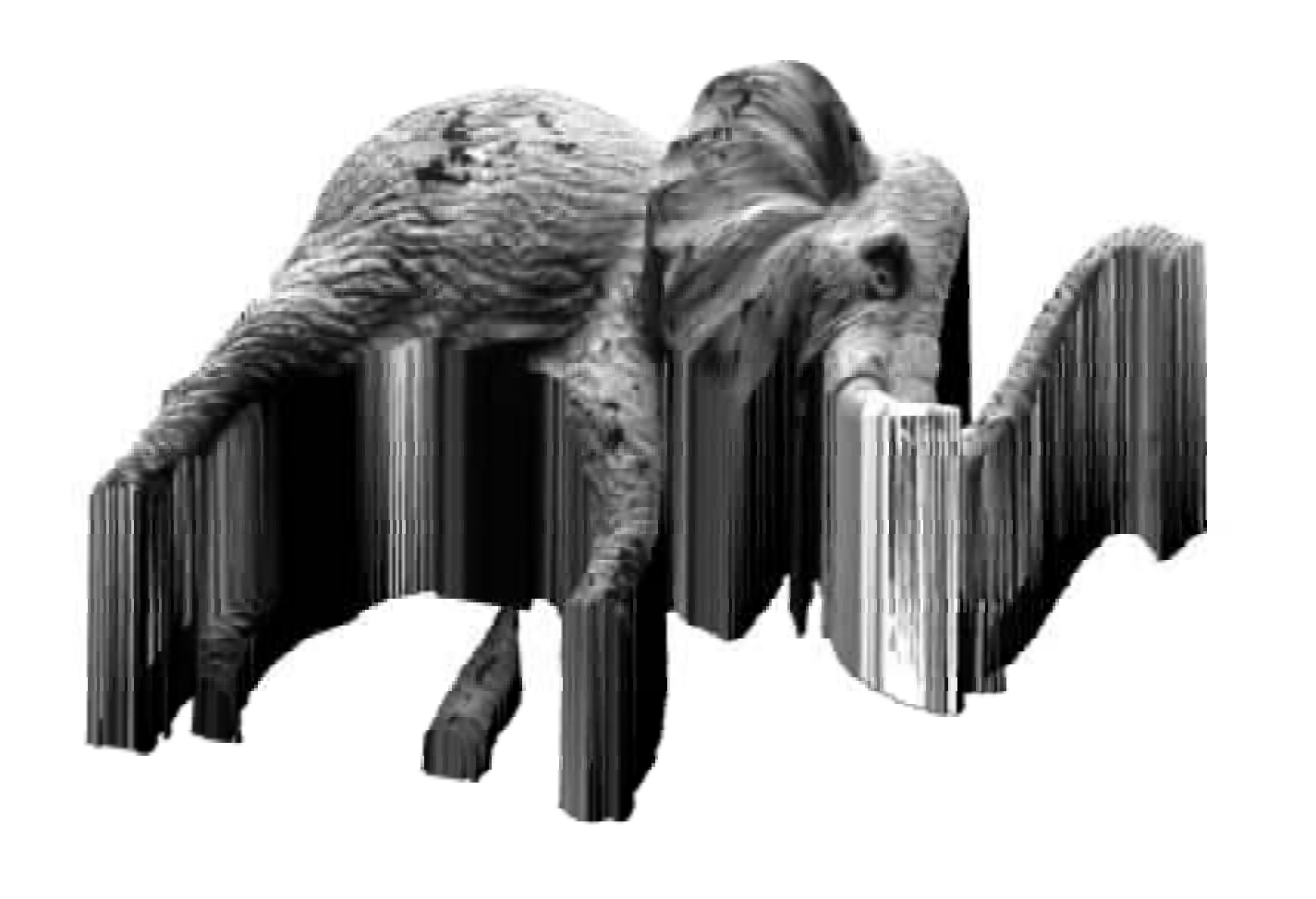}
\includegraphics[width=0.15\linewidth]{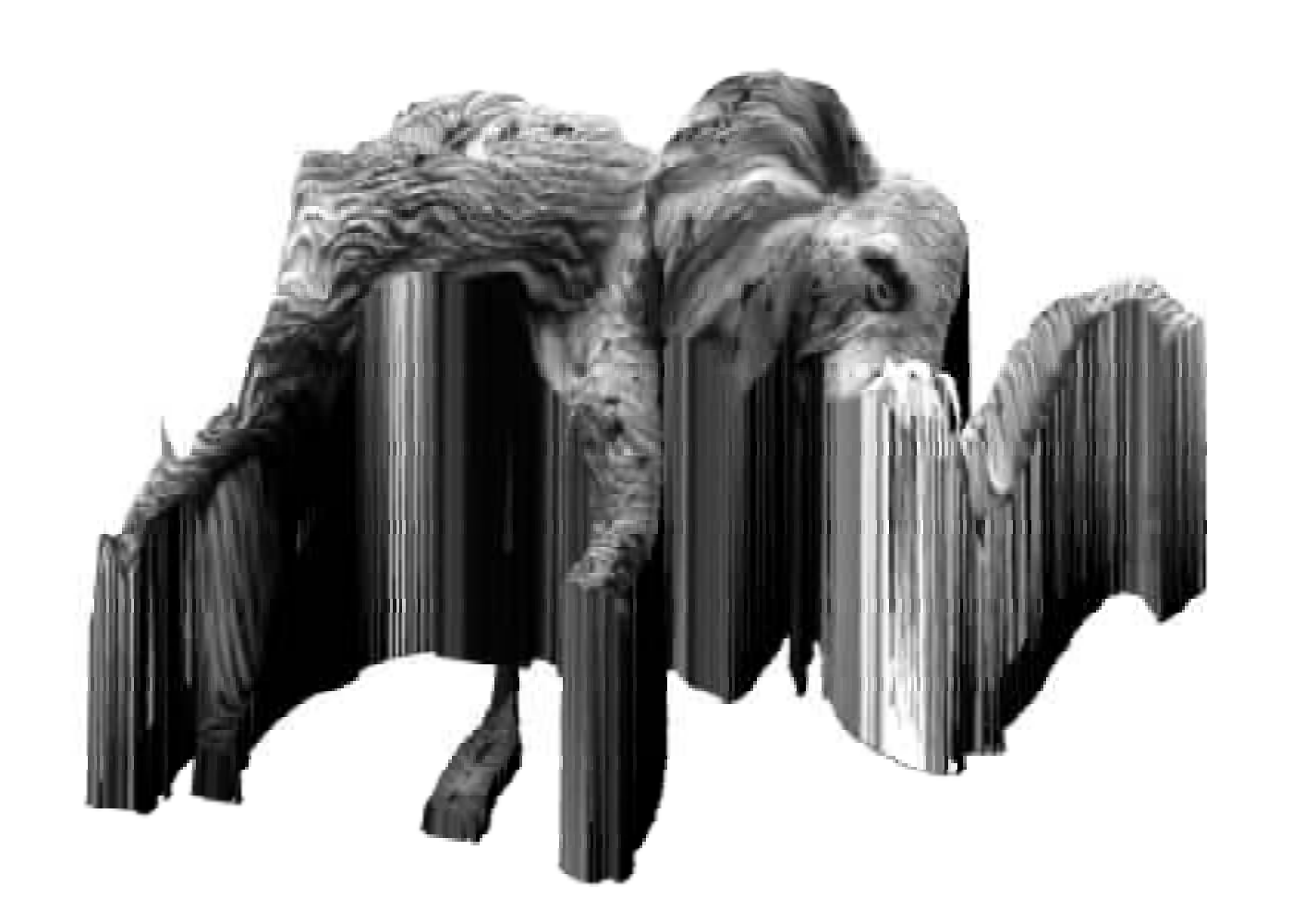}
\includegraphics[width=0.15\linewidth]{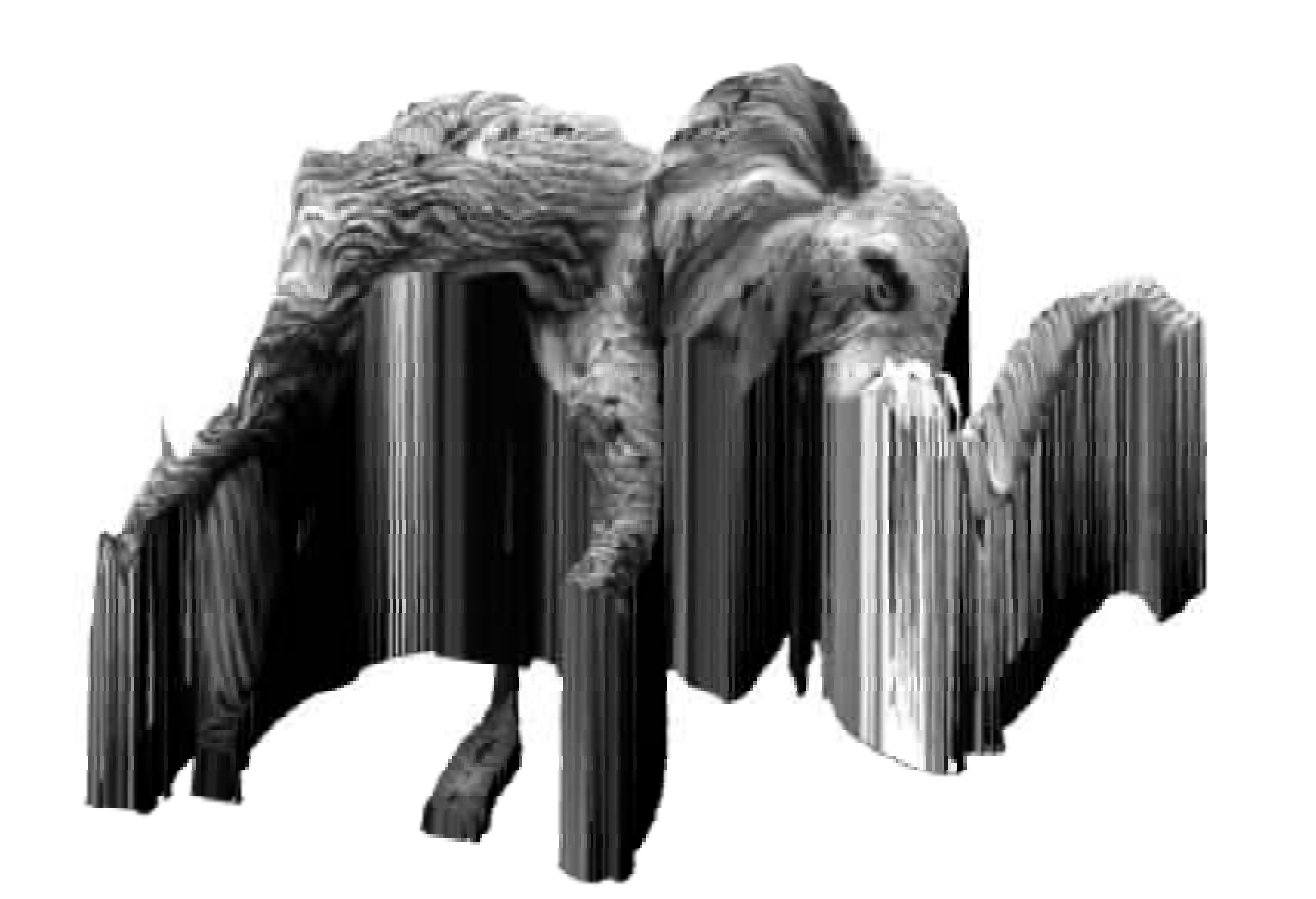} &

\includegraphics[width=0.15\linewidth]{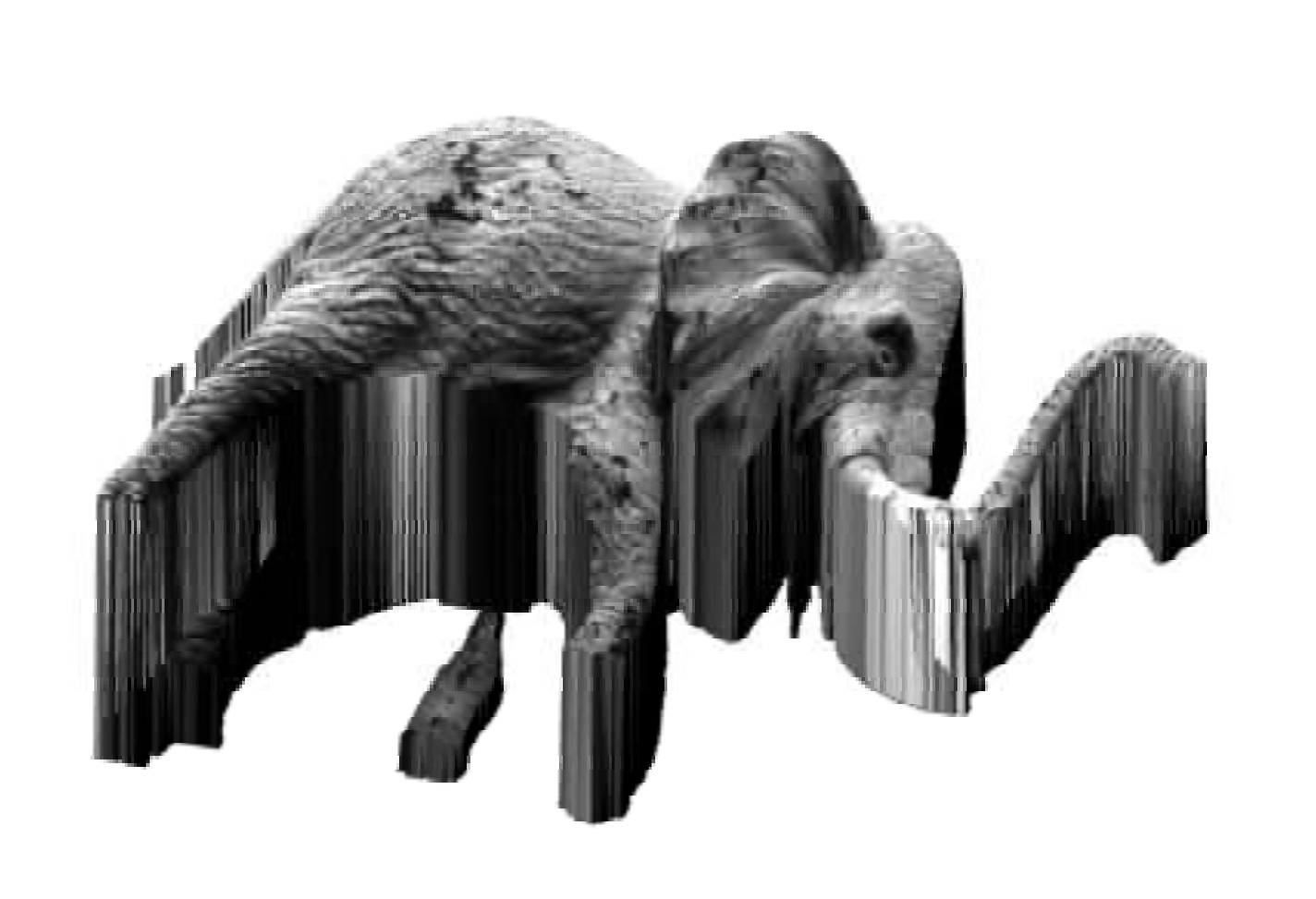}
\includegraphics[width=0.15\linewidth]{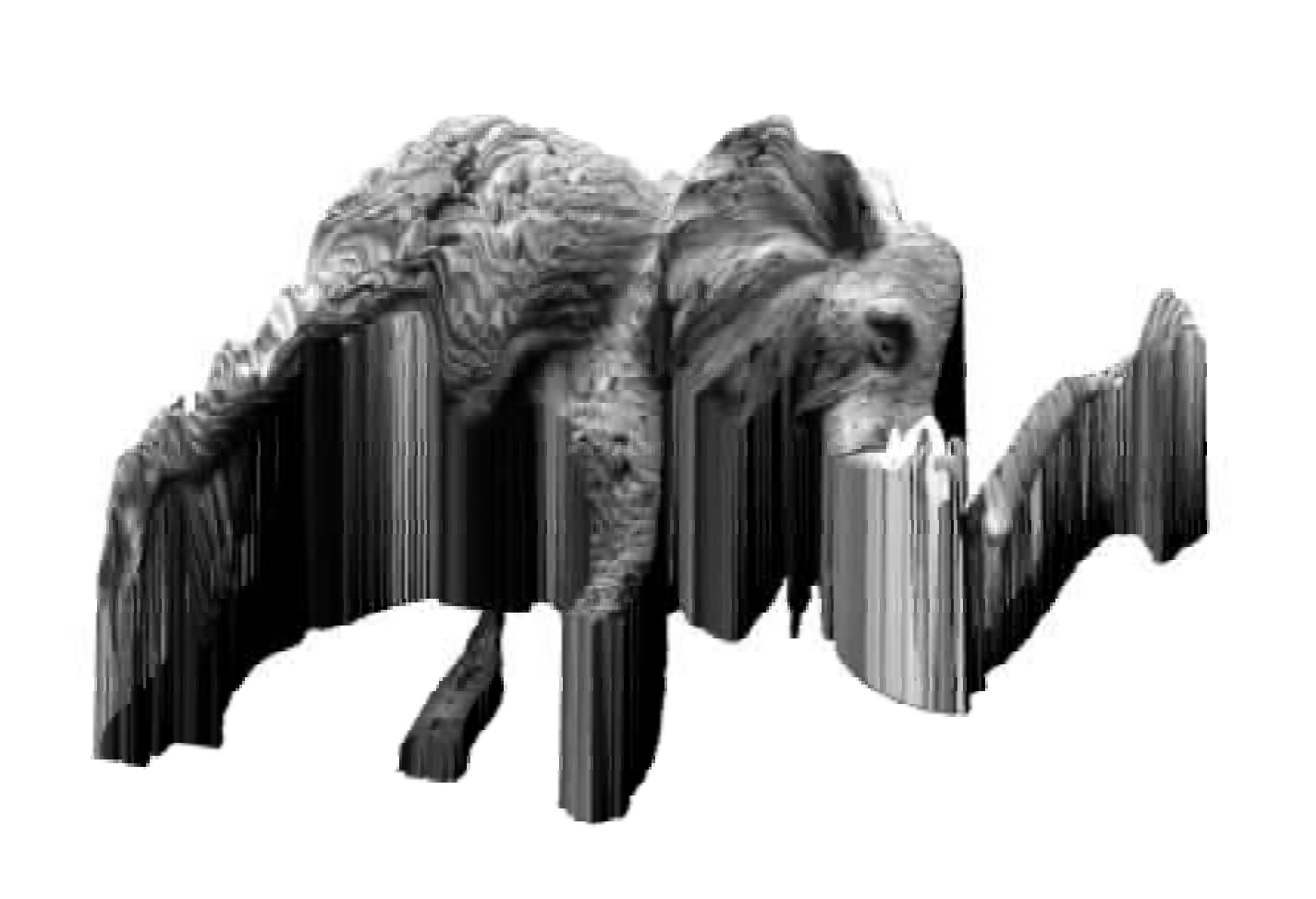}
\includegraphics[width=0.15\linewidth]{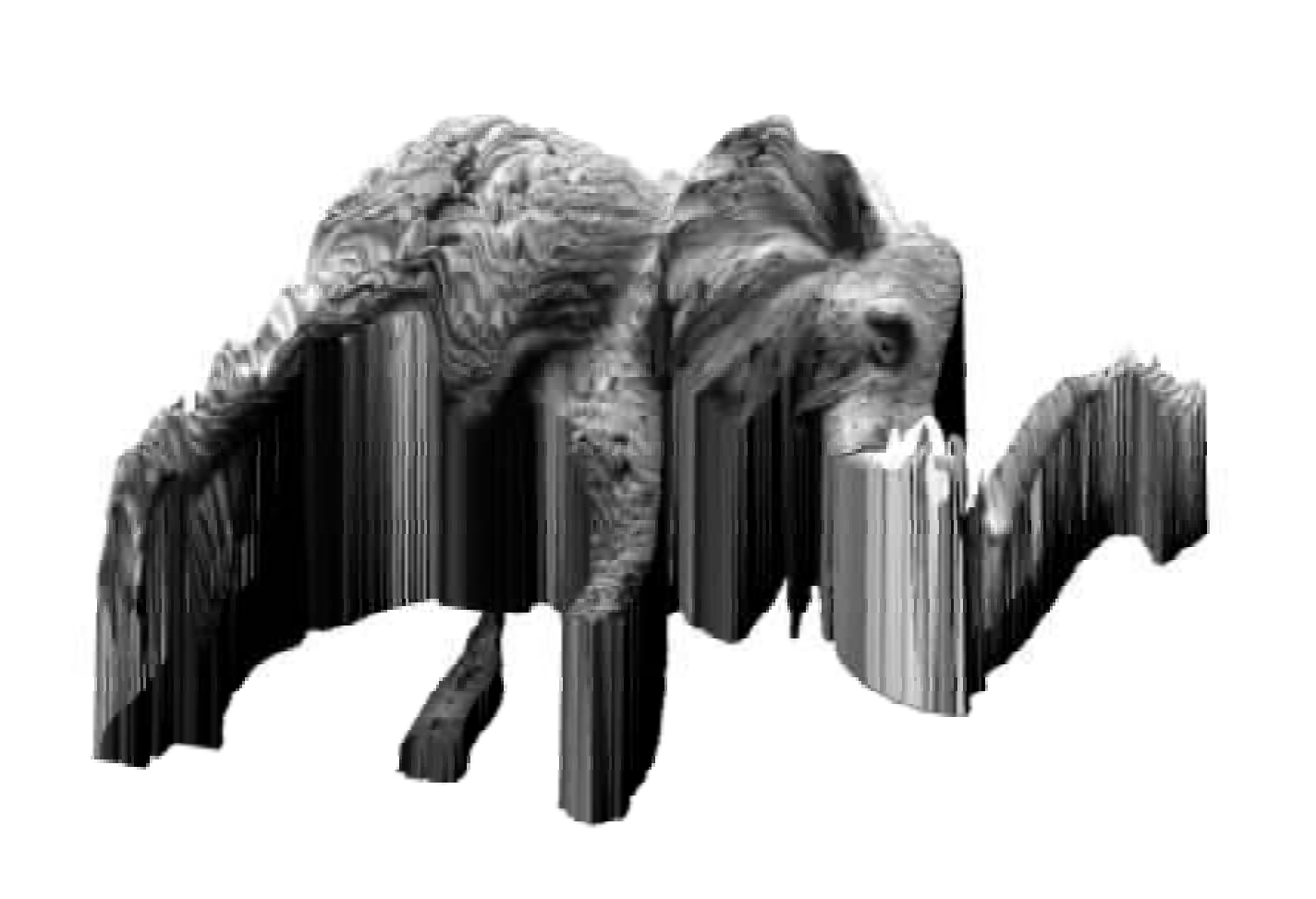} \\

\hline
\end{tabular}
\caption{Reconstructions from image pairs. Each box shows from left to right the laser scan, reconstruction with and without boundary conditions. The top row in each box shows the surface (colormap represents depth values) and the bottom row shows the surface painted with intensity values. Reconstructions are shown for images f and b ($7.5^\circ$)  achieving RMSE 2.14 and 2.12 (top left box) and  images b and f ($7.5^\circ$) achieving RMSE 1.93 and 1.94 (top right box)}
\end{figure}

\begin{figure}
\centering
\begin{tabular}{c|ccc}
GT (e) & d,e,f & c,d,e,f & c,d,e,f,g \\\hline
\includegraphics[width=.2\linewidth]{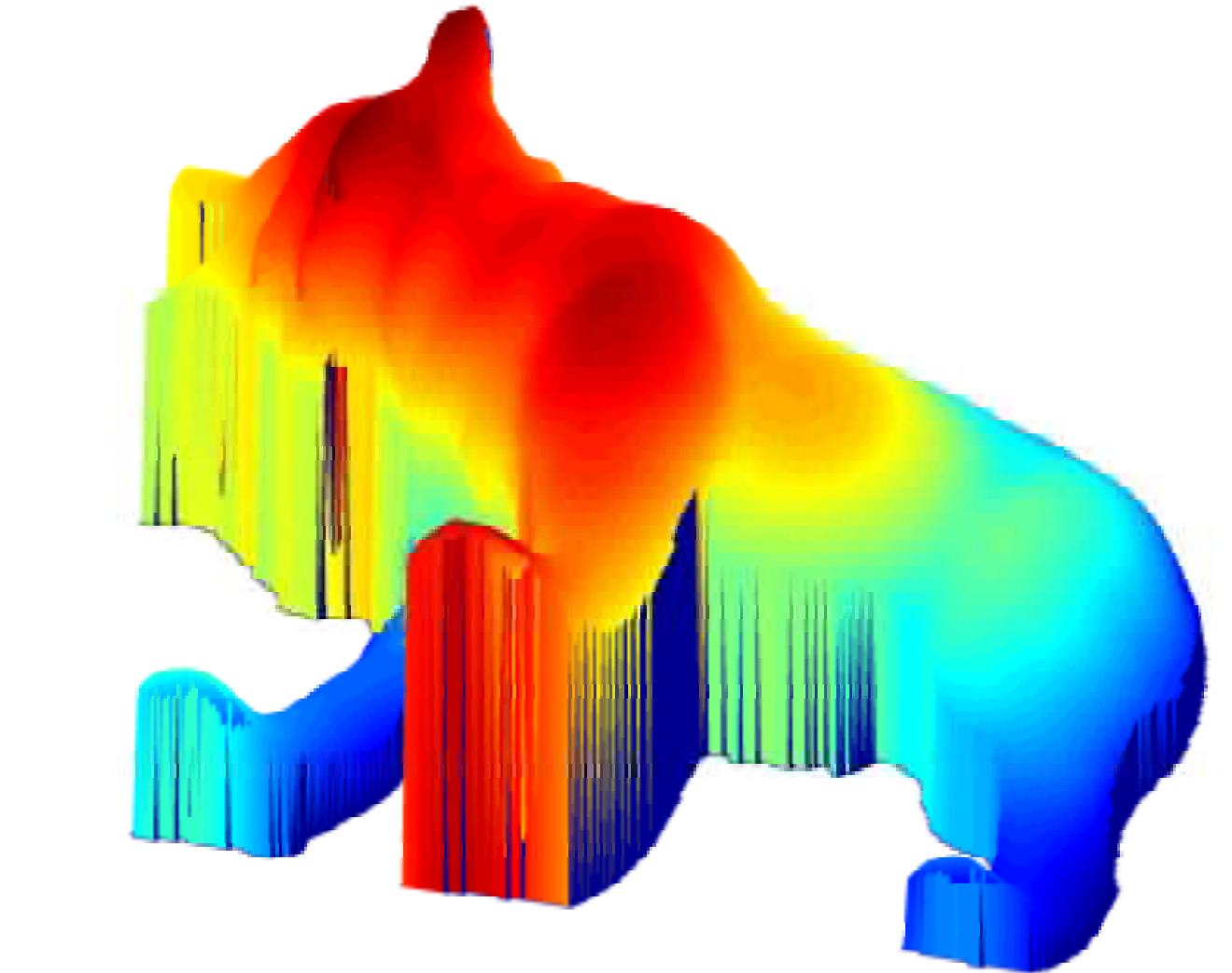}&
\includegraphics[width=.2\linewidth]{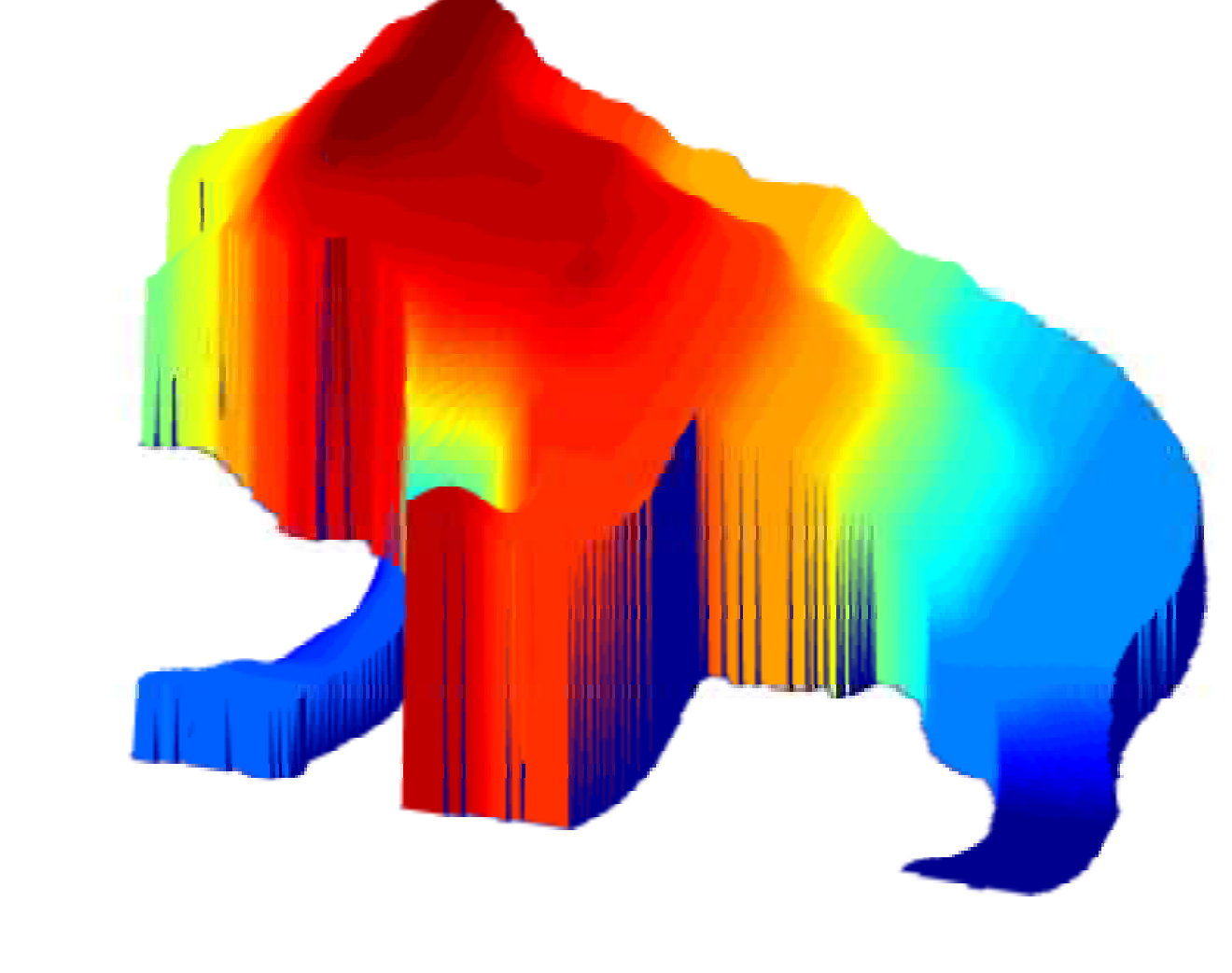}&
\includegraphics[width=.2\linewidth]{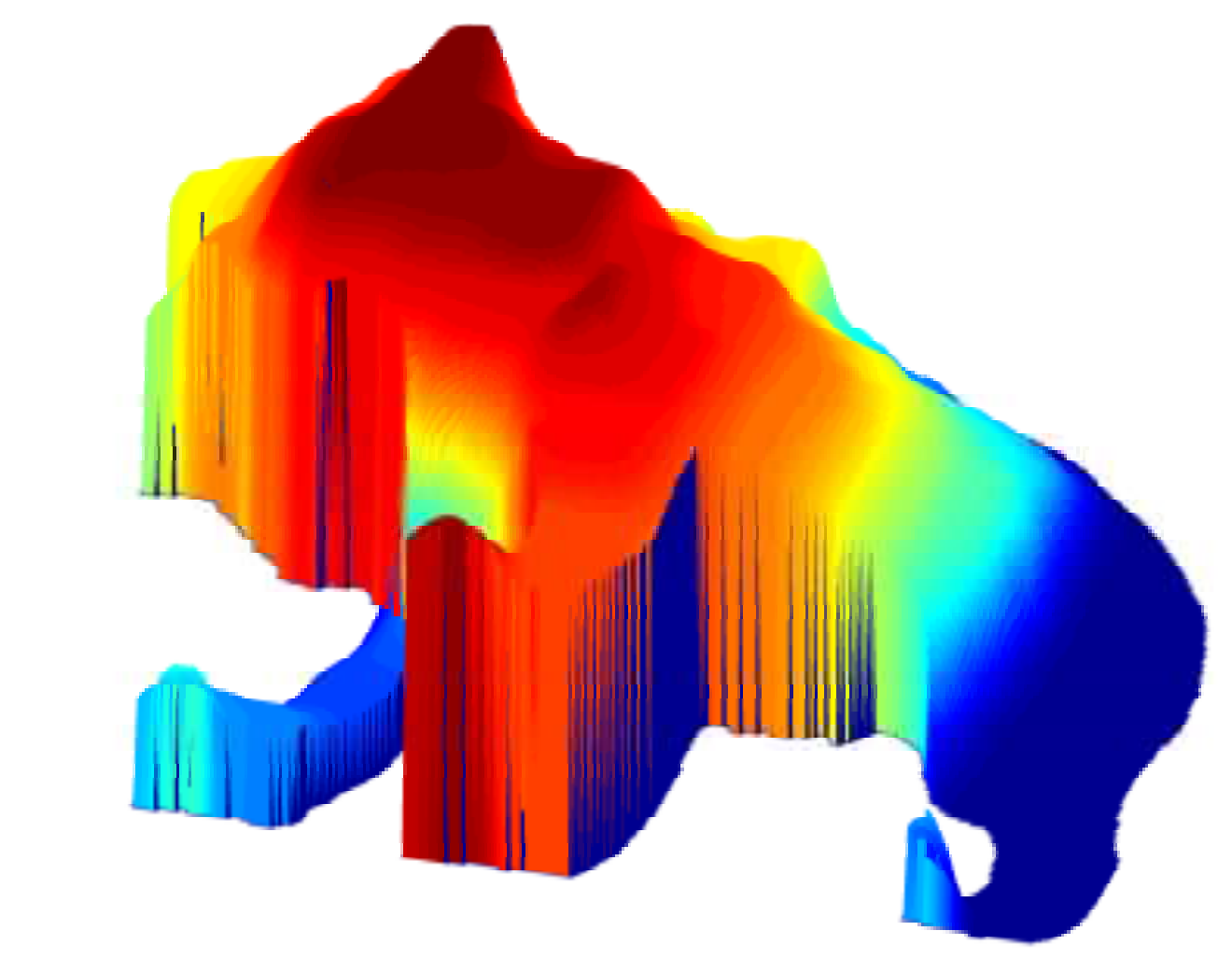}&
\includegraphics[width=.2\linewidth]{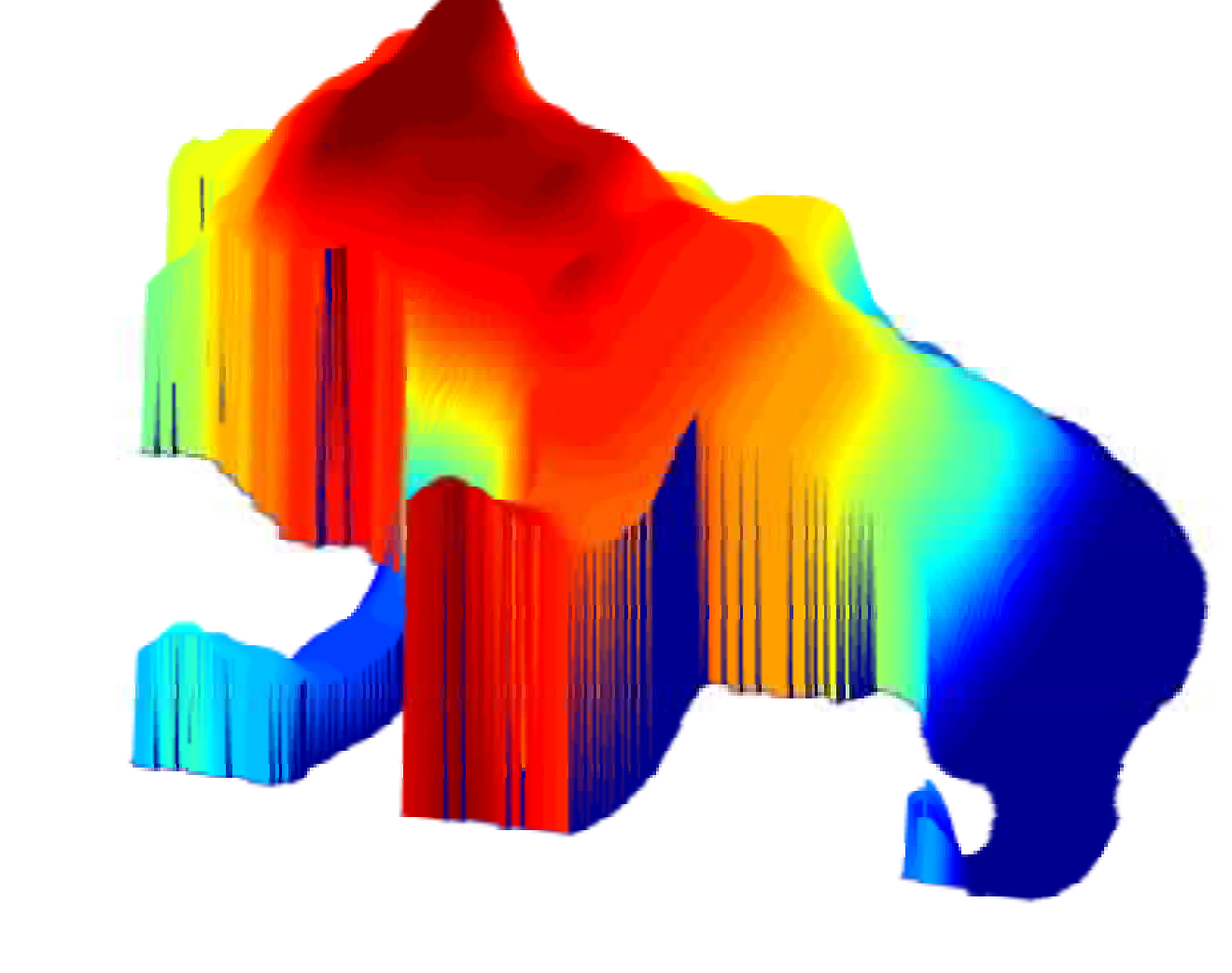}\\
\includegraphics[width=.2\linewidth]{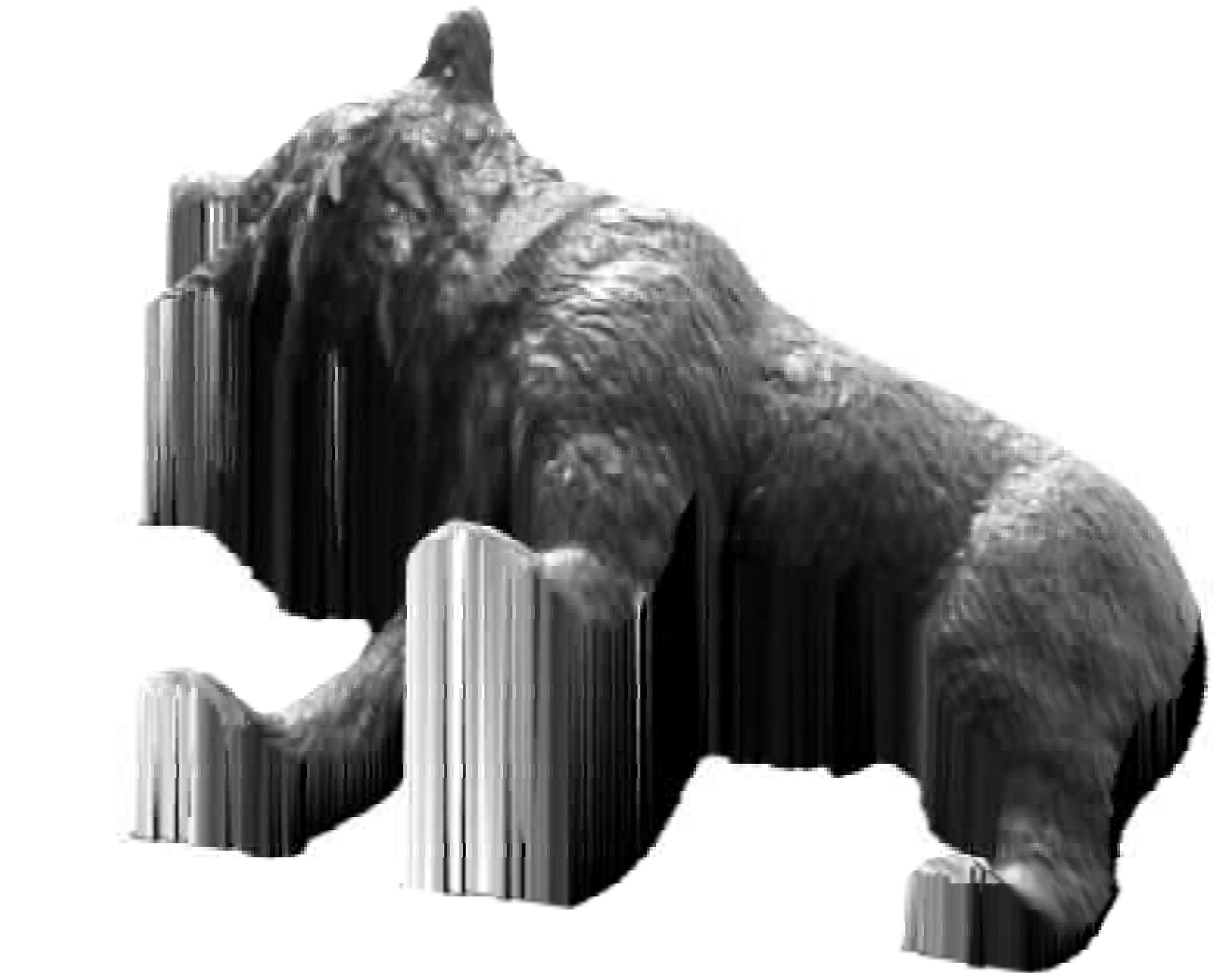}&
\includegraphics[width=.2\linewidth]{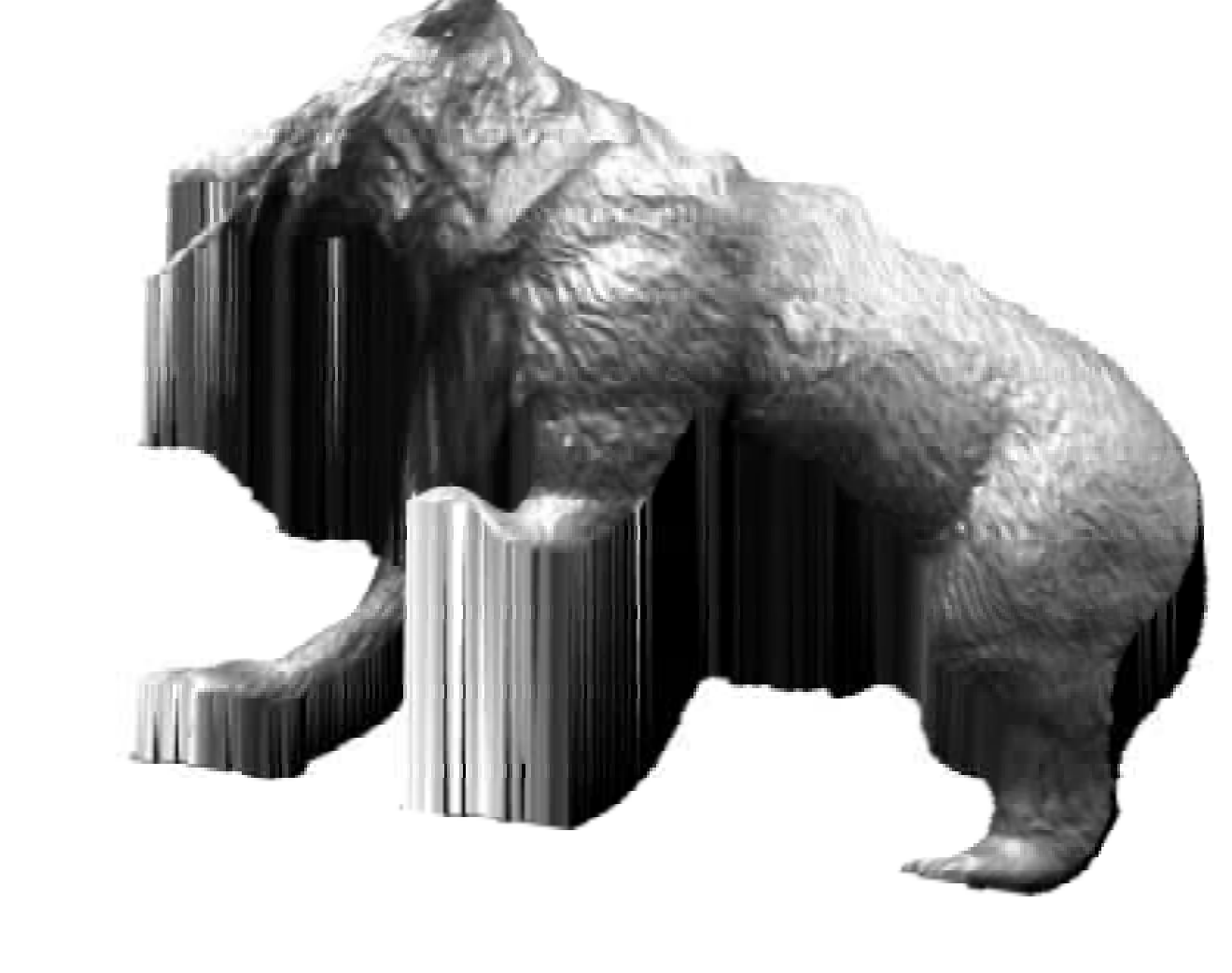}&
\includegraphics[width=.2\linewidth]{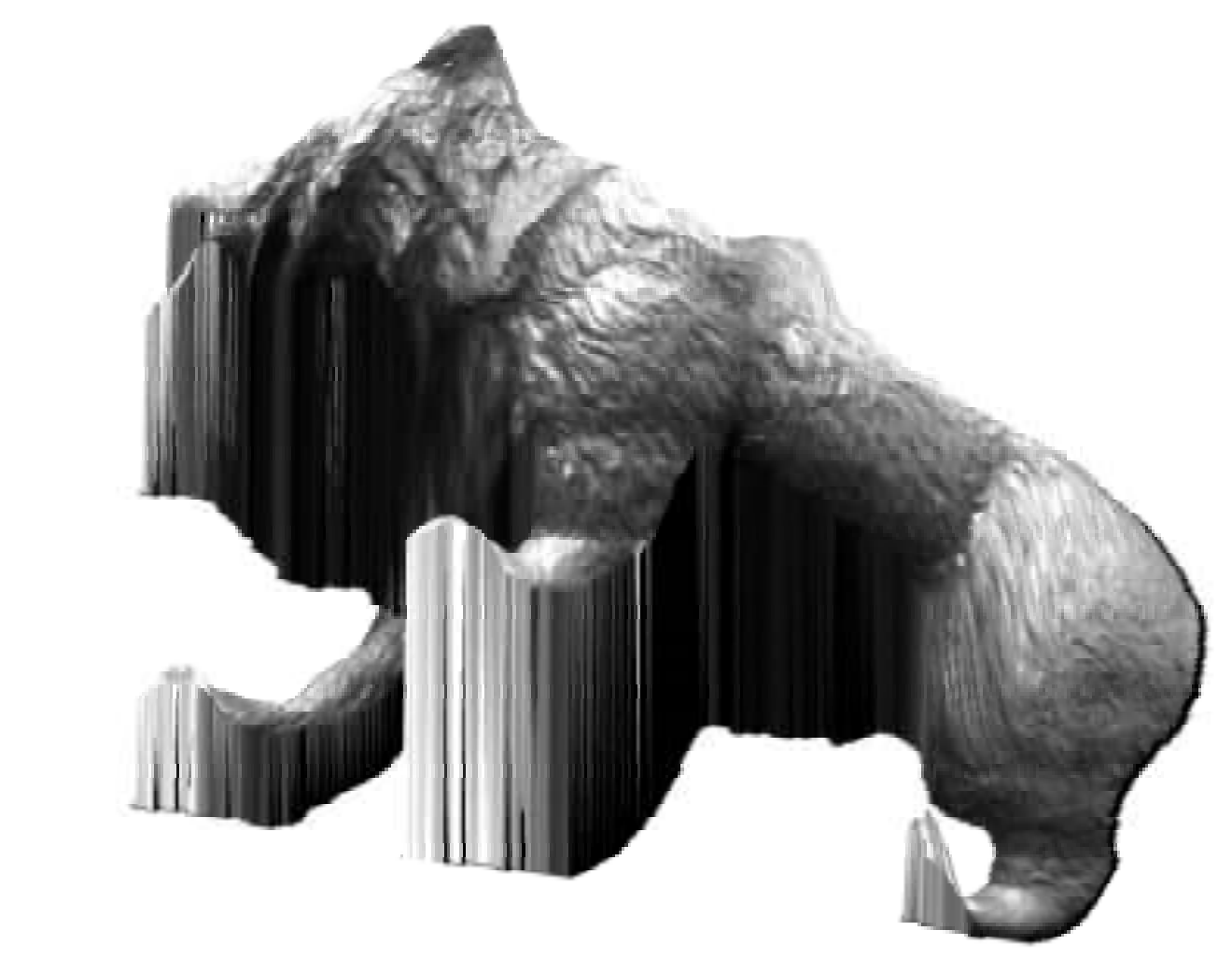}&
\includegraphics[width=.2\linewidth]{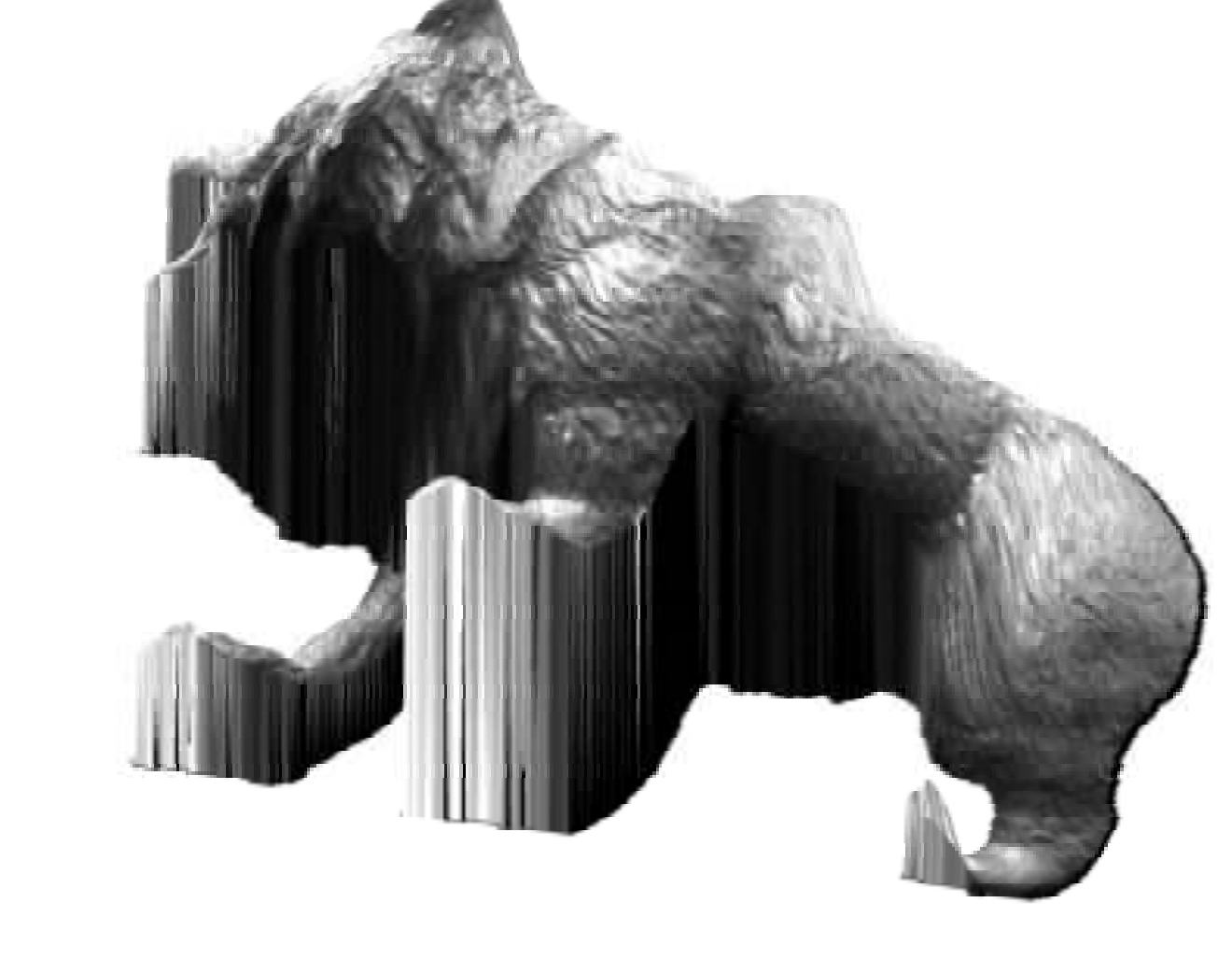}\\\hline
RMSE: & 0.88 & 1.28 & 1.39 \\\hline\hline
% \end{tabular}
%\caption{ Reconstructing Bear, as seen from image e. Resulting RMSE in correspondence (over all images) is shown at the bottom.}
%\label{fig:mi-bear-e1}
%\end{figure}
%
%\begin{figure}
%\centering
%\begin{tabular}{c|ccc}
GT (e) & d,e,g & c,d,e,g & d,e,f,g \\\hline
\includegraphics[width=.2\linewidth]{lighting/BearMI_tiff_3_78__gt_c_r.pdf}&
\includegraphics[width=.2\linewidth]{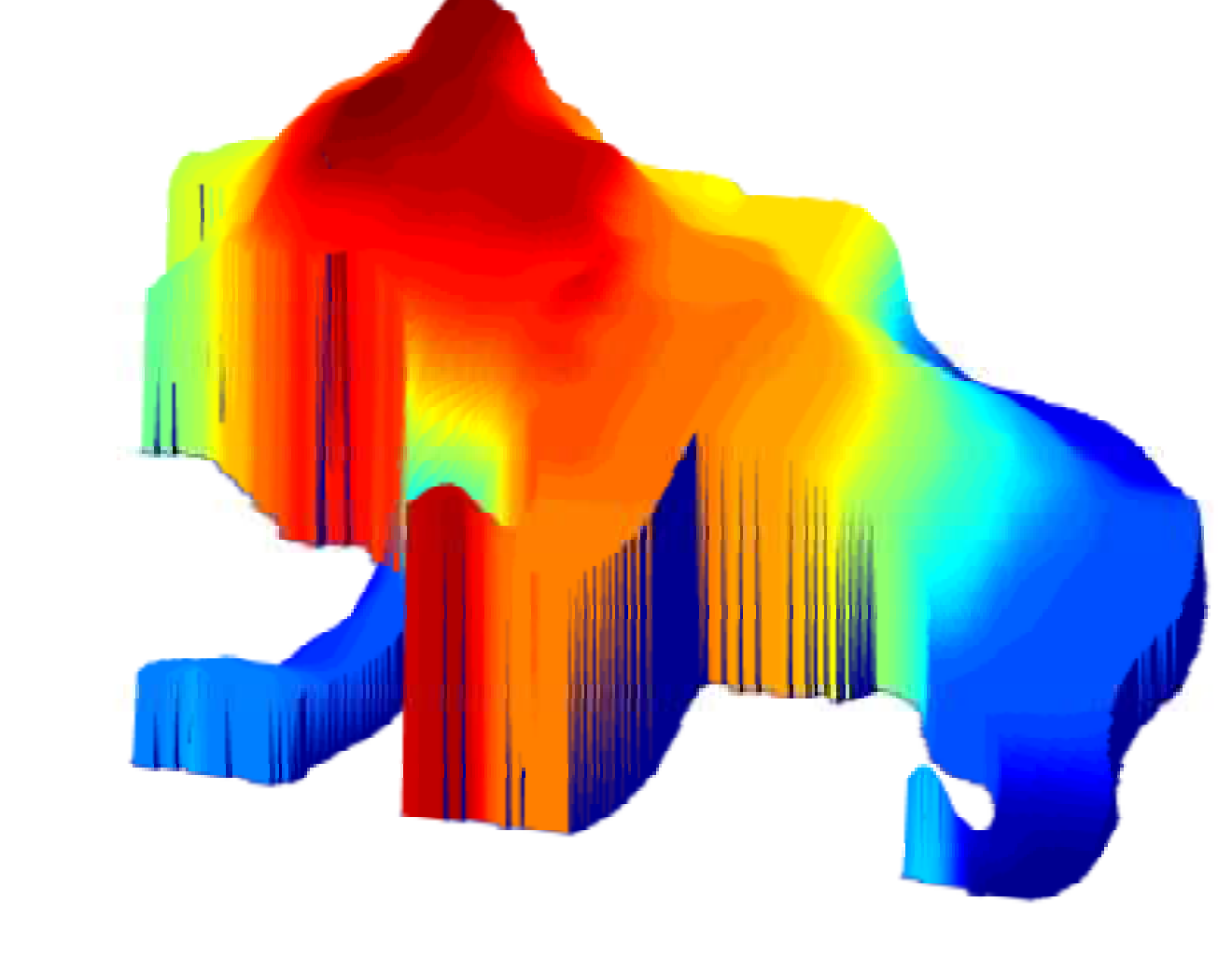}&
\includegraphics[width=.2\linewidth]{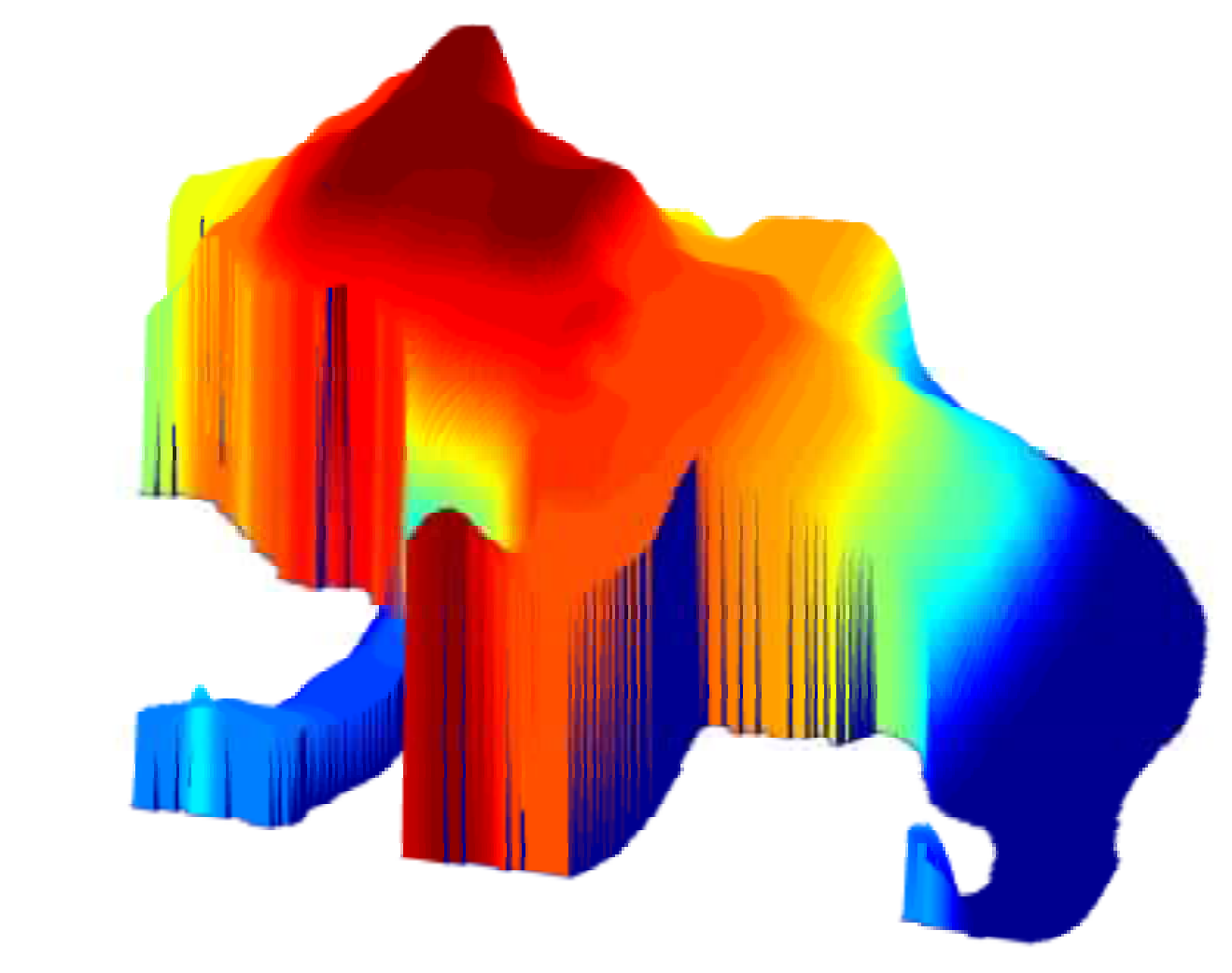}&
\includegraphics[width=.2\linewidth]{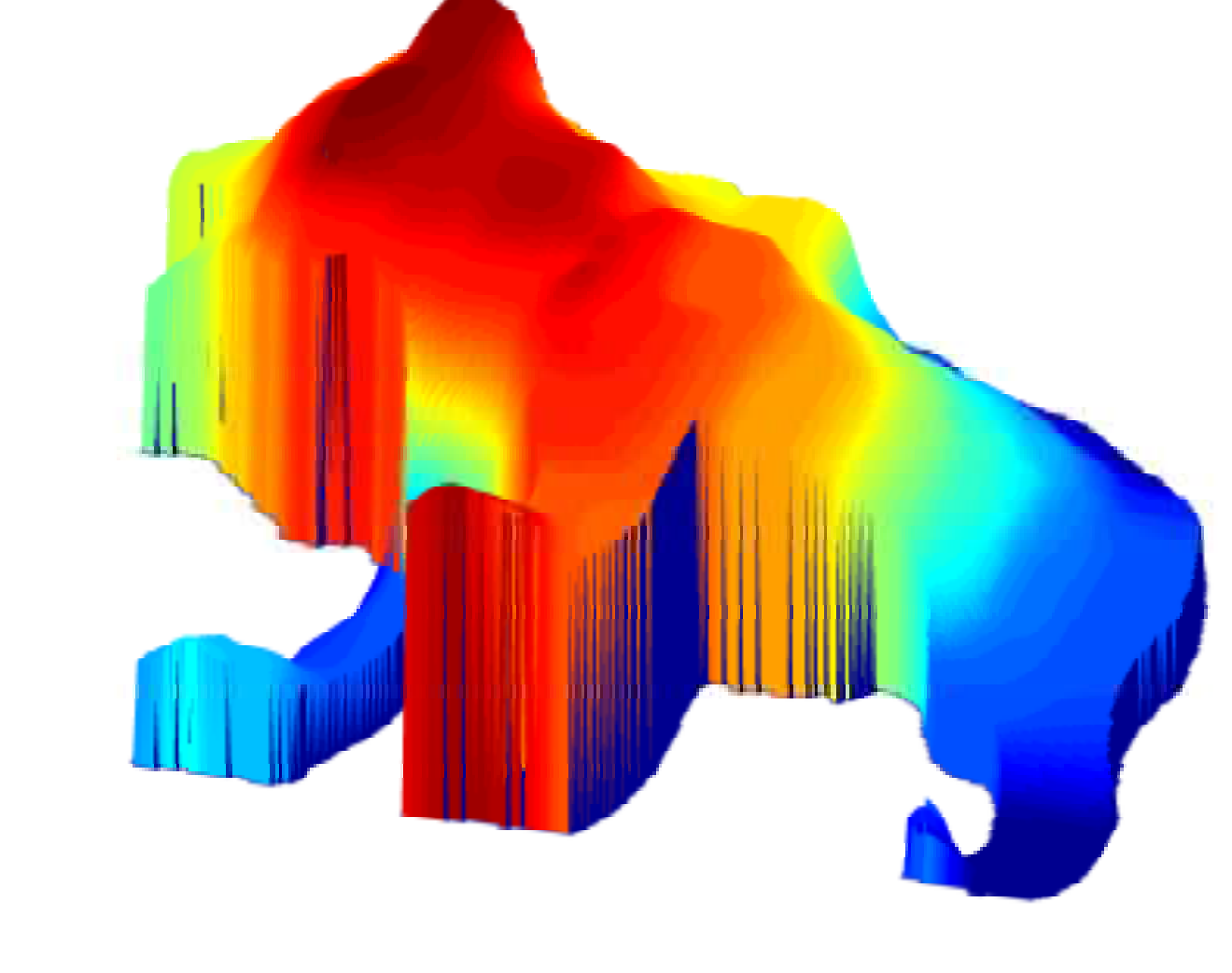}\\
\includegraphics[width=.2\linewidth]{lighting/BearMI_tiff_3_78__gt_map_c_r.pdf}&
\includegraphics[width=.2\linewidth]{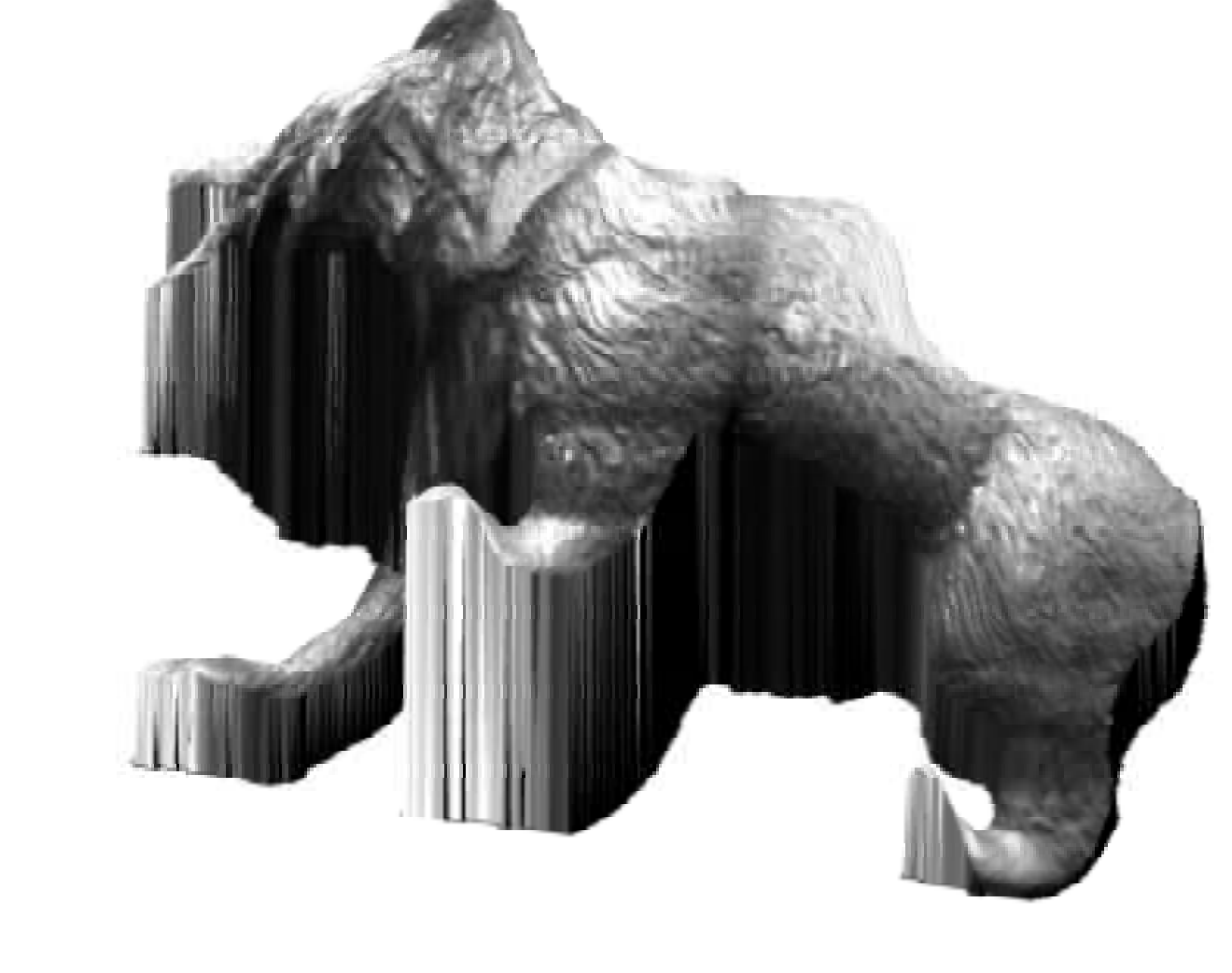}&
\includegraphics[width=.2\linewidth]{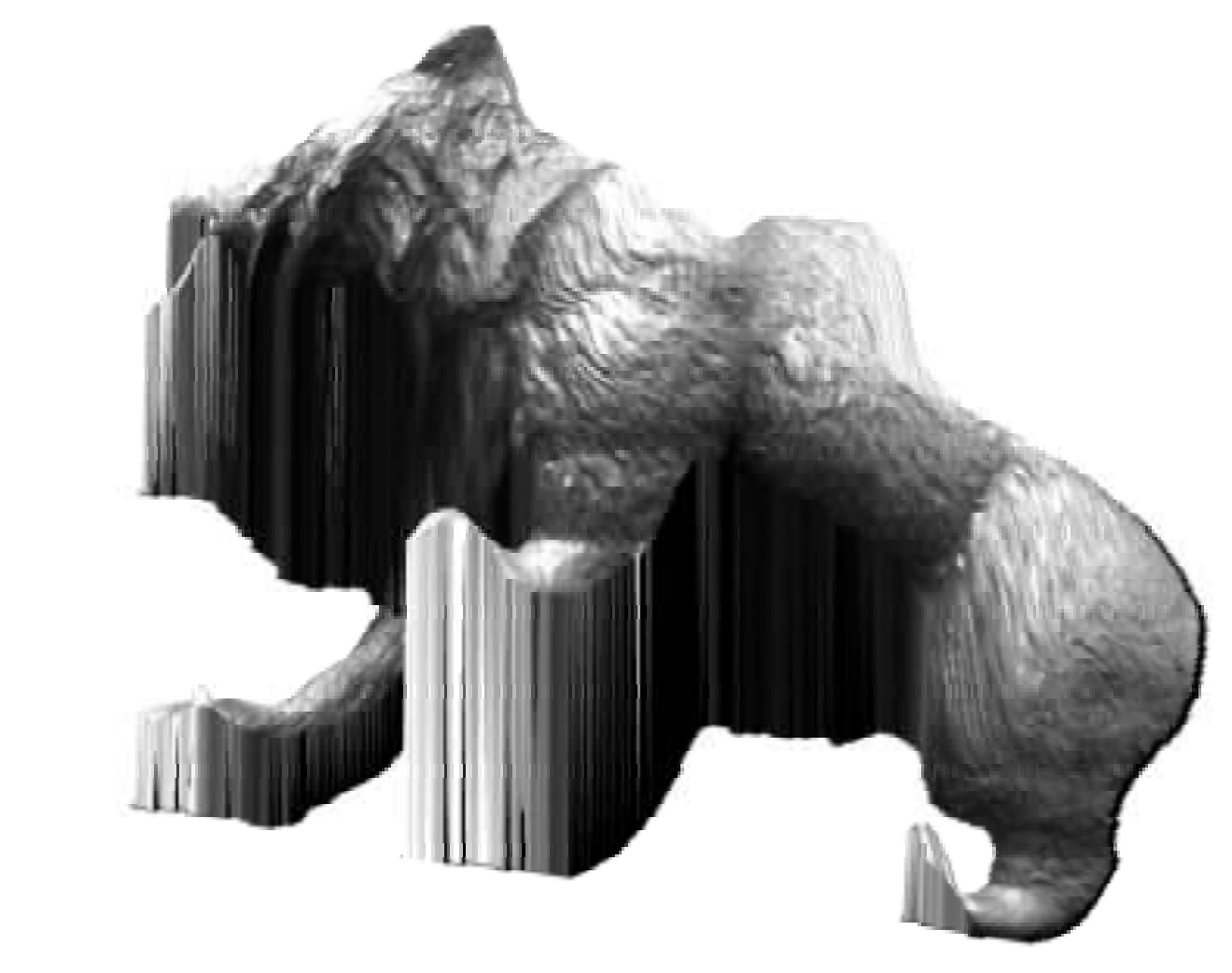}&
\includegraphics[width=.2\linewidth]{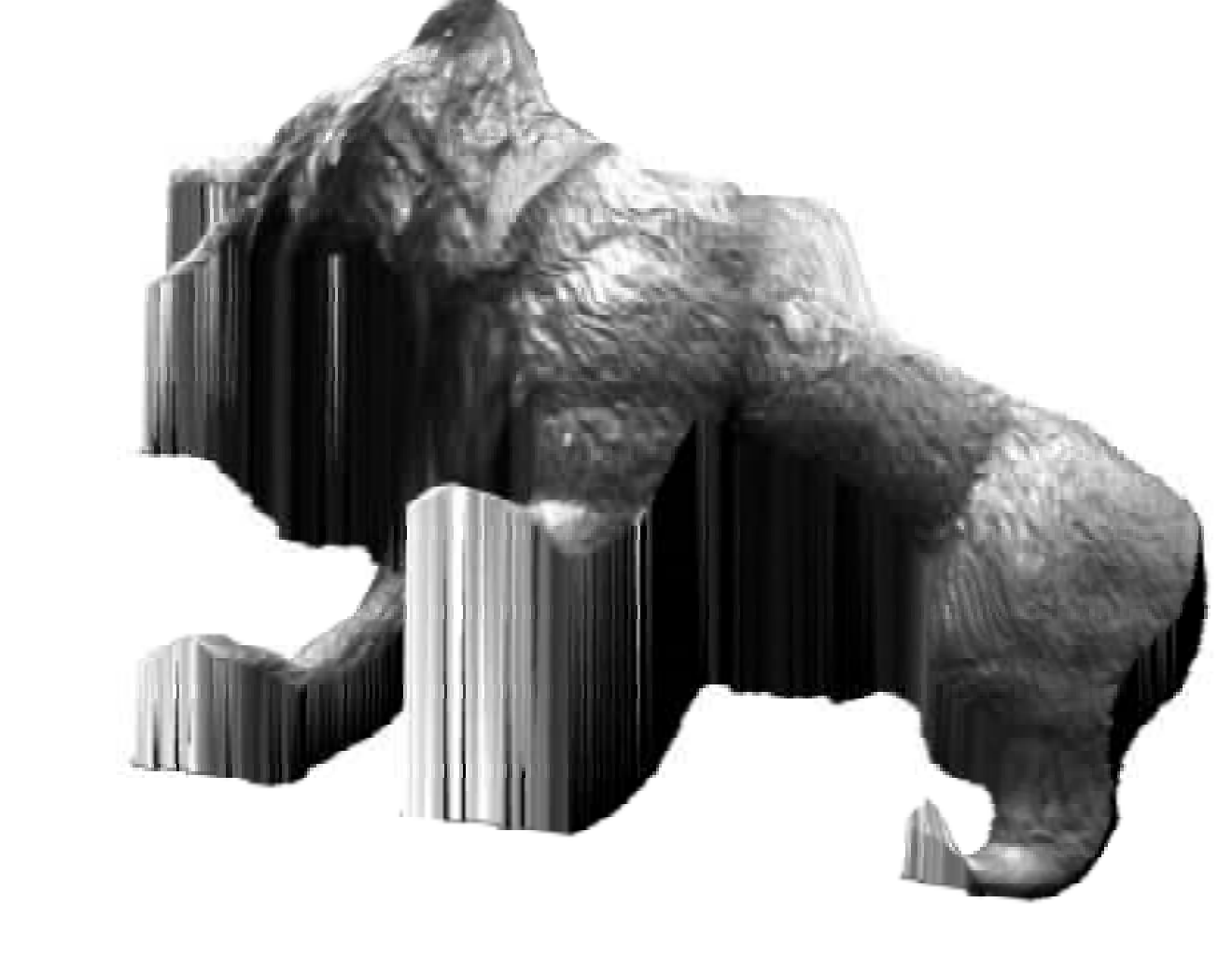}\\\hline
RMSE: & 0.88 & 1.40 & 0.83 \\
\end{tabular}
\caption{Reconstruction from multiple images. Ground truth is shown at left, image labels at the top and RMSE (over all images) at the bottom.}
\label{fig:mi-bear-e2}
\end{figure}

\begin{figure}
\centering
\begin{tabular}{c|ccc}
GT (b) & b,c,f & b,c,d,f & a,b,c,d,e,f \\\hline
\includegraphics[width=.2\linewidth]{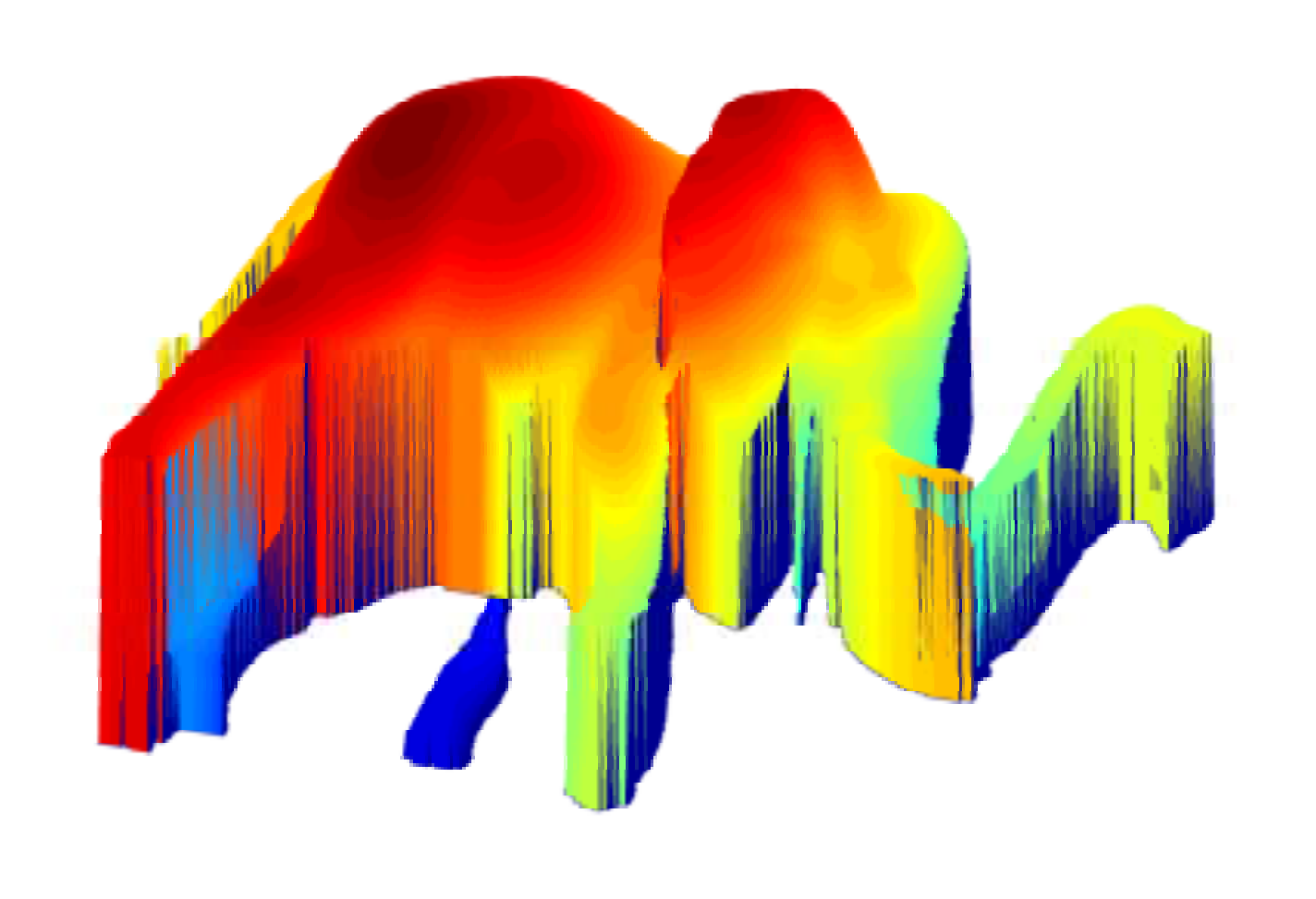}&
\includegraphics[width=.2\linewidth]{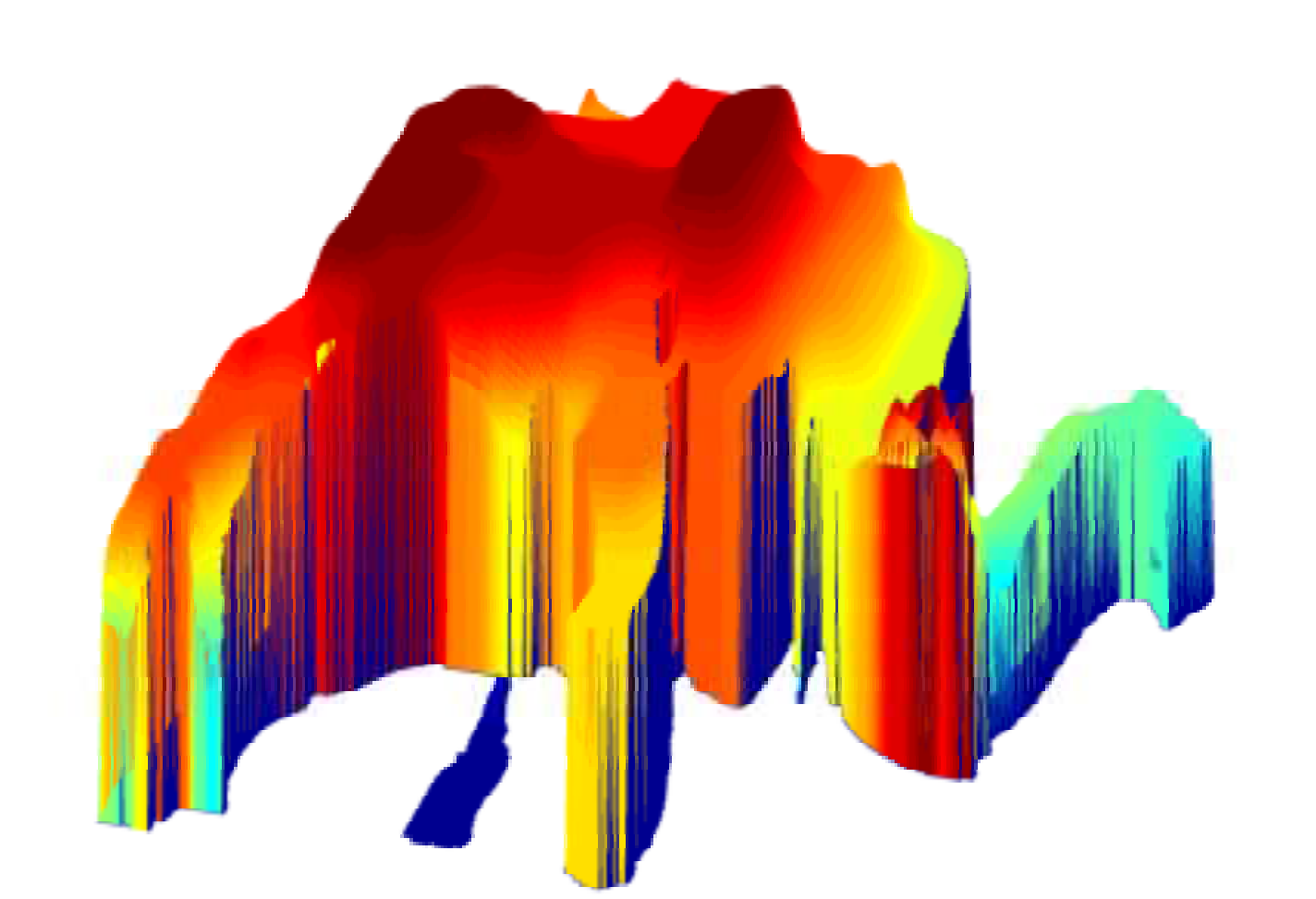}&
\includegraphics[width=.2\linewidth]{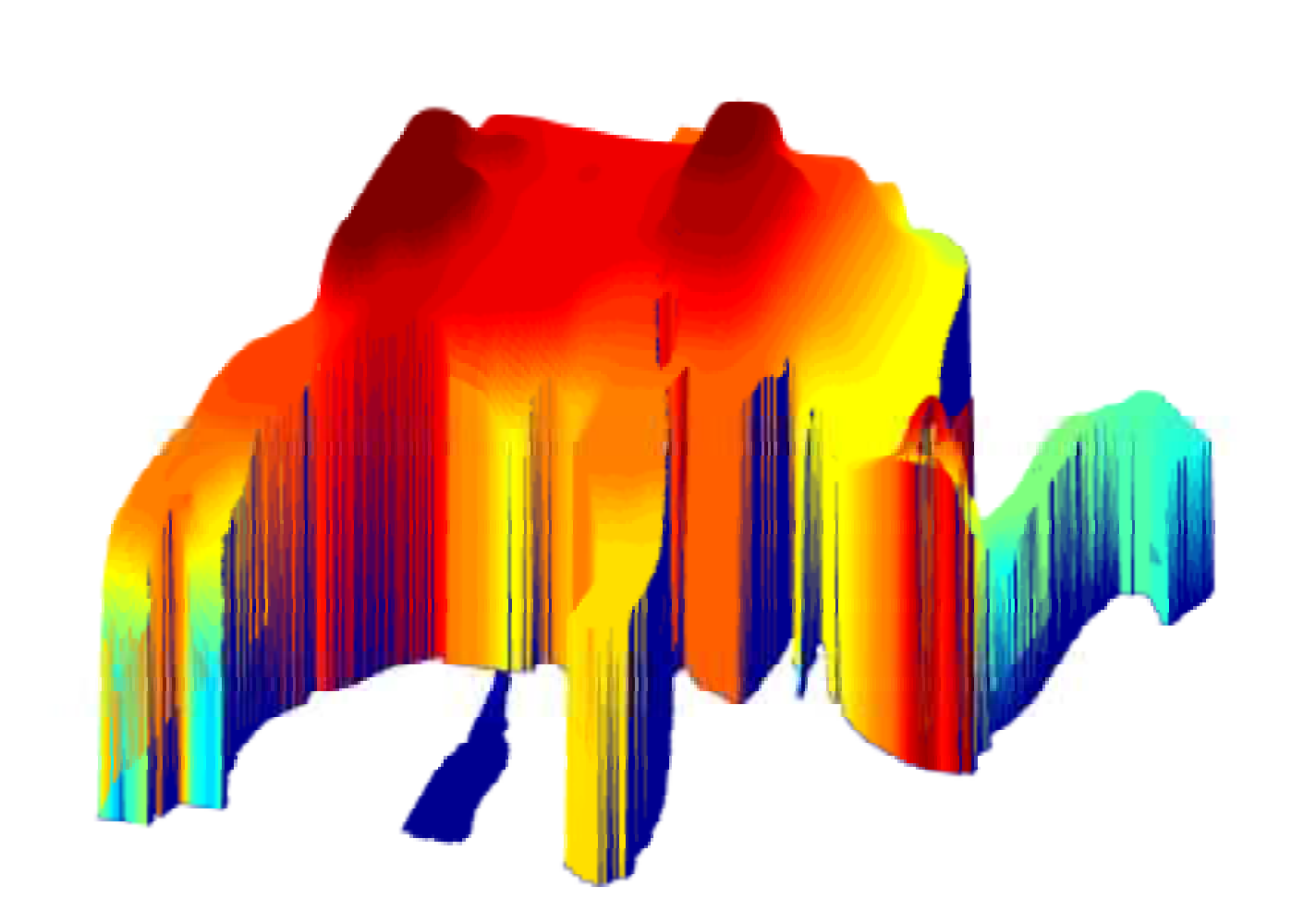}&
\includegraphics[width=.2\linewidth]{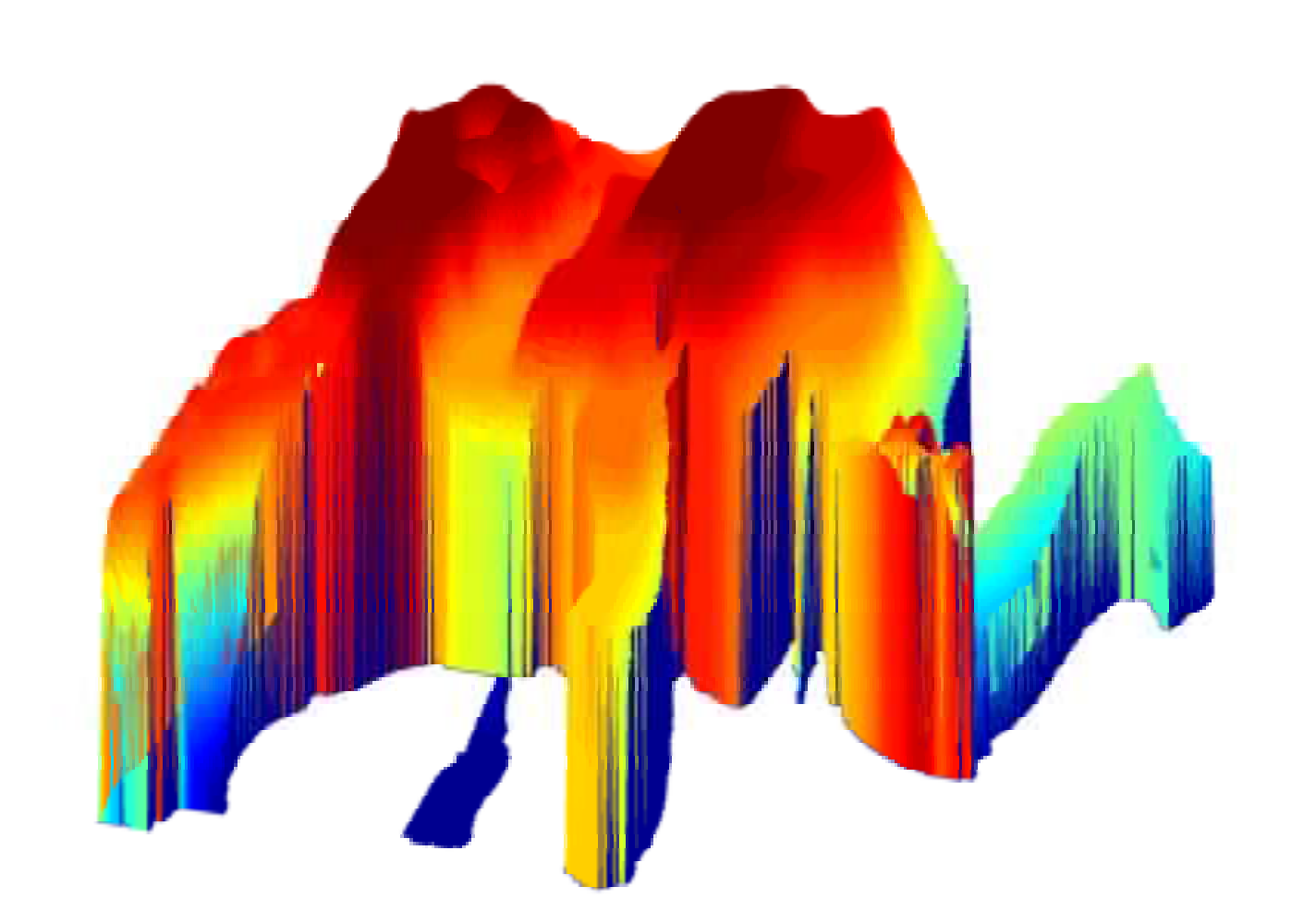}\\
\includegraphics[width=.2\linewidth]{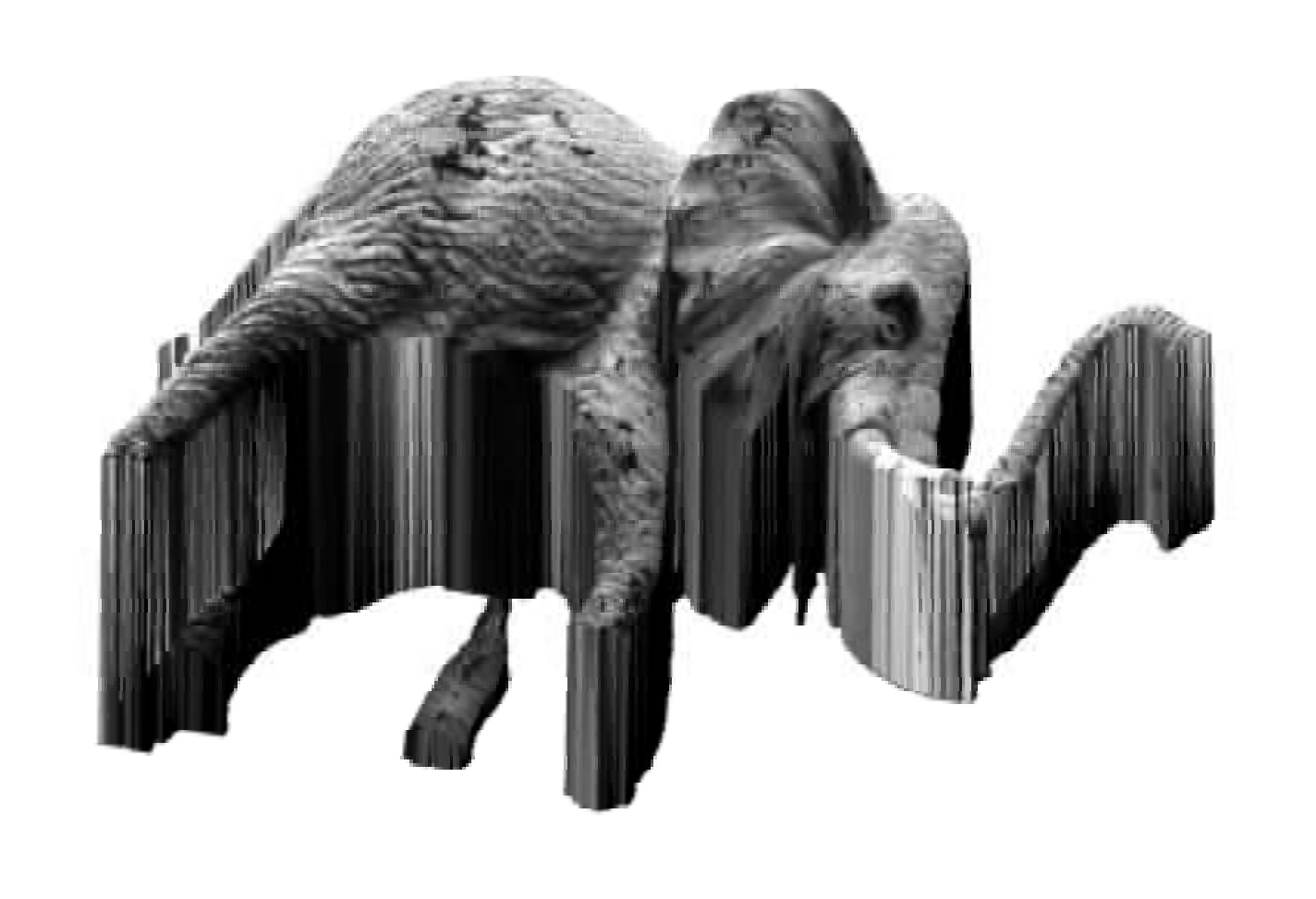}&
\includegraphics[width=.2\linewidth]{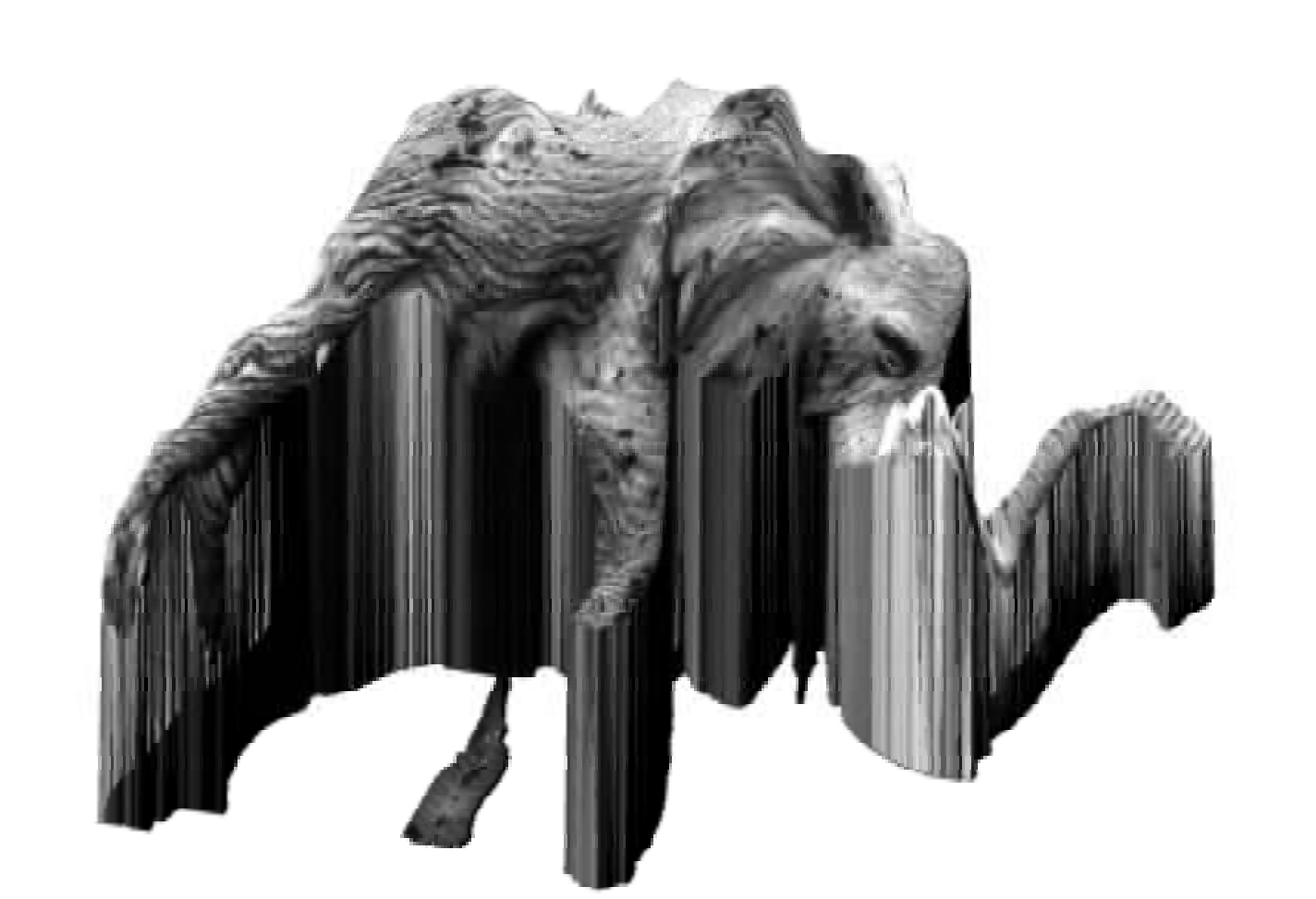}&
\includegraphics[width=.2\linewidth]{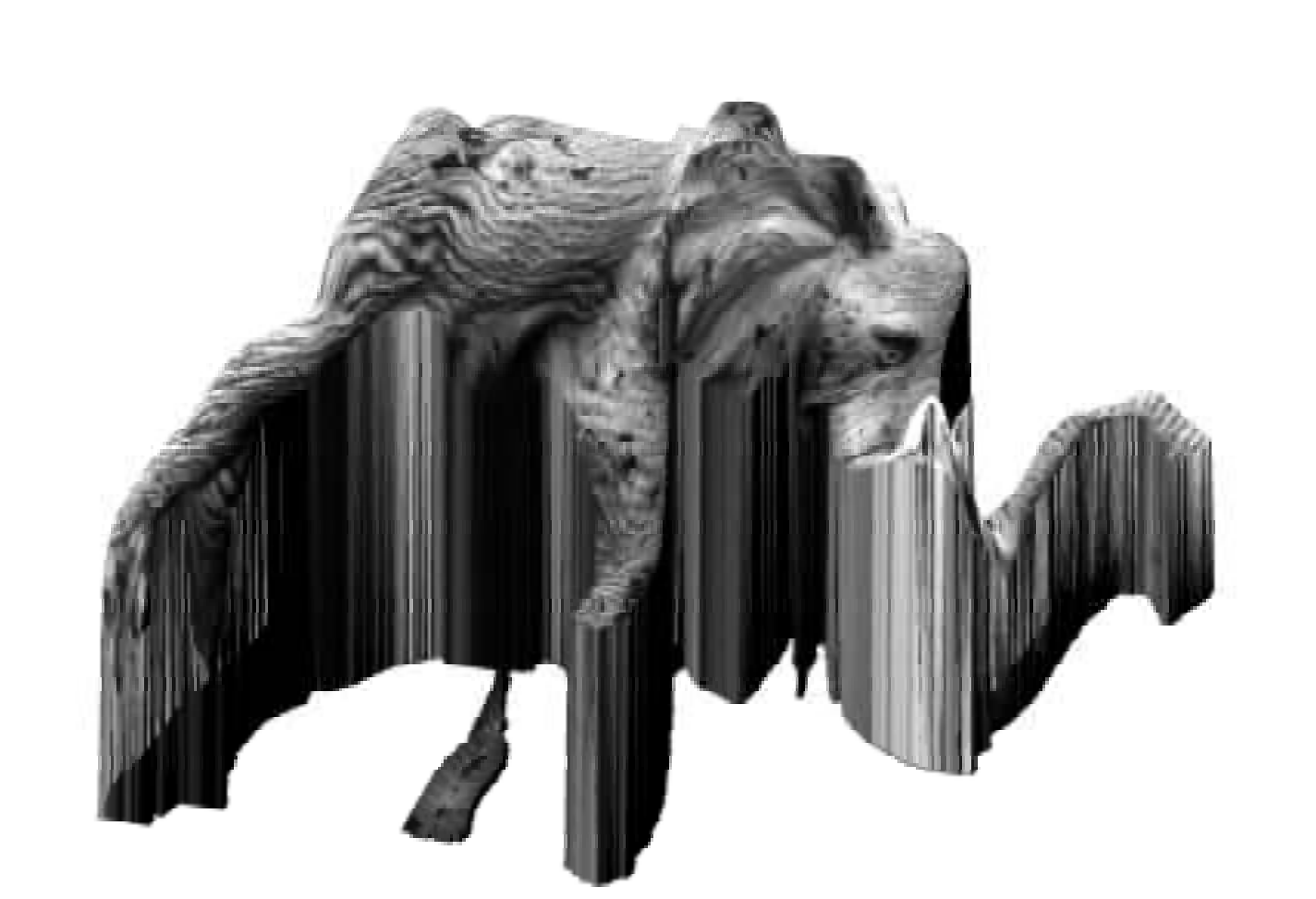}&
\includegraphics[width=.2\linewidth]{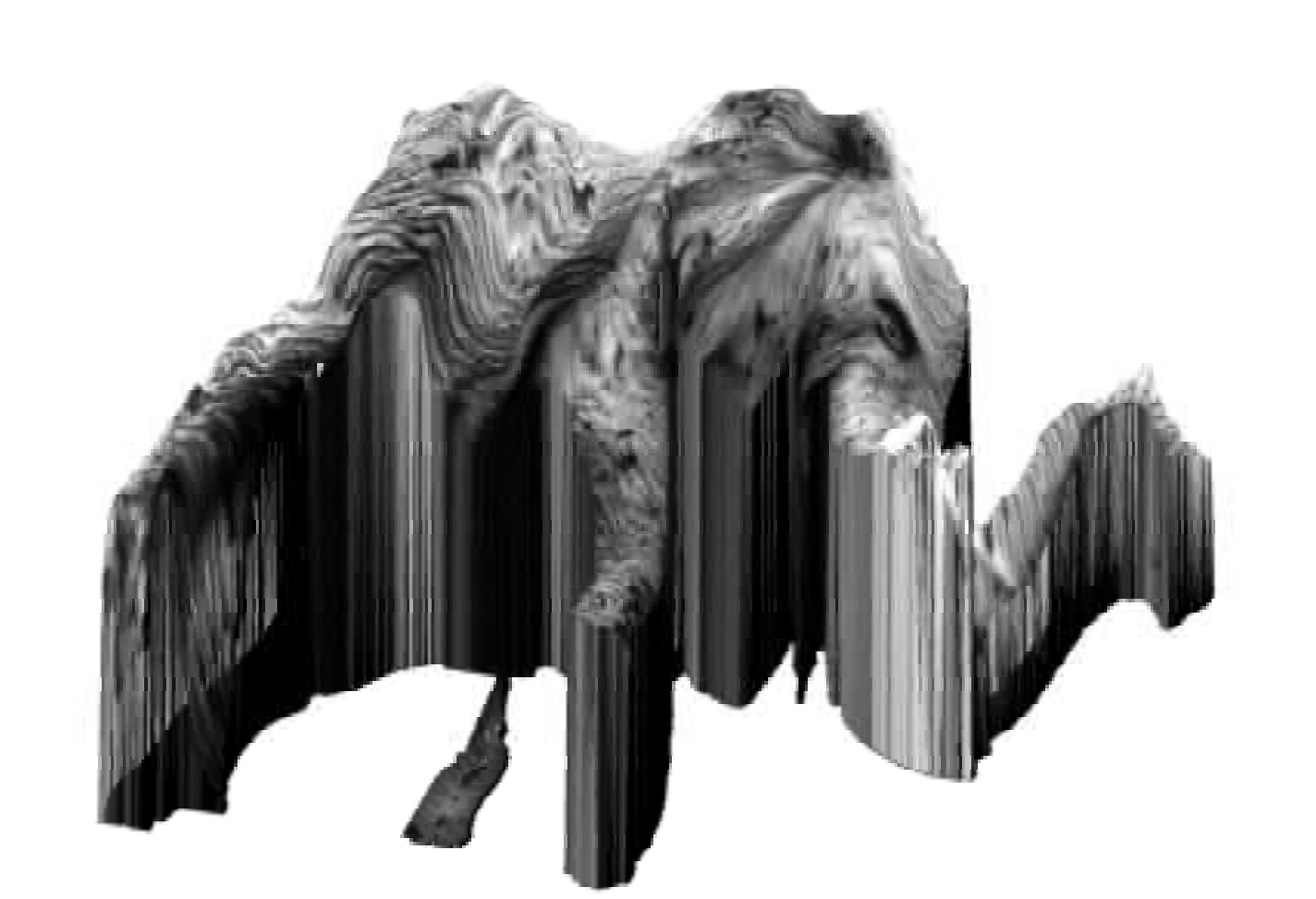}\\\hline
Correspondence error:& 1.78 & 1.44 & 1.69\\
\end{tabular}
\caption{Reconstruction from multiple images. Ground truth is shown at left, image labels at the top and RMSE (over all images) at the bottom.}
\label{fig:mi-elephant-b}
\end{figure}

\begin{figure}
\centering
\begin{tabular}{c|cc}
GT (d) & b,c,d,e,f & a,b,c,d,e,f \\\hline
\includegraphics[width=.23\linewidth]{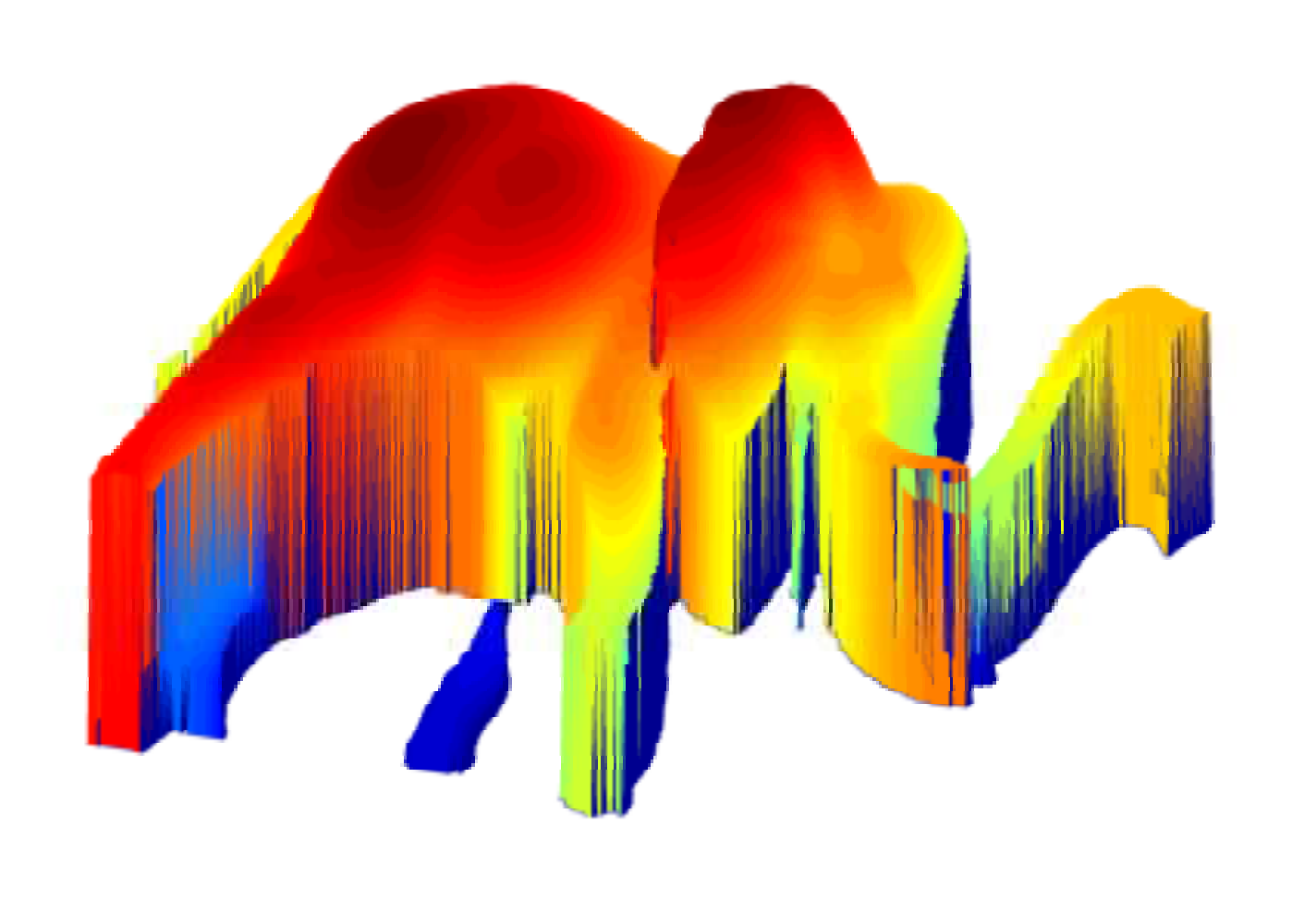}&
\includegraphics[width=.23\linewidth]{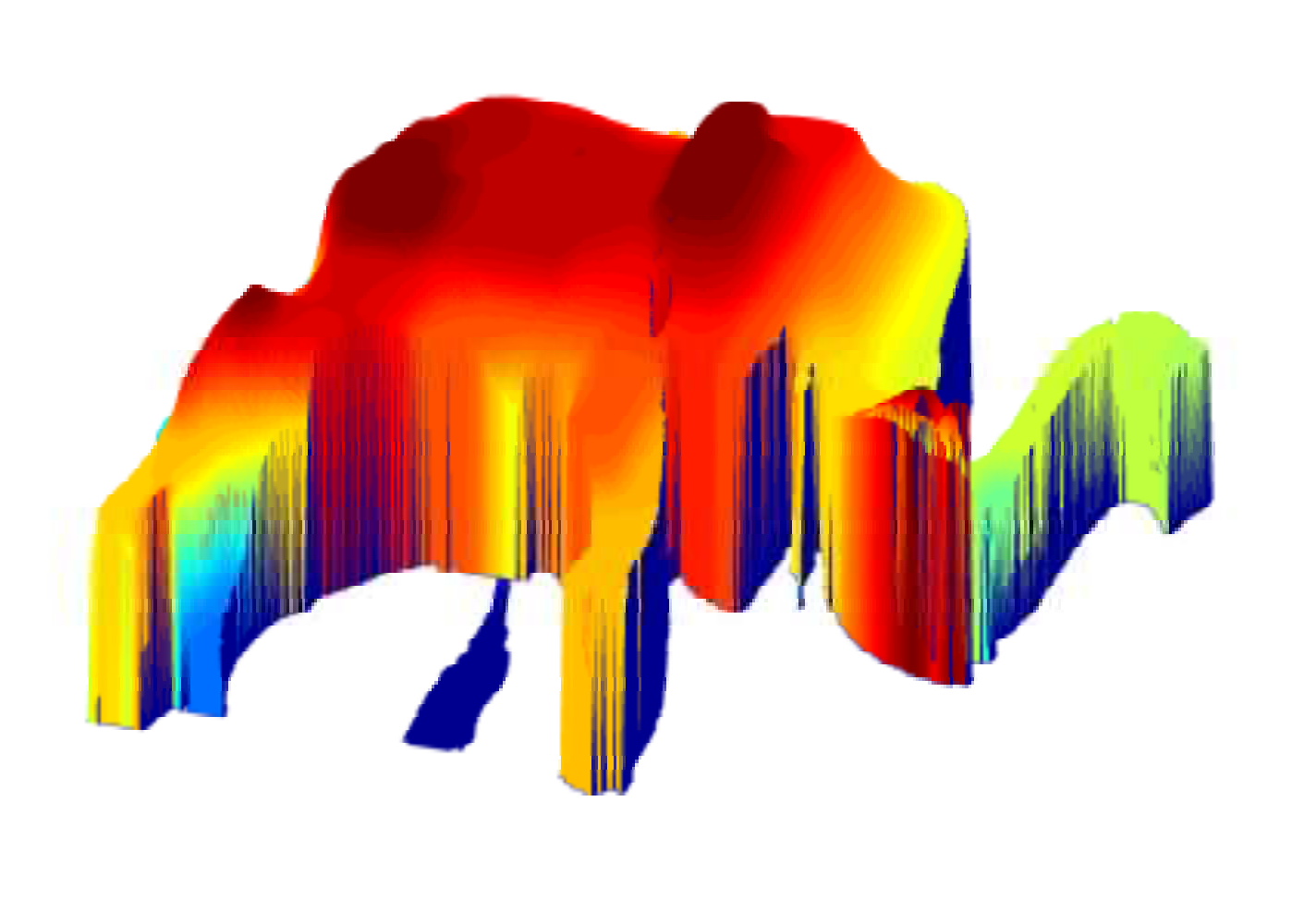}&
\includegraphics[width=.23\linewidth]{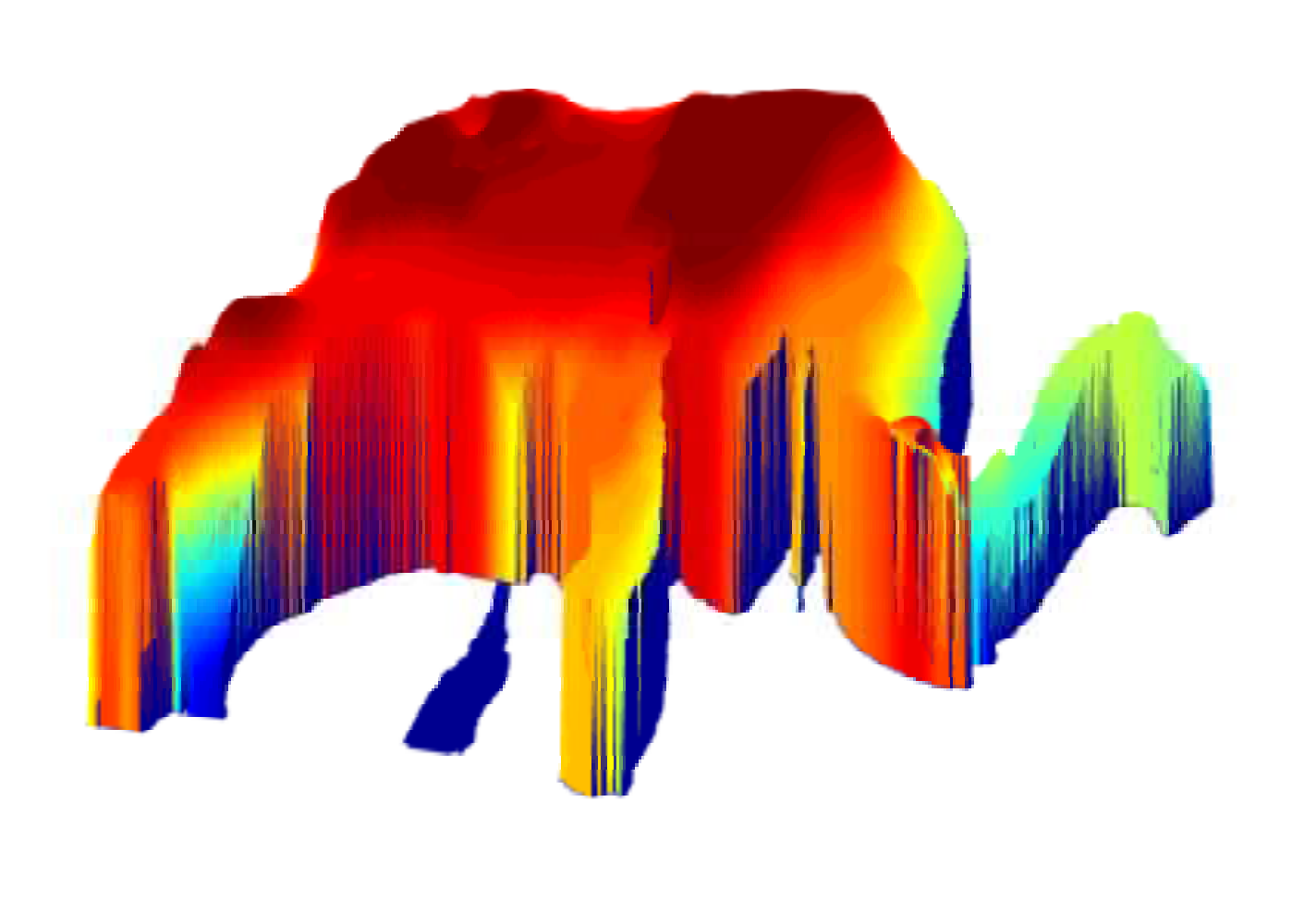}\\
\includegraphics[width=.23\linewidth]{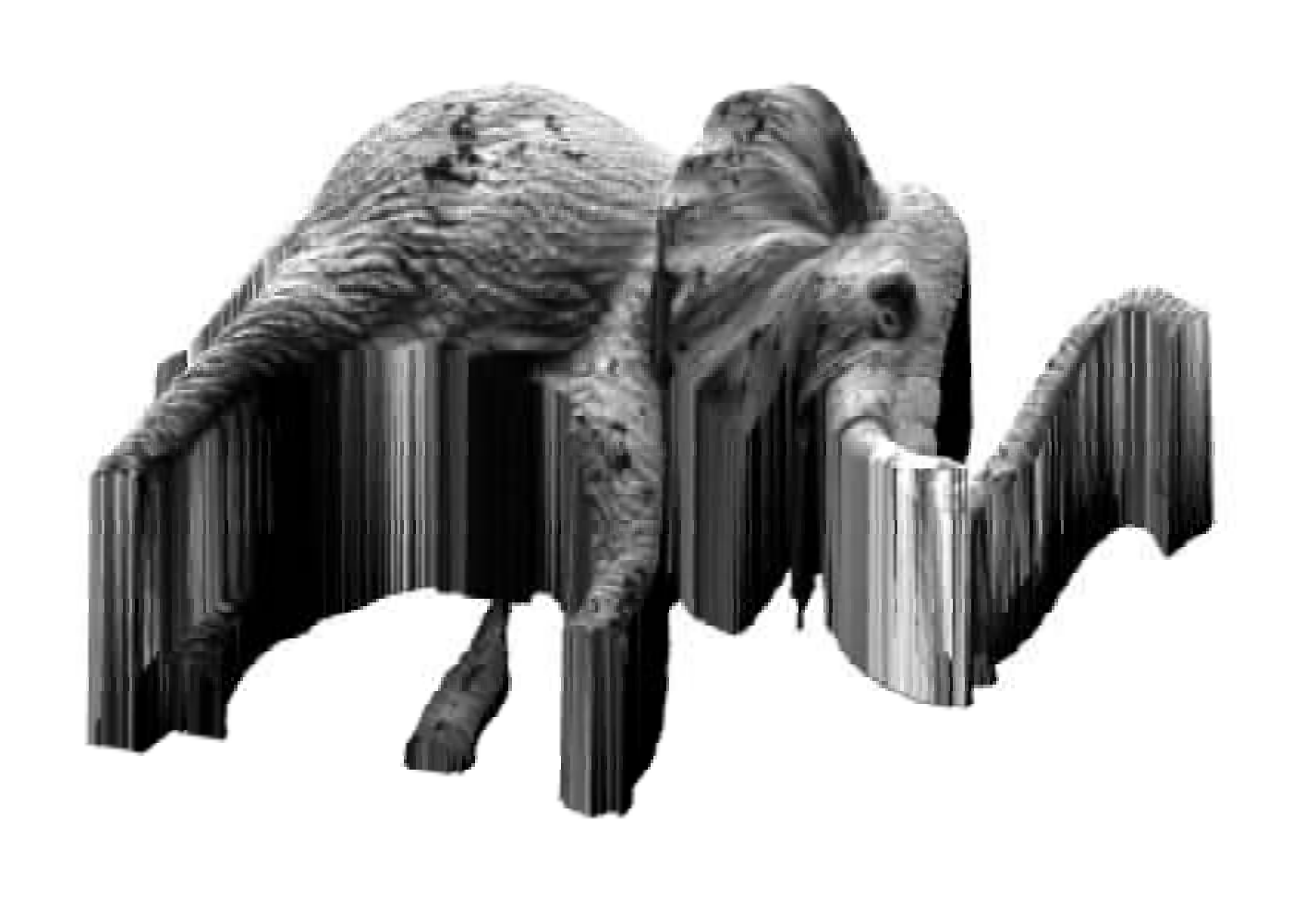}&
\includegraphics[width=.23\linewidth]{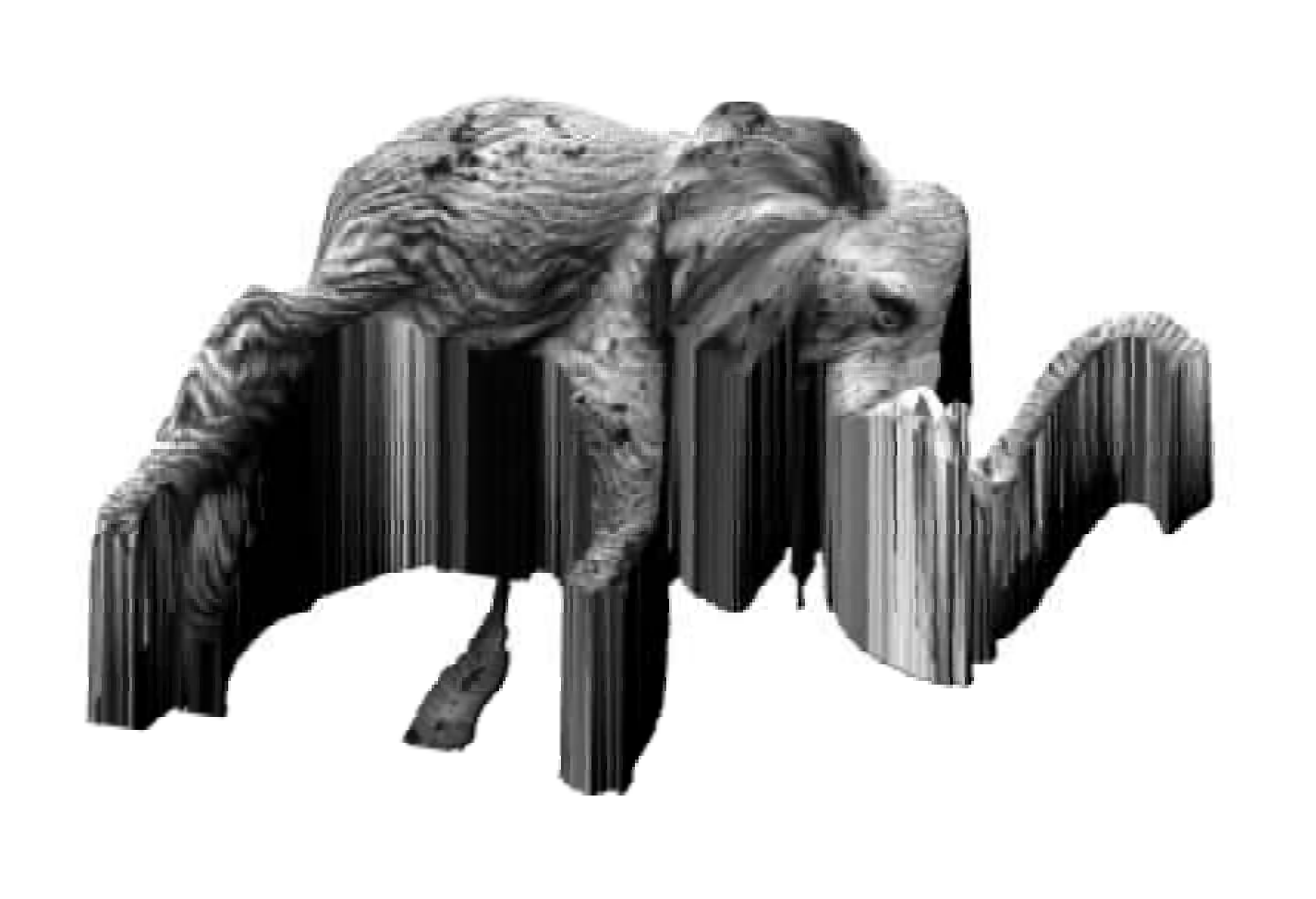}&
\includegraphics[width=.23\linewidth]{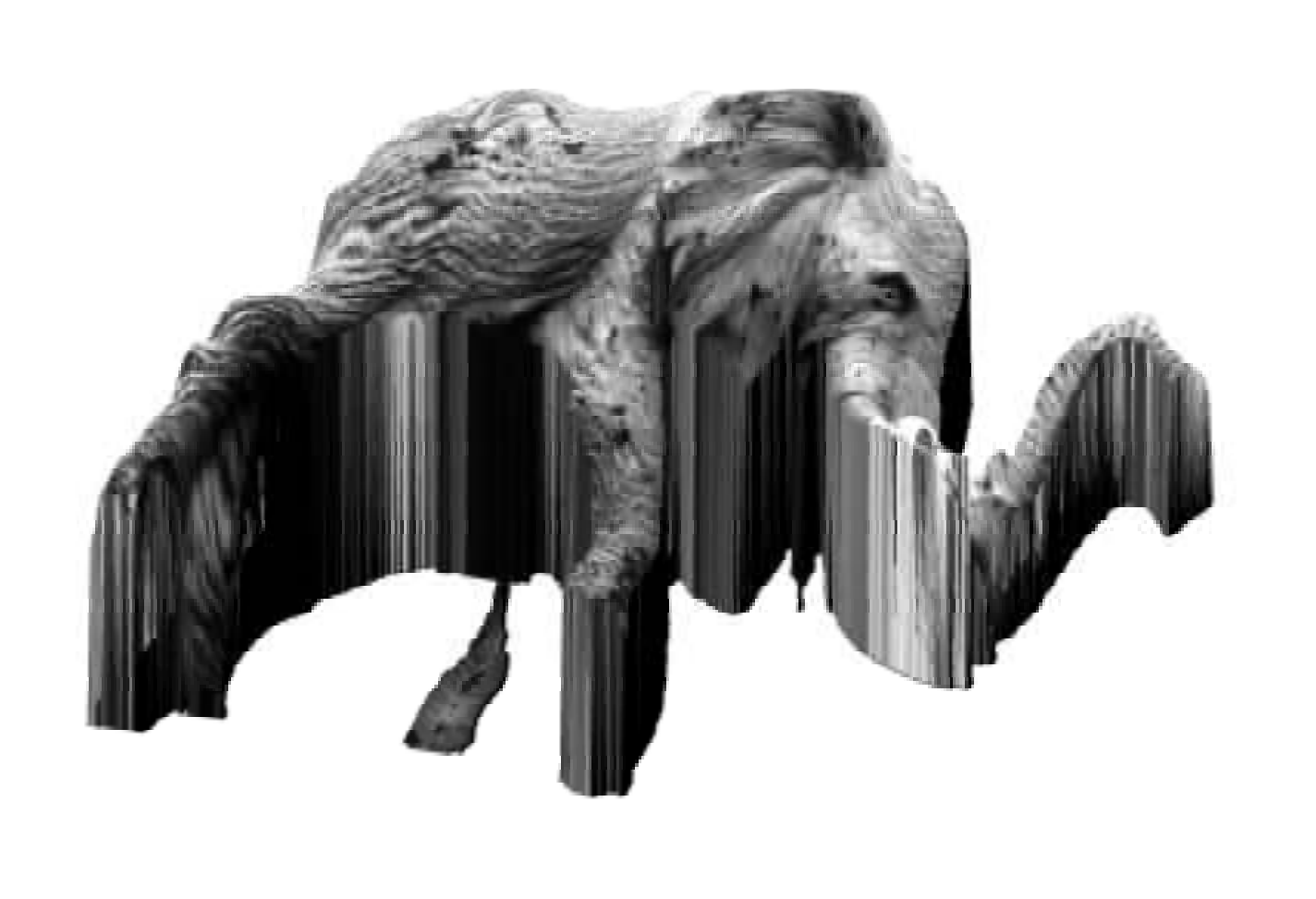}\\\hline
Correspondence error:& 0.69 & 1.30 \\
\end{tabular}
\caption{Reconstruction from multiple images. Ground truth is shown at left, image labels at the top and RMSE (over all images) at the bottom.}
\label{fig:mi-elephant-d}
\end{figure}

%% file: dtf.tex
\vspace{-0.4cm}%
\section{Introduction}
The last decade has seen the meteoric rise in the use of random
field models in computer vision~\cite{Szeliski2008}. Random
fields have been used to model many problems including
foreground-background (fg-bg)
segmentation~\cite{Blake04,Boykov2001}, semantic
segmentation~\cite{he2004multiscalecrf,shotton2007textonboost,winn2006layoutcrf},
stereo~\cite{Boykov2001}, optical
flow~\cite{Baker:ICCV2007,RothBIJCV09,Horn:optical-flow}, and 3D
reconstruction~\cite{SVZ:CVPR00,Vogiatzis:CVPR05}.
Many of these problems can be cast as an image labeling problem, where we are
given an image $\x$ and need to predict labels $\y$.
%, typically one label $y_i$ for each pixel $i \in \calV$.
Random fields provide a way of factorizing the joint distribution
$p(\x,\y)$ or the posterior distribution $p(\y|\x)$ into a product of local
interactions.

% Classic MRF
In the classic Markov random field (MRF) we obtain the posterior distribution
$p(\y|\x)$ by integrating a per-pixel likelihood functions with pairwise
consistency potentials ensuring a smooth solution~\cite{Geman:84,li1995mrf}.
One major advance in the field was to make these smoothness cost dependent on
the local image structure~\cite{Boykov2001},
conditioning parts of the model on the input data. In the last decade, these {\em conditional}
random field (CRF) models~\cite{lafferty2001crf,sutton2007crf,he2004multiscalecrf}
have become popular for their improved ability to capture the relationship between
labels and the image.

A lot of research effort has been devoted at the development of efficient
algorithms for estimating the Maximum a Posteriori (MAP) solution of such
models~\cite{Felzenszwalb2006,Szeliski2008,Kolmogorov2006,komodakis2007fastpd},
and the same is true for algorithms for probabilistic
inference~\cite{wainwright2008graphicalmodels,Koller2009}.
Further, a large number of methods have been proposed to learn the parameters
of random field
models~\cite{AnguelovTCKGHNCVPR2005,SKumarAHEMMCVPR05,sutton2007crf,szummer2008learningcrf,ZhangS05}.
A number of higher order random field formulations have also been proposed
that are able to encode interactions between groups of pixels, and have been
shown to produce much better results~\cite{KohliLTCVPR08,RothBIJCV09}.
However, despite these rapid developments, (most) state-of-the-art random
field CRF models continue to suffer from the following limitations: (1) they
are defined on the basis of a fixed neighborhood structure (except the work of
\cite{KolmogorovB05,RothB07}), and (2) the potentials are assumed to have a
simple parametric form with a pre-specified and fixed dependence on the image
data. While it is relatively easy to think of various ways to overcome these
limitations, the key research challenge is to suggest a model for which
efficient and high-quality training is still tractable.

This paper introduces a new graphical model, the Decision Tree
Field (DTF), which overcomes the above-mentioned limitations of
existing models. We take a simple yet radical view: every interaction
in our model depends on the image, and further, the
dependence is non-parametric.  It is easy to see that even
representing such a model is extremely challenging, since there are
numerous ways of defining a mapping between the image
and the parameters of a unary or pairwise interaction in the graphical
model.

Our model uses \emph{decision trees} to map the image content to
interaction values. Every node of every decision tree in our model
is associated with a set of parameters, which are used to define the
potential functions in the graphical model.
When making predictions on a novel test instance, the leaf node of the
decision tree determines the effective weights.

% Why decision trees?
There are a number of important reasons for the choice of decision trees
to specify the dependence between potentials and image content.
% 1. Non-parametric
Firstly, decision trees are non-parametric and can represent rich functional
relationships if sufficient training data is available.
% 2. Speed
Secondly, the training of decision trees is scalable, both
in the training set size and in that the approach can be parallelized;
recent advances even allows training on graphics processors~\cite{Sharp08}.
Since for most computer vision applications it is well known that the key to
obtaining high predictive performance is the amount of training data, many
recent works use decision trees, or a related variant of it (random
Forests~\cite{breiman2001randomforests}, extremely randomized
trees~\cite{geurts2006extremelyrandomizedtrees}, semantic texton
forest~\cite{shotton2008stf}).
In our context, decision trees give another big advantage: they allow us to efficiently
and jointly learn all parameters in the model. We achieve this
by using a log-concave pseudo-likelihood objective function, which is known to work well
given enough training data because it is a consistent
estimator~\cite{Koller2009}.	% Theorem 20.3

\paragraph{Our Contributions}\mbox{ }\\%
(1) To the best of our knowledge, we propose the first graphical model for
image labeling problems which allows all potential functions to have an
arbitrary dependence on the image data.\\%
(2) We show how the dependence between potential functions and image data can
be expressed via decision trees.\\%
(3) We show how the training of the DTF model, which involves learning of a
large number of parameters, can be performed efficiently.\\%
(4) We empirically demonstrate that DTFs are superior to existing models such
as random forest and common CRFs for applications with complex label
interactions and large neighborhood structures.

\section{Related Work}
\label{sec:related}
There has been relatively little work on learning image-dependent
potential functions, i.e. the ``conditional part'' of a random
field.
Most algorithms for learning the parameters of a random field try to learn
a class-to-class energy table that does not depend on the image
content~\cite{AnguelovTCKGHNCVPR2005,batra2008classspecific,nowozin2009connectivity,szummer2008learningcrf,TaskarCKG05}.
However, there have been few attempts at learning the parameters
of conditional
potentials~\cite{ChoJZKSF10,PrasadZFKT06,gould2009semanticregions}.
Recently, Gould {\em et al.}~\cite{gould2009semanticregions} used a
multiclass logistic regression classifier on a set of manually
selected features, such as the length and orientation of region
boundaries to obtain an image-dependent learned model for pairwise
interactions. Even more recently, Cho {\em et al.}~\cite{ChoJZKSF10}
proposed a model for image restoration whose interactions were dependent
on the semantic meaning of the local image content as predicted by
a classifier. Unlike our work, all the above-mentioned models either target
specific tasks, or assume a particular form for the dependence of the
potentials on the image content. Neither of the above-mentioned approaches is able
to learn a dependency model with thousands or even millions of parameters
which our model can achieve.

% Glesner and Koller paper
Decision trees are popularly used to model unary interactions,
e.g.,~\cite{shotton2011real}; but with two exceptions they have not been
used for pairwise or higher-order interactions.  The first exception
is the paper of Glesner and Koller~\cite{glesner1995beliefnetworks},
where decision trees are used to model conditional probability
tables over many discrete variables in a Bayesian network.  The
difference to our work is that our decision trees evaluate the given
image content, not the state of a random variable, thus requiring no
change to the inference procedure used.

%We believe that a suitable
%combination of these two ideas, \ie using tests based on both image
%content and random variables, is promising, but leave this for
%future work.
%
% Random forest random field
The second exception is the ``random forest random
field''~\cite{payet2010rfrf}.  Despite the similarity in name, the approach is
fundamentally different from ours.  Instead of defining an explicit model as
we do in~(\ref{eqn:p}), Payet and Todorovic~\cite{payet2010rfrf} define the
model distribution implicitly as the equilibrium distribution of a learned
Metropolis-Hastings Markov chain.  The Metropolis-Hastings ratio is estimated
by classification trees.  This is a \emph{clever idea} but comes with a number
of limitations,
% test-time: no choice
i) at test-time there is no choice between different inference methods but one
is bound to using inefficient Markov Chain Monte Carlo (MCMC);
in~\cite{payet2010rfrf} superpixel graphs of few hundred regions are used and
inference takes 30 seconds despite using advanced Swendsen-Wang cuts,
% implicit model
and ii) the model remains \emph{implicit}, such that inspecting the learned
interactions as we will do in Section~\ref{sec:bodyparts}
is not possible.

% Latent crf and deep architectures
In a broader view, our model has a richer representation of complex label
structure.
Deep architectures, such as~\cite{lee2009convolutionaldbn} and latent variable
CRFs, as in~\cite{schnitzspan2010latentcrf}, have the same goal, but use
hidden variables representing the presence of larger entities such as
object parts.  While these models are successful at representing structure,
they are generally difficult to train because their negative log-likelihood
function is no longer convex.  In contrast, by learning powerful
non-parametric conditional interactions we achieve a similar expressive power
but retain convexity of the training problem.

\section{Model}
\label{sec:model} We now describe the details of our model.
Throughout we will refer to $\x \in \calX$ as a given observed image
from the set of all possible images $\calX$. Our goal is to infer a
discrete labeling $\y \in \calY$, where the labeling is per-pixel,
i.e., we have $\y=(y_i)_{i \in \calV}$, $y_i \in \calL$, where all
variables have the same label set $\calL$.
%
% Energy function
We describe the relationship between $\x$ and $\y$ by means of an
\emph{energy function $E$} that decomposes into a sum of
energy functions $E_{t_F}$ over \emph{factors} $F$, where $F$ defines a
subsets of variables. For example, for a pairwise factor it is $|F|=2$.
We have
\vspace{-0.15cm}%
\begin{equation}
    E(\y,\x,\w) = \sum_{F \in \calF} E_{t_F}(y_F,x_F,w_{t_F}).\label{eqn:E}
\vspace{-0.1cm}%
\end{equation}
By $y_F$ we denote the collection $(y_i)_{i \in F}$, and likewise we
write $x_F$ to denote the parts of $\x$ contained in $F$.  While
there may be many different subsets in $\calF$, we assume they are
of few distinct \emph{types} and denote the type of the factor $F$
by $t_F$.  The function $E_{t_F}$ is the same for all factors of
that type, but the variables and image content it acts upon differs.
Furthermore, the function is \emph{parameterized} by means of a
weight vector $w_{t_F}$ to be discussed below.

\begin{figure}[t!]
\centering
\begin{minipage}[b]{5cm}
    \begin{center}
    \includegraphics[width=0.99\linewidth]{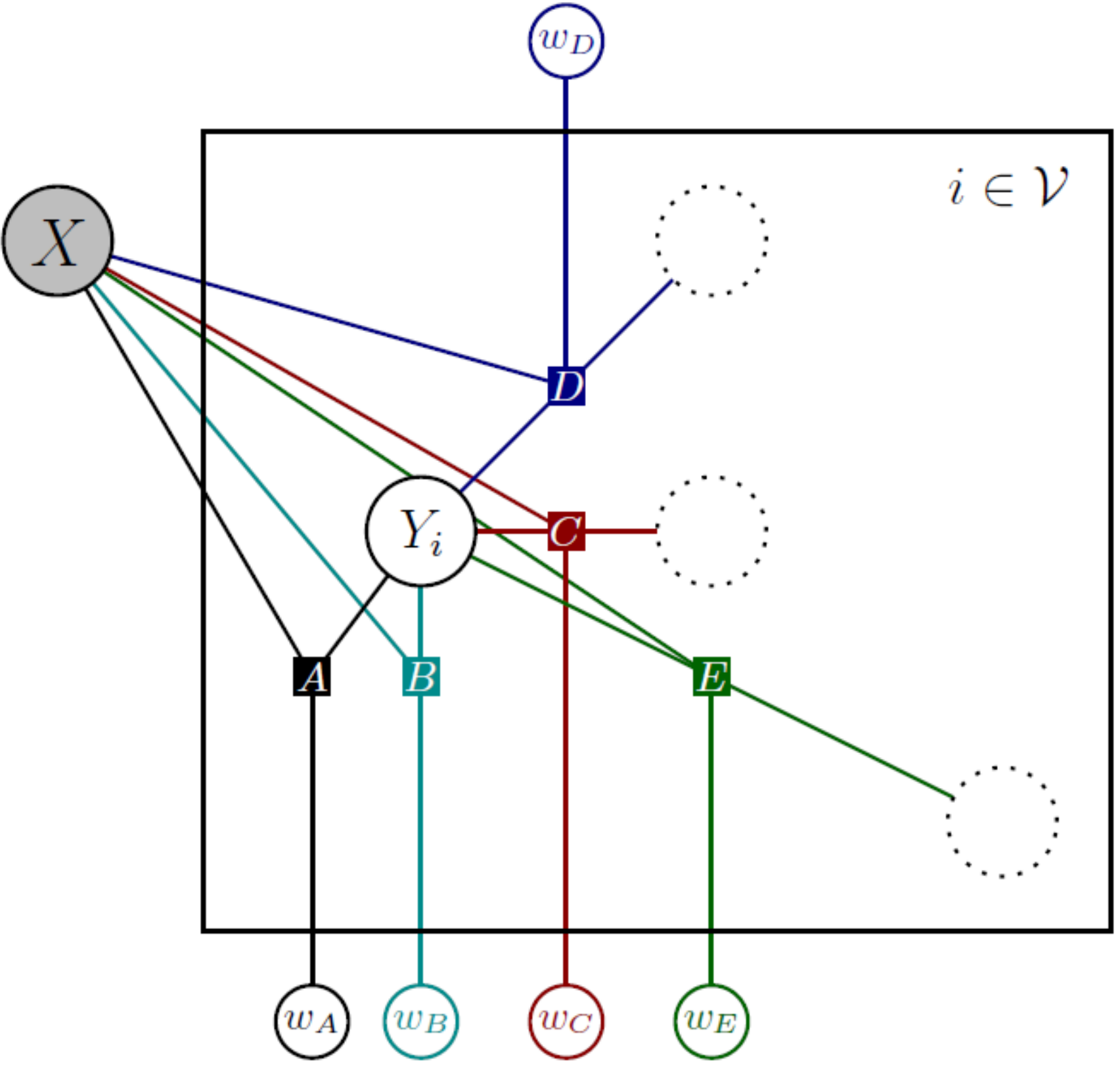}
    \end{center}%
	\vspace{-0.2cm}%
    \caption{Neighborhood structure around each pixel with five different
factor types.}
    \label{fig:nbhd}
\end{minipage}%
\hspace{1cm}%
\begin{minipage}[b]{6cm}
    \begin{center}
    \includegraphics[width=0.99\linewidth]{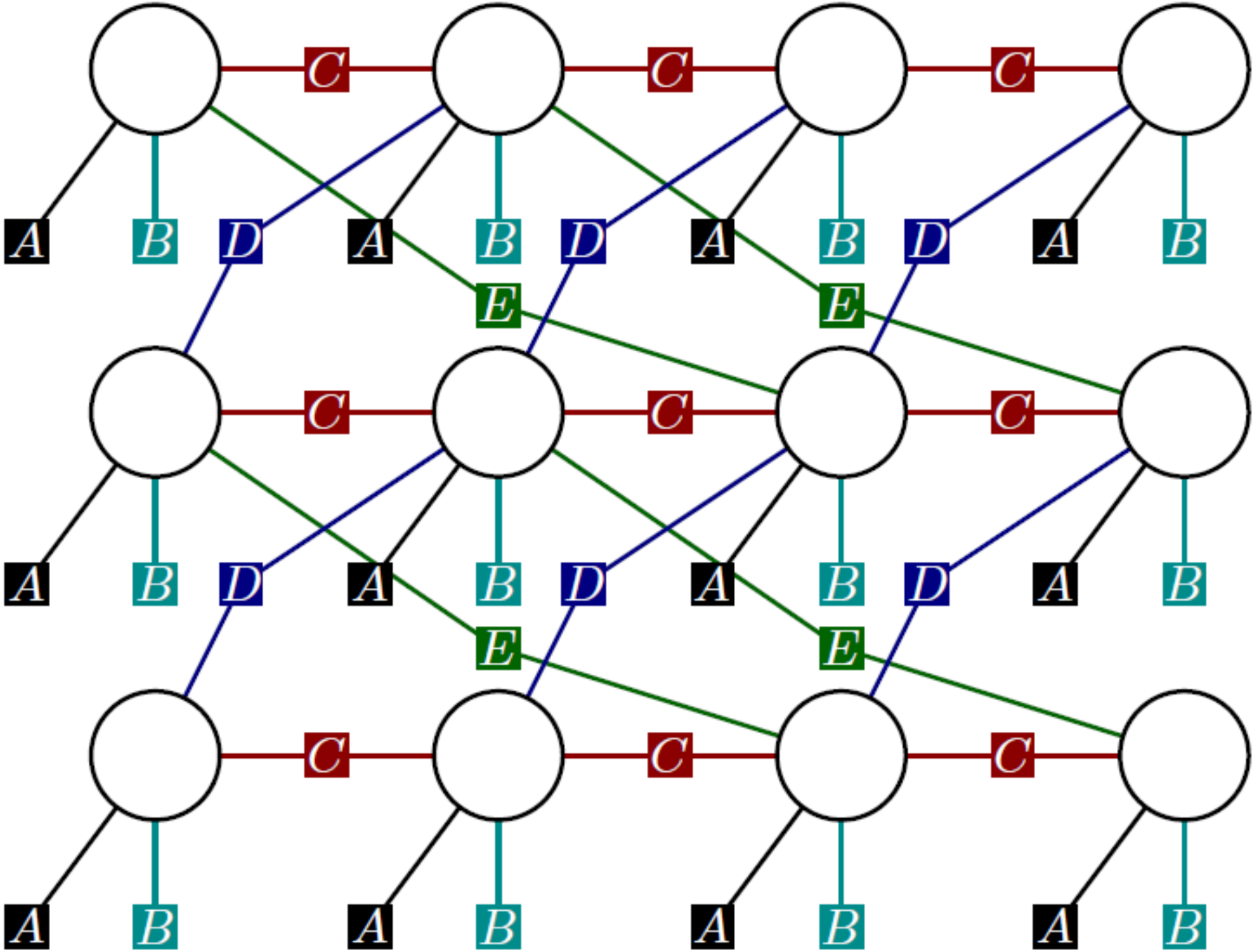}%
    \end{center}%
	\vspace{-0.2cm}%
    \caption{Unrolled factor graph (image size 4-by-3 pixels), dependencies on
$\x$ and $\w$ are not shown.}
    \label{fig:nbhd-unroll}
\end{minipage}%
\vspace{-0.5cm}%
\end{figure}

A visualization of a small factor graph model is shown in
Figure~\ref{fig:nbhd}. It has three pairwise factor types (red,
blue, and green) and two unary factor types (black and turquoise).
All factors depend on the image data $\x$.
Figure~\ref{fig:nbhd-unroll} shows the ``unrolled'' factor graph for
an image of size 4-by-3 pixels, where the basic model structure is repeated
around each pixel $i \in \calV$, and pairwise factors which reach outside the
image range are omitted. In total we have $|\calF|=31$ factors.

% Distribution
The energy function~(\ref{eqn:E}) defines a conditional probability
distribution $p(\y|\x,\w)$ as
\vspace{-0.15cm}%
\begin{equation}
    p(\y|\x,\w) = \frac{1}{Z(\x,\w)} \exp(-E(\y,\x,\w)),
    \label{eqn:p}
\vspace{-0.1cm}%
\end{equation}
where $Z(\x,\w)=\sum_{\y \in \calY} \exp(-E(\y,\x,\w))$ is the normalizing
constant.
% -> CRF
So far, our model is in the general form of a \emph{conditional random
field}~\cite{lafferty2001crf}.
% Special form of E_{t_F}
We now show how to use decision trees for representing $E_{t_F}$
in~(\ref{eqn:E}).

% Figure: path-summing
\begin{floatingfigure}[right]{0.5\linewidth}%
\hspace{-0.2cm}\includegraphics[width=0.5\linewidth]{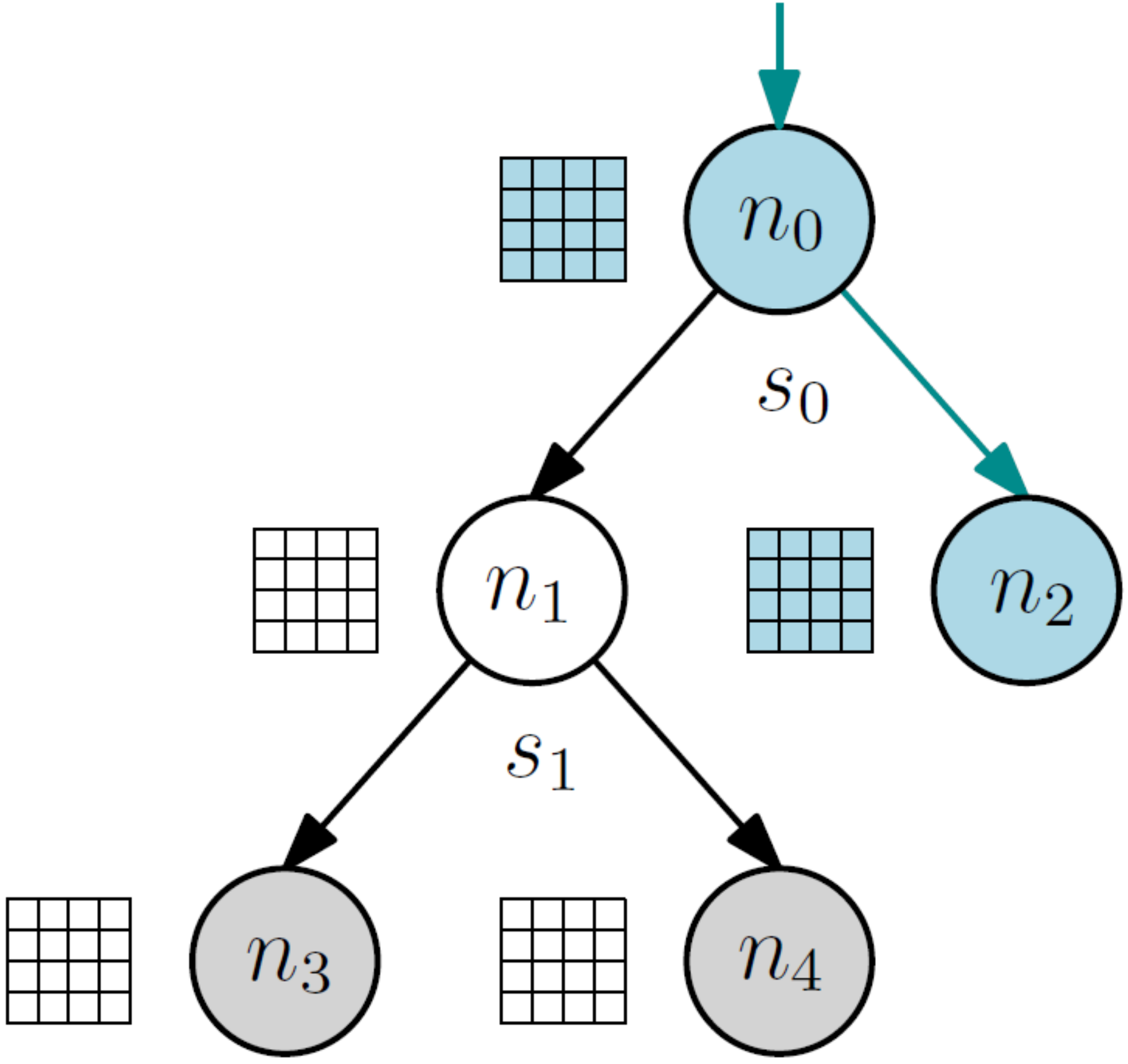}%
\caption{Summation of all energy tables along the path of visited decision
nodes (shaded blue).}
\label{fig:pathsum}%
\end{floatingfigure}%
With each function $E_{t}$ we associate one decision tree.  To evaluate
$E_{t_F}(y_F,x_F,w_{t_F})$, we start at the root of the tree, and perform a
sequence of tests $s$ on the image content $x_F$, traversing the tree to the
left or right.  This process is illustrated in Figure~\ref{fig:pathsum}.
When a leaf node has been reached, we collect the \emph{path} of traversed
nodes from the root node to the leaf node.  With each node $q$ of the tree we
associate a table of energy values $w_{t_F}(q,y_F)$.  Depending on the number
of variables $y_F$ this energy function acts on, the table can be a vector
(unary), a matrix (pairwise), or general $k$-dimensional array (higher order).
We \emph{sum} all the tables along the path taken and compute the energy as
    % Summation of tables along the path
    \[E_{t_F}(y_F,x_F,w_{t_F})=\sum_{q \in \textrm{Path}(x_F)} w_{t_F}(q,y_F),\]
where $\textrm{Path}(x_F)$ denotes the set of nodes taken during evaluating
the tree.
By using one set of weights at each node we can regularize the nodes at the
root of the tree to exert a stronger influence, affecting a large number of
leaves; at test-time we can precompute the summation along each root-to-leaf
path and store the result at each leaf.

% Putting it all together
To compute the overall energy~(\ref{eqn:E}) we evaluate $E_{t_F}$
for all factors $F \in \calF$.  Although the type $t_F$ might be the
same, the function $E_{t_F}$ depends on $x_F$ through the evaluation
of the decision tree.  This allows image-dependent unary, pairwise,
and higher-order interactions. The set $\calF$ is determined by
repeating the same local neighborhood structure for each pixel, as shown in
Figures~\ref{fig:nbhd} and~\ref{fig:nbhd-unroll}.

% Summary and points to discuss
In summary, our model consists of a set of factor types.  Each factor type
contains, (i) the number $k$ of variables it acts on and their relative offsets,
(ii) a single decision tree, and (iii) for each node in the decision tree, a
table of energies of size $\calL^k$.
Given a new image $\x$, for each possible labeling $\y$ we can evaluate
$E(\y,\x,\w)$ by the above procedure.

\subsection{Relation to other Models}
% Generalization
% 1. Per-pixel random forests
The proposed DTF generalizes a number of popular existing image labeling
methods.  If we ignore pairwise and higher-order interactions
in~(\ref{eqn:E}), then the variables are independent and making predictions
for each pixel is the same as evaluating a random forests, as used in
e.g.~\cite{shotton2008stf,TuB10}.
Interestingly, as we will show in the experiments, even in this setting we
still slightly outperform standard random forests since we learn the weights
in each decision node instead of using empirical histograms; this novel
modification improves predictive performance without any test-time overhead
compared to random forests.
% 2. Pairwise simple CRFs
For pairwise interactions we generalize simple CRFs with contrast-sensitive
pairwise potentials such as the popular \emph{GrabCut
system}~\cite{rother2004GC} and
TextonBoost~\cite{shotton2007textonboost}.
% 3. Markov random field
Finally, if for the pairwise interactions we use decision trees of depth one,
such that these interactions do not depend on the image content, then our
model becomes a classic Markov random field prior~\cite{li1995mrf}.

\section{Learning Decision Tree Fields}
\label{sec:learning_inference}
Learning the model involves selecting the neighborhood structure, the decision
trees, and the weights stored in the decision nodes.
During learning we are given an iid set $\{(\x_m,\y^*_m)\}_{m=1,\dots,M}$ of
images $\x_m$ and ground truth labelings $\y^*_m$.  Our goal is to estimate
the parameters $\w$ of our model such as to predict $\y^*_m$ for a given
$\x_m$.  For simplifying the derivation of the learning method, we can treat
the given set of images as if it would be one large collection of pixels as is
done in~\cite{sutton2007crf}.

\subsection{Maximum Likelihood Learning}
For learning the parameters of our model, we need to elaborate on how the
parameters $\w$ define the energy.
One important observation is that for a fixed set of decision trees the energy
function~(\ref{eqn:E}) can be represented such that it is linear in the
parameters $\w$.  To see this, consider a single $E_{t_F}(y_F,x_F,w_{t_F})$
function and define a binary indicator function
\vspace{-0.2cm}%
    \[B_{t_F}(q,z;y_F,x_F)=\left\{\begin{array}{cl}
        1 & \textrm{if $n \in \textrm{Path}(x_F)$ and $z=y_F$},\\
        0 & \textrm{otherwise.}\end{array}\right.\]
Then, we can write the energy $E_{t_F}(y_F,x_F,w_{t_F})$ equivalently as a
function linear in $w_{t_F}$,
\vspace{-0.15cm}%
\begin{equation}
    \sum_{n \in \textrm{Tree}(t_F)} \sum_{z \in \calY_F}
            w_{t_F}(q,z) B_{t_F}(q,z;y_F,x_F).
    \label{eqn:linearform}
\vspace{-0.1cm}%
\end{equation}
%
% Note on non-linear representation
The use of decision trees allows us to represent non-linear functions on $\x$.
Although non-linear in $\x$, by the representation~(\ref{eqn:linearform}) we
can parameterize this function linearly in $w_{t_F}$.
Then, from~(\ref{eqn:linearform}) we see that the gradient has a simple form,
$\nabla_{w_{t_F}(q,z)} E_{t_F}(y_F,x_F,w_{t_F}) = B_{t_F}(q,z;y_F,x_F)$.

% Advantages of linear energy function
Because~(\ref{eqn:E}) is linear in $\w$, the log-likelihood of~(\ref{eqn:p})
is a concave and differentiable function in $\w$~\cite[Corollary
20.2]{Koller2009}.
This means that if computing $Z(\x,\w)$ and the marginal distributions
$p(y_F|\x,\w)$ for all $F \in \calF$ would be tractable, then learning the
parameters by maximum likelihood becomes a convex optimization problem.

We now show how to use efficient approximate likelihood methods to learn all
parameters associated to the decision trees from training data.
For now we assume we are given a fixed set of factor types, including decision
trees, but have to learn the weights/energies associated to the nodes of the
trees.  We will discuss how to learn trees later.

\subsection{Pseudolikelihood}
The pseudolikelihood~\cite{besag1977pseudolikelihood} defines a surrogate
likelihood function that is maximized.  In contrast to the true likelihood
function computing the pseudolikelihood is tractable and very efficient.
% TODO: give consistency result + citation
%
% Pseudolikelihood
The pseudolikelihood is derived from the per-variable conditional
distributions $p(y_i|y^*_{\calV \setminus \{i\}},\x,\w)$.
By defining $\ell_i(\w)=-\log p(y_i|y^*_{\calV \setminus \{i\}},\x,\w)$ we can
write the regularized negative log-pseudolikelihood $\ell_{npl}(\w)$ as the
average $\ell_i$ over all pixels,
\vspace{-0.15cm}%
\begin{equation}
\ell_{npl}(\w) = \frac{1}{|\calV|} \sum_{i \in \calV} \ell_i(\w)
    - \sum_{t} \log p_t(w_t)\label{eqn:npl},
\vspace{-0.1cm}%
\end{equation}
where $p_t(w_t)$ is a \emph{prior distribution} over $w_t$ used to regularize
the weights.  We will use multivariate Normal distributions
$\mathcal{N}(0,\sigma_t I)$, so that $-\log p_t(w_t)$ is of the form
$\frac{1}{2\sigma_t^2} \|w_t\|^2 + C_t(\sigma_t)$ and the constant
$C_t(\sigma_t)$ can be omitted during optimization because it does not depend
on $\w$.
For each factor type $t$ the prior hyperparameter $\sigma_t > 0$ controls the
overall influence of the factor and we need to select a suitable value by
means of a model selection procedure such as cross validation.

Function~(\ref{eqn:npl}) is convex, differentiable, and tractably computable.
For optimizing~(\ref{eqn:npl}) we use the L-BFGS numerical optimization
method~\cite{zhu1997lbfgs}.  To use L-BFGS we need to iteratively compute
$\ell_i(\w)$ and the gradient $\nabla_{w_t} \ell_i(\w)$.
%
% Objective and gradient
The computation of $\ell_i(\w)$ and $\nabla_{w_t} \ell_i(\w)$ is
straightforward
% and yields the expressions shown in Figure~\ref{fig:npl-fig},
\begin{eqnarray}
\ell_i\left(\w\right) & = & \sum_{F \in M(i)}  E_F\left(y^*_F,\x,w_{t_F}\right) \nonumber\\
  & & + \log \sum_{y_i \in \calY_i} \exp \left(
    - \!\! \sum_{F \in M(i)} \!\! E_F\left(y_i,y^*_{\calV \setminus
        \{i\}},\x,w_{t_F}\right)\right)
    \label{eqn:npl-i-obj}\\
\nabla_{w_t} \ell_i(\w) & = &
    \sum_{F \in M_t(i)} \!\!\! \nabla_{w_t} E_F\left(\y^*,\x,w_t\right) \nonumber\\
  & & - \mathbb{E}_{y_i \sim p\left(y_i\left|y^*_{\calV \setminus \{i\}},\x,\w\right.\right)}\left[
        \sum_{F \in M_t(i)} \!\! \nabla_{w_t}
            E_F\left(y_i,y^*_{\calV \setminus \{i\}},\x,w_t\right)
    \right]
    \label{eqn:npl-i-grad}
\end{eqnarray}
where we use $M(i)$ to denote the subset of $\calF$ that involves variable
$y_i$, and $M_t(i)$ likewise but restricted to factors of matching type, i.e.,
$M_t(i)=\{F \in M(i): t_F=t\}$.
By summing~(\ref{eqn:npl-i-obj}) and~(\ref{eqn:npl-i-grad}) over all pixels in
all images, we obtain the objective and its gradient, respectively.
%
%%%\begin{figure}
%%%\fbox{
%%%\begin{minipage}[t]{0.975\linewidth}
%%%\begin{align}
%%%%\addtolength{\fboxsep}{5pt}
%%%\ell_i(\w) &= \sum_{F \in M(i)} \!\! E_F(y^*_F,\x,w_{t_F})
%%%    + \log \sum_{y_i \in \calY_i} \exp \Big(
%%%    - \!\! \sum_{F \in M(i)} \!\! E_F(y_i,y^*_{\calV \setminus
%%%        \{i\}},\x,w_{t_F})\Big),
%%%    \label{eqn:npl-i-obj}\\
%%%\nabla_{w_t} \ell_i(\w) &=
%%%    \sum_{F \in M_t(i)} \!\!\! \nabla_{w_t} E_F(\y^*,\x,w_t)
%%%    - \mathbb{E}_{y_i \sim p(y_i|y^*_{\calV \setminus \{i\}},\x,\w)}\Big[
%%%        \sum_{F \in M_t(i)} \!\! \nabla_{w_t}
%%%            E_F(y_i,y^*_{\calV \setminus \{i\}},\x,w_t)
%%%    \Big].
%%%    \label{eqn:npl-i-grad}
%%%\end{align}
%%%\end{minipage}
%%%}
%%%\caption{Objective and gradient expressions around a single variable
%%%$i \in \calV$ for minimizing the negative log-pseudolikelihood.}
%%%\label{fig:npl-fig}
%%%\end{figure}
% Convergence criterion
When initializing the weights to zero we have approximately $\|\nabla_{\w}
\ell_{npl}(\w)\| \approx 1$.  During optimization we stop when
$\|\nabla_{\w} \ell_{npl}(\w)\| \leq 10^{-4}$, which is the case after around
100-250 L-BFGS iterations, even for models with over a million parameters.

\subsection{Learning the Tree Structure}
Ideally, we would like to learn the neighborhood structure and decision trees
jointly with their weights using a single objective function.
However, whereas the weights are continuous, the set of decision trees is a
large combinatorial set.  We therefore propose to use a simple two-step
heuristic to determine the decision tree structure: we
learn the classification tree using the training samples and the
\emph{information gain} splitting criterion.  This greedy tree construction is
popular and known to work well on image labeling
problems~\cite{shotton2008stf}.
The key parameters are the maximum depth of the tree, the minimum number of
samples required to keep growing the tree, and the type and number of split
features used.  As these settings differ from application to application, we
describe them in the experimental section.
Unlike in a normal classification tree, we store weights at every decision
node and initialize them to zero, instead of storing histograms over classes
at the leaf nodes only.

% Extension to pairwise/higher-order
The above procedure is easily understood for unary interactions, but now show
that it can be extended in a straightforward manner to learn decision trees
for pairwise factors as well.
To this end, if we have a pairwise factor we consider the product set
$\calL \times \calL$ of labels and treat each label pair
$(l_1,l_2)\in\calL\times\calL$ as a single class.  Each training pair of
labels becomes a single class in the product set.  Given a set of such
training instances we learn a classification tree over $|\calL|^2$ classes
using the information gain criterion.  Instead of storing class histograms we
now store weight tables with one entry per element in $\calL\times\calL$.
The procedure extends to higher-order factors in a straightforward way.

% Limitations: overcounting
Once the trees are obtained, we set all their weights to zero and
optimize~(\ref{eqn:npl}).  During optimization the interaction between
different decision trees is taken into account.
This is important because the tree structures are determined
independently and if we were to optimize their weight independently as well,
then we would suffer from overcounting labels during training.
The same overcounting problem would occur if we would want to use the class
histograms at the leaf nodes directly, for example by taking the negative
log-probability as an energy.

\subsection{Complexity of Training}
The complexity to compute the overall objective~(\ref{eqn:npl}) and its
gradient is $O(|\calV|\cdot|\calL|\cdot N)$, where $\calV$ is the set of
pixels in the training set, $\calL$ is the label set, and $N$ is the number of
factors in the neighborhood structure.
Note that this is linear in all quantities, and independent of the order of
the factors.  This is possible only because of the pseudolikelihood
approximation.  Moreover, it is even more efficient than performing a single
sweep of message passing in loopy belief propagation, which has complexity
$O(|\calV|\cdot|\calL|^k\cdot N)$ for factors of order $k \geq 2$.

\subsection{Making Training Efficient}
Training a graphical model on millions of pixels is computationally
challenging.  We have two principled methods to make training efficient.

% 1. Parallelism
First, observe that our training procedure parallelizes in every step: we
train the classification trees in parallel~\cite{Sharp08}.
Likewise, evaluating~(\ref{eqn:npl}) and its gradient is a large summation of
independent terms, which we again compute in parallel with no communication
overhead.

% 2. Subsampling approximation
The second observation is that every step in our training procedure can be
carried out on a subsampled training set.  For classification trees we can
process a subset of pixels, as in~\cite{shotton2008stf}.
Less obvious, we can do the same thing when optimizing our
objective~(\ref{eqn:npl}).  The first term in equation~(\ref{eqn:npl}) takes
the form of an empirical expectation
$\mathbb{E}_{i\sim \mathcal{U}(\calV)}[\ell_i(\w)]$ that can be
approximated both deterministically or by means of stochastic approximation.
We use a deterministic approximation by selecting a fixed subset $\calV'
\subset \calV$ and evaluating
$\ell'_{npl}(\w) = \frac{1}{|\calV'|} \sum_{i\in\calV'} \ell_i(\w) - \sum_t \log p_t(w_t)$.
We select $\calV'$ to be large enough so this computation remains efficient;
typically $|\calV'|$ has a few million elements.\footnote{When sampling
$\calV'$ uniformly at random with replacement from $\calV$, the \emph{law of
large numbers} guarantees the asymptotic correctness of this approximation.}

\subsection{Inference}
We use different inference methods during test-time.  For making maximum
posterior marginal predictions (MPM) we use an efficient Gibbs sampler.
Because the Gibbs sampling updates use the same quantities as used for
computing~(\ref{eqn:npl-i-obj}) we do not have to unroll the graph.
For obtaining approximate MAP predictions, we use the Gibbs sampler with
simulated annealing (SA), again exploiting the model structure.
Both the Gibbs sampler and the SA minimization is explained in the
supplementary materials.
To have a baseline comparison, we also minimize~(\ref{eqn:E}) using
tree-reweighted message passing (TRW) by unrolling the factor graph and using
the implementation of~\cite{Kolmogorov2006}.

\section{Experiments}
\label{sec:exp}
We considered a broad range of applications and report experiments for three
data sets.  One more experiment is reported in the supplementary materials.
The aim is to show that the DTF enables improved performance in challenging
tasks, where a large number of interactions and parameters need to be
considered and these cannot be manually tuned.  Moreover, we show that
conditional pairwise interactions better represent the data and lead to
improved performance.
As the three datasets are quite diverse, they also show the broad
applicability of our system.

\subsection{Conditional Interactions: Snake Dataset}
\begin{floatingfigure}[right]{0.3\linewidth}%
\hspace{-0.2cm}%
\centering
\fbox{
\includegraphics[width=1.25cm]{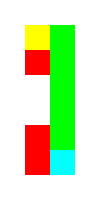}%
}%
\fbox{
\includegraphics[width=1.25cm]{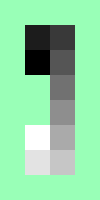}%
}%
\vspace{0.1cm}%
\caption{Input (left), labeling (right).}%
\label{fig:snakes-task}%
\end{floatingfigure}%
% Why? -> weak local evidence
In this experiment we construct a task that has only very weak local evidence
for any particular label and structural information needs to be propagated at
test-time in order to make correct predictions.  Moreover, this structure
is not given but needs to be learned from training data.
% Snake task
Consider Figure~\ref{fig:snakes-task} to the right, illustrating the task.
A ``snake'' shown on the input image is a sequence of adjacent pixels, and the
color in the input image encodes the direction of the next pixel: red means
``go north'', yellow means ``go east'', blue means ``go west'', and green
means ``go south''.  Once a background pixel is reached, the snake ends.  Each
snake is ten pixels long, and each pixel is assigned its own label, starting
from the head (black) to the tail (white).  Knowing about these rules, the
labeling (Figure~\ref{fig:snakes-task}, right) can be perfectly
reconstructed.  Here, however, these rules need to be learned from training
instances.
Of course, in a real system the unary interactions typically provide strong
cues~\cite{shotton2007textonboost,batra2008classspecific}, but we believe that the
task distills the limitations of noisy unary interactions: in this task, for
making perfect predictions, the unary would need to learn about all possible
snakes of length ten, of which there are clearly too many.\footnote{In
general, the number of snakes is equal to the number of fixed-length
self-avoiding walks on a lattice, a number which is conjectured to be
exponential in the length.}

% Setup
We use a standard 4-neighborhood for both the MRF and the DTF models.
The unary decision trees are allowed to look at every pixel in the input
image, and therefore could remember the entire training image.  For
experimental details, please see the supplementary materials.
We use a training set of 200 images, and a test set of 100 images.

\begin{figure}%[ht!]
\vspace{-0.1cm}%
\begin{center}
\hfill
\begin{minipage}[t]{1.0\linewidth}
%	\fbox{%
%    \includegraphics[width=0.125\linewidth]{dtf/snakes_test.png}}%
%    \hspace{0.075cm}%
%    \includegraphics[width=0.125\linewidth]{dtf/snakes_test_GT.png}%
%    \hspace{0.25cm}%
%    \includegraphics[width=0.125\linewidth]{dtf/snakes_unary.png}%
%    \hspace{0.075cm}%
%    \includegraphics[width=0.125\linewidth]{dtf/snakes_MRF.png}%
%    \hspace{0.075cm}%
%    \includegraphics[width=0.125\linewidth]{dtf/snakes_CRF.png}%
%    \hspace{0.25cm}%
%    \includegraphics[width=0.125\linewidth]{dtf/snakes_unary_sample00.png}%
%    \hspace{0.075cm}%
%    \includegraphics[width=0.125\linewidth]{dtf/snakes_unary_sample01.png}%
	\includegraphics[width=0.995\linewidth]{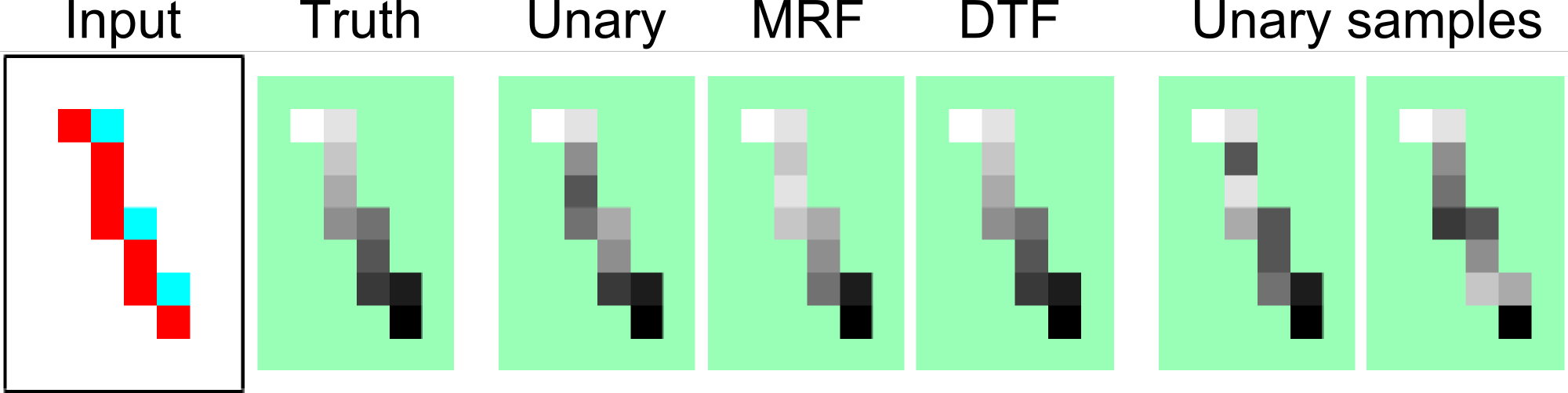}%
	\caption{Predictions on a novel test instance.}%
    \label{fig:snakes-test}%
\end{minipage}%
\hfill%
\end{center}%
\end{figure}
% Results: snakes
\begin{table}%[!hbt]
\vspace{-0.2cm}%
\begin{center}
\begin{small}
\begin{tabular}{lcccc}
 & RF & Unary & MRF & DTF\\
\hline
\hline
Accuracy & 90.3 & 90.9 & 91.9 & \textbf{99.4}\\
Accuracy (tail) & 100 & 100 & 100 & 100\\
Accuracy (mid) & 28 & 28 & 38 & 95\\
\hline
\end{tabular}%
\end{small}%
\end{center}%
\vspace{-0.15cm}%
\caption{Test set accuracies for the snake data set.}
\label{tab:snake-results}
\end{table}

% Results
The results obtained are shown in Table~\ref{tab:snake-results} and
Figure~\ref{fig:snakes-test}.  Here random forests (RF), trained unary
potentials (Unary), and the learned Markov random field (MRF) perform
equally well, at around 91\%.  Upon examining this performance
further, we discovered that while the head and tail labels are labeled with
perfect accuracy, towards the middle segments of the snakes the labeling
error is highest, see Table~\ref{tab:snake-results}.  This is plausible, as
for these labels the local evidence is weakest.
When using conditional pairwise interactions the performance improves to an
almost perfect 99.4\%.  This again makes sense because the pairwise
conditional interactions are allowed to peek at the color-codes at their
neighbors for determining the directionality of the snake.

The predictions are illustrated for a single test instance in
Figure~\ref{fig:snakes-test}.  We see that only the DTF makes a perfect
prediction.  To show the uncertainty of the unary model, we visualize two
samples from the model.

\subsection{Learning Calligraphy: Chinese Characters}

% Result figure: large inpainting task
\begin{figure}%[ht!]
\vspace{-0.3cm}%
\begin{center}
\hfill
\begin{minipage}[t]{1.0\linewidth}
    \includegraphics[width=0.995\linewidth]{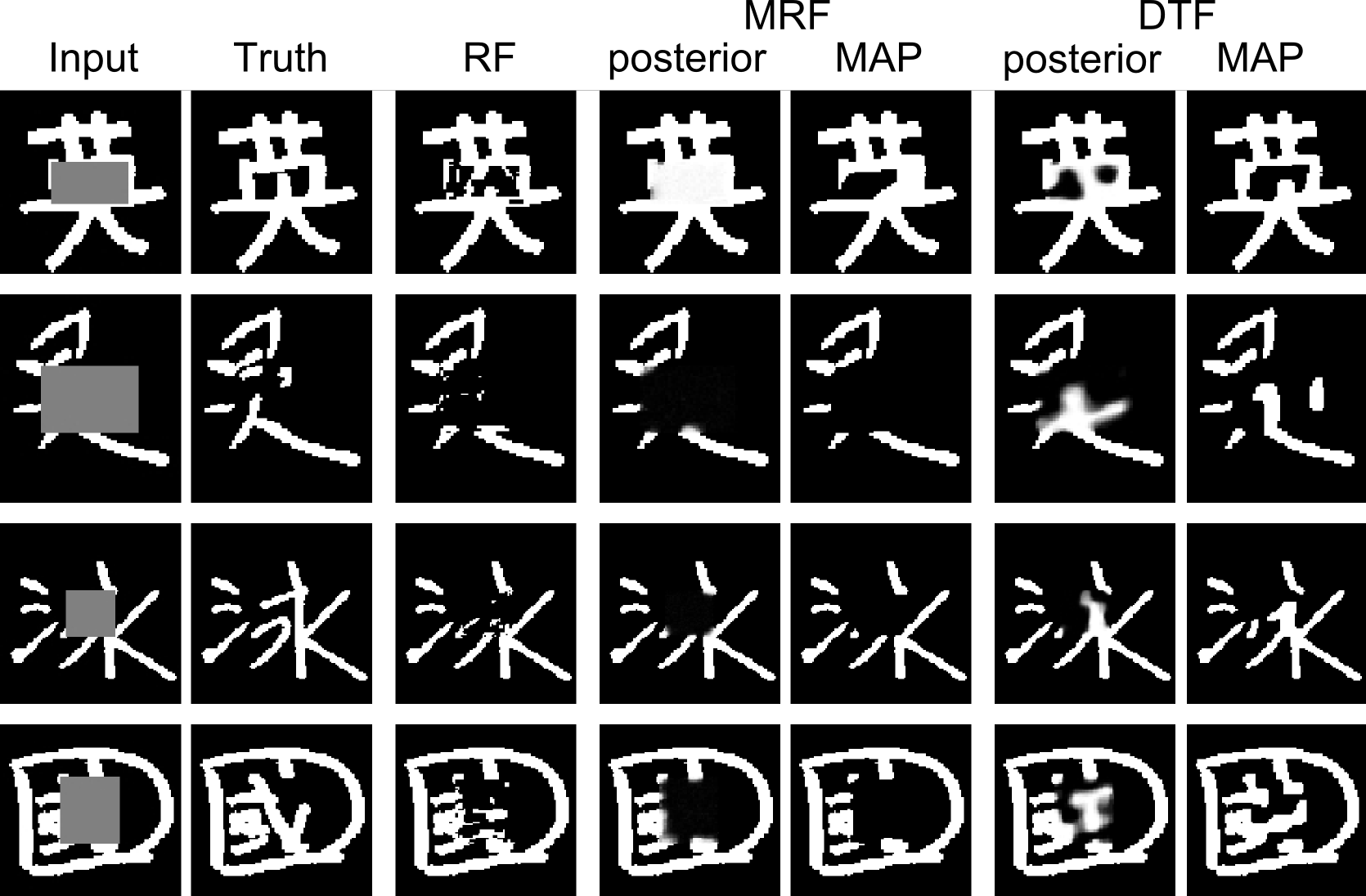}%
	\caption{Test set predictions for the large occlusion case.}
    \label{fig:chinese-inpaint-test}%
\end{minipage}%
\hfill%
\end{center}%
\end{figure}
%
% Interpretation of results
\begin{floatingfigure}[right]{0.5\linewidth}%
\hspace{-0.2cm}\includegraphics[width=0.5\linewidth]{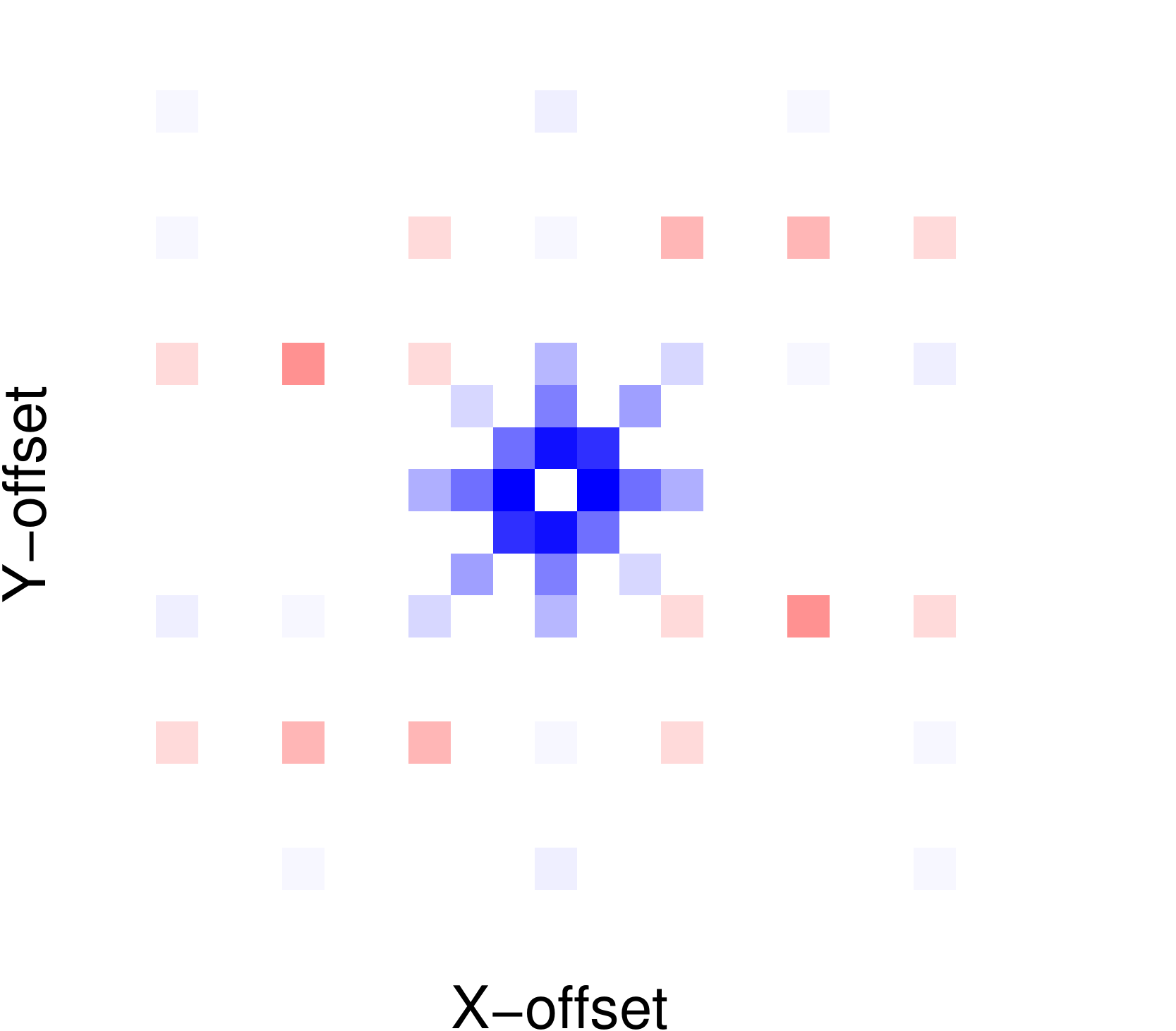}%
\caption{Pairwise associativity strength. Please see text.}
\label{fig:chinese-weights}%
\end{floatingfigure}%

In the previous experiment we have used a standard 4-connected neighborhood
structure.
In this experiment we show that by using larger conditional neighborhoods we
are able to represent shape.
We use the KAIST Hanja2 database of handwritten Chinese characters.
We occlude each character by grey box centered on the image, but with random
width and height.  For more details, please see the supplementary materials.
This is shown in the leftmost
column of Figure~\ref{fig:chinese-inpaint-test}.  We consider two datasets,
one where we have a ``small occlusion'' and one with a ``large occlusion''
box.
Note that most characters in the test set have never been observed in the
training set, but a model that has learned about shape structure of Chinese
characters can still find plausible completions of the input image.
To this end we use one unary factor with a decision tree of depth 15.
Additionally, we use a dense pairwise neighborhood structure of 8-connected
neighbors at one and two pixels distance, plus a sparse set of 27 neighbors
at $\{(-9,0),(-9,3),(-9,6),(-9,9),(-6,0),\dots,(9,9)\}$.  Therefore, each
variable has $2\cdot(24+4+4)=64$ neighboring variables in the model.  For the
pairwise decision trees we use trees of depth one (MRF), or six (DTF).

The results for the large occlusion task are shown in
Figure~\ref{fig:chinese-inpaint-test}.
Qualitatively, they show the difference between a rich connectivity structure
and conditional interactions.  Observe, for example, that the MRF essentially
performs only a smoothing of the results while respecting local stroke-width
constraints, as apparent from the MRF MAP prediction in the first row of
Figure~\ref{fig:chinese-inpaint-test}.
In contrast, the DTF predictions hallucinate meaningful structure that may be
quite different from the ground truth but bears similarity to Chinese
characters.  Note that we achieve this rich structure without the use of any
latent variables.
Because this task is an inpainting task, the quantitative assessment is more
difficult since the task is truly ambiguous.  We therefore report accuracies
only for the small-occlusion case, where a reasonable reconstruction of the
ground truth seems more feasible.
We measure the per-pixel accuracy in the occluded area on the test set.  For
the random forest baseline we obtain $67.74\%$.  The MRF with dense
neighborhood improves this to $75.18\%$ and the DTF obtains $76.01\%$.

% Example of what we learn
As an example of what structure the model is able to learn, consider the
visualization of the MRF pairwise interactions shown in
Figure~\ref{fig:chinese-weights}.
The figure shows for each pairwise interaction the sum of learned diagonal
energies minus the sum of cross-diagonal entries.  If this value is negative
(shown in blue) the interaction is encouraging the pixels to take the same
value.  Red marks interactions that encourage pixels to take different values.
The plot shows that there is a strong local smoothing term, but interestingly
it is not symmetric.  This can be explained by the fact that horizontal
strokes in Chinese characters are typically slanted slightly
upwards~\cite{chen2011chinesecalligraphy}.
Note that we discovered these regularities automatically from the training
data.

\subsection{Accurate body-part detection}\label{sec:bodyparts}
\begin{figure}[t!]
\begin{center}
	\includegraphics[width=0.995\linewidth]{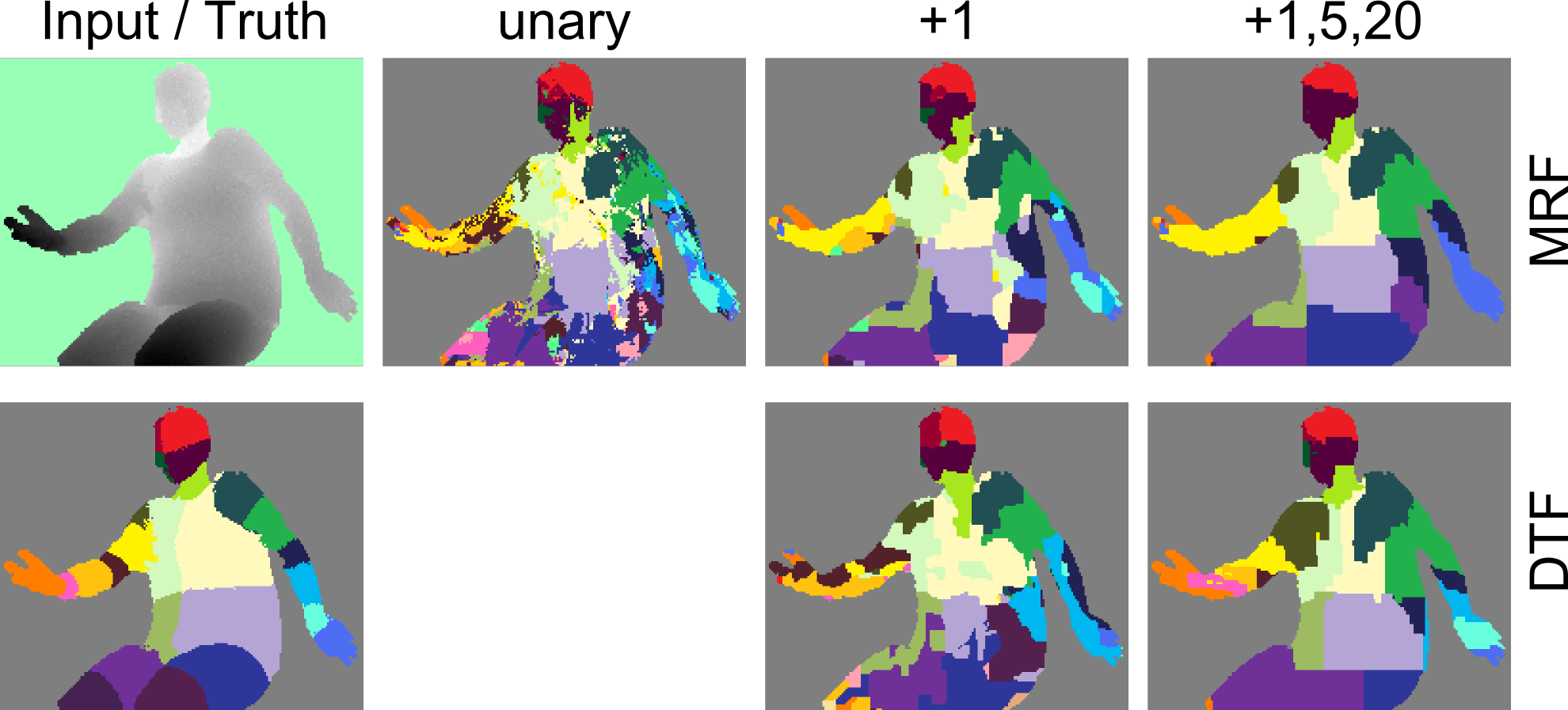}%
\end{center}
\caption{Test recognition results.  MRF (top) and DTF (bottom).}
\label{fig:KinectEx}
\end{figure}

We consider the task of body part classification from depth images, as
recently proposed in~\cite{shotton2011real}.  Given a 2D depth image, and a
foreground mask, the task is to label each pixel as belonging to one of 31
different body parts, as shown in Figure~\ref{fig:KinectEx}.
Despite the variations in pose and body sizes~\cite{shotton2011real} obtains
high-quality recognition results by evaluating a random forest for each pixel,
testing local and global depth disparities.
In this task, the label set has a large amount of structure, but it is not
clear that a sufficiently complex unary classifier, when given the image,
cannot implicitly represent this structure reasonably well.
Here we show that by adding pairwise interactions we in fact improve the
recognition accuracy.  Moreover, once we make the interactions conditional,
accuracy improves even further.

% Experimental setup
The experimental setup is as follows.
We use a subset of the data used in~\cite{shotton2011real}: 30 depth images
for training, and 150 images for testing.
We train 4 unary decision trees for all models.  For the pairwise models, we
use the following neighborhood sizes, (i) ``+1'' for adding a 4-neighborhood
one pixel away, (ii) ``+5'' for an 8-neighborhood five pixels away, and (iii)
``+20'' when adding an 8-neighborhood twenty pixels away.  In the ``+1,5,20''
configuration, each variable has 4+8+8=20 neighbors.
For each of the pairwise interactions we train two trees of depth six.  A
more detailed description of the exact experimental setup can be found in the
supplementary materials.
We measure the results using the same mean per-class accuracy score as used
in~\cite{shotton2011real}.

% Results
The results for 30 training images are shown in
Table~\ref{tab:kinect-results} and one instance is shown in
Figure~\ref{fig:KinectEx}.
%
% Discuss trends in results
Even without adding pairwise interactions, our learned unary weights already
outperform the random forest classifier~\cite{shotton2011real}.
When adding more interactions (+1, +1,20, +1,5,20), the performance increases
because dense pairwise interactions can represent implicit size preferences
for the body parts.
Likewise, when adding conditionality (MRF to DTF), the performance improves.
The best performing model is our DTF with large structure (+1,5,20) and almost
1.5 million free parameters.  It is trained in only 22 minutes and achieves
27.35\% mean per-class accuracy.
%
% Jamie's results, large scale argument
For the same setup of 30 images, the original work~\cite{shotton2011real}
reports a mean per class accuracy of 14.8\%, while achieving an impressive
56.6\% when scaling to 900k training images, trained for one day on a 1000
core cluster.

\begin{figure}
\begin{center}
\hfill
\begin{minipage}[b]{1.9cm}%
    \includegraphics[height=3.4cm]{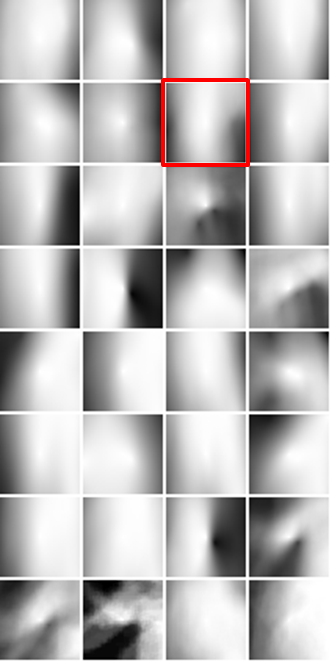}%
\end{minipage}%
\begin{minipage}[b]{1.9cm}%
    \includegraphics[height=1.7cm]{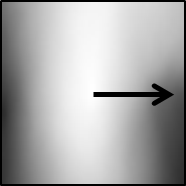}%
	\vspace{0.05cm}\\
    \includegraphics[height=1.7cm]{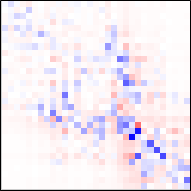}
\end{minipage}%
\begin{minipage}[b]{12.5cm}%
    \includegraphics[height=3.4cm]{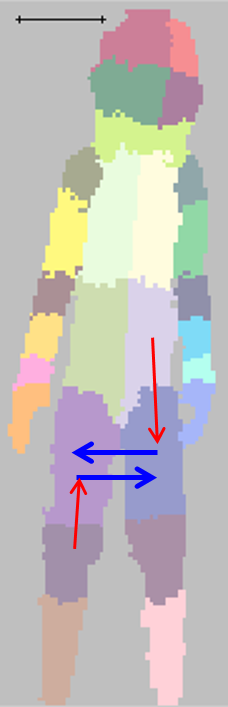}%
    \hspace{0.15cm}%
    \includegraphics[height=3.4cm]{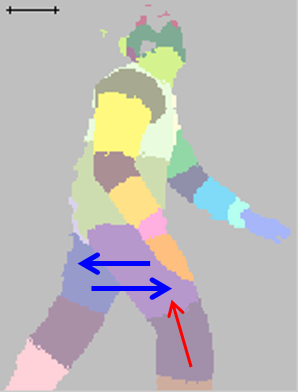}%
    \hspace{0.15cm}%
    \includegraphics[height=3.4cm]{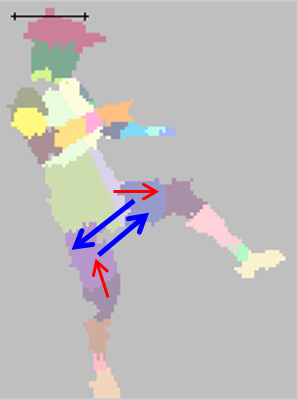}%
    \hspace{0.15cm}%
    \includegraphics[height=3.4cm]{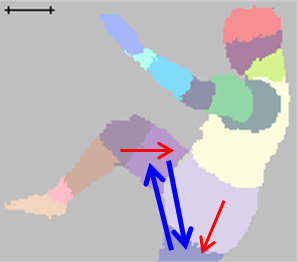}%
\end{minipage}
\hfill%
\end{center}%	
	\caption{
{\bf Learned horizontal interactions:}
The left figure shows the average depth-normalized silhouette reaching one of
the 32 leaf nodes in the learned decision tree.
We select one leaf (marked red, enlarged) and show the corresponding effective
$32\times32$ weight matrix obtained by summing the learned weights along the
path from the root to leaf node.
The conditional interaction can be understood by visualizing the most
attractive ({\color{Blue}blue}) and most repulsive ({\color{Red}red}) elements
in the matrix.  We superimpose arrows for the two most attractive and
repulsive interactions on test images (right).
The first and second pose exemplify how left and right upper parts of the legs
can appear 20 pix to the right of each other in a way that matches the pattern
of the leaf.
While a configuration like shown in the third and fourth pose is plausible, it
does not fit the leaf pattern and thus the interaction is not active.
}%
    \label{fig:kinect-horz}%
\end{figure}

% Example: interaction
An example of a learned pairwise interaction is shown in
Figure~\ref{fig:kinect-horz}, demonstrating that the improved performance of
the DTF can be directly attributed to the more powerful interactions that are
allowed to take the image into account.  We report more results in the
supplementary materials.
\begin{table}[!hbt]
\begin{center}
\begin{small}
\begin{tabular}{llrrrrr}
Model & Measure & \cite{shotton2011real} & unary & +1 & +1,20 & +1,5,20\\
\hline
\hline
MRF	& avg-acc & 14.8 & 21.36 & 21.96 & 23.64 & 24.05\\
	& runtime & 1m & 3m18 & 3m38 & 10m & 10m\\
	& weights & - & 176k & 178k & 183k & 187k\\
%	& weights & 175,520 & 177,568 & 183,264 & 187,360\\
\hline
DTF	& avg-acc & - & - & 23.71 & 25.72 & \textbf{27.35}\\
	& runtime & - & - & 5m16 & 17m & 22m\\
	& weights & - & - & 438k & 951k & 1.47M\\
%	& weights & & 438,112 & 950,880 & 1,469,344\\
\hline
\end{tabular}%
\end{small}%
\end{center}%
\vspace{-0.15cm}%
\caption{Body-part recognition results: mean per-class accuracy, training
time on a single 8-core machine, and number of model parameters.}
\label{tab:kinect-results}
\end{table}

% Kinect figure: 20 pixel horizontal potential

\section{Conclusion}
\label{sec:con}
We have introduced Decision Tree Fields as flexible and accurate models for
image labeling tasks.  This accuracy is achieved by being able to represent
complex image-dependent structure between labels.  Most importantly, this
expressiveness is achieved without the use of latent variables and therefore
we can learn the parameters of our model efficiently by minimizing a convex
function.

%{\footnotesize
%\bibliographystyle{ieee}
%\bibliography{paper}
%}
%
%\end{document}

%% file: approximations_main.tex
%\part{Approximations}
\part{Approximate Optimization}
\label{part:approx}

In the previous part several different applications in the domain of computer vision were presented to demonstrate the
enhanced descriptive power gained by considering arbitrary energies.
However, this gain comes with a price tag: existing optimization algorithms no longer provide good approximations in practice.
This part of my work addresses this issue.
It concentrates around the second axis of this thesis, which focuses on practical methods and approaches for approximating the minimization of challenging arbitrary discrete energies.

In particular, Chapter~\ref{cp:CC} proposes a discrete optimization approach to the correlation clustering functional presented in Chapters~\ref{cp:negaff-sketch} and~\ref{cp:negaff-cc}.
This approach scales gracefully with the number of variables, better than existing approaches (\cite{Vitaladevuni2010}).
In fact, we show that our discrete approach to the CC optimization can handle energies defined over hundreds of thousands of variables, as arise in e.g., image segmentation (Sec.~\ref{sec:cc-app-uimos}).
This is by far more variables than any other existing method can handle.

Chapter~\ref{cp:multiscale} concludes this part with a more general perspective on discrete optimization.
This new perspective is inspired by multiscale approaches and suggests to cope with the NP-hardness of discrete optimization using the {\em multiscale landscape} of the energy function.
Defining and observing this multiscale landscape of the energy, I propose methods to explore and exploit it to derive coarse-to-fine optimization framework.
This new perspective gives rise to a unified multiscale framework for discrete optimization.
The proposed multiscale approach is applicable to a diversity of discrete energies, both smoothness-encouraging as well as arbitrary, contrast-enhancing functions.

%---------------------------------------------------------------------------%
\chapter[Correlation Clustering Optimization]{Correlation Clustering Optimization\protect\footnotemark{}}\protect\footnotetext{This is joint work with Meirav Galun}
\label{cp:CC}

The focus of this chapter is the optimization of the Correlation Clustering functional which combines positive and negative affinities between the data points.
The main contribution of this chapter are new optimization algorithms which can cope with large scale problems
($>100K$ variables) that are infeasible using existing methods.
We draw an analogy between optimizing this functional and the well known Potts energy minimization.
This analogy allows us to suggest several new optimization algorithms,
which exploit the intrinsic ``model-selection" capability of the functional
to automatically recover the underlying number of clusters.
We compare our algorithms to existing methods on both synthetic and real data.

\input{cc}

%---------------------------------------------------------------------------%
\chapter[Discrete Multiscale Optimization]{Discrete Multiscale Optimization\protect\footnotemark{}}\protect\footnotetext{This is joint work with Meirav Galun. It was published in the $5^{th}$ NIPS workshop on optimization for machine learning, \citeyear{Bagon2012opt}.}  	
\label{cp:multiscale}
This chapter presents a unified multiscale framework for discrete energy minimization
that directly acts on the energy.
Our approach utilizes algebraic multiscale principles to efficiently explore the discrete solution space.
The main goal of our framework is to improve optimization performance for challenging, non-submodular energies for which current methods provide unsatisfactory approximations.
Furthermore, the ability to derive a multiscale pyramid directly from an energy makes our framework application independent.
Two important implications rise from this independence:
(i)~One no longer needs to tailor a multiscale scheme to suit each different application.
(ii)~Our framework makes it trivial to turn existing single scale optimization algorithms into powerful multiscale methods.
Our framework gives rise to two complementary energy coarsening strategies:
one in which coarser scales involve fewer variables,
and a more revolutionary one in which the coarser scales involve fewer discrete labels.
We empirically evaluated our unified framework on a variety of both non-submodular and submodular energies, including energies from the Middlebury benchmark.

\input{ms}

% Copies of papers published by the student may be attached to the Final Report as appendices.

%% file: cc.tex
\section{Introduction}
\label{sec:cc-opt-intro}

Optimizing CC is tightly related to many graph partitioning formulations (\cite{Nowozin2009}), however it is
known to be NP-hard (\cite{Bansal2004}).
Existing methods derive convex continuous relaxations to approximately optimize the CC functional.
However, these algorithms do not scale beyond a few thousands of variables.
See for example, the works of \cite{Nowozin2009,Bagon2010,Vitaladevuni2010,Glasner2011}.
%Bagon \etal\ \shortcite{Bagon2010}, Vitaladevuni and Basri \shortcite{Vitaladevuni2010}, and Glasner~\etal\ %\shortcite{Glasner2011}.

This work suggests a new perspective on the CC functional, showing its analogy to the known {\em Potts model}.
This new perspective allows us to leverage on recent advances in discrete optimization to propose new CC optimization algorithms.
We show that our algorithms scale to large number of variables ($>100K$), and in fact can tackle tasks that were {\bf infeasible in the past}, e.g., applying CC to pixel-level image segmentation.
In addition, we provide a {\em rigorous statistical interpretation} for the CC functional and justify its intrinsic model selection capability.
Our algorithms exploit this ``model selection" property to automatically recover the underlying number of clusters $k$.

%
%-------------------------------------------------------------------------------------------------------------------%

%-------------------------------------------------------------------------------------------------------------------%
%
\section{CC Optimization: Continuous Perspective}\label{sec:existing-cc-optimization}

%---------------------------------------------%
% existing optimization methods
Optimizing the correlation clustering functional (Eq.~(\ref{eq:CorrClust})) is NP-hard (\cite{Bansal2004}).
Instead of solving {\bf directly} for a partition $U$, existing methods optimize {\bf indirectly} for the binary adjacency matrix $X=UU^T$, i.e., $X_{ij}=1$ iff $i$ and $j$ belong to the same cluster.
By introducing the binary adjacency matrix the quadratic objective (w.r.t. $U$): $-\sum_{ij} W_{ij}\uut$ becomes linear (w.r.t. $X$): $-\sum_{ij} W_{ij}X_{ij}$.
The connected components of $X$, after proper rounding, are the resulting clusters, and the number of clusters $k$ naturally emerges.
Indirect optimization methods must ascertain that the feasible set consists only of ``decomposable" $X$: $X=UU^T$.
This may be achieved either by posing semi-definite constraints on $X$ (\cite{Vitaladevuni2010}), or by introducing large number of linear inequalities (\cite{Demaine2003,Vitaladevuni2010}).
These methods take a continuous and convex relaxation approach to approximate the resulting functional.
This approach allows for nice theoretical properties due to the convex optimization at the cost of a very restricted scalability.

Solving for $X$ requires $\sim n^2$ variables instead of only \mbox{$\sim n$} when solving directly for $U$.
Therefore, these methods scale poorly with the number of variables $n$, and in fact, they cannot handle more than a few hundreds of variables.
In summary, these methods suffer from two drawbacks: (i)~recovering $U$ from $X$ is highly susceptible to noise and more importantly (ii)~it is {\em infeasible} to solve large scale problems by these methods.

%-------------------------------------------------------------------------------------------------------------------%
%
\section{Our New Perspective on CC}
\label{sec:perspective}

Existing methods view the CC optimization in the context of convex relaxation and build upon methods and approaches that are common practice in this field of continuous optimization.
We propose an alternative perspective to the CC optimization:
{\em viewing it as a discrete energy minimization}.
This new perspective allows us to build upon recent advances in discrete optimization and propose efficient and direct CC optimization algorithms.
More importantly, the resulting algorithms solve {\em directly} for $U$, and thus scales significantly better with the number of variables.

We now show how to cast the CC functional of Eq.(\ref{eq:CorrClust}) as a discrete pair-wise conditional random field (CRF) energy.
For ease of notation, we describe a partition $U$ using a labeling vector $L\in\left\{1,2\ldots\right\}^n$: $l_i = c$ iff  $U_{ic}=1$.
A general form of pair-wise CRF energy is $E\left(L\right)=\sum_i E_i\left(l_i\right) + \sum_{ij} E_{ij}\left(l_i, l_j\right)$ (\cite{Boykov2001}). Discarding the unary term ($\sum_i E_i\left(l_i\right)$), and taking the pair-wise term to be $W_{ij}$ if $l_i \ne l_j$ we can re-write the CC functional as a CRF energy:
\begin{eqnarray}
\mathcal{E}_{CC}\left(L\right) = \sum_{ij} W_{ij} \cdot \mathbbm{1}_{\left[l_i\ne l_j\right]} \label{eq:CorrClustCRF}
\end{eqnarray}
This is a Potts model.
Optimizing the CC functional can now be interpreted as searching for a MAP assignment for the energy (\ref{eq:CorrClustCRF}).

The resulting Potts energy has three unique characteristics, each posing a challenge to the optimization process:\\*
\noindent(i)~{\bf Non-submodular:} The energy is non-submodular.
The notion of submodularity is the discrete analogue of convexity from continuous optimization (\cite{Lovasz1983}).
Optimizing a non-submodular energy is NP-hard, even for the binary case (\cite{Rother2007}).\\*
\noindent(ii)~{\bf Unknown number of labels:}
Most CRF energies are defined for a fixed and known number of labels.
Thus, the search space is restricted to $L\in\left\{1,\ldots,k\right\}^n$.
When the number of labels $k$ is unknown the search space is by far larger and more complicated.\\*
\noindent(iii)~{\bf No unary term:} There is no unary term in the energy.
The unary term plays an important role in guiding the optimization process (\cite{Szeliski2008}).
Moreover, a strong unary term is crucial when the energy in non-submodular (\cite{Rother2007}).

There exist examples of CRFs in the literature that share some of these characteristics (e.g., non-submodular \cite{Rother2007,kolmogorov2005}, unknown number of labels \cite{Isack2011,Bleyer2010}).
Yet, to the best of our knowledge, no existing CRF exhibits all these three challenges at once.
More specifically, we are the first to handle non-submodular energy that has no unary term.
Therefore, we cannot just use ``off-the-shelf" Potts optimization algorithms,
but rather modify and improve them to cope with the three challenges posed by the CC energy.

%
%-------------------------------------------------------------------------------------------------------------------%

%-------------------------------------------------------------------------------------------------------------------%
%
%----------------------------------------------------------------------%
\begin{algorithm}[t!]
%\caption{Correlation Clustering -- $\alpha$-expand\label{alg:a-expand}}
\caption{Expand-and-Explore}\label{alg:a-expand}
\DontPrintSemicolon
% \LinesNumbered
\SetKw{KwInit}{Init}
\SetKwFunction{KwExpand}{Expand}
\SetKwFunction{Weinberg}{$\mathcal{E}_{CC}$}

\KwIn{Affinity matrix $W\in\mathbb{R}^{n\times n}$}
\KwOut{Labeling vector $L\in\left\{1,2,\ldots\right\}^n$}
\BlankLine

\KwInit{$L_i\leftarrow 1$, $i=1,\ldots,n$}\tcp*[f]{initial labeling}\;
\Repeat{$L$ is unchanged}{
    \For{$\alpha\leftarrow1$ ; $\alpha\le\#L+1$ ; $\alpha++$}{
%    $\alpha \leftarrow 1$ \;
%    \While {$\alpha\le\#L+1$}{
        $L \leftarrow $ \KwExpand{$\alpha$}\;
        % $\alpha++$\;
    } % while \alpha
} % repeat
\BlankLine

$\#L$ denotes the number of different labels in $L$.\;
\KwExpand{$\alpha$}: expanding $\alpha$ using QPBOI.\;
By letting $\alpha = \#L+1$ the algorithm ``expand" and explore an empty label. This may affect the number of labels $\#L$.\;
%\KwEnergy{$L$} is the correlation clustering score of labeling $L$ according to Eq.~(\ref{eq:CorrClust}).\;
\end{algorithm}
%----------------------------------------------------------------------%

\section{Our Large Scale CC Optimization}
\label{sec:alg}

In this section we adapt known discrete energy minimization algorithms to cope with the three challenges posed by the CC energy.
We derive three CC optimization algorithms that stem from either large move making algorithms ($\alpha$-expand and $\alpha\beta$-swap of \cite{Boykov2001}), or Iterated Conditional Modes (ICM) of \cite{Besag1986}.
Our resulting algorithms scale gracefully with the number of variables $n$, and solve CC optimization problems that were {\em infeasible} in the past.
Furthermore, our algorithms take advantage of the intrinsic model selection capability of the CC functional (Sec.~\ref{sec:cc-theory}) to robustly recover the underlying number of clusters.

\subsection{Improved large move making algorithms}
\cite{Boykov2001} introduced a very effective method for multi-label energy minimization that makes large search steps by iteratively solving binary sub-problems.
There are two large move making algorithms: $\alpha$-expand and $\alpha\beta$-swap that differ by the binary sub-problem they solve.
$\alpha$-expand consider for each variable whether it is better to retain its current label or flip it to label $\alpha$.
The binary step of $\alpha\beta$-swap involves only variables that are currently assigned to labels $\alpha$ or $\beta$, and consider whether it is better to retain their current label or switch to either $\alpha$ or $\beta$.
Defined for submodular energies, the binary step in these algorithms is solved using graph-cut.

We propose new optimization algorithms: {\em Expand-and-Explore} and {\em Swap-and-Explore}, 
%inspired by $\alpha$-expand and $\alpha\beta$-swap, 
that can cope with the challenges of the CC energy.
(i)~For the binary step we use a solver that handles non-submodular energies.
(ii)~We incorporate ``model selection" into the iterative search to recover the underlying number of clusters $k$.
(iii)~In the absence of unary term, a good initial labeling is provided to the non-submodular binary solver.

Binary non-submodular energies can be approximated by an extension of graph-cuts: QPBO (\cite{Rother2007}).
When the binary energy is non-submodular QPBO is not guaranteed to provide a labeling for all variables.
Instead, it outputs only a partial labeling. How many variables are labeled depends on the amount of non-submodular pairs and the relative strength of the unary term for the specific energy.
When no unary term exists in the energy QPBO leaves most of the variables unlabeled. To circumvent this behavior we use the ``improve" extension of QPBO (denoted by QPBOI): This extension is capable of improving an initial labeling to find a labeling with lower energy (\cite{Rother2007}).
In the context of expand and swap algorithms a natural initial labeling for the binary steps is to use the current labels of the variables and use QPBOI to improve on it, ensuring the energy does not increase during iterations.

To overcome the problem of finding the number of clusters $k$ our algorithms do not iterate over a fixed number of labels, but explore an ``empty" cluster in addition to the existing clusters in the current solution. Exploring an extra empty cluster allows the algorithms to optimize over all solutions with any number of clusters $k$. The fact that there is no unary term in the energy makes it straight forward to perform. Alg.~\ref{alg:a-expand} and Alg.~\ref{alg:ab-swap} presents our {\em Expand-and-Explore} and {\em Swap-and-Explore} algorithms in more detail.
%----------------------------------------------------------------------%
\begin{algorithm}[t!]
%\caption{Correlation Clustering -- $\alpha\beta$-swap\label{alg:ab-swap}}
\caption{Swap-and-Explore}\label{alg:ab-swap}
\DontPrintSemicolon
% \LinesNumbered
\SetKw{KwInit}{Init}
\SetKwFunction{KwSwap}{Swap}
\SetKwFunction{KwEnergy}{$\mathcal{E}_{CC}$}

\KwIn{Affinity matrix $W\in\mathbb{R}^{n\times n}$}
\KwOut{Labeling vector $L\in\left\{1,2,\ldots\right\}^n$}
\BlankLine

\KwInit{$L_i\leftarrow 1$, $i=1,\ldots,n$}\tcp*[f]{initial labeling}\;
\Repeat{$L$ is unchanged}{
    \For{$\alpha\leftarrow1$ ; $\alpha\le\#L$ ; $\alpha++$}{
        \For{$\beta\leftarrow\alpha$ ; $\beta\le\#L+1$ ; $\beta++$}{
            $L \leftarrow $ \KwSwap{$\alpha$, $\beta$}\;
        } % For \beta
    } % For \alpha
%    $\alpha, \beta \leftarrow 1$ \;
%    \While {$\alpha\le\#L$, $\alpha\le\beta\le\#L+1$}{
%        $L \leftarrow $ \KwSwap{$\alpha$, $\beta$}\;
%    } % while \alpha, \beta
} % repeat
\BlankLine

$\#L$ denotes the number of different labels in $L$.\;
\KwSwap{$\alpha$, $\beta$}: swapping labels $\alpha$ and $\beta$ using QPBOI.\;
By letting $\beta = \#L+1$ the algorithm explore new number of clusters, this may affect the number of labels $\#L$.\;
%\KwEnergy{$L$} is the correlation clustering score of labeling $L$ according to Eq.~(\ref{eq:CorrClust}).\;
\end{algorithm}
%----------------------------------------------------------------------%

\subsection{Adaptive-label ICM}
Another discrete energy minimization method that we modified to cope with the three challenges of the CC optimization is ICM (\cite{Besag1986}).
It is a point-wise greedy search algorithm.
Iteratively, each variable is assigned the label that minimizes the energy, conditioned on the current labels of all the other variables.
ICM is commonly used for MAP estimation of energies with a {\em fixed} number of labels.
Here we present an {\em adaptive-label ICM}: using the ICM conditional iterations
%of labeling a single variable while all other labels are kept fixed, to
we adaptively determine the number of labels $k$.
Conditioned on the current labeling, we assign each point to the cluster it is most attracted to, or to a singleton cluster if it is repelled by all.

\ % leave a blank line here

In this section we proposed a new perspective on CC optimization.
Interpreting it as MAP estimation of Potts energy allows us to propose a variety of efficient optimization methods\footnote{Matlab implementation available at: \protect\url{http://www.wisdom.weizmann.ac.il/~bagon/matlab.html}.}:
\begin{itemize}
  \item Swap-and-Explore (with binary step using QPBOI)
  \item Expand-and-Explore (with binary step using QPBOI)
  \item Adaptive-label ICM
\end{itemize}

% PROPERTIES:
% Solves directly for U - does not require post process
% low resources - compact representation of U - can handle hundreds of thousands of variables
%
Our proposed approach has the following advantages:\\*
\noindent(i)~It solves only for $n$ integer variables.
This is by far less than the number of variables required by existing methods described in Sec.~\ref{sec:existing-cc-optimization}, which require $\sim n^2$ variables of the adjacency matrix $X=UU^T$. It makes our approach capable of dealing with large number of variables ($>100K$) and suitable for pixel-level image segmentation.\\*
\noindent(ii)~The algorithms solve directly for the cluster membership of each point, thus there is no need for rounding scheme to extract $U$ from the adjacency matrix $X$.\\*
\noindent(iii)~The number of clusters $k$ is optimally determined by the algorithm and it does not have to be externally supplied like in many other clustering/segmentation methods.

In their work \cite{elsner2009} proposed a greedy algorithm to optimize the CC functional over complete graphs. Their algorithm is in fact an ICM method presented outside the proper context of CRF energy minimization, and thus does not allow to generalize the concept of discrete optimization to more recent optimization methods.

%
%-------------------------------------------------------------------------------------------------------------------%

%-------------------------------------------------------------------------------------------------------------------%
%
\section{Experimental Results} \label{sec:synthetic}

This section evaluates the performance of our proposed optimization algorithms using both synthetic and real data.
We compare to both existing discrete optimization algorithms that can handle multi-label non-submodular energies (TRW-S of \cite{Kolmogorov2006} and BP of \cite{Pearl1988}\footnote{Since these two algorithms work only with pre-defined number of clusters $k$, we over-estimate $k$ and report only the number of {\em non empty} clusters in the solution.}), and to existing state-of-the-art CC optimization method of \cite{Vitaladevuni2010}.
Since existing CC optimization methods do not scale beyond several hundreds of variables, extremely small matrices are used in the following experiments.
Interactive segmentation results (Sec.~\ref{sec:cc-app-uimos}) evaluated our method on large scale problems.

\subsection{Synthetic data}
\begin{figure}
\centering
\begin{tabular}{cc}
(a) Energy (lower=better) & (b) Recovered $k$ (GT in dashed) \\
\includegraphics[width=.25\linewidth]{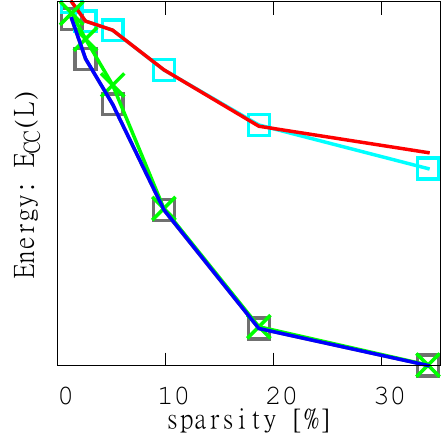}&
\includegraphics[width=.25\linewidth]{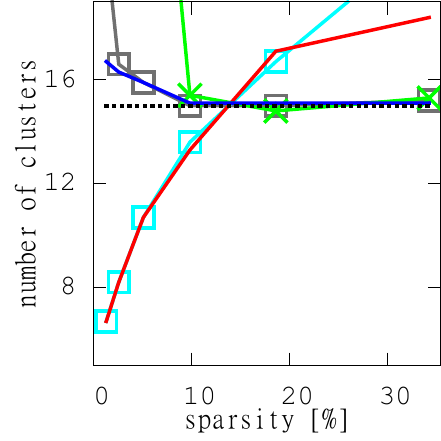}\\
&\\
(c) Purity & (d) Run time \\
\includegraphics[width=.25\linewidth]{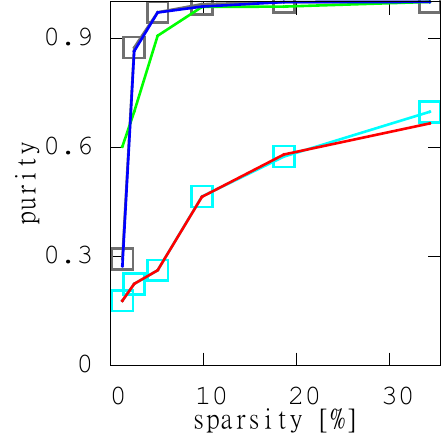}&
\includegraphics[width=.25\linewidth]{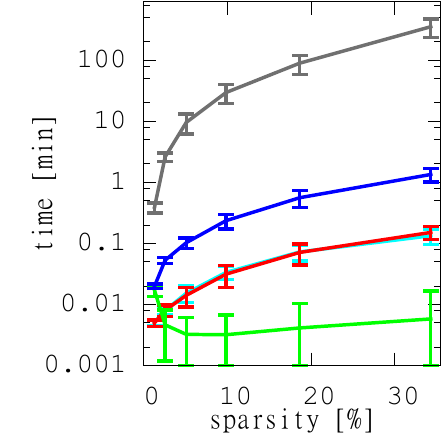}\\
\multicolumn{2}{c}{Legend:{\color{swap} Swap-and-Explore}, {\color{expand} Expand-and-Explore}, {\color{icm} ICM}, {\color{trws} TRW-S}, {\color{bp} BP}
}
\end{tabular}
\caption{{\bf Synthetic results: }{\em Graphs comparing (a)~Energy at convergence. (b)~Recovered number of clusters. (c)~Purity of resulting clusters. (d)~Run time of algorithms (log scale).
{\color{trws}TRW-S} and {\color{bp}BP} are almost indistinguishable, as are {\color{swap}Swap} and {\color{expand}Expand} in most of the plots.
}}
\label{fig:synth-res}
\end{figure}

This experiment uses synthetic affinity matrices $W$ to compare our algorithms to existing Potts optimization algorithms.
The synthetic data have 750 variables randomly assigned to 15 clusters with different sizes
(ratio between larger to smaller cluster: $\sim\times5$). For each variable we sampled roughly the same number of neighbors: of which $\sim25\%$ are from within the cluster and the rest from the other clusters.
We corrupted the clean ground-truth adjacency matrix with $20\%$ noise affecting both the sign of $W_{ij}$ and the certainty (i.e., $\left|W_{ij}\right|$). Overall the resulting percent of positive (submodular) connections is $\sim30\%$.

We report several measurements for these experiments: run-time, energy ($\mathcal{E}_{CC}$), purity of the resulting clusters and the recovered number of clusters $k$ for each of the algorithms as a function of the sparsity of the matrix $W$, i.e., percent of non-zero entries. Each experiment was repeated $10$ times with different randomly generated matrices.

Fig.~\ref{fig:synth-res} shows results of the synthetic experiments.
Existing multi-label approaches ({\color{trws}TRW-S} and {\color{bp}BP}) do not perform too well: higher $\mathcal{E}_{CC}$, lower purity and incorrect recovery of $k$.
This demonstrates the difficulty of the energy minimization problem that has no unary term and many non-submodular pair-wise terms. These results are in accordance with the observations of \cite{kolmogorov2005} when the energy is hard to optimize.

For our large move making algorithms, {\color{expand}Expand-and-Explore} provides marginally better clustering results than the {\color{swap}Swap-and-Explore}. However, its relatively slow running time makes it infeasible for large CC problems\footnote{This difference in run time between  Expand and Swap can be explained by looking at the number of variables involved in each of the binary steps carried out: For the expand algorithm, each binary step involves almost all the variables, while the binary swap move considers only a small subset of the variables.}.
A somewhat surprising result of these experiments shows that for matrices not too sparse (above $10\%$), {\color{icm} adaptive-label ICM} performs surprisingly well. In fact, it is significantly faster than all the other methods and manages to converge to the correct number of clusters with high purity and low energy.

From these experiments we conclude that {\color{swap}Swap-and-Explore} (Alg.~\ref{alg:ab-swap}) is a very good choice of optimization algorithm for the CC functional. However, when the affinity matrix $W$ is not too sparse, it is worth while giving our {\color{icm}adaptive-label ICM} a shot.

%--------------------------------------------------------%
\subsection{Co-clustering data}

The following experiment compares our algorithms with a state-of-the-art CC optimization method, R-LP, of \cite{Vitaladevuni2010}.
For this comparison we use affinity matrices computed for co-segmentation.
The co-segmentation problem can be formulated as a correlation clustering problem with super pixels as the variables (\cite{Glasner2011}).

We obtained 77 affinity matrices, courtesy of \cite{Glasner2011}, used in their experiments.
The number of variables in each matrix ranges from 87 to 788,
Their sparsity (percent of non-zero entries) ranges from $6\%$ to $50\%$,
and there are roughly the same number of positive (submodular) and negative (non-submodular) entries.

Table~\ref{tab:comp-daniel} shows the ratio between our energy and the energy of R-LP method. The table also shows the percent of matrices for which our algorithms found a solution with lower energy than R-LP.
The results show the superiority of our algorithms to existing multi-label energy minimization approaches (TRW-S and BP). Furthermore, it is shown that our methods are in par with existing state-of-the-art CC optimization method on small problems.
However, unlike existing methods, our algorithms can be applied to problems {\em two orders of magnitude larger}.
Optimizing directly for $U$ not only did not compromise the performance of our method, but also allows us to handle large scale CC optimization, as demonstrated in the next section.

\begin{table}
\centering
\begin{tabular}{c||c|c|c||c|c}
&\multicolumn{3}{c||}{Ours}& & \\
& Swap & Expand & ICM & TRWS & BP  \\
\hline \hline
Energy ratio & $98.6$   & $98.4$   & $77.4$    & $83.8$   &  $83.6$  \\
 (\%)        & $\pm1.4$ & $\pm1.9$ & $\pm23.9$ & $\pm5.4$ & $\pm6.3$  \\
\hline
Strictly lower& \multirow{2}{*}{15\%} &   \multirow{2}{*}{11.7\%} & \multirow{2}{*}{0} & \multirow{2}{*}{0} & \multirow{2}{*}{0} \\
$\left(>100\%\right)$ & & & & &\\
\end{tabular}
\caption{{\bf Comparison to \protect\cite{Glasner2011}:}
{\em Ratio between our energy and of Glasner~\etal:
Since all energies are negative, higher ratio means lower energy.
Ratio higher than $100\%$ means our energy is better than Glasner~\etal.
Bottom row shows the percentage of cases where each method got strictly lower energy than Glasner~\etal.}}
\label{tab:comp-daniel}
\end{table}

%
%-------------------------------------------------------------------------------------------------------------------%

%-------------------------------------------------------------------------------------------------------------------%
%
\section{Conclusion}\label{sec:concl}

This work suggests a new perspective on the functional, viewing it as a discrete Potts energy.
The resulting energy minimization presents three challenges: (i)~the energy is non submodular, (ii)~the number of clusters is not known in advance, and (iii)~there is no unary term.
We proposed new energy minimization algorithms that can successfully cope with these challenges.

%
%-------------------------------------------------------------------------------------------------------------------%

%\section*{Acknowledgments}
%The authors would like to thank these people for their fruitful and insightful remarks:
%Ronen Basri, Michal Irani, Boaz Nadler, Shiv Vitaladevuni, Daniel Glasner, Stella Yu, Tal Hassner and Lena Gorelick.
%
%\bibliographystyle{my_aistats}
%\bibliography{negaff}
%
%\end{document}

%% file: ms.tex
\newlength{\origtabcolsep}
\setlength{\origtabcolsep}{\tabcolsep}

\section{Introduction}

\begin{figure}[t]

\centering
\framebox[\linewidth][c]{
\parbox{\linewidth}{
\centering
\includegraphics[width=.75\linewidth]{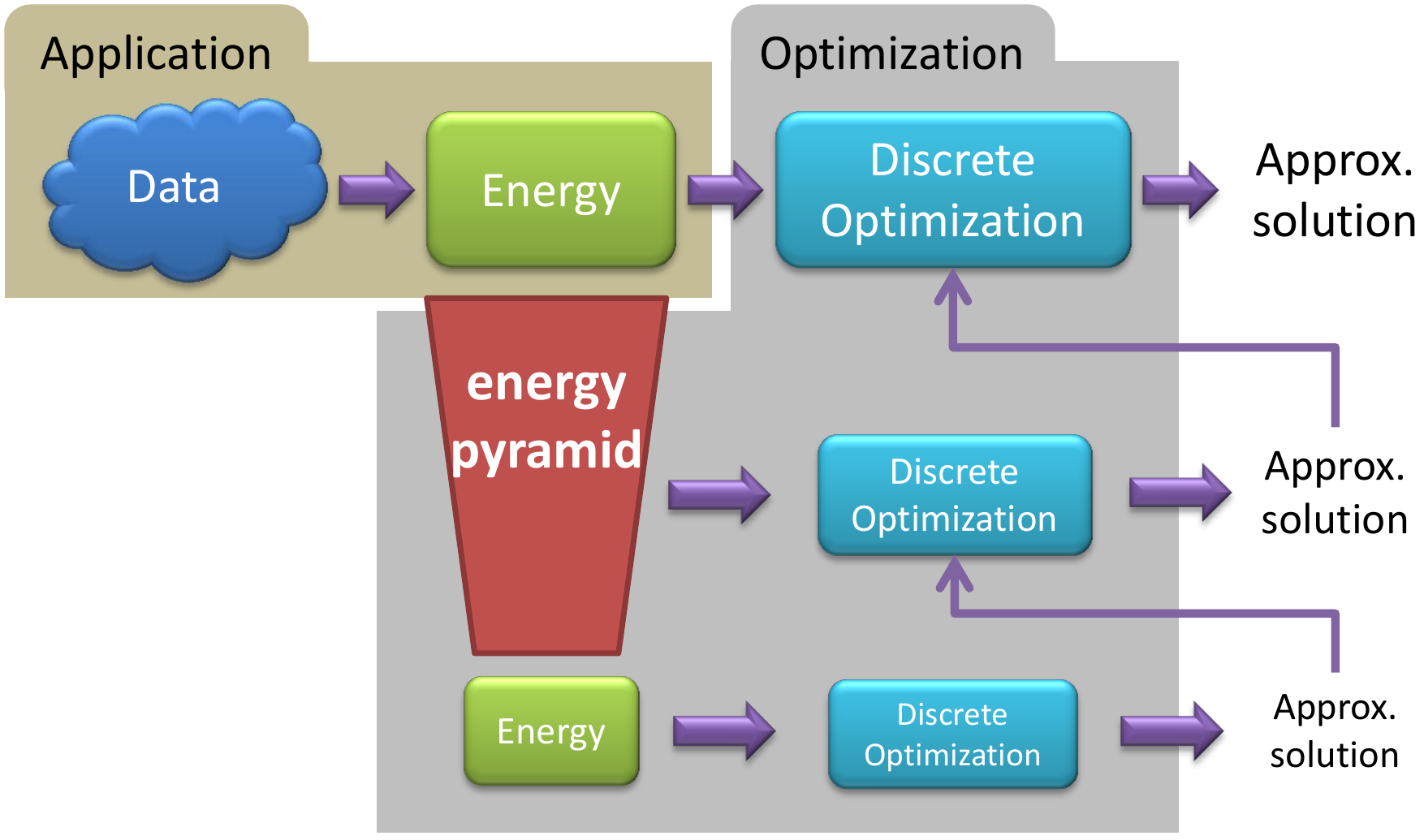}
}}
\vspace*{1mm}
\caption{
{\bf A Unified multiscale framework:}
{\em
We derive multiscale representation of the energy itself = energy pyramid.
Our multiscale framework is unified in the sense that different problems with different energies share the same multiscale scheme, making our framework widely applicable and general.
}}
\label{fig:multiscale-schemes}
\end{figure}

%------------------------------------------------------------------------------------%

Discrete energy minimization is ubiquitous in computer vision, and spans a variety of problems such as segmentation, denoising, stereo, etc.
Unfortunately, apart from the submodular binary case, minimizing these energies is known to be NP-hard.
A lot of effort is recently put on developing algorithms for approximate discrete optimization for ever more challenging energies: multi-label, non-submodular, etc. (e.g., \cite{Szeliski2008,Kolmogorov2006,Bagon2012}).

Discrete energies may be grossly divided into two categories: submodular (regular) energies and non-submodular energies.
Submodular energies are characterized by smoothness-encouraging pair-wise (and higher order) terms.
These energies reflect the ``piece-wise constant" prior that is very popular and common in computer vision applications.
For this reason most of the effort and research regarding discrete optimization, in the context of computer vision, focuses on these energies with encouraging results.
In practice, despite the NP-hardness of these energies, algorithms were developed that provide solutions with energies close to global optimum (e.g., \cite{Kolmogorov2006,Boykov2001}).
Therefore we consider this type of energies as ``easy to optimize".

In contrast, non-submodular energies are characterized by contrast-encouraging pair wise terms.
These energies may be encountered when the parameters of the energy are learned (e.g., \cite{Nowozin2011}),
or when different functionals are used (e.g., \cite{Bagon2012,Glasner2011}).
When it comes to optimization it is generally considered a more challenging task to optimize a non-submodular energies.
Since these examples of non-submodular energies are only recently explored, their optimization receives less attention, and consequently, the existing optimization methods provide approximations that may be quite unsatisfactory.
We consider these energies as ``hard to optimize".

Algorithms for discrete energy minimization may also be classified into two categories: primal methods and dual methods.
Primal methods act directly on the discrete variables in the label space to minimize the energy (e.g., \cite{Besag1986,Boykov2001}).
In contrast, dual methods formulate a dual problem to the energy and maximize a lower bound to the sought energy (e.g., \cite{Kolmogorov2006}).
Dual methods are recently considered more favorable since they not only provide an approximate solution, but also provide a lower bound on how far this solution is from the global optimum.
Furthermore, if a labeling is found with energy equals to the lower bound a certificate is provided that the global optimum was found.

Since most of the relevant discrete optimization problems are NP-hard, one can only provide an {\em empirical} evaluation of how well a given algorithm approximates representative instances of these energies.
For the submodular, ``easy to optimize", energies it was shown (by \cite{Szeliski2008}) that dual methods tend to  provide better approximations with very tight lower bounds.

% P2 - multiscale
\parpic[r][r]{\includegraphics[width=.45\linewidth]{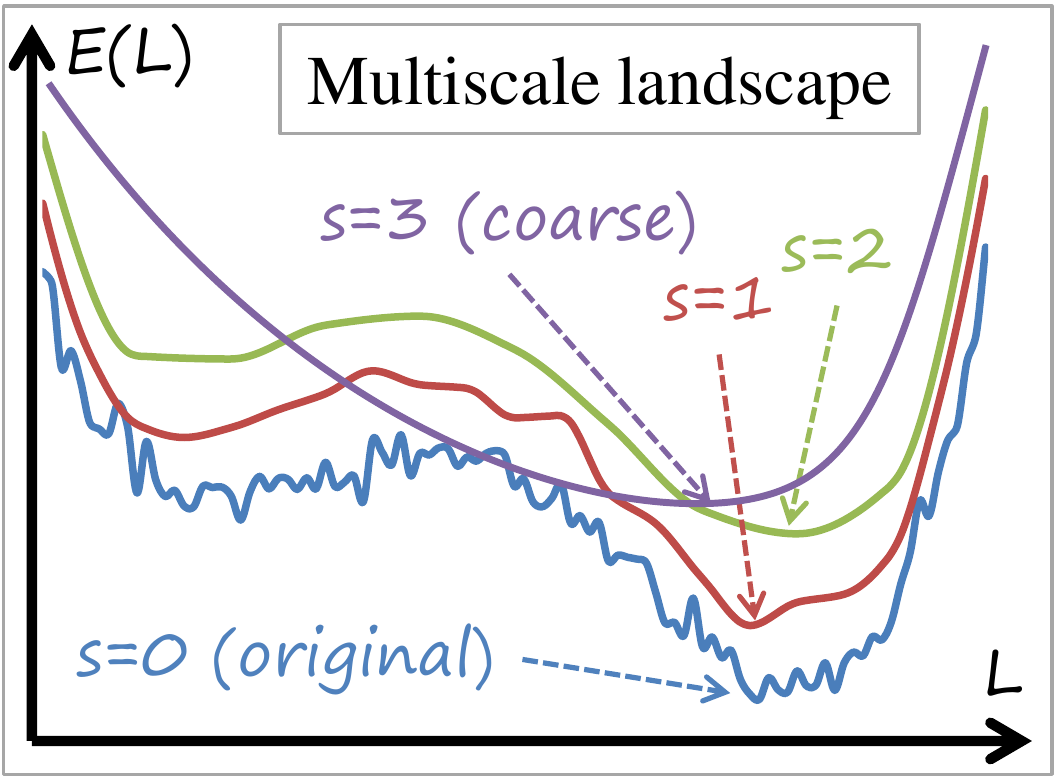}}
But what makes discrete energy minimization such a challenging endeavor?
The fact that this minimization
% Discrete optimization is a very challenging endeavor, since it
implies an exploration of an exponentially large search space makes it such a challenging task.
%Coping with this challenge calls for a sophisticated search strategy.
One way to alleviate this difficulty is to use multiscale search.
The illustration on the right shows
%On the right we illustrate
a toy ``energy" $E(L)$ at different scales of detail.
Considering only the original scale ($s=0$), it is very difficult to suggest an effective exploration/optimization method.
However, when looking at coarser scales ($s=1,\ldots,3$) of the energy an interesting phenomenon is revealed.
At the coarsest scale ($s=3$) the large basin of attraction emerges, but with very low accuracy.
As the scales become finer ($s=2,\ldots,0$), one ``loses sight" of the large basin, but may now ``sense" more local properties with higher accuracy.
We term this well known phenomenon as the {\em multiscale landscape} of the energy.
This multiscale landscape phenomenon encourages coarse-to-fine exploration strategies:
starting with the large basins that are apparent at coarse scales,
and then gradually and locally refining the search at finer scales.

For more than three decades the vision community has focused on the multiscale pyramid of the {\em image} (e.g., \cite{Lucas1981,Burt1983}).
There is almost no experience and no methods that
apply a multiscale scheme directly to the discrete energy.

In this paper we present a novel unified discrete multiscale optimization scheme that acts {\em directly} on the energy
%That is, we propose a method that exposes the multiscale landscape of the {\em energy itself}
(Fig.~\ref{fig:multiscale-schemes}).
Our approach allows for an efficient exploration of the discrete solution space through the construction of an energy pyramid.
Moreover, our multiscale framework is application independent:
different problems with different energies {\em share the same} multiscale scheme, making our framework widely applicable and general.

Performing empirical evaluations of non-submodular energies minimization lead us to conclude that when it comes to hard to optimize non-submodular energies, primal methods tend to provide better approximations than dual methods.
Motivated by this observation, we formulate out multiscale framework in the primal space (i.e., expressing it in terms of the variables and labels directly).
Our multiscale framework becomes the core of the optimization process allowing for existing ``off-the-shelf" primal optimization algorithms to efficiently exploit the multiscale landscape of the energy and achieves significantly lower energies faster.

This work makes several contributions: % \*
\renewcommand{\labelenumi}{(\roman{enumi})}
\begin{enumerate}
  \item A novel unified multiscale framework for discrete optimization.
A wide variety of optimization problems, including segmentation, stereo, denoising, correlation-clustering, and others share the same multiscale framework.

  \item Energy-aware coarsening scheme.
Variable aggregation takes into account the underlying structure of the energy itself, thus efficiently and directly exposes its multiscale landscape.

  \item Coarsening the labels.
Our formulation allows for variable aggregation as well as for label coarsening.
This yields an energy pyramid with  fewer {\em labels} at the coarser scales.

  \item Integrating existing single-scale optimization algorithms into our multiscale framework.
We achieve significantly lower energy assignments on diverse computer vision energies, including challenging non-submodular examples.

  \item Optimizing hard non-submodular energies.
Using several classes of non-submodular energies, we empirically exemplify the superiority of primal methods.
We further show how combining in our multiscale framework single-scale primal optimization methods achieve increased optimization performance on these challenging problems.
\end{enumerate}
\renewcommand{\labelenumi}{\arabic{enumi}.}

%%%\noindent(i)~A novel unified multiscale framework for discrete optimization.
%%%A wide variety of optimization problems, including segmentation, stereo, denoising, correlation-clustering, and others share the same multiscale framework.\*
%%%
%%%\noindent(ii)~Energy-aware coarsening scheme.
%%%Variable aggregation takes into account the underlying structure of the energy itself, thus efficiently and directly exposes its multiscale landscape. \*
%%%
%%%\noindent(iii)~Coarsening the labels.
%%%Our formulation allows for variable aggregation as well as for label coarsening.
%%%This yields an energy pyramid with  fewer {\em labels} at the coarser scales.\*
%%%
%%%\noindent(iv)~Integrating existing single-scale optimization algorithms into our multiscale framework.
%%%We achieve significantly lower energy assignments on diverse computer vision energies, including challenging non-submodular examples.\*
%%%
%%%\noindent(v)~Optimizing hard non-submodular energies.
%%%Using several classes of non-submodular energies, we empirically exemplify the superiority of primal methods.
%%%We further show how combining in our multiscale framework single-scale primal optimization methods achieve increased optimization performance on these challenging problems.

%\noindent(v)~Exploring the energy coarse-to-fine provides lower energy value at shorter run time.

\subsection{Related work}
There are very few works that apply multiscale schemes directly to the energy.
A prominent example for this approach is that of \cite{Felzenszwalb2006}, that provide a coarse-to-fine belief propagation scheme restricted to regular diadic pyramid.
A more recent work is that of \cite{Komodakis2010} that provides an algebraic multigrid formulation for discrete optimization in the dual space.
However, despite his general formulation \citeauthor{Komodakis2010} only provides examples using regular diadic grids of easy to optimize submodular energies.

%They significantly improved the performance of belief propagation by coarse-to-fine message passing.
The work of \cite{Kim2011} proposes a two-scales scheme mainly aimed at improving run-time of the optimization process.
Their proposed coarsening strategies can be interpreted as special cases of our unified framework.
We analyze their underlying assumptions (Sec.~\ref{sec:local-correlations}), and suggest better methods for efficient exploration of the multiscale landscape of the energy.

A different approach for discrete optimization suggests
large move making algorithms (e.g., \cite{Boykov2001,Swendsen1987}).
We experimentally show how plugging such methods within our multiscale framework improves optimization results.
These methods do not scale gracefully with the number of labels.
\cite{Lempitsky2007} proposed a method to exploit known properties of the metric between the labels to allow for faster minimization of the energy.
However, their method is restricted to energies with clear and known label metrics and requires training.
In contrast, our framework addresses this issue via a principled scheme that builds an energy pyramid with a decreasing number of {\em labels} without prior training and with fewer assumptions on the labels interactions.

%------------------------------------------------------------------------------------%
\section{Multiscale Energy Pyramid}
\label{sec:unified}

% Notations
In this work we consider discrete pair-wise minimization problems, defined over a (weighted) graph $\left(\VV, \EE\right)$, of the form:
\begin{eqnarray}
E\left(L\right)&=&\sum_{i\in\VV} \psi_i\left(l_i\right) + \sum_{\left(i,j\right)\in\EE} w_{ij}\cdot \psi\left(l_i,l_j\right) \label{eq:GenEng}
\end{eqnarray}
where $\VV$ is the set of variables, $\EE$ is the set of edges, and the solution is discrete: $L\in\left\{1,\ldots,l\right\}^n$, with $n$ variables taking $l$ possible labels.
Many problems in computer vision are cast in the form of~(\ref{eq:GenEng}) (see \cite{Szeliski2008}).
Furthermore, we do not restrict the energy to be submodular, and our framework is
also applicable to more challenging non-submodular energies.

Our aim is to build an energy pyramid with a decreasing number of degrees of freedom.
The key component in constructing such a pyramid is the interpolation method.
The interpolation maps solutions between levels of the pyramid,
and defines how to approximate the original energy with fewer degrees of freedom.
We propose a novel principled energy aware interpolation method.
The resulting energy pyramid exposes the multiscale landscape of the energy making low energy assignments apparent at coarse levels.

%however, the trivial manner of interpolating continuous variable assignments does not apply when it comes to interpolating discrete assignments.
%
%However, interpolation in the continuous space is not naturally applicable to interpolation in the discrete space.
%For example, a linear interpolation of two discrete values, which have a certain semantic interpretation, may result with completely irrelevant semantic interpretation.

However, it is counter intuitive to directly interpolate discrete values,
since they usually have only semantic interpretation.
%especially if no order is assumed over the labels.
Therefore, we substitute an assignment $L$
by a binary matrix $U\in\left\{0,1\right\}^{n\times l}$.
The rows of $U$ correspond to the variables, and the columns corresponds to labels:
$U_{i,\alpha}=1$ iff variable $i$ is labeled ``$\alpha$" ($l_i=\alpha$).
This representation allows us to interpolate discrete solutions, as will be shown in the subsequent sections.

Expressing the energy (\ref{eq:GenEng}) using $U$ yields a relaxed quadratic representation (along the lines of \cite{Anand2000}) that forms the basis for our multiscale framework derivation:
\begin{eqnarray}
E\left(U\right)&=&Tr\left(DU^T+WUVU^T\right) \label{eq:EngU} \\
      & \mbox{s.t.} & U\in\left\{0,1\right\}^{n\times l},\ \sum_{\alpha=1}^l U_{i\alpha}=1 \label{eq:const-U}
\end{eqnarray}
where $W$ is sparse with entries $\left\{w_{ij}\right\}$, $D\in\mathbb{R}^{n\times l}$ s.t. $D_{i,\alpha}\deff \psi_i(\alpha)$, and $V\in\mathbb{R}^{l\times l}$ s.t. $V_{\alpha,\beta}\deff \psi\left(\alpha,\beta\right)$, $\alpha,\beta\in\left\{1,\ldots,l\right\}$.
A detailed derivation of~(\ref{eq:EngU}) can be found in Sec.~\ref{sec:ms-deriv-U}.

An energy over $n$ variables with $l$ labels is now parameterized by $\left(n, l , D, W, V\right)$.

We first describe the energy pyramid construction for {\em a given} interpolation matrix $P$,
and defer the detailed description of our novel interpolation to Sec.~\ref{sec:matrix-P}.

\subsubsection*{Energy coarsening by variables}

Let $\left(n^f, l, D^f, W^f, V\right)$ be the fine scale energy.
We wish to generate a coarser representation $\left(n^c, l, D^c, W^c, V\right)$ with $n^c<n^f$.
This representation approximates $E\left(U^f\right)$ using fewer {\em variables}: $U^c$ with only $n^c$ rows.

Given an interpolation matrix  $P\in\left[0,1\right]^{{n^f}\times{n^c}}$ s.t. $\sum_jP_{ij}=1$ $\forall i$, it maps coarse to fine assignments through:
\begin{eqnarray}
U^f & \approx & PU^c \label{eq:interp}
\end{eqnarray}

For any fine assignment that can be approximated by a coarse assignment $U^c$
we may plug (\ref{eq:interp}) into~(\ref{eq:EngU}) yielding:
\begin{eqnarray}
E\left(U^f\right) & = & Tr\left(D^f{U^f}^T+W^fU^fV{U^f}^T\right) \nonumber \\
& \approx & Tr\left(D^f{U^c}^TP^T+W^fPU^cV{U^c}^TP^T\right) \nonumber \\
& = & Tr\Big(\underbrace{\left(P^TD^f\right)}_{\mbox{\normalsize $ \deff D^c$ }}{U^c}^T + \underbrace{\left(P^TW^fP\right)}_{\mbox{\normalsize $\deff W^c$}}U^cV{U^c}^T\Big) \nonumber \\
& = & Tr\left(D^c{U^c}^T+W^cU^cV{U^c}^T\right) \nonumber \\
& = & E\left(U^c\right)  \label{eq:EngC}
\end{eqnarray}
We have generated a coarse energy $E\left(U^c\right)$
parameterized by $\left(n^c, l, D^c, W^c, V\right)$ that approximates the fine energy $E(U^f)$.
This coarse energy is {\em of the same form} as the original energy allowing us to apply the coarsening procedure recursively to construct an energy pyramid.

\subsubsection*{Energy coarsening by labels}
So far we have explored the reduction of the number of degrees of freedom by reducing the number of {\em variables}.
However, we may just as well look at the problem from a different perspective: reducing the search space by decreasing the number of {\em labels} from $l_f$ to $l_c$ ($l_c<l_f$).
It is a well known fact that optimization algorithms (especially large move making, e.g., \cite{Boykov2001}) suffer from significant degradation in performance as the number of {\em labels} increases (\cite{Bleyer2010}).
Here we propose a novel principled and general framework for reducing the number of labels at each scale.

Let $\left(n, l^f, D^{\hat{f}}, W, V^{\hat{f}}\right)$ be the fine scale energy.
Looking at a  different interpolation matrix $\hat{P}\in\left[0,1\right]^{\mbox{$l^f\times l^c$}}$,
we may interpolate a coarse solution by $U^{\hat{f}} \approx U^{\hat{c}}\hat{P}^T$.
This time the interpolation matrix $\hat{P}$ acts on the {\em labels}, i.e., the {\em columns} of $U$.
The coarse labeling matrix $U^{\hat{c}}$ has the same number of rows (variables), but fewer columns (labels).
We use $\hat{\Box}$
% the ``hat"
notation to emphasize that the coarsening here affects the labels rather than the variables.

Coarsening the labels yields:
\begin{equation}
E\left(U^{\hat{c}}\right) = Tr\left( \left(D^{\hat{f}}\hat{P}\right)\mbox{$U^{\hat{c}}$}^T + WU^{\hat{c}} \left(\hat{P}^TV^{\hat{f}}\hat{P}\right)\mbox{$U^{\hat{c}}$}^T\right)
\label{eq:EngC-V}
\end{equation}
Again, we end up with the same type of energy, but this time it is defined over a smaller number of discrete labels:
$\left(n, l^c, D^{\hat{c}}, W, V^{\hat{c}}\right)$,
where $D^{\hat{c}} \deff D^{\hat{f}}\hat{P}$ and $V^{\hat{c}} \deff \hat{P}^T V^{\hat{f}} \hat{P}$.

\

The main theoretical contribution of this work is encapsulated in the
multiscale ``trick" of equations~(\ref{eq:EngC}) and~(\ref{eq:EngC-V}).
This formulation forms the basis of our unified framework allowing us to coarsen the energy {\em directly} and exploits its multiscale landscape for efficient exploration of the solution space.
This scheme moves the multiscale completely to the optimization side and makes it independent of any specific application.
We can practically now approach a wide and diverse family of energies using {\em the same} multiscale implementation.

The effectiveness of the multiscale approximation of~(\ref{eq:EngC}) and~(\ref{eq:EngC-V}) heavily depends on the interpolation matrix $P$ ($\hat{P}$ resp.).
Poorly constructed interpolation matrices will fail to expose the multiscale landscape of the functional.
In the subsequent section we
describe our principled energy-aware method for computing it.

%------------------------------------------------------%

\section{Energy-aware Interpolation}
\label{sec:matrix-P}

\begin{figure}
\centering
\includegraphics[width=.4\linewidth]{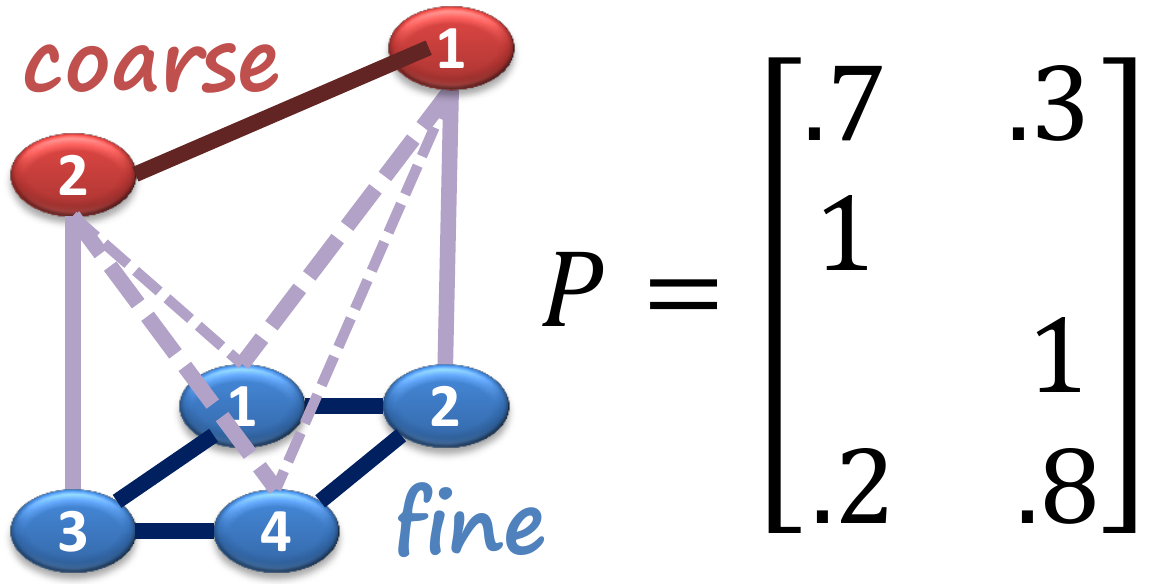}
\caption{
{\bf Interpolation as soft variable aggregation:}
{\em {\color{fine}fine} variable {\color{fine}1}, {\color{fine}2} and {\color{fine}4} are aggregated into {\color{coarse}coarse} variable {\color{coarse}1}, while {\color{fine}fine} variables {\color{fine}1},{\color{fine}3} and {\color{fine}4} are aggregated into {\color{coarse}coarse} variable {\color{coarse}2}. Soft aggregation allows for {\color{fine}fine} variables to be influenced by few {\color{coarse}coarse} variables,
e.g.: {\color{fine}fine} variable {\color{fine}1} is a convex combination of $.7$ of {\color{coarse}1} and $.3$ of {\color{coarse}2}.
Hard aggregation is a special case where $P$ is a binary matrix.
In that case each fine variable is influenced by exactly one coarse variable.}
}
\label{fig:multiscale}
\end{figure}

In this section we use terms and notations for variable coarsening ($P$),
however the motivation and methods are applicable for label coarsening ($\hat{P}$) as well due to the similar algebraic structure of~(\ref{eq:EngC}) and~(\ref{eq:EngC-V}).
% the interchangeability of $W$ and $V$ in~(\ref{eq:EngU}).

Our energy pyramid approximates the original energy using a decreasing number of degrees of freedom,
thus excluding some solutions from the original search space at coarser scales.
Which solutions are excluded is determined by the interpolation matrix $P$.
{\bf A desired interpolation does not exclude low energy assignments at coarse levels}.

The matrix $P$ can be interpreted as an operator that aggregates fine-scale variables into coarse ones (Fig.~\ref{fig:multiscale}).
Aggregating fine variables $i$ and $j$ into a coarser one excludes from the search space all assignments for which $l_i\ne l_j$.
This aggregation is undesired if assigning $i$ and $j$ to different labels yields low energy.
However, when variables $i$ and $j$ are {\em strongly correlated} by the energy (i.e., assignments with $l_i=l_j$ yield low energy),
aggregating them together efficiently allows exploration of low energy assignments.
{\bf A desired interpolation aggregates $i$ and $j$ when $i$ and $j$ are strongly correlated by the energy}.

%------------------------------------------------------%

\subsection{Measuring energy-aware correlations}
\label{sec:local-correlations}
We provide two correlations measures, one used in computing variable coarsening ($P$) and the other used for label coarsening ($\hat{P}$).

%------------------------------------------------------%
\noindent{\bf Energy-aware correlations between variables:}
A reliable estimation for the correlations between the variables allows us to construct a desirable $P$ that aggregates strongly correlated variables.
A na\"{\i}ve approach would assume that neighboring variables are correlated (this assumption underlies \cite{Felzenszwalb2006}). This assumption clearly does not hold in general and may lead to an undesired interpolation matrix $P$.
\cite{Kim2011} proposed several ``closed form formulas" for energy-aware variable grouping.
However, their formulas take into account either the unary term or the pair-wise term.
Indeed it is difficult to decide which term dominates and how to fuse these two terms together.
Therefore, there is no ``closed form" method that successfully integrates both of them.

As opposed to these ``closed form" methods, we propose a novel empirical scheme for correlation estimation.
Empirical estimation of the correlations naturally accounts for and integrates the influence of both the unary and the pair-wise terms.
Moreover, our method, inspired by \cite{Ron2011,Livne2011}, extends to all energies (\ref{eq:EngU}): submodular, non-submodular, metric $V$, arbitrary $V$, arbitrary $W$, energies defined over regular grids and arbitrary graphs.

Variables $i$ and $j$ are correlated by the energy when $l_i=l_j$ yields relatively low energy value.
To estimate these correlations we empirically generate several ``locally" low energy assignments,
and measure the label agreement between neighboring variables $i$ and $j$.
We use Iterated Conditional Modes (ICM) of \cite{Besag1986} to obtain locally low energy assignments:
Starting with a random assignment, ICM chooses, at each iteration and for each variable, the label
yielding the largest decrease of the energy function, conditioned on the labels assigned to its neighbors.

Performing $t=10$ ICM iterations for $K=10$ random initializations provides $K$ locally low energy assignments $\left\{L^k\right\}_{k=1}^K$.
Our empirical dissimilarity between $i$ and $j$ is given by $d_{ij}=\frac{1}{K}\sum_k V_{l^k_i,l^k_j}$,
and their correlation is given by $c_{ij}=\exp\left(-\frac{d_{ij}}{\sigma}\right)$,
with $\sigma \propto \max V$.

It is interesting to note that strong correlation between variables $i$ and $j$ usually implies that the pair-wise term binding them together ($\phiij$) is a smoothness-preserving type of relation.
We assume that even for challenging energies with many contrast-enhancing pair-wise terms, there are still significant amount of smoothness-preserving terms to allow for effective coarsening.

%Then we perform generating several relatively low energy assignments.
%
%What happens to a solution $L$ if we start with a random assignment and employ a few Iterated Conditional Modes (ICM) \cite{Besag1986} iterations?
%
%Each ICM
%These ICM iterations ...
%
%This will definitely not be a good approximation for the global optimum.
%However, since ICM tracks the local behavior of the solution space, the resulting solution $L$ provides a good indication to the {\em local} correlations between neighboring variables.
%
%Specifically, our empirical dissimilarity between $i$ and $j$ is given by $d_{ij}=V_{l_i,l_j}$.
%To get a reliable estimation we average $d_{ij}$ over $K$ random initializations.
%Finally, the correlation between $i$ and $j$ is given by $c_{ij}=\exp\left(-\frac{d_{ij}}{\sigma}\right)$,
%with $\sigma \propto \max V$.
%Throughout our experiments we perform $10$ ICM iterations, with $K=10$ random initializations.

%------------------------------------------------------%

\noindent{\bf Energy-aware correlations between labels:}
Correlations between labels are easier to estimate, since this information is explicit in the matrix $V$ that encodes the ``cost" (i.e., dissimilarity) between two labels.
Setting
$\hat{c}_{\alpha,\beta}\propto \left(\hat{V}_{\alpha,\beta}\right)^{-1}$,
we get a ``closed-form" expression for the correlations between labels.

%------------------------------------------------------%

\subsection{From correlations to interpolation}
\label{sec:amg-p}

Using our measure for the variable correlations, $c_{ij}$, we follow the Algebraic Multigrid (AMG) method of \cite{Brandt1986} to compute an interpolation matrix $P$ that softly aggregates strongly correlated variables.
% The computed $P$ uses soft aggregation to allow for more flexibility in the coarse approximation.

We begin by selecting a set of coarse representative variables $\VV^c\subset \VV^f$,
such that every variable in $\VV^f \backslash \VV^c$ is strongly correlated with $\VV^c$.
That is, every variable in $\VV^f$ is either in $\VV^c$ or is {\em strongly correlated} to other variables in $\VV^c$.
A variable $i$ is considered strongly correlated to $\VV^c$ if $\sum_{j\in\VV^c}c_{ij} \ge \beta \sum_{j\in\VV^f} c_{ij}$.
$\beta$ affects the coarsening rate, i.e., the ratio $n^c/n^f$,
smaller $\beta$ results in a lower ratio.
We perform this selection greedily and sequentially, starting with $\VV^c=\emptyset$ adding $i$ to $\VV^c$ if it is not yet strongly correlated to $\VV^c$.

Given the selected coarse variables $\VV^c$,
$I(j)$ maps indices of variables from fine to coarse:
$I(j)$ is the coarse index of the variable whose fine index is $j$ (in Fig.~\ref{fig:multiscale}: $I(2)=1$ and $I(3)=2$).
The interpolation matrix $P$ is defined by:
\begin{equation}
P_{iI(j)} = \left\{
\begin{array}{cl}
c_{ij}                & i\in\VV^f\backslash\VV^c,\ j\in\VV^c\\
1                      & i\in\VV^c, j=i\\
0                      & \mbox{otherwise}\\
\end{array}
\right. \label{eq:entries-of-P}
\end{equation}
We further prune rows of $P$ leaving only $\delta$ maximal entries.
Each row is then normalized to sum to 1.
Throughout our experiments we use $\beta=0.2$ ($\hat{\beta}=0.75$), $\delta=3$ ($\hat{\delta}=2$) for computing $P$ ($\hat{P}$ resp.).

%------------------------------------------------------------------------------------%
\section{Unified Discrete Multiscale Framework}
\label{sec:pipeline}

%------------------------------------------------------------------------------------%
\begin{algorithm}[t]
\caption{Discrete multiscale optimization. \label{alg:multiscale}}
\DontPrintSemicolon
%\LinesNumbered
\SetKw{KwInit}{Init}
\SetKw{KwOpt}{Refine}
\SetKw{KwCoarse}{Coarsen}
\KwIn{Energy $\left(\VV^0, D^0, W^0, V\right)$.}
\KwOut{$U^0$}
\KwInit{$s\leftarrow 0$}\tcp{fine scale}
\tcp{Energy pyramid construction:}
\While{$\abs{\VV^s} \ge 10$} {
    Estimate pair-wise correlations $c_{ij}$ at scale $s$ (Sec.~\ref{sec:local-correlations}).\;
    Compute interpolation matrix $P^s$ (Sec.~\ref{sec:amg-p}).\;
    Derive coarse energy $\left(\VV^{s+1}, D^{s+1}, W^{s+1}, V\right)$ (Eq.~\ref{eq:EngC}).\;
    $s++$\;
}
\tcp{Coarse-to-fine optimization:}
\While{$s\ge0$} {
    $U^s\leftarrow$ \KwOpt{$(\tilde{U}^s)$}\;
    $\tilde{U}^{s-1} = P^sU^s$\tcp{interpolate a solution}\label{line:refine}
    $s--$\;
}
where \KwOpt{$(\tilde{U}^s)$} uses an ``off-the-shelf" algorithm to optimize the energy $\left(\VV^{s}, D^{s}, W^{s}, V\right)$ with $\tilde{U}^s$ as an initialization.\;
\end{algorithm}
%------------------------------------------------------------------------------------%

So far we have described the different components of our multiscale framework.
Alg.~\ref{alg:multiscale} puts them together into a multiscale minimization scheme.
Given an energy $\left(\VV, D, W, V\right)$,
our framework first works fine-to-coarse to compute interpolation matrices $\left\{P^s\right\}$ that generates an ``energy pyramid".
Typically we end up at the coarsest scale with less than $10$ variables.
As a result, exploring the energy at this scale is robust to the initial assignment of the single-scale method used\footnote{In practice we use ``winner-take-all" initialization as suggested by \cite[\S3.1]{Szeliski2008}.}.
%At finer scales the interpolated coarse assignments allows the single-scale methods to avoid getting stuck at local minima, resulting with low energy assignments.
%
%In a coarse-to-fine manner we explore the solution space:

Starting from the coarsest scale, a coarse solution at scale $s$ is interpolated to a finer scale $s-1$.
At the finer scale it serves as a good initialization for an ``off-the-shelf" single-scale optimization that refines this interpolated solution.
%The interpolated coarse assignments allows the single-scale methods to avoid getting stuck at local minima, resulting with low energy assignments.
These two steps are repeated for all scales from coarse to fine.
%
%
%
%
%
%
%Then, in a coarse-to-fine manner we explore the solution space, using an ``off-the-shelf" single-scale optimization algorithm at each scale.
%The gist of multiscale methods is the ``crosstalk" between the scales: coarse solutions are interpolated to efficiently guide the exploration at finer scales.
%Typically at the coarsest scale we have less than $10$ variables.
%As a result, exploring the energy at this scale is robust to the initial assignment of the single-scale method used\footnote{In practice we use ``winner-take-all" initialization as suggested by \cite[\S3.1]{Szeliski2008}.}.
%At finer scales the interpolated coarse assignments allows the single-scale methods to avoid getting stuck at local minima, resulting with low energy assignments.

The interpolated solution $\tilde{U}^{s-1}$, at each scale,
might not satisfy the binary constraints~(\ref{eq:const-U}).
We round each row of $\tilde{U}^{s-1}$ by setting the maximal element to $1$ and the rest to $0$.

The most computationally intensive modules
%Computationally, the most dominating modules
of our framework are the empirical estimation of the variable correlations and the single-scale optimization used to refine the interpolated solutions.
The complexity of the correlation estimation is $O\left(\abs{\EE}\cdot l\right)$, where $\abs{\EE}$ is the number of non-zero elements in $W$ and $l$ is the number of labels.
However, it is fairly straightforward to parallelize this module.

It is now easy to see how our framework generalizes \cite{Felzenszwalb2006}, \cite{Komodakis2010} and \cite{Kim2011}:
They are restricted to hard aggregation in $P$.
\cite{Felzenszwalb2006} and \cite{Komodakis2010} use a multiscale pyramid, however their variable aggregation is not energy-aware, and is restricted to diadic pyramids.
On the other hand, \cite{Kim2011} have limited energy-aware aggregation, applied to a two level only ``pyramid".
They only optimize at the coarse scale and cannot refine the solution on the fine scale.

%------------------------------------------------------------------------------------%
\section{Experimental Results}
\label{sec:ms-results}
Our experiments has two main goals: first, to stress the difficulty of approximating non-submodular energies and to show the advantages of primal methods for this type of minimization problems.
The other goal is to demonstrate how our unified multiscale framework improved the performance of existing single-scale primal methods.

We evaluated our multiscale framework on a diversity of discrete optimization tasks\footnote{code available at \url{www.wisdom.weizmann.ac.il/~bagon/matlab.html}.}: ranging from challenging non-submodular synthetic and co-clustering energies to low-level submodular vision energies such as denoising and stereo.
In addition we provide a comparison between the different methods for measuring variable correlations that were presented in Sec.~\ref{sec:local-correlations}. We conclude with a label coarsening experiment.
In all of these experiments we minimize a {\em given} publicly available benchmark energy,
{\em we do not attempt to improve on the energy formulation itself}.

We use ICM (\cite{Besag1986}), $\alpha\beta$-swap and $\alpha$-expansion (large move making algorithms of \cite{Boykov2001}) as representative single-scale ``off-the-shelf" primal optimization algorithms.
To help large move making algorithms to overcome the non-submodularity of some of these energies we augment them with QPBO(I) of \cite{Rother2007}.

We follow the protocol of \cite{Szeliski2008} that uses the {\em lower bound} of TRW-S (\cite{Kolmogorov2006}) as a baseline for comparing the performance of different optimization methods for different energies.
We report how close the results (in percents) to the lower bound: {\bf closer to $100\%$ is better}.

We show a remarkable improvement for ICM combined in our multiscale framework compared with a single-scale scheme.
For the large move making algorithms there is a smaller but consistent improvement of the multiscale over a single scale scheme.
TRW-S is a dual method and is considered state-of-the-art for discrete energy minimization \cite{Szeliski2008}.
However, we show that when it comes to non-submodular energies it struggles behind the large move making algorithms and even ICM.
For these challenging energies, multiscale gives a significant boost in optimization performance.

%------------------------------------------------------------------------------------%
\begin{table}
\centering
\setlength{\tabcolsep}{1mm}
\begin{tabular}{c||c|c||c|c||c|c||c}
 \multirow{3}{*}{$\lambda$} & \multicolumn{2}{c||}{ICM} & \multicolumn{2}{c||}{Swap} & \multicolumn{2}{c||}{Expand} & \multirow{2}{*}{TRW-S} \\
 & \multirow{2}{*}{{\color{ours}Ours}}  & single  &  \multirow{2}{*}{{\color{ours}Ours}} &single & \multirow{2}{*}{{\color{ours}Ours}}  & single & \\
 & & scale & & scale & & scale & \\\hline \hline
$5$ & {\color{ours}$112.6\%$} & $115.9\%$ & {\color{ours}$108.9\%$} & $110.0\%$ & {\color{ours}$110.5\%$} & $110.0\%$ & $116.6\%$ \\
$10$ & {\color{ours}$123.6\%$} & $130.2\%$ & {\color{ours}$118.5\%$} & $120.2\%$ & {\color{ours}$121.5\%$} & $121.0\%$ & $134.6\%$ \\
$15$ & {\color{ours}$127.1\%$} & $135.8\%$ & {\color{ours}$122.1\%$} & $124.1\%$ & {\color{ours}$124.6\%$} & $125.1\%$ & $138.3\%$ \\
\end{tabular}
\caption{ {\bf Synthetic results (energy):}
{\em
Showing percent of achieved energy value relative to the lower bound (closer to $100\%$ is better) for ICM, $\alpha\beta$-swap, $\alpha$-expansion and TRW-S
for varying strengths of the pair-wise term ($\lambda=5,\ldots,15$, stronger $\rightarrow$ harder to optimize.)}
}
\setlength{\tabcolsep}{\origtabcolsep}
\label{tab:res-synthetic}
\end{table}

\subsection{Synthetic}
We begin with synthetic {\em non-submodular} energies defined over a 4-connected grid graph of size $50\times50$ ($n=2500$), and $l=5$ labels.
The unary term $D \sim \mathcal{N}\left(0,1\right)$.
The pair-wise term $V_{\alpha\beta}=V_{\beta\alpha} \sim \mathcal{U}\left(0, 1\right)$ ($V_{\alpha\alpha}=0$) and $w_{ij}=w_{ji} \sim \lambda \cdot \mathcal{U}\left(-1,1\right)$.
The parameter $\lambda$ controls the relative strength of the pair-wise term,
stronger (i.e., larger $\lambda$) results with energies more difficult to optimize (see \cite{Kolmogorov2006}).
Table~\ref{tab:res-synthetic} shows results, averaged over 100 experiments.

The resulting synthetic energies are non-submodular (since $w_{ij}$ may become negative).
For these challenging energies, state-of-the-art dual method (TRW-S) performs rather poorly\footnote{We did not restrict the number of iterations, and let TRW-S run until no further improvement to the lower bound is made.} (worse than single scale ICM) and there is a significant gap between the lower bound and the energy of the actual primal solution provided.
Among the primal methods used,
These results motivate our focusing on primal methods, especially $\alpha\beta$-swap.

%------------------------------------------------------------------------------------%
\begin{figure}
\centering
\newlength{\cipwidth}
\setlength{\cipwidth}{.11\linewidth}
\setlength{\tabcolsep}{0.5mm}
\begin{tabular}{c|c||c|c|c|c||c|c}
\multirow{3}{*}{GT} & \multirow{3}{*}{Input} & \multicolumn{2}{c|}{ICM} & \multicolumn{2}{c||}{QPBO} &\multirow{3}{*}{TRW-S} & Sim. \\
& & \multirow{2}{*}{{\color{ours} Ours}} & single & \multirow{2}{*}{{\color{ours} Ours}} & single &  & Ann.\\
& &                                      & scale  &                                      & scale  & & \\
\includegraphics[width=\cipwidth]{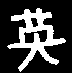}&
\includegraphics[width=\cipwidth]{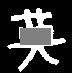}&
\includegraphics[width=\cipwidth]{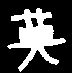}&
\includegraphics[width=\cipwidth]{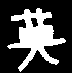}&
\includegraphics[width=\cipwidth]{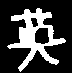}&
\includegraphics[width=\cipwidth]{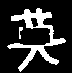}&
\includegraphics[width=\cipwidth]{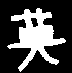}&
\includegraphics[width=\cipwidth]{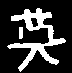}\\
\includegraphics[width=\cipwidth]{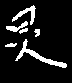}&
\includegraphics[width=\cipwidth]{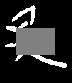}&
\includegraphics[width=\cipwidth]{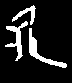}&
\includegraphics[width=\cipwidth]{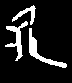}&
\includegraphics[width=\cipwidth]{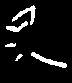}&
\includegraphics[width=\cipwidth]{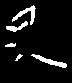}&
\includegraphics[width=\cipwidth]{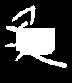}&
\includegraphics[width=\cipwidth]{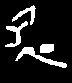}\\
\includegraphics[width=\cipwidth]{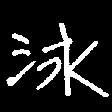}&
\includegraphics[width=\cipwidth]{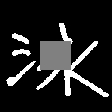}&
\includegraphics[width=\cipwidth]{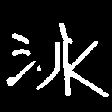}&
\includegraphics[width=\cipwidth]{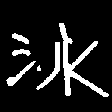}&
\includegraphics[width=\cipwidth]{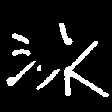}&
\includegraphics[width=\cipwidth]{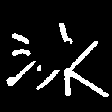}&
\includegraphics[width=\cipwidth]{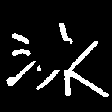}&
\includegraphics[width=\cipwidth]{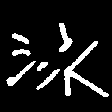}\\
\includegraphics[width=\cipwidth]{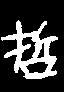}&
\includegraphics[width=\cipwidth]{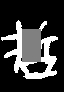}&
\includegraphics[width=\cipwidth]{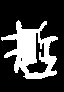}&
\includegraphics[width=\cipwidth]{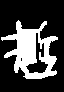}&
\includegraphics[width=\cipwidth]{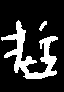}&
\includegraphics[width=\cipwidth]{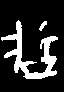}&
\includegraphics[width=\cipwidth]{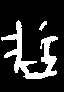}&
\includegraphics[width=\cipwidth]{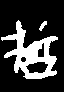}\\
\includegraphics[width=\cipwidth]{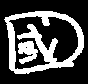}&
\includegraphics[width=\cipwidth]{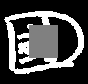}&
\includegraphics[width=\cipwidth]{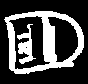}&
\includegraphics[width=\cipwidth]{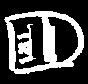}&
\includegraphics[width=\cipwidth]{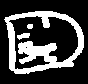}&
\includegraphics[width=\cipwidth]{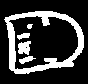}&
\includegraphics[width=\cipwidth]{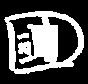}&
\includegraphics[width=\cipwidth]{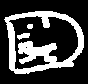}\\
\includegraphics[width=\cipwidth]{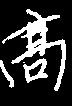}&
\includegraphics[width=\cipwidth]{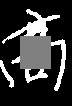}&
\includegraphics[width=\cipwidth]{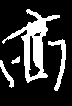}&
\includegraphics[width=\cipwidth]{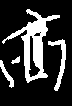}&
\includegraphics[width=\cipwidth]{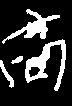}&
\includegraphics[width=\cipwidth]{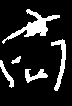}&
\includegraphics[width=\cipwidth]{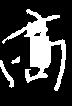}&
\includegraphics[width=\cipwidth]{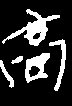}
\end{tabular}
\setlength{\tabcolsep}{\origtabcolsep}
\caption{{\bf Chinese characters inpainting:}
{\em Visualizing some of the instances used in our experiments.
Columns are (left to right):
The original character used for testing.
The input partially occluded character.
ICM and QPBO results both our multiscale and single scale results.
Results of TRW-S and results of \cite{Nowozin2011} obtained with a  very long run of simulated annealing (using Gibbs sampling inside the annealing).}}
\label{fig:sebastian-dtf}
\end{figure}

\begin{figure}
\centering
\comment{
\begin{tabular}{cc}
\includegraphics[width=.45\linewidth]{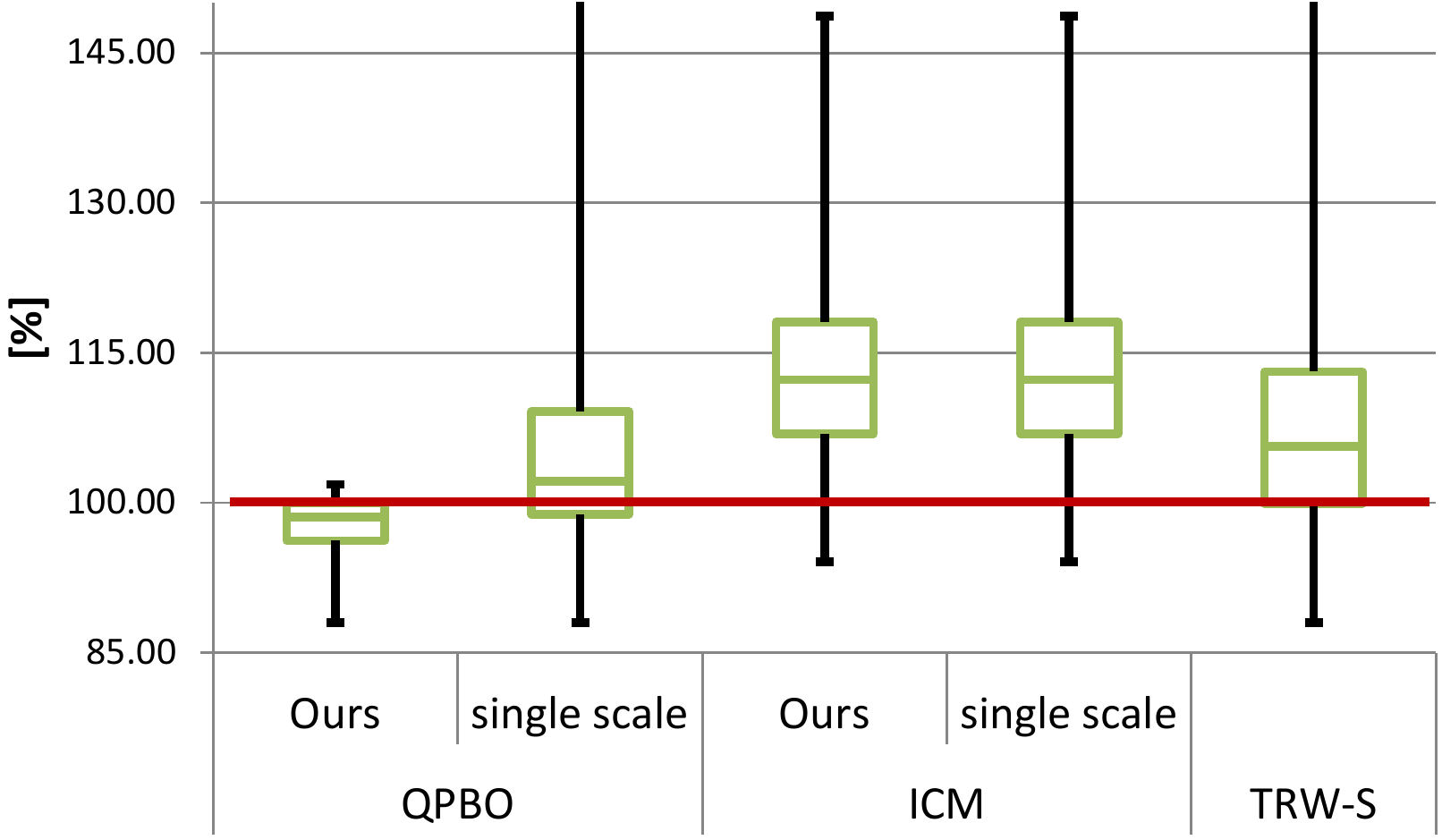}&
\includegraphics[width=.45\linewidth]{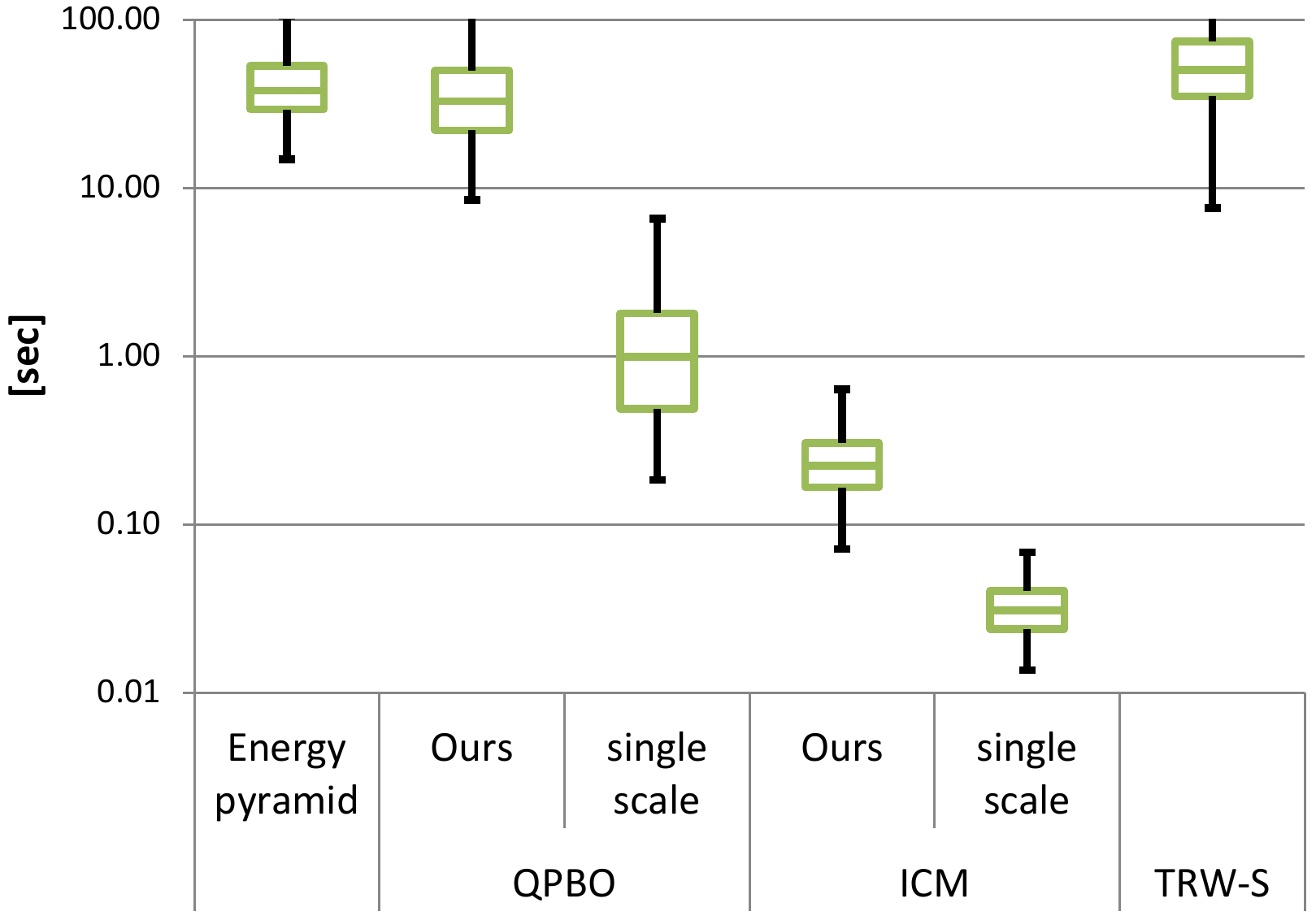}\\
(a) & (b)
\end{tabular}
}
\includegraphics[width=.6\linewidth]{ms/binary_dtf_energy_cropped.pdf}
\caption{ {\bf Energies of Chinese characters inpainting:}
{\em Box plot showing 25\%, median and 75\% of the resulting energies relative to reference energies of \cite{Nowozin2011} (lower than $100\%$ = lower than baseline).
Our multiscale approach combined with QPBO achieves consistently better energies than baseline, with very low variance.
TRW-S improves on only 25\% of the instances with very high variance in the results.}
}
\label{fig:sebastian-dtf-graphs}
\end{figure}

\begin{table}
\centering
\begin{tabular}{c||c|c||c|c||c}
 & \multicolumn{2}{c||}{ICM} & \multicolumn{2}{c||}{QPBO} & \multirow{2}{*}{TRW-S} \\
 & {\color{ours}Ours} & single-scale & {\color{ours}Ours} & single-scale &  \\\hline
(a) &  {\color{ours}$114.0\%$} & $114.0\%$ & {\color{ours}$97.8\%$} & $106.2\%$ & $108.6\%$ \\\hline
(b) & {\color{ours}$7.0\%$} & $7.0\%$ &{\color{ours}$77.0\%$} & $34.0\%$ & $25.0\%$ \\\hline
\end{tabular}
\caption{{\bf Energies of Chinese characters inpainting:}
{\em table showing
(a) mean energies for the inpainting experiment relative to baseline of \cite{Nowozin2011} (lower is better, less than $100\%$ = lower than baseline).
(b) percent of instances for which strictly lower energy was achieved.
}}\label{tab:dtf-binary}
\end{table}

\subsection{Chinese character inpainting}

We further experiment with learned binary energies of \cite[\S5.2]{Nowozin2011}\footnote{available at \url{www.nowozin.net/sebastian/papers/DTF_CIP_instances.zip}.}.
These 100 instances of non-submodular pair-wise energies are defined over a 64-connected grid.
These energies were designed and trained to perform the task of learning Chinese calligraphy, represented as a complex, non-local binary pattern.
Despite the very large number of parameters involved in representing such complex energies, learning is conducted very efficiently using Decision Tree Field (DTF).
The main challenge in these models becomes the inference at test time.

Our experiments show how approaching these challenging energies using our unified multiscale framework allows for better approximations.
Table~\ref{tab:dtf-binary} and Fig.~\ref{fig:sebastian-dtf} compare our multiscale framework to single-scale methods acting on the primal binary variables.
Since the energies are binary, we use QPBO instead of large move making algorithms.
We also provide an evaluation of a dual method (TRW-S) on these energies.
In addition to the quantitative results, Fig.~\ref{fig:sebastian-dtf-graphs} provides a visualization of some of the instances of the restored Chinese characters.

These ``real world" energies highlight the advantage primal methods has over dual ones when it comes to challenging non-submodular energies.
It is further clear that significant improvement is made by our multiscale framework.

%------------------------------------------------------------------------------------%
\begin{table}
\centering
\setlength{\tabcolsep}{.5mm}
\begin{tabular}{c||c|c||c|c||c|c||c}
 & \multicolumn{2}{c||}{ICM} & \multicolumn{2}{c||}{Swap} & \multicolumn{2}{c||}{Expand} & TRW-S\\
               & \multirow{2}{*}{{\color{ours}Ours}}     & single  & \multirow{2}{*}{{\color{ours}Ours}} & single &  \multirow{2}{*}{{\color{ours}Ours}} & single & \\
               &                          & scale &          & scale &   & scale &  \\
\hline \hline
(a) & {\color{ours}$99.9\%$} & $177.7\%$ & {\color{ours}$99.8\%$} & $101.5\%$ & {\color{ours}$99.8\%$} & $101.6\%$ & $176.2\%$ \\\hline
(b) & {\color{ours}$55.6\%$} & $0.0\%$ & {\color{ours}$71.8\%$} & $15.5\%$ & {\color{ours}$70.8\%$} & $11.6\%$ & $0.5\%$ \\\hline
\comment{(c) & {\color{ours}$68.2$} & $0.8$ & {\color{ours}$1387.6$} & $11787.7$ & {\color{ours}$11676.4$} & $42084.5$ & $84.6$ \\\hline}
\end{tabular}
\caption{ {\bf Co-clustering results: }
{\em
Baseline for comparison are state-of-the-art results of \cite{Glasner2011}.
(a) We report our results as percent of the baseline: smaller is better, lower than $100\%$ even outperforms state-of-the-art.
(b) We also report the fraction of energies for which our multiscale framework outperform state-of-the-art.
\comment{(c) run times. pyramid construction  $230.3$ milisec.}
} }
\label{tab:cocluster-res}
\end{table}

\subsection{Co-clustering}

The problem of co-clustering addresses the matching of superpixels within and across frames in a video sequence.
Following \cite[\S6.2]{Bagon2012}, we treat co-clustering as a discrete minimization of {\em non-submodular} Potts energy.
We obtained 77 co-clustering energies, courtesy of \cite{Glasner2011}, used in their experiments.
The number of variables in each energy ranges from 87 to 788.
Their sparsity (percent of non-zero entries in $W$) ranges from $6\%$ to $50\%$,
The resulting energies are non-submodular, have no underlying regular grid, and are very challenging to optimize \cite{Bagon2012}.

Table~\ref{tab:cocluster-res} compares our discrete multiscale framework combined with ICM and $\alpha\beta$-swap.
For these energies we use a different baseline: the state-of-the-art results of \cite{Glasner2011} obtained by applying specially tailored convex relaxation method
(We do not use the lower bound of TRW-S here since it is far from being tight for these challenging energies).
% in this challenging case).
Our multiscale framework improves state-of-the-art for this family of challenging energies.

%------------------------------------------------------------------------------------%
\subsection{semi-metric energies}
We further applied our multiscale framework to optimize less challenging semi-metric energies.
We use the diverse low-level vision MRF energies from the Middlebury benchmark \cite{Szeliski2008}\footnote{Available at \url{vision.middlebury.edu/MRF/}.}.

\begin{figure}
\centering
\newlength{\stwidth}
\setlength{\stwidth}{.11\linewidth}
\begin{tabular}{cc||cc||cc||c}
\multicolumn{2}{c||}{ICM} & \multicolumn{2}{c||}{Swap} & \multicolumn{2}{c||}{Expand} & Ground \\
{\color{ours}Ours} & Single scale & {\color{ours}Ours} & Single scale & {\color{ours}Ours} & Single scale & truth \\ \hline
\includegraphics[width=\stwidth]{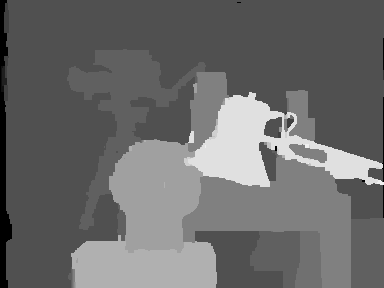}&
\includegraphics[width=\stwidth]{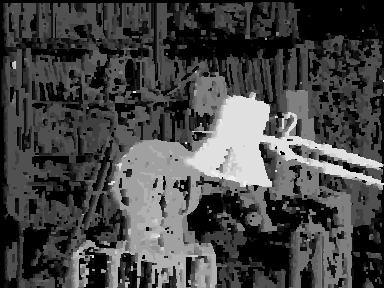}&
\includegraphics[width=\stwidth]{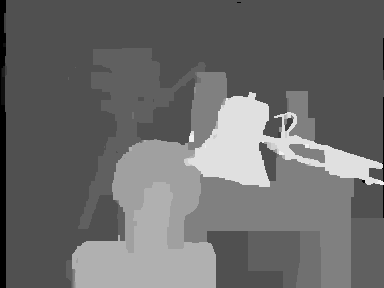}&
\includegraphics[width=\stwidth]{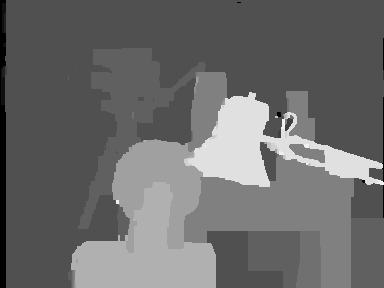}&
\includegraphics[width=\stwidth]{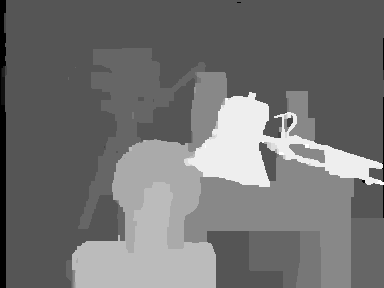}&
\includegraphics[width=\stwidth]{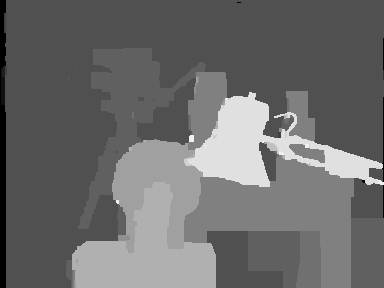}&
\includegraphics[width=\stwidth]{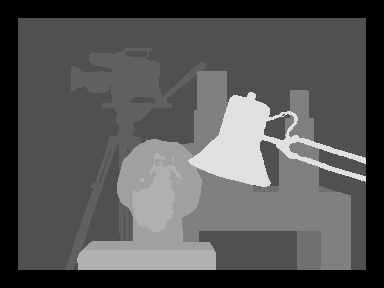}\\
\includegraphics[width=\stwidth]{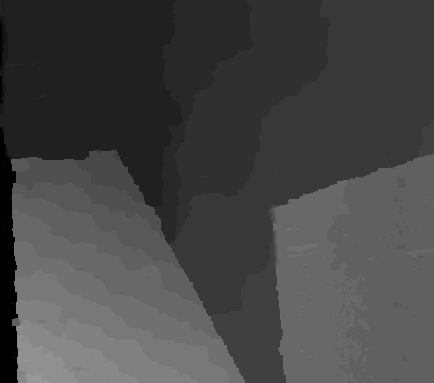}&
\includegraphics[width=\stwidth]{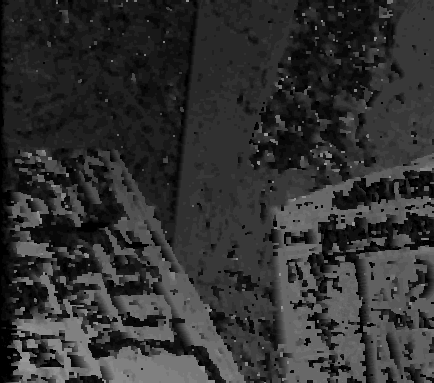}&
\includegraphics[width=\stwidth]{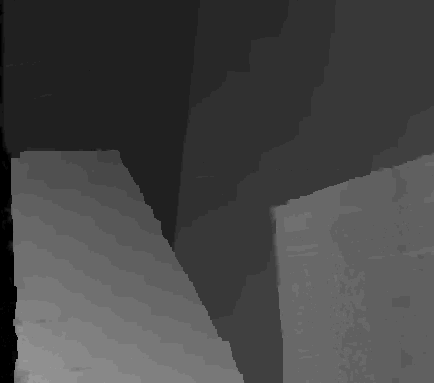}&
\includegraphics[width=\stwidth]{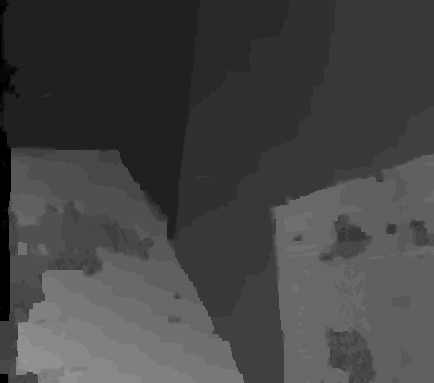}&
\includegraphics[width=\stwidth]{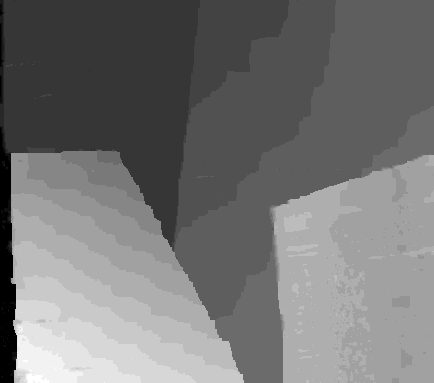}&
\includegraphics[width=\stwidth]{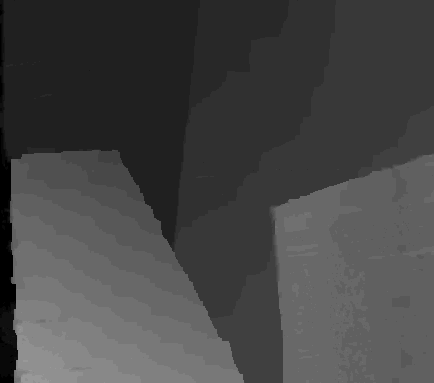}&
\includegraphics[width=\stwidth]{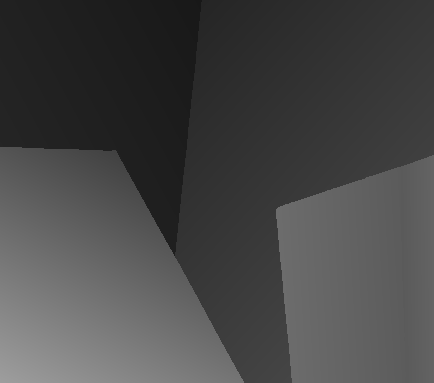}\\
\includegraphics[width=\stwidth]{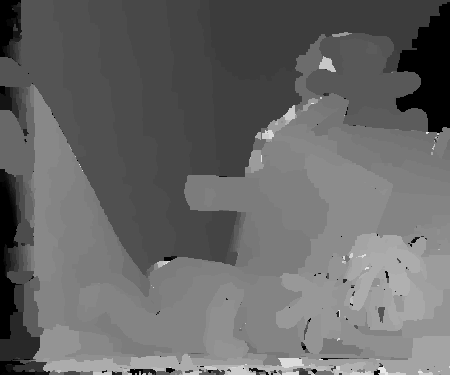}&
\includegraphics[width=\stwidth]{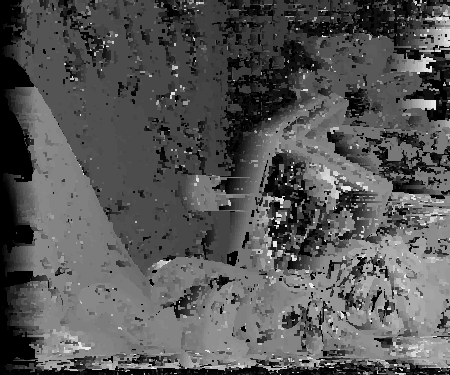}&
\includegraphics[width=\stwidth]{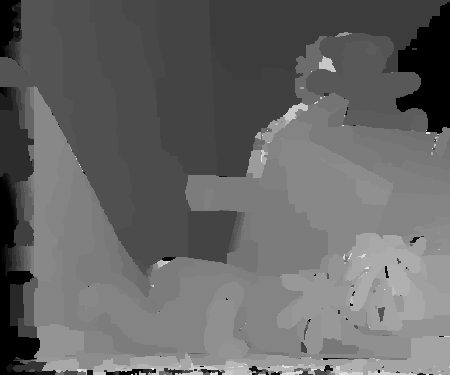}&
\includegraphics[width=\stwidth]{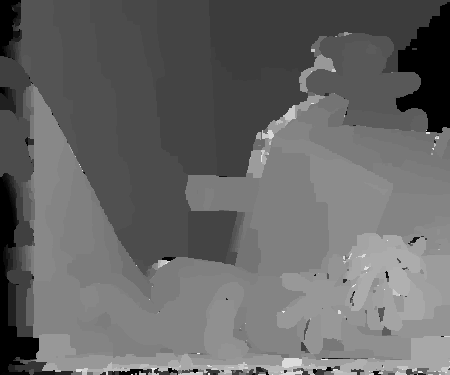}&
\includegraphics[width=\stwidth]{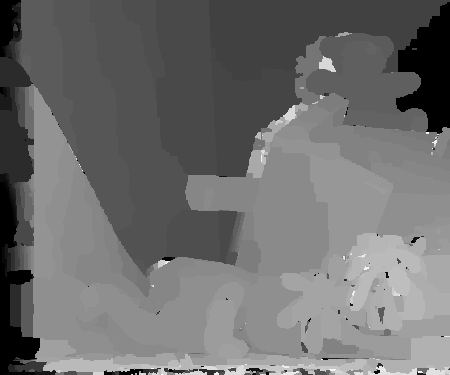}&
\includegraphics[width=\stwidth]{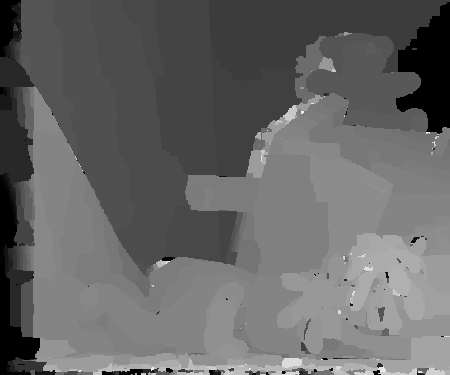}&
\includegraphics[width=\stwidth]{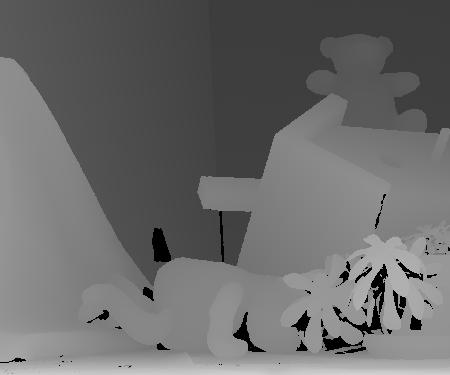}\\
\end{tabular}
\caption{ {\bf Stereo:}
{\em Note how our multiscale framework drastically improves ICM results.
visible improvement for $\alpha\beta$-swap can also be seen in the middle row (Venus).
Numerical results for these examples are shown in Table~\ref{tab:stereo-res}.
%Energies from \cite{Szeliski2008}.
}}
\label{fig:res-stereo}
\end{figure}

\begin{figure}
\centering
\newlength{\dnwidth}
\setlength{\dnwidth}{.11\linewidth}
\begin{tabular}{c||cc||cc||cc}
Input & \multicolumn{2}{c||}{ICM} & \multicolumn{2}{c}{Swap} & \multicolumn{2}{c}{Expand}\\
 & {\color{ours}Ours} & Single scale & {\color{ours}Ours} & Single scale & {\color{ours}Ours} & Single scale\\ \hline
\includegraphics[width=\dnwidth]{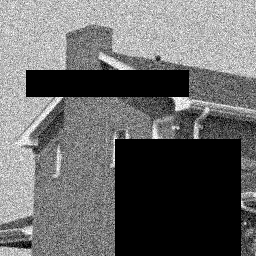}&
\includegraphics[width=\dnwidth]{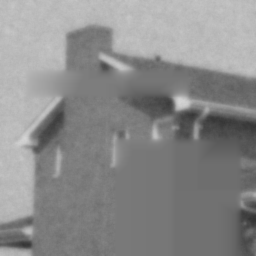}&
\includegraphics[width=\dnwidth]{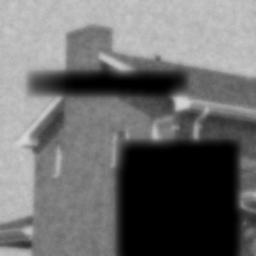}&
\includegraphics[width=\dnwidth]{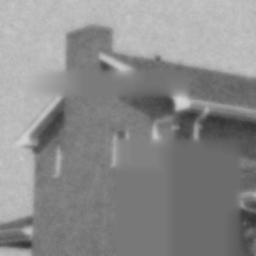}&
\includegraphics[width=\dnwidth]{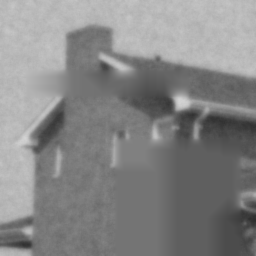}&
\includegraphics[width=\dnwidth]{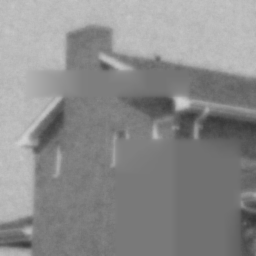}&
\includegraphics[width=\dnwidth]{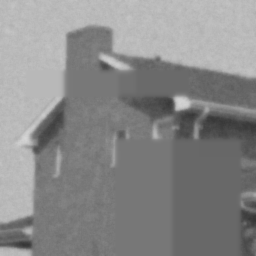}\\
\includegraphics[width=\dnwidth]{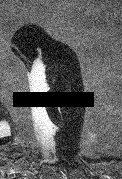}&
\includegraphics[width=\dnwidth]{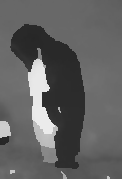}&
\includegraphics[width=\dnwidth]{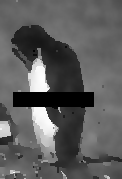}&
\includegraphics[width=\dnwidth]{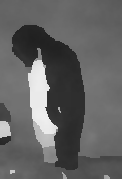}&
\includegraphics[width=\dnwidth]{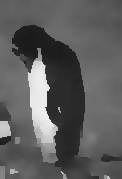}&
\includegraphics[width=\dnwidth]{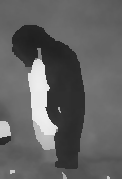}&
\includegraphics[width=\dnwidth]{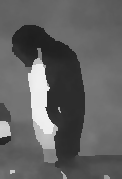}\\
\end{tabular}
\caption{ {\bf Denoising and inpainting:}
{\em
Single scale ICM is unable to cope with inpainting: performing local steps it is unable to propagate information far enough to fill the missing regions in the images.
On the other hand, our multiscale framework allows ICM to perform large steps at coarse scales and successfully fill the gaps.
Numerical results for these examples are shown in Table~\ref{tab:denoise-res}.
%Energies from \cite{Szeliski2008}.
}}
\label{fig:res-denoise}
\end{figure}

For these semi-metric energies, TRW-S (single scale) performs quite well and in fact, if enough iterations are allowed,
its lower bound converges to the global optimum.
As opposed to TRW-S, large move making and ICM do not always converge to the global optimum.
Yet, we are able to show a significant improvement for primal optimization algorithms when used within our multiscale framework.
Tables~\ref{tab:stereo-res} and~\ref{tab:denoise-res}
%  \ref{tab:object-seg} \ref{tab:stereo-res} \ref{tab:denoise-res} \ref{tab:segmentation-res} and
and
Figs.~\ref{fig:res-stereo} and~\ref{fig:res-denoise}
% \ref{fig:object-seg} \ref{fig:res-stereo} \ref{fig:res-denoise} \ref{fig:res-segmentation}
show our multiscale results for the different submodular energies.

One of the conclusions of the Middlebury challenge was that ICM is no longer a valid candidate for optimization.
Integrating ICM into our multiscale framework puts it back on the right track.

\comment{Table~\ref{tab:variable-runtime} exemplifies how our framework improved running times for two difficult energies (``Penguin" denoising and ``Venus" stereo).}

\begin{table}
\centering
\setlength{\tabcolsep}{.5mm}
\begin{tabular}{c||c|c||c|c||c|c}
 & \multicolumn{2}{c||}{ICM} & \multicolumn{2}{c||}{Swap}& \multicolumn{2}{c}{Expand}\\
               & {\color{ours}Ours}      & single scale & {\color{ours}Ours}   & single scale & {\color{ours}Ours}   & single scale\\
\hline \hline
Tsukuba & {\color{ours}$102.8\%$} &$653.4\%$ &{\color{ours}$100.2\%$} &$100.5\%$ &{\color{ours}$100.1\%$} &$100.3\%$ \\ \hline
Venus & {\color{ours}$112.3\%$} &$405.1\%$ &{\color{ours}$102.8\%$} &$128.7\%$ &{\color{ours}$102.7\%$} &$102.8\%$ \\ \hline
Teddy & {\color{ours}$102.5\%$} &$234.3\%$ &{\color{ours}$100.4\%$} &$100.8\%$ &{\color{ours}$100.3\%$} &$100.5\%$ \\ \hline
\end{tabular}
\caption{ {\bf Stereo:}
{\em
Showing percent of achieved energy value relative to the lower bound (closer to $100\%$ is better).
Visual results for these experiments are in Fig.~\ref{fig:res-stereo}.
Energies from \cite{Szeliski2008}.}}
\label{tab:stereo-res}
\end{table}

\begin{table}
\centering
\setlength{\tabcolsep}{.5mm}
\begin{tabular}{c||c|c||c|c||c|c}
 & \multicolumn{2}{c||}{ICM} & \multicolumn{2}{c||}{Swap} & \multicolumn{2}{c}{Expand}\\
               & {\color{ours}Ours}      & single scale & {\color{ours}Ours}   & single scale & {\color{ours}Ours}   & single scale\\
\hline \hline
House & {\color{ours}$100.5\%$} &$111.3\%$ &{\color{ours}$100.4\%$} &$100.9\%$ &{\color{ours}$102.3\%$} &$103.4\%$ \\ \hline
Penguin & {\color{ours}$106.9\%$} &$132.9\%$ &{\color{ours}$104.6\%$} &$111.3\%$ &{\color{ours}$104.0\%$} &$103.7\%$ \\ \hline
\end{tabular}
\caption{ {\bf Denoising and inpainting:}
{\em
Showing percent of achieved energy value relative to the lower bound (closer to $100\%$ is better).
Visual results for these experiments are in Fig.~\ref{fig:res-denoise}.
Energies from \cite{Szeliski2008}.} }
\label{tab:denoise-res}
\end{table}

\comment{
\begin{table}
\centering
\begin{tabular}{c|c|c||c|c}
Energy &  \#variables & \#variables & {\color{ours}Ours} & single \\
       &  (finest)     & (coarsest)     & {\color{ours}(multiscale)}     & scale \\ \hline
%       &             &              & [sec]                          & [sec] \\ \hline
Penguin &  $21,838$   & $5$          & {\color{ours}$103.7\%$}       & $111.3\%$ \\
        &             &              & {\color{ours}$20.8$ + $95.1$} [sec]  & $253.7$ [sec]\\ \hline
Venus   &  $166,222$  & $6$          & {\color{ours}$102.8\%$}        & $128.7\%$ \\
        &             &              & {\color{ours}$54.7$ + $30.7$} [sec]  & $130.1$ [sec]\\
\end{tabular}
\caption{{\bf Running times for variable coarsening ($\alpha\beta$-swap):}
{\em Examples of typical running times (in seconds). For multiscale we report the runtime for constructing the pyramid and the overall time it took to optimize coarse-to-fine.
Note that the reported times are of our unoptimized serial Matlab implementation.}}
\label{tab:variable-runtime}
\end{table}
}

%------------------------------------------------------------------------------------%
\subsection{Comparing variable correlation estimation methods}

As explained in Sec.~\ref{sec:matrix-P} the correlations between the variables are the most crucial component in constructing an effective multiscale scheme.
In this experiment we compare our energy-aware correlation measure (Sec.~\ref{sec:local-correlations}) to three methods proposed by \cite{Kim2011}: ``unary-diff", ``min-unary-diff" and ``mean-compat".
These methods estimate the correlations based either on the unary term or the pair-wise term, but {\em not both}.
We also compare to an energy-agnostic measure, that is $c_{ij}=1$ $\forall i,j$,
this method underlies \cite{Felzenszwalb2006}.
We use ICM within our framework to evaluate the influence these methods have on the resulting multiscale performance for four representative energies.

\begin{figure}
\centering
\includegraphics[width=.8\linewidth]{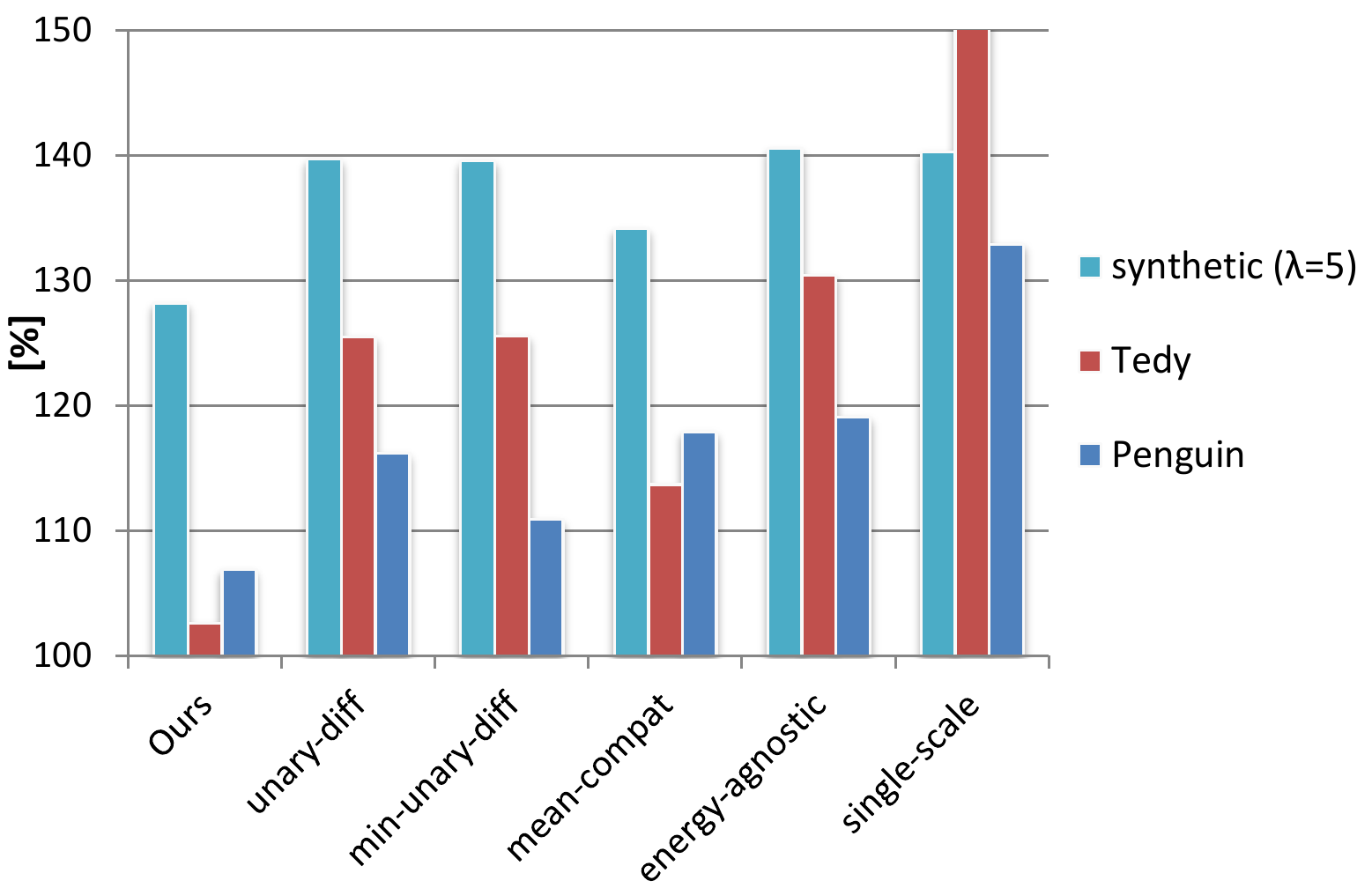}
\caption{{\bf Comparing correlation measures:}
{\em
Graphs showing percent of lower bound (closer to $100\%$ is better)
for different methods of computing variable-correlations.
Some of the bars are cropped at $\sim150\%$.
Our energy-aware measure consistently outperforms all other methods.
}}
\label{fig:comapre-weights}
\end{figure}

Fig.~\ref{fig:comapre-weights} shows percent of lower bound for the different energies.
Our measure consistently outperforms all other methods, and successfully balances between the influence of the unary and the pair-wise terms.

%------------------------------------------------------------------------------------%
\subsection{Coarsening labels}
$\alpha\beta$-swap
does not scale gracefully with the number of labels.
%is very sensitive to the number of labels.
Coarsening an energy in the labels domain (i.e., same number of variables, fewer labels) proves to significantly improve performance of $\alpha\beta$-swap, as shown in Table~\ref{tab:coarsening-v}.
For these examples constructing the energy pyramid took only milliseconds, due to the ``closed form" formula for estimating label correlations.

Our principled framework for coarsening labels improves $\alpha\beta$-swap performance for these energies.

\begin{table}
\centering
\begin{tabular}{c|c|c||c|c}
\multirow{2}{*}{Energy} & \#labels &  \#labels   & \multirow{2}{*}{{\color{ours}Ours}}& single \\
                        & (finest)   &  (coarsest)   &                                   & scale \\
\hline
Penguin     & \multirow{2}{*}{256} & \multirow{2}{*}{67} & {\color{ours}$103.6\%$}   & $111.3\%$ \\
(denoising) &                      &                     & {\color{ours}$128$} [sec] & $253$ [sec] \\ \hline
Venus       & \multirow{2}{*}{20}  & \multirow{2}{*}{4} & {\color{ours}$106.0\%$}    & $128.7\%$ \\
(stereo)    &                      &                    & {\color{ours}$100$} [sec]  & $130$ [sec]\\
\end{tabular}
%We improved both energy and run-time.} }
\caption{ {\bf Coarsening labels ($\alpha\beta$-swap):}
{\em
Working coarse-to-fine in the labels domain. We use 5 scales
with coarsening rate of $\sim0.7$.
Number of variables is unchanged.
Table shows percent of achieved energy value relative to the lower bound (closer to $100\%$ is better), and running times.}}
\label{tab:coarsening-v}
\end{table}

%------------------------------------------------------------------------------------%
\section{Conclusion}

This work presents a unified multiscale framework for discrete energy minimization
that allows for efficient and {\em direct} exploration of the multiscale landscape of the energy.
We propose two paths to expose the multiscale landscape of the energy:
one in which coarser scales involve fewer and coarser {\em variables},
and another in which the coarser levels involve fewer {\em labels}.
We also propose adaptive methods for energy-aware interpolation between the scales.
Our multiscale framework significantly improves optimization results for challenging energies.

Our framework provides the mathematical formulation that ``bridges the gap" and relates multiscale discrete optimization and algebraic multiscale methods used in PDE solvers (e.g., \cite{Brandt1986}).
This connection allows for methods and practices developed for numerical solvers to be applied in multiscale discrete optimization as well.

%\begin{acknowledgements}
%We would like to thank Maria Zontak and Daniel Glasner for their insightful remarks and discussions.
%\end{acknowledgements}
%
%% BibTeX users please use one of
%%\bibliographystyle{spmpsci}      % mathematics and physical sciences
%%\bibliographystyle{spphys}       % APS-like style for physics
%\bibliography{disc_amg}   % name your BibTeX data base
%
%
%\end{document}
% end of file

%------------------------------------------------------------------------------------%

\section*{Appendix}
\begin{subappendices}

\section{Derivation of eq.~(\ref{eq:EngU})}
\label{sec:ms-deriv-U}

In this work we consider discrete pair-wise minimization problems of the form:

~\hfill $E\left(L\right)=\sum_{i\in\VV} \psi_i\left(l_i\right) + \sum_{\left(i,j\right)\in\EE} w_{ij}\cdot \psi\left(l_i,l_j\right)$ \hfill (\ref{eq:GenEng})

%\begin{eqnarray*}
%E\left(L\right)&=&\sum_{i\in\VV} \psi_i\left(l_i\right) + \sum_{\left(i,j\right)\in\NN} w_{ij}\cdot \psi\left(l_i,l_j\right) (\ref{eq:GenEng})
%\end{eqnarray*}

Using the following parameterizations: $D\in\mathbb{R}^{n\times l}$ s.t. $D_{i,\alpha}\deff \psi_i(\alpha)$, and $V\in\mathbb{R}^{l\times l}$ s.t. $V_{\alpha,\beta}\deff \psi\left(\alpha,\beta\right)$
we claim that (\ref{eq:GenEng}) is equivalent to:

~\hfill $E\left(U\right)=Tr\left(DU^T+WUVU^T\right)$ \hfill (\ref{eq:EngU})

~\hfill s.t. $U\in\left\{0,1\right\}^{n\times l},\ \sum_{\alpha=1}^l U_{i\alpha}=1$ \hfill (\ref{eq:const-U})

%\begin{eqnarray*}
%E\left(U\right)&=&Tr\left(DU^T+WUVU^T\right) \hfill (\ref{eq:EngU})  \\ %\label{eq:EngU} \\
%      & \mbox{s.t.} & U\in\left\{0,1\right\}^{n\times l},\ \sum_{\alpha=1}^l U_{i\alpha}=1 \hfill (\ref{eq:const-U}) % \label{eq:const-U}
%\end{eqnarray*}
Assuming both $V$ and $W$ are symmetric\footnote{if they are not we need to be slightly more careful with transposing them, but roughly similar expression can be derived.}.
Looking at the different components in (\ref{eq:EngU}):
\[
\left[VU^T\right]_{\alpha j} = \sum_\beta V_{\alpha \beta}U_{j \beta}
\]
\begin{eqnarray*}
\left[UVU^T\right]_{ij} &=& \sum_\alpha U_{i\alpha}\left[VU^T\right]_{\alpha j} \\
&=& \sum_\alpha U_{i\alpha} \sum_\beta V_{\alpha \beta}U_{j \beta}\\
&=& \sum_{\alpha\beta} V_{\alpha\beta}U_{i\alpha}U_{j\beta} \\
&=& \sum_{\alpha\beta} \psi\left(\alpha,\beta\right)U_{i\alpha}U_{j\beta} \\
&=& \psi\left(l_i, l_j\right)
\end{eqnarray*}
Looking at the trace of the second term:
\begin{eqnarray}
Tr\left(WUVU^T\right) &=& \sum_i \left[WUVU^T\right]_{ii} \nonumber\\
&=& \sum_i \sum_j w_{ij} \left[UVU^T\right]_{ji} \nonumber\\
&=& \sum_i \sum_j w_{ij} \psi\left(l_j, l_i\right) \nonumber\\
&=& \sum_{ij} w_{ij} \psi\left(l_i, l_j\right) \label{eq:pair-wise}
\end{eqnarray}
As for the unary term:
\begin{eqnarray}
Tr\left(DU^T\right) &=& \sum_i \left[DU^T\right]_{ii} \nonumber\\
&=& \sum_i \sum_\alpha D_{i\alpha}U_{i\alpha} \nonumber\\
&=& \sum_i \sum_\alpha \psi_i\left(\alpha\right)U_{i\alpha} \nonumber\\
&=& \sum_i \psi_i\left(l_i\right) \label{eq:ms-unary}
\end{eqnarray}
Putting (\ref{eq:ms-unary}) and (\ref{eq:pair-wise}) together we get:
\begin{eqnarray*}
Tr\left(DU^T+WUVU^T\right)&=&Tr\left(DU^T\right)+Tr\left(WUVU^T\right)\\
&=& \sum_i \psi_i\left(l_i\right) + \sum_{ij} w_{ij} \psi\left(l_i, l_j\right) \quad \square
\end{eqnarray*}

%\hfill $\square$

Note that the diagonal of $W$ is assumed to be zero: this is a reasonable assumption as $w_{ii}$ represents an interaction of variable $i$ with itself.
This type of interaction is well represented by the unary term $D_i$.
When coarsening the energy it may happen that $W^c$ will no longer have zeros on the diagonal.
This case may arise when a single coarse variable represents neighboring fine scale variables.
In that case the fine scale pair-wise interaction should be absorbed into the coarse scale unary term.
It is easy to see that the term $w_{ii}V_{\alpha\alpha}$ should be added to the unary term $D_{i\alpha}$, whenever $W$ has non zeros on the diagonal.
After this rectification, the non-zeros entries on the diagonal of $W$ can be set to zero.

%------------------------------------------------------------------------------------%
\section{More General Energy Functions}

This chapter~(\ref{cp:multiscale}) has focused on the construction of an energy pyramid for energies of the form~(\ref{eq:GenEng}).
However, this form does not cover all possible pair-wise energies.
We have used this slightly restricted form throughout the paper since it emphasizes the simplicity of the algebraic derivation of our multiscale framework.
Nevertheless, our multiscale framework can be as easily applied to more general energies.

%------------------------------------------------------------------------------------%
\subsection{Pair-wise}

A general from for pair-wise energy over a graph $\GG=\left(\VV,\EE\right)$ can be written as
\begin{equation}
E\left(L\right) = \sum_{i\in\VV} \varphi_i\left(l_i\right) + \sum_{\left(i,j\right)\in\EE} \varphi_{ij}\left(l_i,l_j\right)
\label{eq:gen-pair-energy}
\end{equation}

In this more general form, the pair-wise term can be entirely different for each pair: $\varphi_{ij}\left(l_i,l_j\right)$.
This is in contrast to the energy~(\ref{eq:GenEng}) where the pair-wise terms differ only by the scaling factor $w_{ij}$ of a {\em single} fixed term $\varphi\left(l_i,l_j\right)$.
The Photomontage energies of \cite[\S4.2]{Szeliski2008} are an example of such general pair-wise energies.

Instead of using a pair of matrices $V$ and $W$ to parameterize the pair-wise terms, we use a collection of matrices $\left\{V^{ij}\right\}_{ij\in\EE}$ for the same purpose in the general settings.
Each matrix $V^{ij}$ is of size $l\times l$ and is defined as $V^{ij}_{\alpha\beta} \deff \varphi_{ij}\left(\alpha,\beta\right)$.

A general energy is now parameterized by $\left(n, l, D, \left\{V^{ij}\right\}_{ij\in\EE} \right)$.

\noindent{\bf Coarsening variables:}
The computation of the interpolation matrix $P$ is unchanged.
ICM can be applied to energies of the form~(\ref{eq:gen-pair-energy}) to estimate agreement between neighboring variables.
These agreements are then used to compute the interpolation matrix $P$ (Sec.~\ref{sec:matrix-P}).

In order to write the coarsening of the pair-wise term we need to introduce some notations:
Let $i,j$ be variables at the fine scale, and $I,J$ denote variables in the coarse scale.
An entry, $P_{iI}$, in the interpolation matrix indicates how fine scale variable $i$ is affected by a coarse scale variable $I$.
Given an interpolation matrix $P$ we can coarsen the energy by
\begin{eqnarray}
D^c & \deff & P^T D^f \nonumber \\
\mbox{$V^{IJ}$}_{\alpha\beta} & \deff & \sum_{ij\in\EE^f} \mbox{$V^{ij}$}_{\alpha\beta} P_{iJ}P_{jJ}
\label{eq:gen-pair-coarse-variables}
\end{eqnarray}
The coarse graph, $\EE^c$, is defined by all the non-zero pair-wise terms $\mbox{$V^{IJ}$}$.
The coarser energy is now parameterized by
\begin{equation}
\left(n^c, l, D^c, \left\{\mbox{$V^{IJ}$}\right\}_{ij\in\EE^c}\right) \nonumber
\end{equation}

\noindent{\bf Coarsening labels:}
In this case, the computation of the interpolation matrix $\hat{P}$ is not trivial,
since we no longer have a single matrix $V$ to derive the agreements from.
We leave this issue of deriving an efficient interpolation matrix for the general case for future work.
However, given a matrix $\hat{P}$ the coarsening of labels can be done easily:
\begin{eqnarray}
D^{\hat{c}} &\deff& D^{\hat{f}}\hat{P} \nonumber \\
\mbox{$V^{ij}$}^{\hat{c}} &\deff& \hat{P}^T \mbox{$V^{ij}$}^{\hat{f}} \hat{P} \;\; \forall ij\in\EE
\label{eq:gen-pair-coarse-labels}
\end{eqnarray}
yielding a coarser energy parameterized by
\begin{equation}
\left(n, l^c, D^{\hat{c}}, \left\{\mbox{$V^{ij}$}^{\hat{c}}\right\}_{ij\in\EE}\right) \nonumber
\end{equation}
with the same number of variables, $n$, but fewer labels $l^c<l^f$.

\

It is fairly straight forward to see that the energy~(\ref{eq:GenEng}) is a special case of the more general form~(\ref{eq:gen-pair-energy}) and so the coarsening of variables and labels of Eq.~(\ref{eq:EngC}) and~(\ref{eq:EngC-V}) can be seen as special cases of Eq.~(\ref{eq:gen-pair-coarse-variables}) and~(\ref{eq:gen-pair-coarse-labels}) resp.

%------------------------------------------------------------------------------------%
\subsection{High-order energies}
Discrete energies may involve terms that are beyond pair-wise: that is, describe interaction between sets of variables. These energies are often referred to as high-order energies.
Examples of such energies can be found in e.g., \cite{Kohli2009,Rother2009}.

A high order energy is defined over a hyper-graph $\left(\VV,\calS\right)$ where the hyper-edges are subsets of variables $s\in\calS$ s.t. $s\subseteq\VV$.
\begin{equation}
E\left(L\right) = \sum_{i\in\VV} \varphi_i\left(l_i\right) + \sum_{s\in\calS} \varphi_{s}\left(\left\{l_i\left|i\in s\right.\right\}\right)
\label{eq:high-ord-energy}
\end{equation}

Where the high-order terms $\varphi_{s}$ are $\abs{s}$-way discrete functions:
\begin{equation}
\varphi_s:\left\{1,\ldots,l\right\}^{\abs{s}}\rightarrow\mathbb{R}
\end{equation}

A high-order energy of the form~(\ref{eq:high-ord-energy}) can be parameterized using tensors.
Each high-order term, $\varphi_s$, is parameterized by a $\abs{s}$-order tensor
\begin{equation}
V^s_{\alpha,\beta,\ldots,\zeta} \deff \varphi_s\left( l_i = \alpha, l_j=\beta,\ldots, l_m = \zeta \left| s=\left\{i,j,\ldots,m\right\} \right. \right)
\end{equation}

A high-order energy~(\ref{eq:high-ord-energy}) is now parameterized by $\left(n, l, D, \left\{V^s\right\}_{s\in\calS}\right)$.

\noindent {\bf Coarsening variables:}
The computation of the interpolation matrix $P$ is unchanged.
ICM can be applied to energies of the form~(\ref{eq:high-ord-energy}) to estimate agreement between neighboring variables.
These agreements are then used to compute the interpolation matrix $P$ (Sec.~\ref{sec:matrix-P}).

Given an interpolation matrix $P$ we can coarsen the energy by
\begin{eqnarray}
D^c & \deff & P^T D^f \nonumber \\
V^{S=\left\{I,J,\ldots,M\right\}} & \deff  &  P_{iI}\cdot P_{jJ}\cdot\ldots  V^{\left\{s=i,j,\ldots,m\right\}} \; \; \forall s\in\calS^f
\end{eqnarray}
For all coarse variables $I,J,\ldots,M$ with non-zero entry $P_{iI}, P_{jJ},\ldots$.
These non-zeros interactions in $P$ defines the coarse scale hyper-edges in $\calS^c$.
Note that when two (or more) fine-scale variable $i,j,\ldots$ are represented by the same coarse variable $I$, then the size of $S$ (the coarse scale hyper-edge) is reduced relative to the size of $s$ (the fine scale hyper-edge).

\noindent{\bf Coarsening labels:}
In this case, the computation of the interpolation matrix $\hat{P}$ is not trivial,
since we no longer have a clear representation of the interactions between the different labels.
We leave this issue of deriving an efficient interpolation matrix for the general case for future work.
However, given a matrix $\hat{P}$ the coarsening of labels can be done easily.
Using $\calpha,\cbeta$ to denote coarse scale labels

\begin{eqnarray}
D^{\hat{c}} &\deff& D^{\hat{f}}\hat{P} \nonumber \\
\mbox{$V^{s}$}^{\hat{c}}_{\calpha,\cbeta,\ldots} &\deff& \sum_{\alpha,\beta,\ldots}\mbox{$V^{s}$}^{\hat{f}}_{\alpha,\beta,\ldots} \hat{P}_{\alpha\calpha} \cdot \hat{P}_{\beta\cbeta} \cdot \ldots
\label{eq:hi-ord-coarse-labels}
\end{eqnarray}
yielding a coarser energy parameterized by
\begin{equation}
\left(n, l^c, D^{\hat{c}}, \left\{\mbox{$V^{s}$}^{\hat{c}}\right\}_{s\in\calS}\right) \nonumber
\end{equation}
with the same number of variables, $n$, but fewer labels $l^c<l^f$.

\end{subappendices}

%% file: discussion.tex
\part{Discussion}  	

In this work I motivated the use of challenging discrete energies, beyond semi-metric, for different tasks in computer vision.
I conclude that despite the computational challenges posed by stepping away form the well studied smoothness-encouraging energies, the gain in terms of expressiveness and quality of solution is worth while.
For some applications these energies describe better the underlying process and therefore better suited for modeling.
Moreover, practical algorithms and approaches were proposed in this work to cope with the resulting challenging optimization tasks.

Considering the task of unsupervised clustering as an example, using the arbitrary Potts energy of the Correlation Clustering functional allows not only to model the partition of the data into different clusters, but also to recover the underlying number of clusters.
This recovery of the number of cluster is an issue usually evaded by many clustering algorithms that assume this information is somehow given to them as an input.
This thesis showed how the challenging optimization of the CC functional lies at the core of several diverse applications, such as: sketching the common, unsupervised face identification and interactive segmentation.

Focusing on the CC optimization example, this work also proposed several practical approximation algorithms.
The CC functional was casted as a special case of an arbitrary Potts energy minimization.
This discrete formulation of the functional allows to adapt and utilize known methods for discrete optimization resulting with CC optimization algorithms capable of handling large number of variables (two orders of magnitude more than existing methods).

Another example of challenging arbitrary energy is the task of 3D shape reconstruction.
The problem of 3D reconstruction from two images with common lighting conditions and known object motion with respect to static camera is discretized to form a challenging energy.
This energy emerges from the induced integrality constraints on the recovered surface.
It is defined over a very large label space (a few thousand of labels).
Despite the challenging pair-wise terms and the large label space successful approximation scheme was proposed and adapted for this task.

When it comes to learning the parameters of discrete energies for various tasks in computer vision,
it so happens that the underlying data may be better explained by contrast enhancing terms, as in the Chinese characters and the body parts.
In fact when the learning procedure is not restricted it may prefer contrast-enhancing functions as better depictions for the underlying statistical behavior of the training set.
Therefore, restricting the learned model to only submodular or semi-metric energies might severely compromise its ability to generalize to novel exemplars.

Coping with the optimization of the resulting challenging energies is a difficult task.
Yet, this work proposes several methods that provide good empirical approximations for a variety of practical energies.
Our unified discrete multiscale framework succeeds in providing good approximations, in practice, for a variety of challenging energies.
Our multiscale framework exploits the underlying {\em multiscale landscape} of the given energy and exposes it to allow for an efficient and effective coarse-to-fine optimization process.
Formulating the coarsening procedure in a clear algebraic formulation allows us to propose two paths to expose the multiscale landscape of the energy:
one in which coarser scales involve fewer and coarser {\em variables},
and another in which the coarser levels involve fewer {\em labels}.
We also propose adaptive methods for energy-aware interpolation between the scales.
This approach exploits the underlying {\em multiscale landscape} of the given energy and exposes it to allow for an efficient and effective coarse-to-fine optimization process.
Our framework provides the mathematical formulation that ``bridges the gap" and relates multiscale discrete optimization and algebraic multiscale methods used in PDE solvers (e.g., \cite{Brandt1986}).
This connection allows for methods and practices developed for numerical solvers to be applied in multiscale discrete optimization as well.